\PassOptionsToPackage{table}{xcolor}
\documentclass[preprint,12pt]{elsarticle}

\usepackage{amssymb}
\usepackage{amsmath}
\usepackage{enumitem}  % compact, customizable lists [web:40][web:41]

\usepackage{multirow}
\usepackage{xcolor}%
\usepackage{booktabs}
\usepackage{tablefootnote}
\usepackage{subcaption}
\usepackage{nicematrix}
\usepackage[hidelinks]{hyperref}

\usepackage{booktabs}
\usepackage{siunitx}
\usepackage{float}
\usepackage{subcaption}
\usepackage{bbm}

\usepackage{ulem}   % for \uline
\makeatletter
\newcommand\footnoteref[1]{\protected@xdef\@thefnmark{\ref{#1}}\@footnotemark}
\makeatother

\journal{Expert Systems with Applications}

\begin{document}

\begin{frontmatter}

\title{BenchSeg: A Large-Scale Dataset and Benchmark for Multi-View Food Video Segmentation}

\author[1,4]{Ahmad AlMughrabi*}

\author[1,4]{Guillermo Rivo}

\author[1,4]{Carlos Jiménez-Farfán}

\author[1,4]{Umair Haroon}

\author[1]{Farid Al-Areqi}

\author[tum]{Hyunjun Jung}

\author[tum]{Benjamin Busam} 

\author[2,5]{Ricardo Marques}

\author[1,3,5]{Petia Radeva}

\affiliation[1]{organization={Matemàtiques i Informàtica, Universititat de Barcelona},
            addressline={Gran Via de les Corts Catalanes, 585, L'Eixample}, 
            city={Barcelona},
            postcode={08007}, 
            country={Spain}}

\affiliation[2]{organization={Department of Engineering, Pompeu Fabra University},
            addressline={Carrer de Tànger, 122-140}, 
            city={Barcelona},
            postcode={08018}, 
            country={Spain}}

\affiliation[3]{organization={Institut de Neurosciències, Universititat de Barcelona},
            addressline={Passeig de la Vall d’Hebron, 171}, 
            city={Barcelona},
            postcode={08035}, 
            country={Spain}}

\affiliation[tum]{organization={Photogrammetry and Remote Sensing, Technical University of Munich},
            addressline={Arcisstraße 21}, 
            city={München},
            postcode={80333}, 
            country={Germany}}

\affiliation[4]{orgnization={Equal Contributions}}
\affiliation[5]{orgnization={Equal Supervision}}

%% Abstract
\begin{abstract}
Food image segmentation is a critical task for dietary analysis, enabling accurate estimation of food volume and nutrients. However, current methods suffer from limited multi-view data and poor generalization to new viewpoints. We introduce BenchSeg, a novel multi-view food video segmentation dataset and benchmark. BenchSeg aggregates 55 dish scenes (from Nutrition5k, Vegetables \& Fruits, MetaFood3D, and FoodKit) with 25,284 meticulously annotated frames, capturing each dish under free 360° camera motion. We evaluate a diverse set of 20 state-of-the-art segmentation models (e.g., SAM-based, transformer, CNN, and large multimodal) on the existing FoodSeg103 dataset and evaluate them (alone and combined with video-memory modules) on BenchSeg. Quantitative and qualitative results demonstrate that while standard image segmenters degrade sharply under novel viewpoints, memory-augmented methods maintain temporal consistency across frames. Our best model based on a combination of SeTR-MLA+XMem2 outperforms prior work (e.g., improving over FoodMem by ~2.63\% mAP), offering new insights into food segmentation and tracking for dietary analysis. In addition to frame-wise spatial accuracy, we introduce a dedicated temporal evaluation protocol that explicitly quantifies segmentation stability over time through continuity, flicker rate, and IoU drift metrics. This allows us to reveal failure modes that remain invisible under standard per-frame evaluations. We release BenchSeg to foster future research. The project page including the dataset annotations and the food segmentation models can be found at \footnote{https://amughrabi.github.io/benchseg}.
\end{abstract}

\begin{graphicalabstract}
\centering
\includegraphics[trim={2.3cm 1.9cm 2.3cm 2cm},clip,width=.85\linewidth]{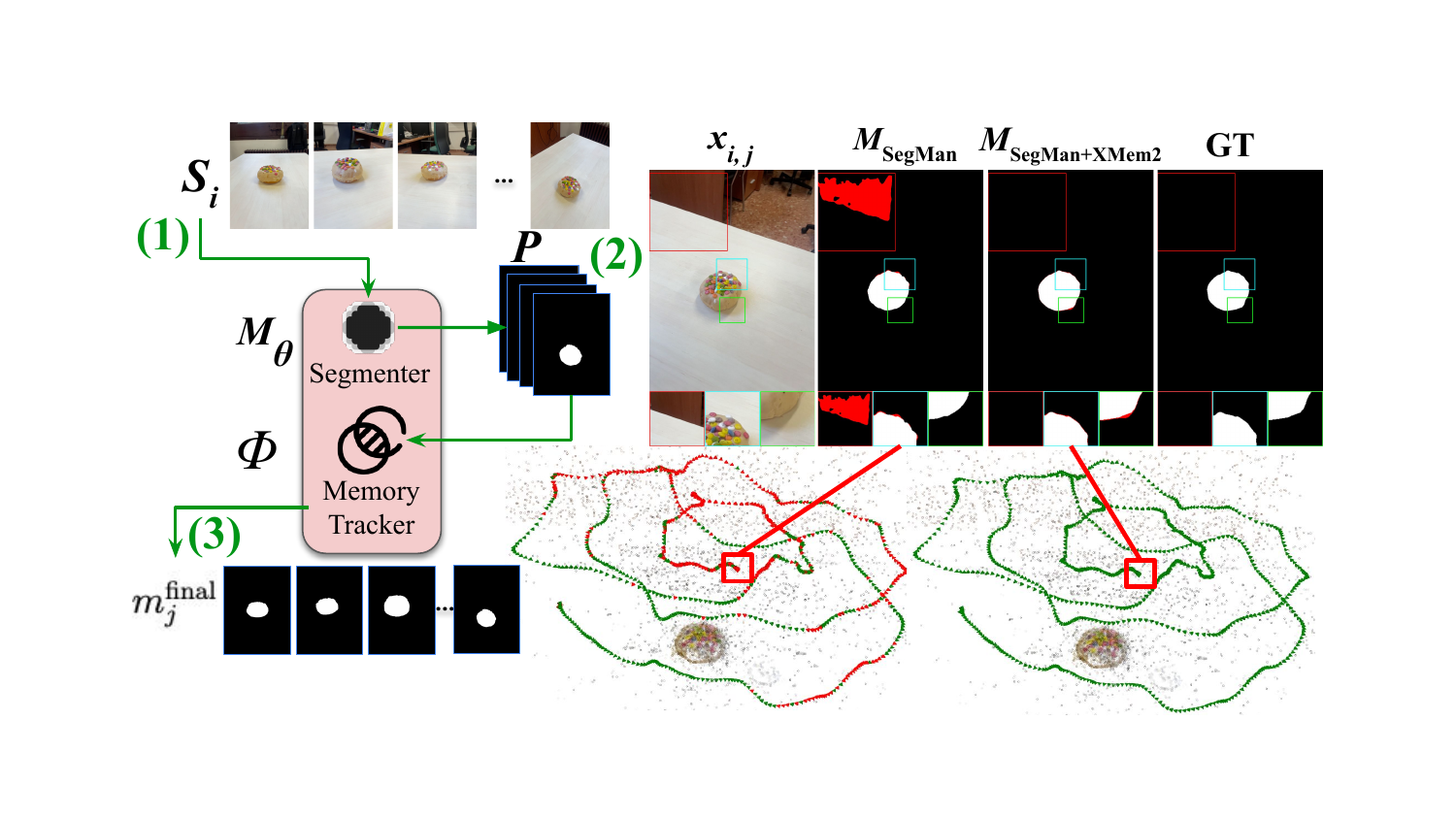}
\centering
\includegraphics[trim={0cm 0cm 8cm 0cm},clip,width=.85\linewidth]{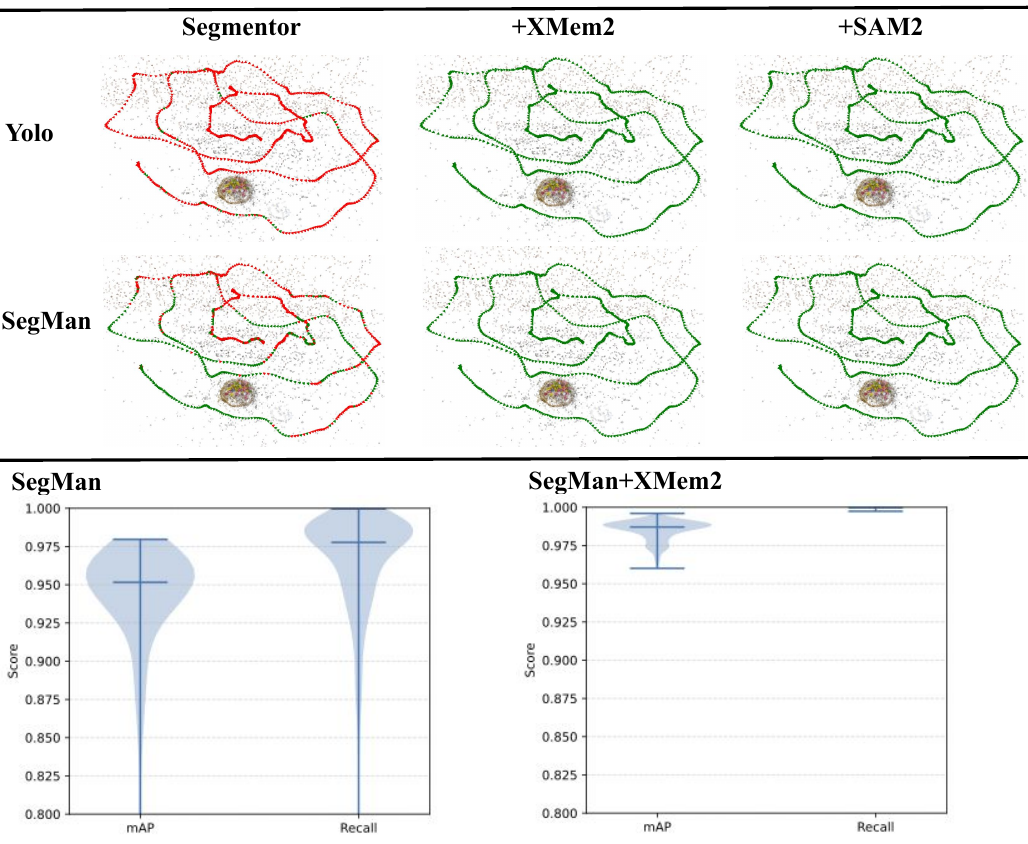}

\end{graphicalabstract}

\begin{highlights}
\item We introduce BenchSeg, a multi-scene food segmentation benchmark comprising \textbf{25,284 manually annotated} frames across 55 dishes, curated from the four public food datasets.

\item BenchSeg provides free-motion, hemispherical-coverage video sequences annotations, enabling \textbf{rigorous evaluation of generalization} across diverse camera trajectories. 
% \pr{the images are coming from the public datasets, the reviewer will argue that this is not our contribution}

\item We benchmark \textbf{20 state-of-the-art} architectures and hybrid segmentation--tracking systems under a unified cross-dataset protocol.

\item We empirically evaluate cross-dataset generalization and robustness to unseen camera poses as all models are trained solely on FoodSeg103, ensuring that the evaluation isolates cross-dataset generalization and robustness to unseen camera poses.

\item We perform a comprehensive quantitative evaluation using \textbf{$\mathrm{mAP}$, Recall, Precision, F1, IoU, and Accuracy}, demonstrating that most models exhibit significant degradation when exposed to unfamiliar viewpoints and motion patterns. 

\item We introduce a set of temporal stability metrics to quantify flicker, continuity, and segmentation drift in video segmentation.

\item We show that hybrid 2D segmentor and memory-based tracking models, where per-frame masks are temporally propagated via a memory module, achieve the most stable performance across datasets, revealing promising future directions. 

\item We provide detailed comparisons of computational efficiency, including model size, memory footprint, and inference speed, to help deployment-oriented choices for dietary assessment workflows.
\end{highlights}

%% Keywords
\begin{keyword}

food video segmentation \sep dietary assessment \sep benchmark dataset \sep video-based segmentation \sep memory-augmented models \sep cross-dataset generalization



\end{keyword}

\end{frontmatter}

\section{Introduction}
\label{sec:introduction}

Accurate and reliable segmentation of food items in images and videos is a foundational component of automated dietary assessment systems. Segmentation delineates the spatial extent of edible items, enabling downstream tasks such as recognition, portion estimation, and nutritional analysis; errors at this stage propagate through the pipeline and systematically degrade subsequent estimates. While recent advances in deep learning have produced highly capable single-image segmentors, the specific demands of dietary applications—including large intra-class variation, occlusion by utensils or hands, complex tableware, and unconstrained capture conditions—place stringent requirements on both spatial completeness and temporal stability \cite{wu2021large,thames2021nutrition5k}.

Existing food segmentation datasets and benchmarks have contributed significantly to progress, yet remain limited with respect to the multi-view, video-centric scenarios encountered in practical dietary monitoring. Datasets such as FoodSeg103 \cite{wu2021large} provide dense ingredient-level annotations for still images and have served as key resources for training modern architectures. Larger-scale collections incorporating video or multi-view capture, including Nutrition5k (N5k) \cite{thames2021nutrition5k}, Vegetables \& Fruits (V\&F) \cite{steinbrener2023learning}, MetaFood3D (MTF) \cite{chen2024metafood3d}, and FoodKit (FKit) \cite{haroon2025vole}, increase scene diversity but often lack dense per-frame annotations or do not emphasize free-motion camera trajectories that induce strong viewpoint variation \cite{thames2021nutrition5k, steinbrener2023learning, chen2024metafood3d, haroon2025vole}. As a result, models trained and validated primarily on static or canonical-view imagery are rarely exposed to the appearance deformations, occlusions, and specularities that arise when users record short sweeps or handheld clips of meals.

This discrepancy manifests as a concrete generalization problem. Single-frame segmentors that achieve high accuracy on in-distribution, canonical-view images frequently produce fragmented, inconsistent, or incomplete masks when applied frame-by-frame to free-motion videos. Common failure modes include: (i) missing thin or foreshortened components visible only from certain angles, (ii) abrupt label switching and mask flicker across adjacent frames, and (iii) spurious inclusion of non-food objects with similar color or texture. These behaviors suggest that architectural scaling alone—e.g., larger backbones or more expressive attention mechanisms—is insufficient unless training data and evaluation protocols explicitly address multi-view variability and temporal coherence \cite{zheng2021rethinking, liu2021swin}.

To mitigate these issues, recent work has explored memory-augmented and hybrid segmentation–tracking paradigms. Two-stage approaches that combine a strong per-frame segmentor with a temporal propagation module (e.g., space–time memory networks) exploit redundancy across frames to recover missing regions and enforce consistency \cite{bekuzarov2023xmem++, almughrabi2025foodmem}. Promptable or foundation models, such as SAM-based methods, provide high-quality instance masks in single images and, when coupled with temporal aggregation, can further improve per-frame quality \cite{lan2023foodsam, ravi2024sam}. Nevertheless, open questions remain regarding the relative benefits of different backbone families, the robustness of memory propagation under large viewpoint changes, and the trade-offs between accuracy, temporal stability, and runtime in realistic dietary scenarios.

Importantly, most existing evaluations remain predominantly frame-centric, treating each frame independently. While such metrics capture instantaneous spatial accuracy, they fail to reflect temporal artifacts such as flickering, mask fragmentation, and gradual drift, which strongly affect user perception in real-world video scenarios. Consequently, methods with similar per-frame accuracy may exhibit drastically different temporal behavior, motivating the need for explicit stability-oriented evaluation criteria.

\subsection{From Datasets to Benchmarks: Evaluation as Diagnosis}
\label{subsec:benchmark_philosophy}

The transition from a static dataset to a formalized benchmark requires a shift from data accumulation to systematic failure analysis. In contemporary computer vision research, a benchmark is defined not merely by its scale, but by its capacity to expose the boundary conditions under which algorithms succeed or fail. In food-centric video understanding, these boundary conditions include extreme viewpoint variation, occlusion, appearance deformation, and temporal inconsistency—failure modes that are largely invisible to static-image evaluations.

Guided by these principles, BenchSeg is designed as a multidimensional evaluation benchmark that goes beyond single-metric or purely frame-wise performance assessment. Rather than reporting only aggregate spatial accuracy, our goal is to enable diagnostic analysis of method behavior across multiple complementary axes. In particular, we treat temporal stability as a first-class evaluation dimension, alongside conventional measures of spatial segmentation quality. While standard benchmarks implicitly assume that strong frame-wise accuracy correlates with stable behavior over time, this assumption often fails in practice. Methods with comparable per-frame performance may exhibit markedly different temporal characteristics, including flickering, discontinuities, and gradual degradation. To explicitly capture these phenomena, BenchSeg introduces dedicated temporal stability metrics that quantify continuity, abrupt changes, and drift across frames. This design allows us to systematically expose failure modes that remain invisible under static or frame-centric evaluation protocols, providing a more realistic and actionable assessment of real-world video segmentation behavior. As summarized in Table~\ref{tab:benchmarking_standards}, BenchSeg aligns with the defining characteristics of mature benchmarks by providing standardized evaluation protocols, explicit multi-view consistency metrics, and a large diagnostic suite of over 20 heterogeneous baselines. This design allows us to quantify not only peak performance, but also the robustness, failure modes, and generalization limits of modern food video segmentation systems.

\begin{table*}[htb]
\centering
\tiny
\setlength{\tabcolsep}{1pt}
\caption{Chronological comparison of BenchSeg with representative segmentation and food-computing benchmarks (2021--2026). Our benchmark is distinguished by the synthesis of ontological depth, multi-view invariance, and comprehensive diagnostic baselines.}
\label{tab:benchmarking_standards}
\begin{tabular}{@{}lcccccccc@{}}
\toprule
\textbf{Benchmark} & \textbf{Year} & \textbf{Food} & \textbf{Video} & \textbf{Ontology$^1$} & \textbf{Protocol} & \textbf{Multi-View} & \textbf{Baselines} & \textbf{Diagnostics} \\ \midrule
\rowcolor{gray!12}FoodSeg103 \cite{wu2021large} & 2021 & \checkmark &  & \checkmark & \checkmark &  & 5 &  \\
Nutrition5k \cite{thames2021nutrition5k} & 2021 & \checkmark & \checkmark &  &  & \checkmark & 3 &  \\
\rowcolor{gray!12}VIPSeg \cite{miao2022large} & 2022 &  & \checkmark & \checkmark & \checkmark &  & 8 & \checkmark \\
FoodSAM \cite{lan2023foodsam} & 2023 & \checkmark &  &  & \checkmark &  & 4 &  \\
\rowcolor{gray!12}V\&F \cite{steinbrener2023learning} & 2023 & \checkmark & \checkmark & \checkmark &  &  & 2 &  \\
FoodMem \cite{almughrabi2025foodmem} & 2024 & \checkmark & \checkmark & \checkmark & \checkmark &  & 6 &  \\
\rowcolor{gray!12}MetaFood3D \cite{chen2024metafood3d} & 2024 & \checkmark & \checkmark &  &  & \checkmark & 4 &  \\
MeViS \cite{ding2025mevis} & 2025 &  & \checkmark & \checkmark & \checkmark &  & 12 & \checkmark \\
\rowcolor{gray!12}FoodKit \cite{haroon2025vole} & 2025 & \checkmark & \checkmark & \checkmark &  &  & 5 &  \\ \midrule
\textbf{Ours} & \textbf{2026} & \textbf{\checkmark} & \textbf{\checkmark} & \textbf{\checkmark} & \textbf{\checkmark} & \textbf{\checkmark} & \textbf{20} & \textbf{\checkmark} \\ \bottomrule
\end{tabular}

\vspace{0.5em}
\tiny
\raggedright
$^1$ Ontology refers to binary edible-region definition.
\end{table*}

Motivated by these gaps—and by the absence of benchmarks that explicitly probe multi-view and temporal failure modes—we introduce \textbf{BenchSeg}, a large-scale benchmark for multi-view food video segmentation. BenchSeg is designed to: (i) provide dense, per-frame food masks across free-motion, hemispherical capture trajectories drawn from multiple existing collections, and (ii) enable systematic evaluation of both image-only and hybrid segmentation–tracking methods with respect to generalization across camera poses and scene variation. In doing so, the benchmark supports rigorous comparison of contemporary CNN- and transformer-based architectures, promptable foundation models, and memory-aware pipelines, while explicitly surfacing the failure modes that must be addressed for real-world dietary monitoring.

\section{Related work}
\label{sec:related_work}
Accurate food segmentation is a foundational component of automated dietary assessment systems because it enables downstream tasks such as portion recognition, food classification, and temporal tracking of consumption. Prior work on food segmentation has proceeded along two complementary directions: (1) construction of domain-specific datasets and benchmarks tailored to food imagery, and (2) development of segmentation architectures and video-tracking methods that address the challenges posed by food appearance, occlusion, and scene variability. Below, we summarize the most relevant contributions in these areas and identify the gaps that motivate the BenchSeg benchmark.

\subsection{Related Datasets}
Several large-scale image-level datasets have become standard for training and evaluating food segmentation models. FoodSeg103 provides fine-grained, ingredient-level annotations across hundreds to thousands of dish images and has been widely used to train semantic and instance segmentors \cite{wu2021large}. N5K supplies multi-view captures of meals and has been used for multi-view analysis and dietary tasks, but in many cases, pixel-level segmentation annotations are are unavailable or provided only for a limited subset of images and are not uniformly available for free-motion video sequences, which restricts its suitability for evaluating temporally consistent image-level segmentation models \cite{thames2021nutrition5k}. The V\&F collection, MTF, and FKit further expand domain coverage by including specific classes (produce-oriented scenes), multi-view/3D captures, and controlled-turntable scans; however, these datasets either lack dense segmentation mask annotations or provide them only for restricted scenarios, constraining their use for comprehensive segmentation evaluation \cite{steinbrener2023learning,chen2024metafood3d,haroon2025vole}.

\begin{table}[t]
\centering
\tiny
\setlength{\tabcolsep}{1pt}
\caption{Summary of food digitization datasets. The table reports dataset size, available modalities, camera pose information, acquisition setup, and whether free camera motion is supported. A \checkmark indicates availability. Methods are sorted in descending order according to publication year.}
\label{tab:food-datasets}

\begin{tabular}{l l l c c c c c c c c l}
\toprule
\textbf{Year$\blacktriangledown$} & \textbf{Dataset} & \textbf{Items / Categories}
& \textbf{Mask}
& \multicolumn{5}{c}{\textbf{Modalities}}
& \textbf{Pose}
& \textbf{Free}
& \textbf{Setup} \\

\cmidrule(lr){5-9}
 &  &  &  & RGB & Depth & IMU & LiDAR & CT &  &  &  \\

\midrule
\rowcolor{gray!12}2025 & Fkit$^1$ & 21 household food objects
&  & \checkmark & \checkmark & \checkmark &  & 
& \checkmark & \checkmark & Mobile \\

2024 & MTF$^2$ & 637--743 items, 108--131 classes
& auto & \checkmark & \checkmark &  &  & 
& \checkmark & \checkmark & Scanner \\

\rowcolor{gray!12}2024 & SimpleFood45 & 12--45 food types
&  & \checkmark & \checkmark &  &  & 
& \checkmark &  & Scanner \\

2024 & FruitNeRF & Various fruits
&  & \checkmark &  &  &  & 
& \checkmark &  & -- \\

\rowcolor{gray!12}2024 & MozzaVID & 149 mozzarella samples
&  &  &  &  &  & \checkmark
&  &  & -- \\

2024 & MozzaVID & 591--37,824 slices
&  &  &  &  &  & \checkmark
&  &  & -- \\

\rowcolor{gray!12}2024 & AmodalAppleSize & 3,925 Fuji + 2,731 Elstar apples
&  & \checkmark & \checkmark &  &  & 
& \checkmark &  & Orchard \\

2023 & NutritionVerse-3D & 105 / 52 synthetic / scanned models
&  & \checkmark & \checkmark &  &  & 
&  &  & Synthetic \\

\rowcolor{gray!12}2023 & MADIMA23 & Meals and individual items
&  & \checkmark & \checkmark & \checkmark & \checkmark & 
&  & \checkmark & Multi-sensor \\

2023 & V\&F$^3$ & $\sim$55 items, 11 classes
&  &  &  & \checkmark &  & 
&  & \checkmark & Handheld \\

\rowcolor{gray!12}2021 & Real Food Dataset & 50 fixed + 416 handheld dishes
&  & \checkmark & \checkmark &  &  & 
&  & \checkmark & Mobile \\

2021 & N5K$^4$ & $\sim$5,000 dishes
&  &  & \checkmark &  &  & 
&  &  & Video \\

\rowcolor{gray!12}2020 & PFuji-Size & $\sim$615 Fuji apples
&  & \checkmark & \checkmark &  &  & 
& \checkmark & \checkmark & Mobile \\

2019 & KFuji & 967 images, 12,839 apples
&  & \checkmark & \checkmark &  &  & 
& \checkmark &  & IR \\

\midrule
\rowcolor{gray!12}
\textbf{--} & \textbf{Ours$^{1,2,3,4}$} & \textbf{25,284 images, 55 dishes}
& \checkmark & \checkmark & \checkmark & \checkmark &  & 
& \checkmark & \checkmark & Video \\

\bottomrule
\end{tabular}

\vspace{2pt}
\raggedright
\tiny{
$^1$Fkit,\;
$^2$MTF,\;
$^3$V\&F,\;
$^4$N5K datasets are included  with 25,284 annotations within our benchmark.
}
\end{table}

Table \ref{tab:food-datasets} summarizes representative food-digitization datasets. N5k \cite{thames2021nutrition5k} provides controlled turntable captures with high geometric fidelity, while MTF \cite{he2024metafood}, V\&F \cite{steinbrener2023learning}, Fkit \cite{haroon2025vole} and MADIMA23 \cite{abdur2023comparative} enable free-motion, multi-sensor capture with volume measurements, but there are no food segmentation annotations. Other datasets, such as FruitNeRF \cite{meyer2024fruitnerf}, PFuji-Size \cite{gene2021pfuji}, KFuji \cite{gene2019kfuji}, MozzaVID \cite{pieta2024mozzavid}, and AmodalAppleSize \cite{gene2024amodalapplesize_rgb}, focus on handheld, single-image, or CT/RGBD scans with varying support for pose, free-motion, and volumetric ground truth. Despite these valuable resources, most existing datasets are either image-centric (lacking dense temporal  image-level segmentation annotations) or limited in viewpoint diversity to canonical top-down or narrow angular ranges; consequently, they do not fully represent the free-motion, hemispherical capture scenarios common in realistic dietary-recording settings.

\subsection{Related Methodologies}
On the method side, early food segmentation efforts adapted classical semantic- and instance-segmentation architectures (e.g., FPN\cite{Kirillov_2019_CVPR}, DeepLab-family models, and attention-augmented CNNs\cite{huang2019ccnet}) to the food domain, often obtaining strong performance when training and test distributions are closely matched \cite{wu2021large,huang2019ccnet}. More recent transformer-based backbones (e.g., Swin \cite{liu2021swin}, SeTR \cite{zheng2021rethinking}) and high-resolution refinement networks (e.g., BiRefNet) have further improved per-frame mask quality and boundary precision \cite{zheng2021rethinking,liu2021swin,zheng2024bilateral}. These architectures perform well on controlled or top-down images but are susceptible to domain shift when camera poses, lighting conditions, or food presentation differ from those in the training examples.

A distinct and growing line of work leverages foundation and promptable models to improve segmentation flexibility across visual domains. The Segment Anything Model (SAM) \cite{kirillov2023segment} demonstrated that high-quality, category-agnostic mask proposals can be produced from a variety of prompts; adaptations of SAM for food (collectively referred to as FoodSAM-style approaches  \cite{lan2023foodsam}) fuse semantic predictions with SAM proposals to recover fine-grained food masks in single images \cite{kirillov2023segment,lan2023foodsam}. Large multimodal models (LMMs) and food-specialized LMM adaptations (e.g., FoodLMM) \cite{yin2025foodlmm} show potential for jointly reasoning about ingredient semantics and mask prediction, but their segmentation performance on unconstrained, multi-view food videos is still nascent \cite{yin2025foodlmm}.

Video and temporal-consistency methods address limitations of frame-by-frame segmentation by exploiting inter-frame information. Memory-augmented video object segmentation methods, such as XMem \cite{cheng2022xmem} and its variants, store representative frame features and use them to propagate masks reliably across long sequences \cite{bekuzarov2023xmem++}. DEVA \cite{cheng2023tracking} and other decoupled frameworks separate image-level segmentation from class-agnostic temporal propagation, enabling “track-anything” behavior with minimal video-specific training \cite{cheng2023tracking}. In the food domain, the FoodMem \cite{almughrabi2025foodmem} pipeline combined a transformer-based image segmenter (SeTR) \cite{zheng2021rethinking} with a memory tracker (XMem2 \cite{bekuzarov2023xmem++}) to substantially reduce mask flicker and improve completeness in 360° food videos \cite{almughrabi2025foodmem}. These hybrid two-stage strategies are compelling because they compensate for weaknesses in single-frame predictions by enforcing temporal coherence.

While temporal consistency has been explored in the context of tracking through metrics such as identity switches or trajectory continuity, these measures primarily focus on object identity preservation rather than mask-level stability. In contrast, our proposed metrics explicitly target segmentation behavior over time, capturing phenomena such as flickering and gradual degradation that are not adequately reflected by conventional tracking or frame-wise IoU-based evaluations.

 % \pr{We are in the methodology section!:} 
Despite progress, important gaps remain. First, most benchmarks do not jointly allow to evaluate (a) the ability of models to generalize across widely varying camera poses, (b) the effectiveness of promptable or foundation models in continuous video, and (c) the interplay between initial segmentation quality and temporal-aware propagation. Second, computational and memory costs of high-performing pipelines (e.g., SAM-based or memory-augmented methods) have not been systematically reported in food-specific evaluations, yet they are crucial for practical dietary applications on mobile or embedded platforms. Finally, prior datasets typically lack the combination of dense per-frame masks, hemispherical capture coverage, and cross-dataset origin that stresses generalization in realistic dietary scenarios.

Taken together, the literature suggests that robust video-based food segmentation requires: (i) training and evaluation data that reflect free-motion viewpoint variation, (ii) methods that combine strong per-frame segmentation with temporally-aware propagation, and (iii) careful measurement of both accuracy and operational costs. BenchSeg is designed to address these needs by providing multi-scene, multi-view, per-frame mask annotations and by serving as a unified benchmark for comparing static segmentors, promptable or foundation approaches, and memory-enhanced hybrids under identical evaluation criteria.

%% Use \subsection commands to start a subsection.

\section{Proposed Benchmark: BenchSeg with 25,284 Annotations}
\label{sec:methodology}

We present a unified grounded benchmark for food segmentation and temporally coherent mask propagation. Our formulation emphasizes abstraction over dataset-specific metrics—i.e., metrics that are tied to a particular dataset’s labels or evaluation protocol—focusing instead on operations and model structures that are reproducible across datasets and tasks.  
\begin{figure}[htb]
    \centering
    \includegraphics[trim={2.3cm 1.9cm 2.3cm 2cm},clip,width=1.0\linewidth]{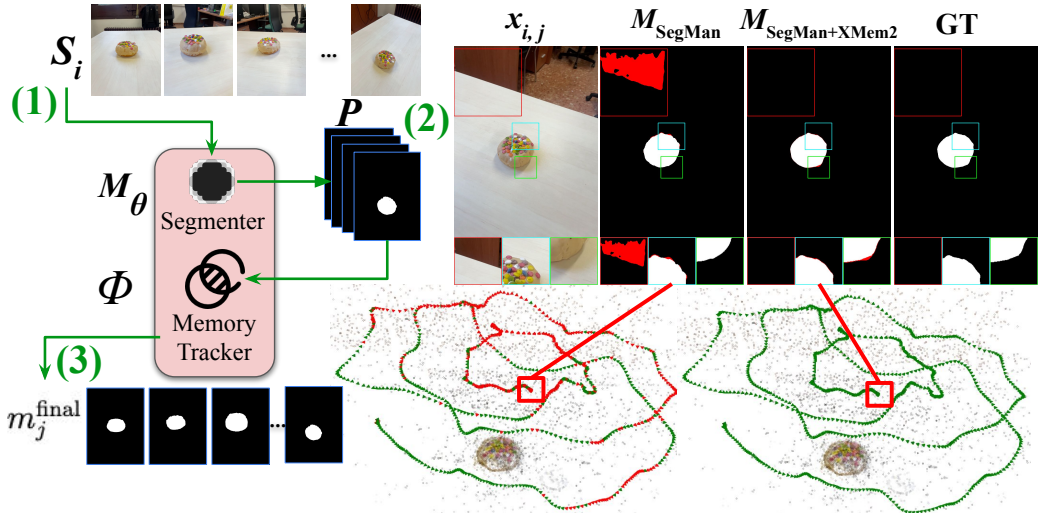}
    \caption{Overview of the proposed three-stage food segmentation methodology: (1) keyframe segmentation generates initial masks, (2) temporal propagation transfers them across non-key frames using stored features, and (3) late fusion refines masks by combining propagated and predictions, enabling a reproducible and temporally coherent food-segmentation process. Camera poses shown in green indicate cases where the matching accuracy threshold $mAP \geq 95\%$ is satisfied; poses in red denote those falling below this threshold.}
    \label{fig:methodology}
\end{figure}

Conceptually, we describe a common \emph{evaluation abstraction} for video segmentation systems as a three-stage process: First, per-frame  food segmentation models generate preliminary predictions for selected keyframes. Second, temporal propagation distributes these masks to all non-key frames, leveraging stored features and temporal correlations (i.e.,  mask propagation). Third, optional late fusion integrates fresh per-frame predictions with propagated masks to correct errors and reduce drift (mask refinement). This design formalizes temporal segmentation as a sequence of well-defined operators acting on images and mask embeddings, providing a general, reproducible methodology for temporally coherent food segmentation, as shown in Fig.~\ref{fig:methodology}. 

Mathematically, let the benchmark dataset be denoted by $\mathcal{D} = \{S_i\}_{i=1}^S$, where each scene denoted by $S_i$ is an ordered sequence of image–mask pairs $S_i = \{(x_{i,j}, m_{i,j})\}_{j=1}^{n_i}$. Here, $x_{i,j} \in \mathbb{R}^{H \times W \times 3}$ is an RGB observation and $m_{i,j} \in \{0,1\}^{H \times W}$ is its corresponding foreground mask. Each scene captures a single food instance, potentially across a hemispherical sweep of camera viewpoints, while preserving the temporal order of frames to enable sequence-level reasoning. Annotations were verified via double-blind review on representative subsets, with disagreements adjudicated by a senior annotator and annotation guidelines refined iteratively.

\subsection{Temporal Stability Metrics}
\label{sec:temporal_metrics}

Let $\mathrm{IoU}_t = \mathrm{IoU}(\hat m_t, m_t)$ denote the frame-wise IoU between the predicted mask $\hat m_t$ and ground truth $m_t$ at time $t$.
We define temporal stability over a sequence of length $T$ using changes in $\mathrm{IoU}_t$:

\paragraph{Continuity}
Given a quality threshold $\gamma$ (we use $\gamma=0.5$), continuity measures the fraction of consecutive frames whose IoU stays above $\gamma$:
\begin{equation}
C_\gamma = \frac{1}{T-1}\sum_{t=2}^{T} \mathbbm{1}\{\mathrm{IoU}_{t-1}\ge \gamma \ \wedge\ \mathrm{IoU}_t \ge \gamma\}.
\end{equation}

\paragraph{Flicker Rate}
Given a drop threshold $\delta$ (we use $\delta=0.2$), flicker measures abrupt quality degradations:
\begin{equation}
FR_\delta = \frac{1}{T-1}\sum_{t=2}^{T} \mathbbm{1}\{\mathrm{IoU}_{t-1} - \mathrm{IoU}_t > \delta\}.
\end{equation}

\paragraph{IoU Drift and Volatility}
We quantify gradual change as the mean absolute IoU difference
\begin{equation}
\Delta \mathrm{IoU} = \frac{1}{T-1}\sum_{t=2}^{T} |\mathrm{IoU}_{t} - \mathrm{IoU}_{t-1}|,
\end{equation}
and volatility as $\sigma(\mathrm{IoU}_1,\dots,\mathrm{IoU}_T)$.

All temporal metrics are first averaged per-scene and then macro-averaged across scenes within each partition.

\subsection{Annotation Protocol and Quality Control}
All segmentation masks within the BenchSeg dataset were meticulously annotated through a polygon-based labeling interface. Annotators were instructed to delineate the entire visible area of each food item, including regions that were partially obstructed, while deliberately excluding non-edible objects such as utensils, packaging, and tableware.

To guarantee the consistency of annotations, comprehensive written guidelines were developed, featuring representative examples of ambiguous cases, including reflections, transparent containers, and overlapping food items. Each annotator underwent a calibration phase prior to engaging in large-scale labeling tasks.

A subset comprising 4,000 frames was independently annotated by two separate annotators. Discrepancies were resolved via an adjudication process involving a third reviewer. The level of inter-annotator agreement was assessed using mAP and IoU metrics, with the corresponding results documented in Table~\ref{tab:quantify_annotation_consistency}.
\begin{table}[htb]
\centering
\caption{Quantifying annotation consistency across datasets reported as mean $\pm$ standard deviation.}
\begin{tabular}{lcc}
\toprule
Dataset & mAP & Recall \\
\midrule
FKit & $0.9642 \pm 0.0064$ & $0.9998 \pm 0.0007$ \\
\rowcolor{gray!12}MTF     & $0.9723 \pm 0.0055$ & $0.9943 \pm 0.0010$ \\
N5K     & $0.9270 \pm 0.0108$ & $0.9997 \pm 0.0008$ \\
\rowcolor{gray!12}V\&F    & $0.9471 \pm 0.0172$ & $0.9986 \pm 0.0030$ \\
\bottomrule
\end{tabular}
\label{tab:quantify_annotation_consistency}
\end{table}
\subsubsection*{Annotators and tooling}
Annotations were produced by 3 annotators (trained undergraduates) using the LabelMe \cite{russell2008labelme} polygon interface. Annotators completed a calibration session (240 hours) and used a written guideline (available on the project page) describing foreground definitions, handling of reflections/transparent containers, and rules for occlusions.

\subsubsection*{Double annotation and adjudication}
A subset of 4,000 frames ($\approx$ 16\% of the corpus) was independently labeled by two annotators; disagreements were resolved by a senior annotator. Inter-annotator agreement was measured using per-image mAP and IoU; Table~\ref{tab:quantify_annotation_consistency} reports mean ± std per source dataset. For reproducibility, we will publish: (i) the guideline PDF, (ii) anonymized annotator IDs and time per frame statistics, and (iii) the adjudicated vs. original masks for a random 500-frame sample.

\section{Experimental Results}
\label{sec:experimental_results}
We train 20 state-of-the-art models on FoodSeg103 and evaluate them on BenchSeg, specifying essential details for the subset of the BenchSeg dataset. The evaluation reported here quantifies segmentation quality and computational cost for all tested methods on the BenchSeg partitions. We report per-partition average precision $\mathrm{mAP}$ and Recall together with model size, runtime (speed), and peak memory usage. Table~\ref{tab:results} contains the complete results for the N5k, V\&F, MTF and FKit partitions.

\subsection{BenchSeg: Dataset Analysis}
BenchSeg is designed to evaluate segmentation models. The BenchSeg benchmark integrates four heterogeneous datasets—FKit, MTF, N5k, and V\&F—each exhibiting distinct structural and statistical properties, such as differences in image resolution, object size distribution, number of classes, and background variability. This diversity is deliberate: it enables a controlled examination of segmentation model generalization under varying scene sizes, visual variability, and distributional imbalance. Below, we analyze the datasets both individually and as a unified corpus, with emphasis on image volume, scene granularity, and intra-dataset variability.

\paragraph{\textbf{FKit}} is the largest component of the benchmark, comprising 20,606 images across 21 scenes, with an average of 981.24 ± 142.69 images per scene (ranging from 715 to 1,209 images per scene, illustrating the variability in scene representation). Its large volume and high per-scene density make it particularly valuable for evaluating temporal consistency and robustness to fine-grained appearance variations. The scene distribution exhibits a strong skewness, ranging from chocolate\_panettone (1209 images) to yellow\_cane (715 images). This long-tailed distribution introduces intentional challenges for models trained on balanced datasets, highlighting resilience to class-frequency imbalance.

\paragraph{\textbf{MTF}} contributes 1,749 images across 13 scenes, with a mean of 134.54 ± 82.65 images per scene. Unlike FKit, MTF exhibits a bimodal distribution: most scenes contain approximately 200 images, while a subset contains only 30 images. This structured imbalance allows for assessing how segmentation accuracy degrades in low-data scenes, making it an ideal testbed for studying underrepresented viewpoints and sparse trajectories.

\paragraph{\textbf{N5k}} is considerably more uniform, providing 621 images across 10 scenes, with an average of 62.10 ± 1.97 images per scene. Its extremely low variance (stdv. = 1.97) indicates nearly equal per-scene sampling, making N5k particularly suitable for controlled generalization studies where scene size is not a confounding factor. Because N5k dishes exhibit substantial visual complexity, the uniform sampling ensures consistent difficulty across scenes.

\paragraph{\textbf{V\&F}} offers 2,308 images across 11 scenes, with a mean of 209.82 ± 18.76 images per scene. This moderate-scale dataset strikes a balance between N5k’s uniformity and FoodKit’s scale, presenting steady variability across produce-centric scenes. The controlled variation in lighting and object geometry makes V\&F a strong foundation for analyzing performance on natural, non-prepared food items.

\begin{figure}[htb]
    \centering
    \setlength{\tabcolsep}{1pt}
    \begin{tabular}{cccc}
        \includegraphics[trim={1cm 0 1cm 0},clip,width=0.25\textwidth]{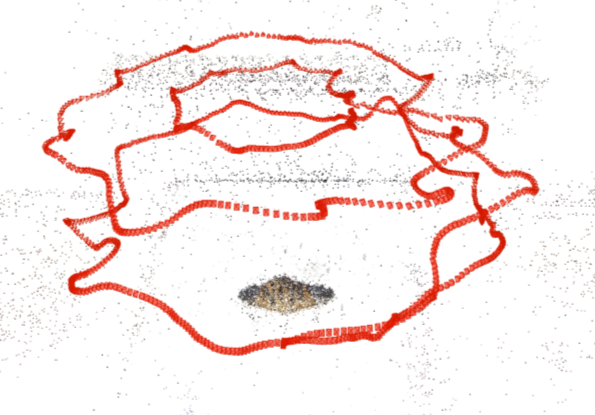}
        &
        \includegraphics[trim={2cm 0 2cm 0},clip,width=0.25\textwidth]{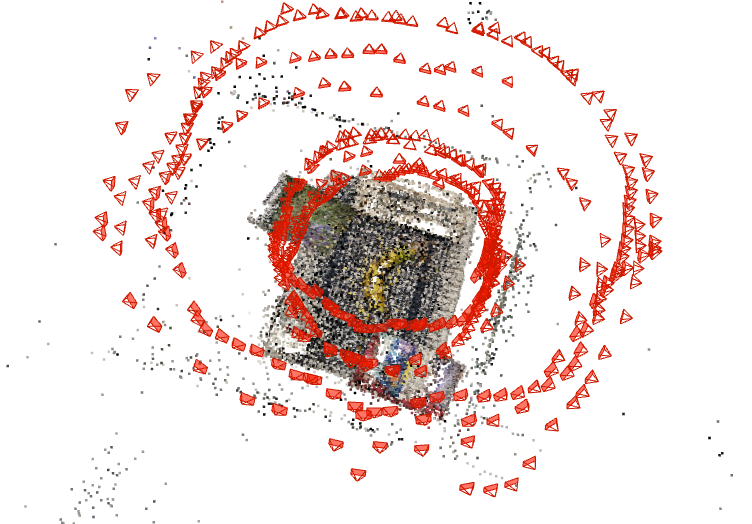} &
        \includegraphics[trim={3cm 3cm 3cm 0},clip,width=0.25\textwidth]{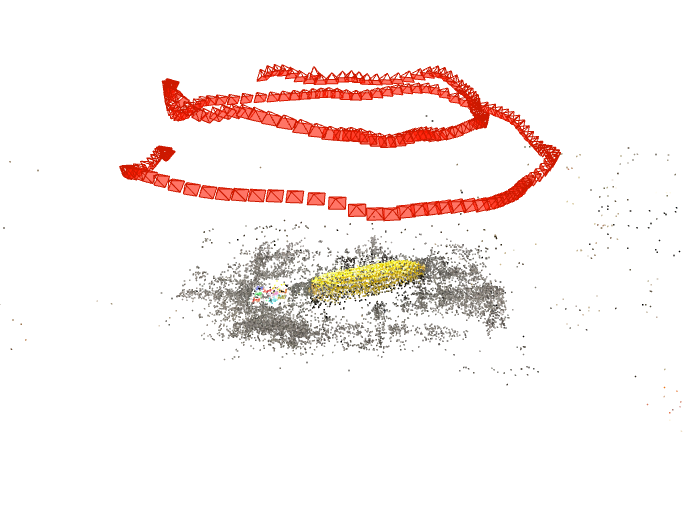}   
        &
       \includegraphics[trim={1cm 3cm 1cm 0},clip,width=0.25\textwidth]{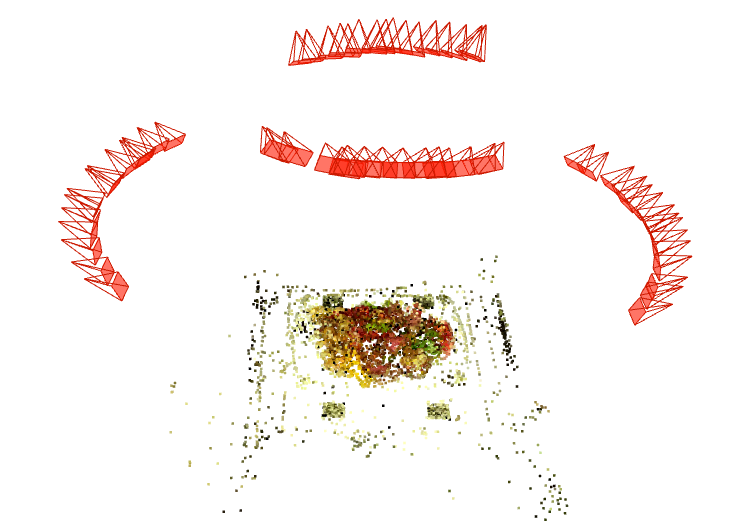}
    \end{tabular}
    \caption{For illustration, camera locations and orientations were estimated by Colmap for various bounded and unbounded scenes from the V\&F, N5k, FKit, and MTF datasets. The first scene on the left is derived from the FoodKit dataset; the second scene is taken from the Vegetables and Fruits dataset; the third scene originates from the MTF dataset; and the final scene is obtained from the N5k dataset.}
    \label{fig:cameralocations}
\end{figure}

\begin{figure}[htb]
    \centering
    \setlength{\tabcolsep}{1pt}
    \begin{tabular}{ccc}
        \includegraphics[width=0.33\textwidth]{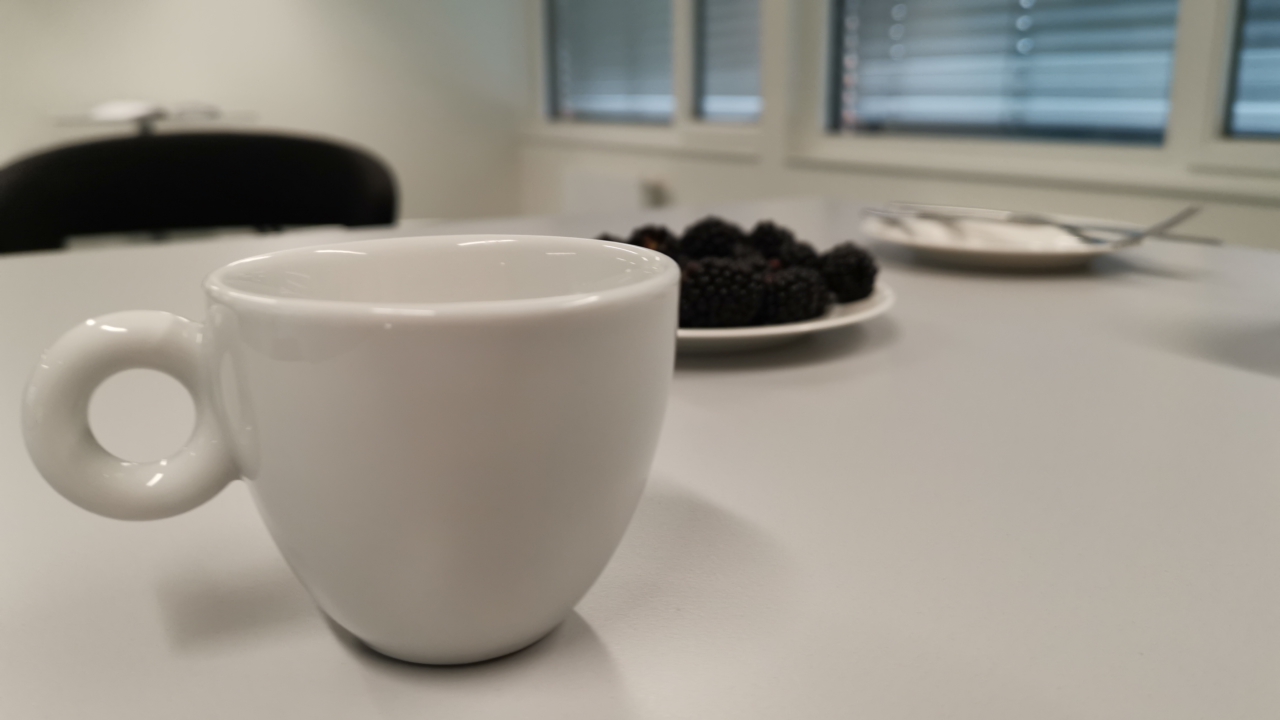}
        &
        \includegraphics[width=0.33\textwidth]{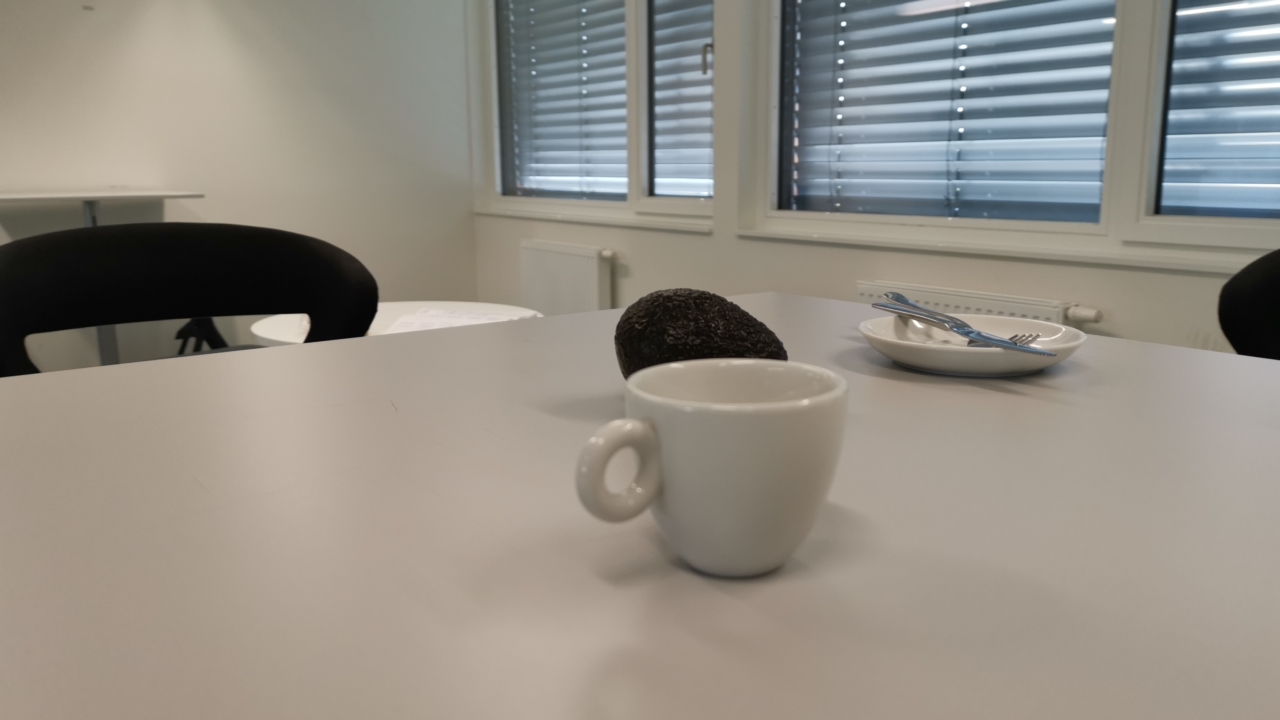}
        &
        \includegraphics[width=0.33\textwidth]{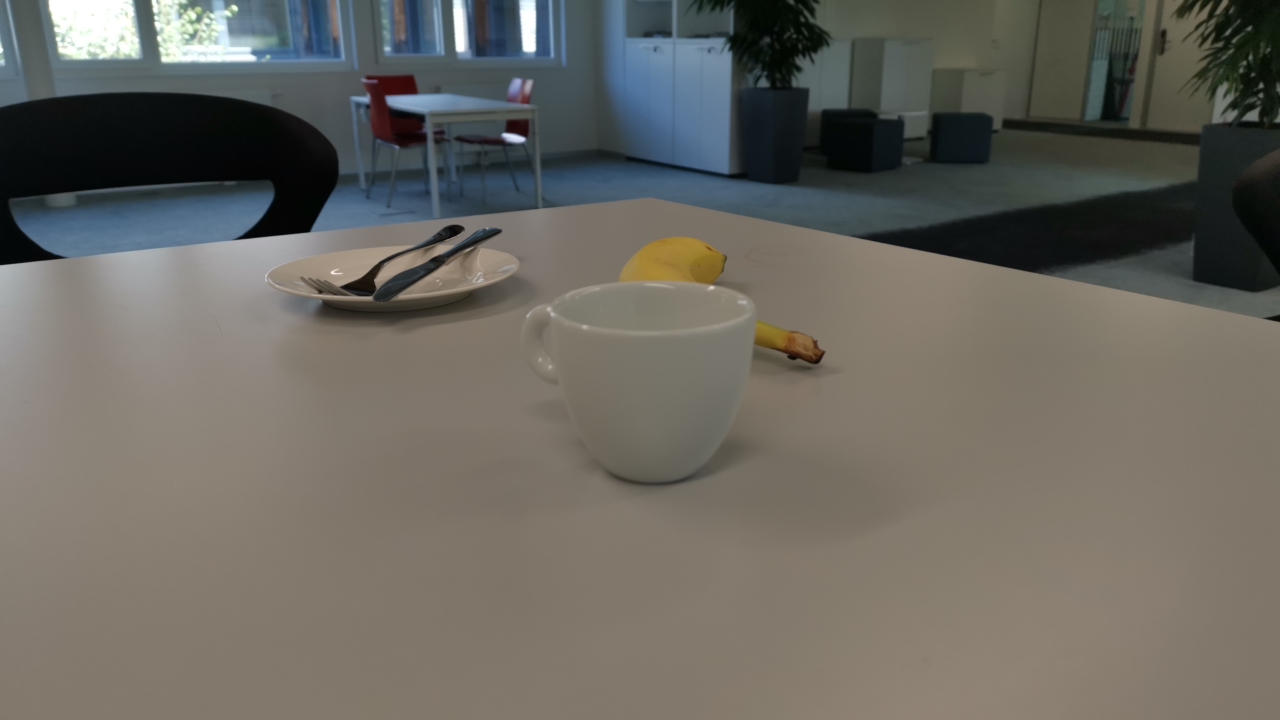}
        \\
        \includegraphics[width=0.33\textwidth]{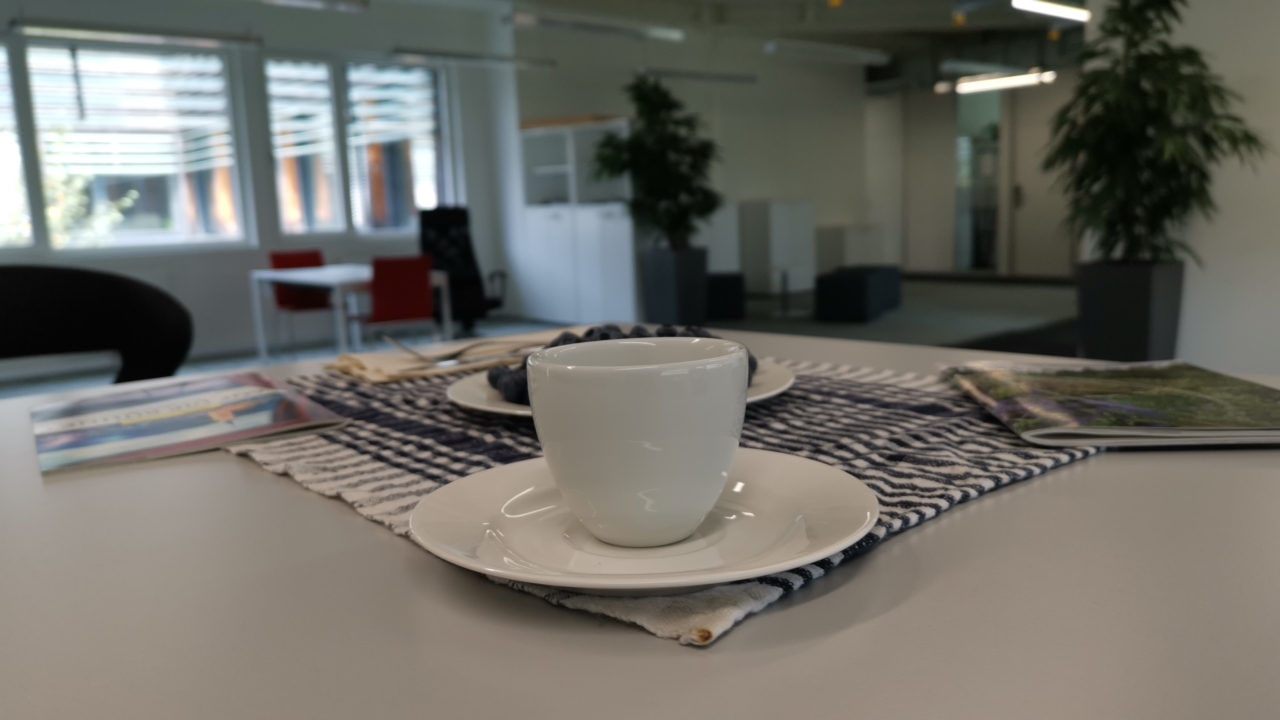} 
        & 
        \includegraphics[width=0.33\textwidth]{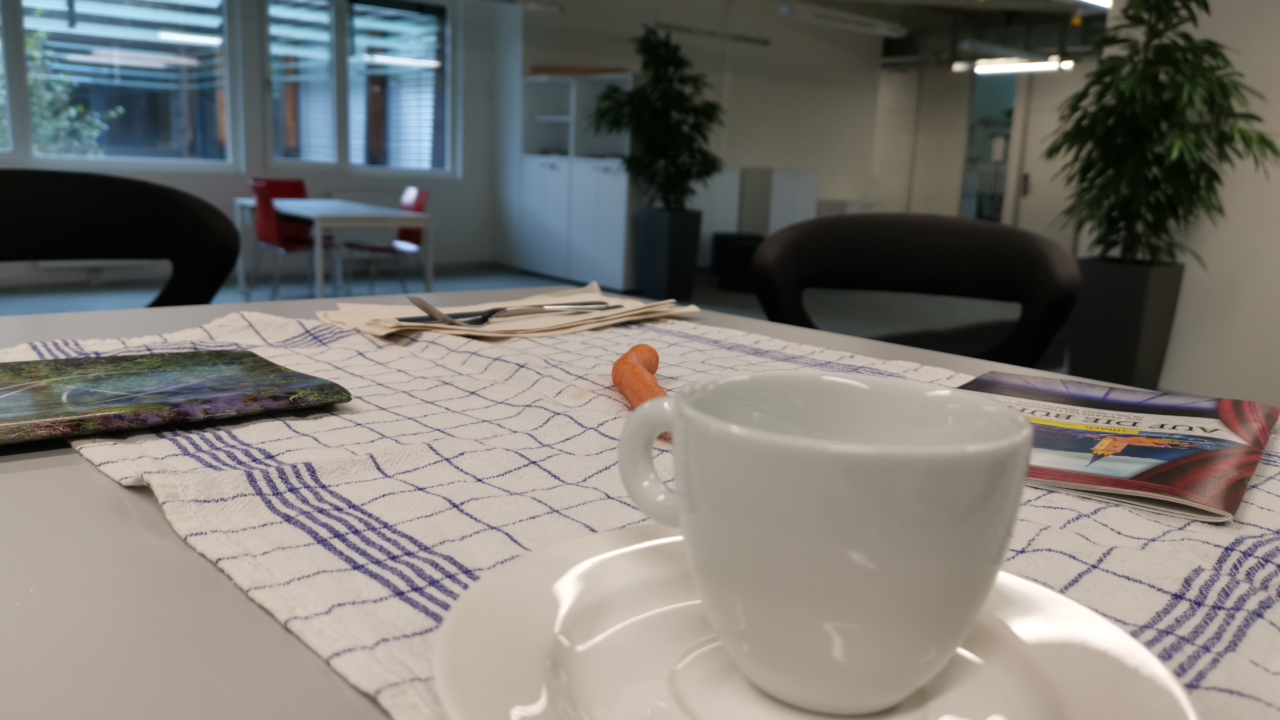} 
        & 
        \includegraphics[width=0.33\textwidth]{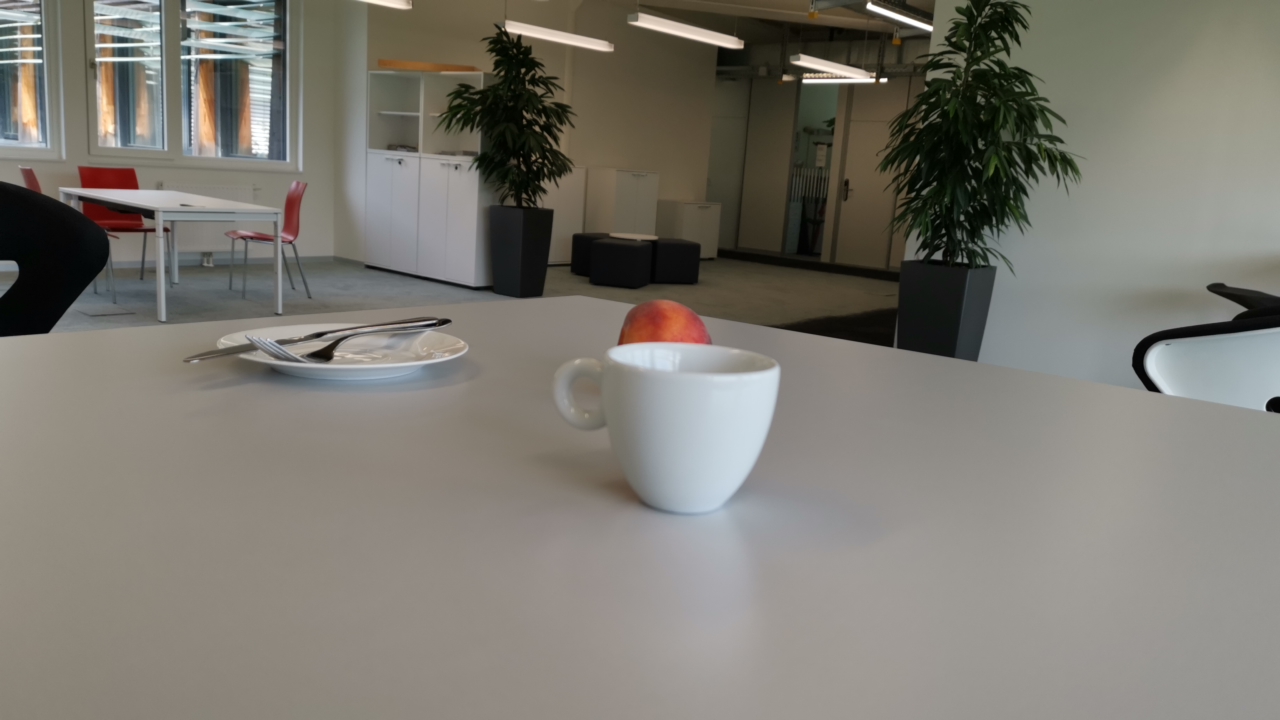} 
    \end{tabular}
    \caption{Examples of the occlusion issues present in the V\&F dataset are highlighted. The figure shows that an obstacle can obscure the food object depending on the camera position.}
    \label{fig:occlusion_samples}
\end{figure}

\begin{figure}[htb]
    \centering
    \setlength{\tabcolsep}{1pt}
    \begin{tabular}{ccc}
        \includegraphics[width=0.33\textwidth]{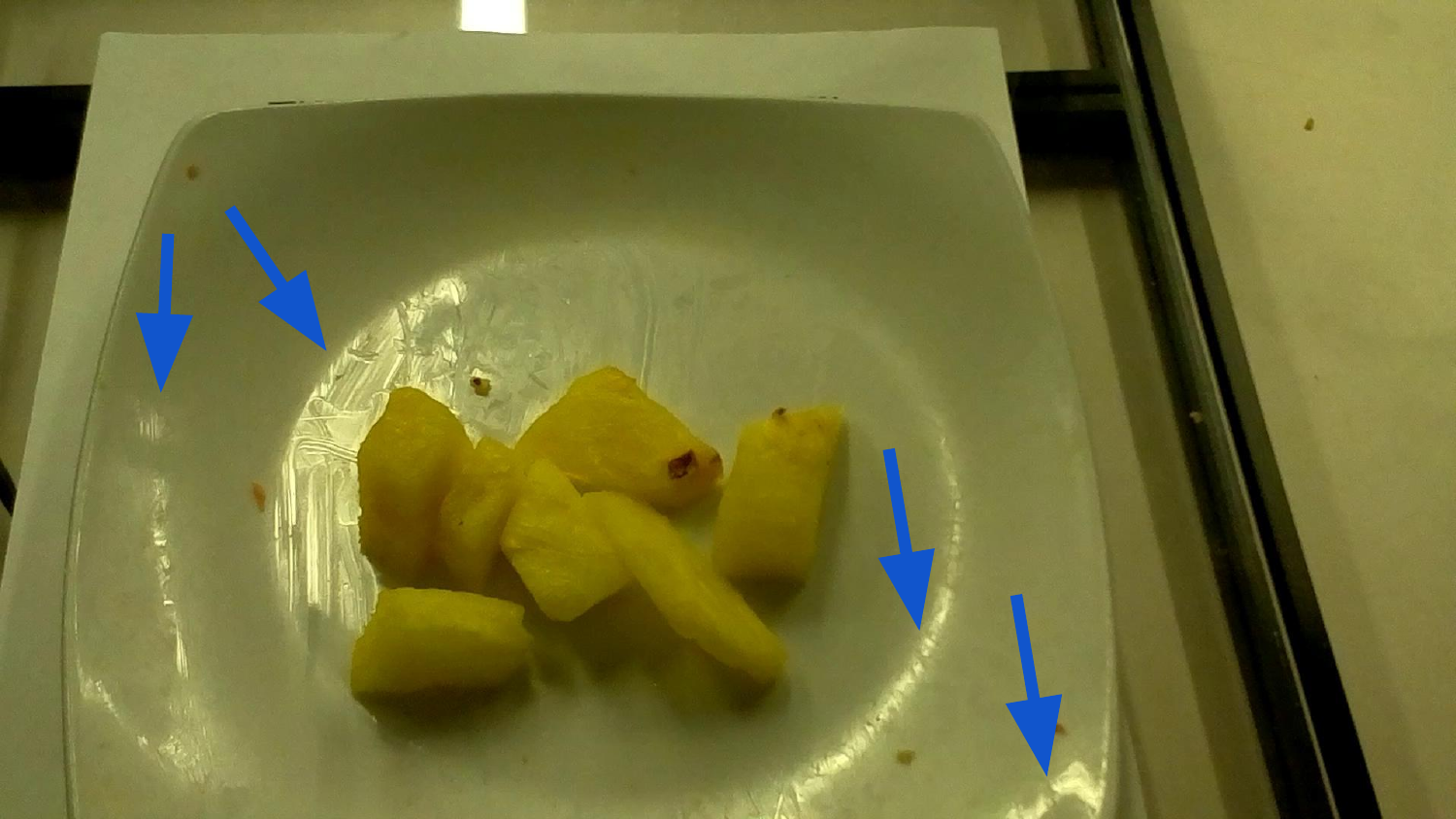}
        &
        \includegraphics[width=0.33\textwidth]{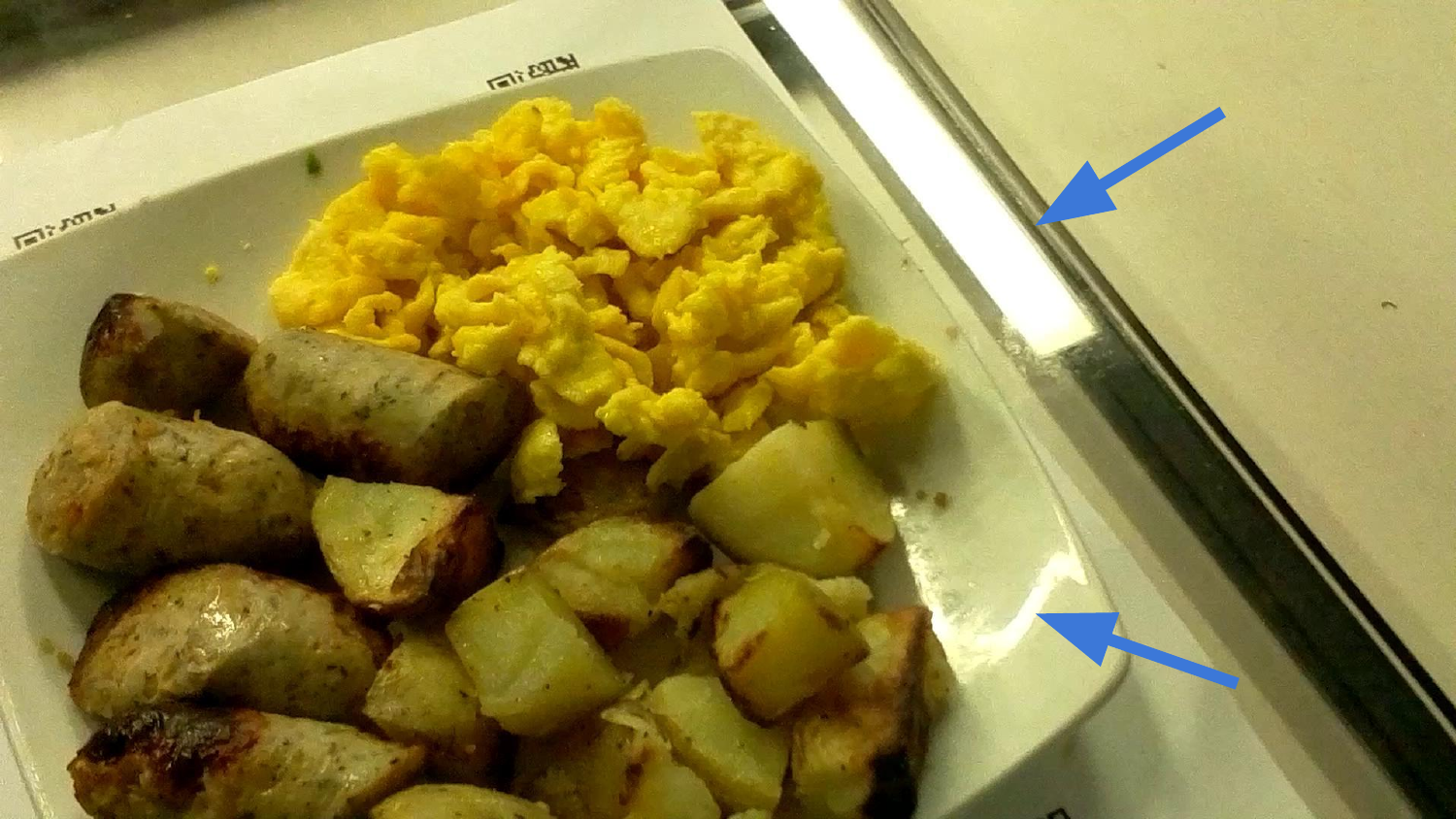}
        &
        \includegraphics[width=0.33\textwidth]{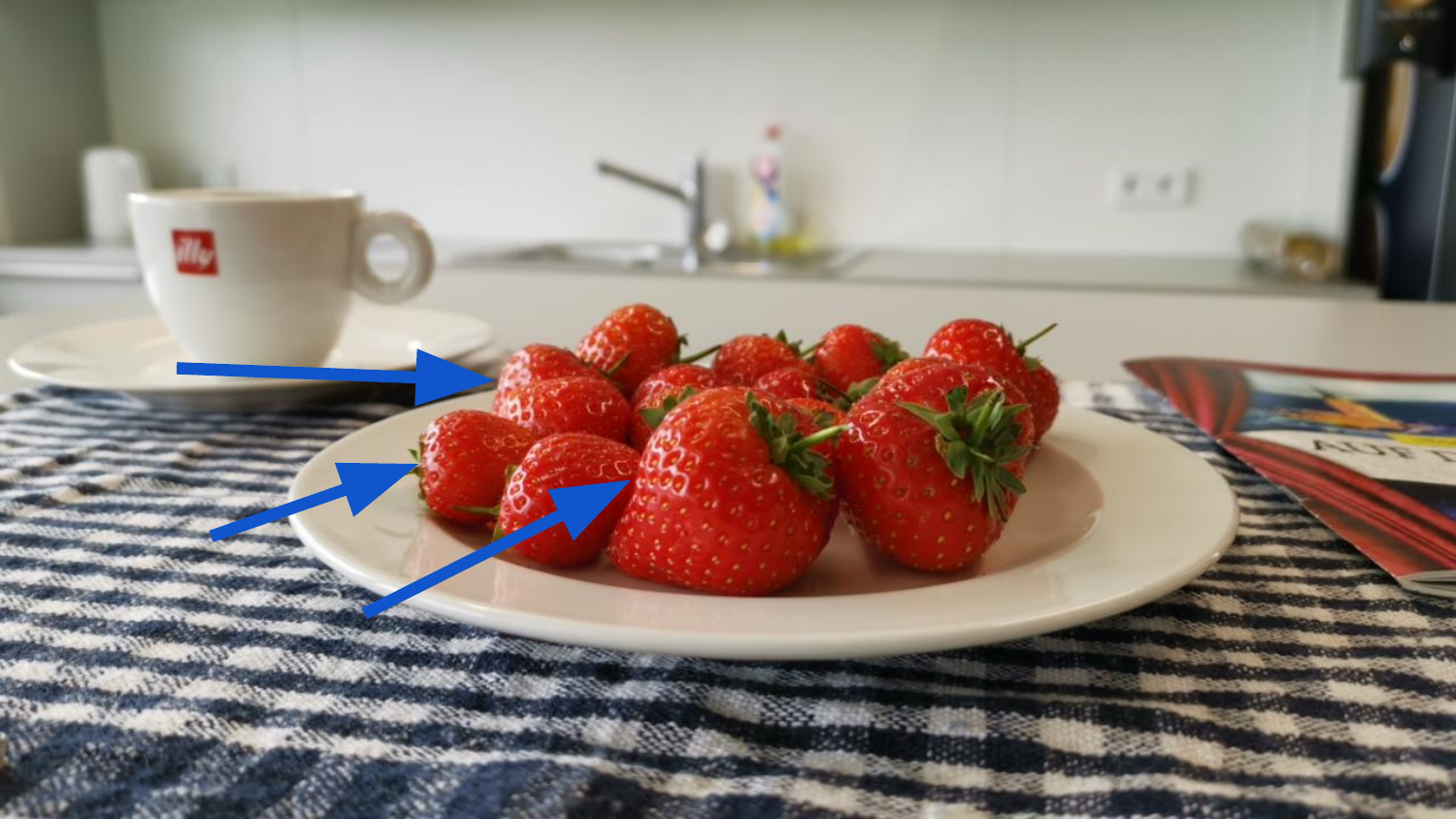}
        \\
        \includegraphics[width=0.33\textwidth]{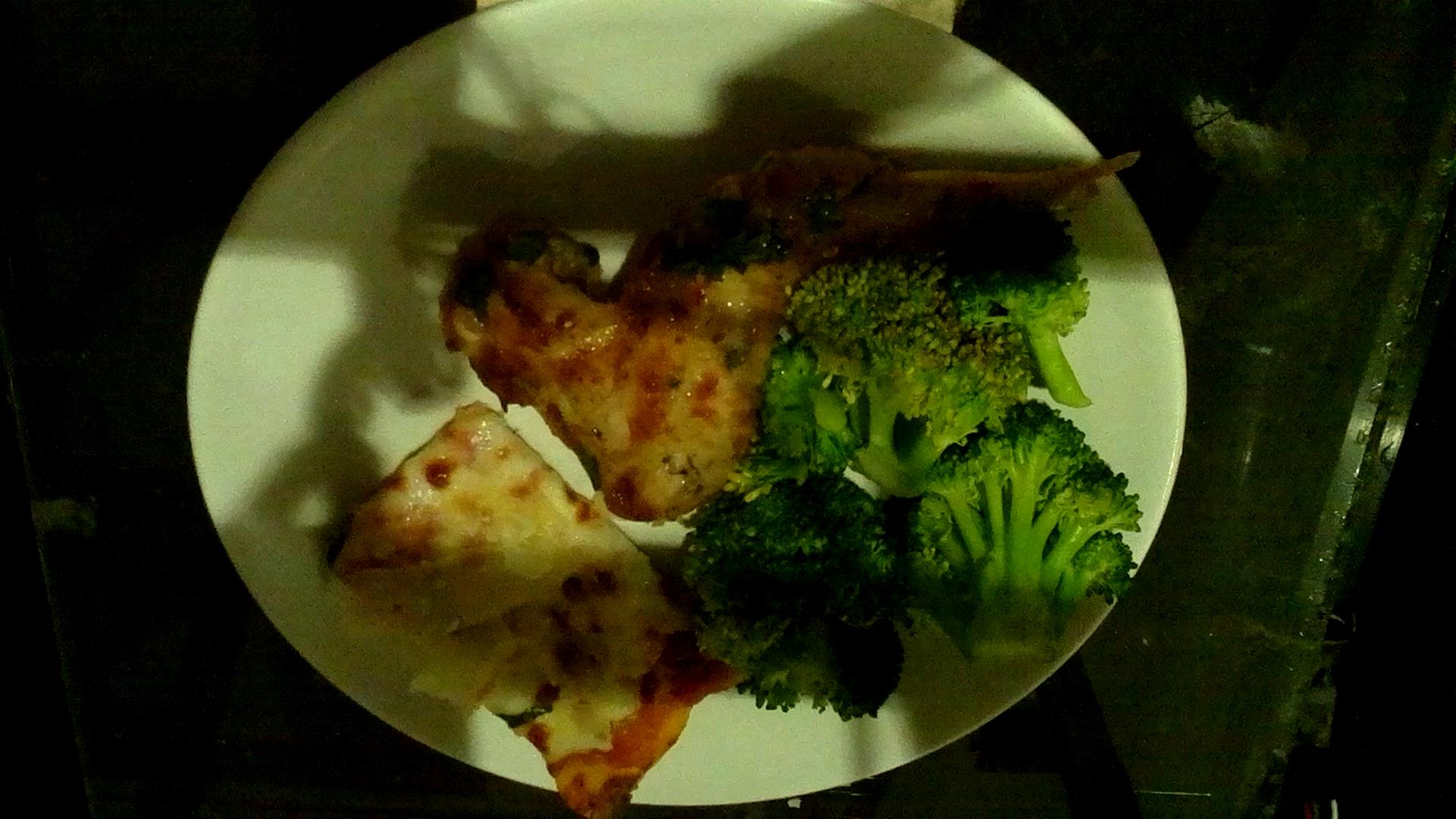} 
        & 
        \includegraphics[width=0.33\textwidth]{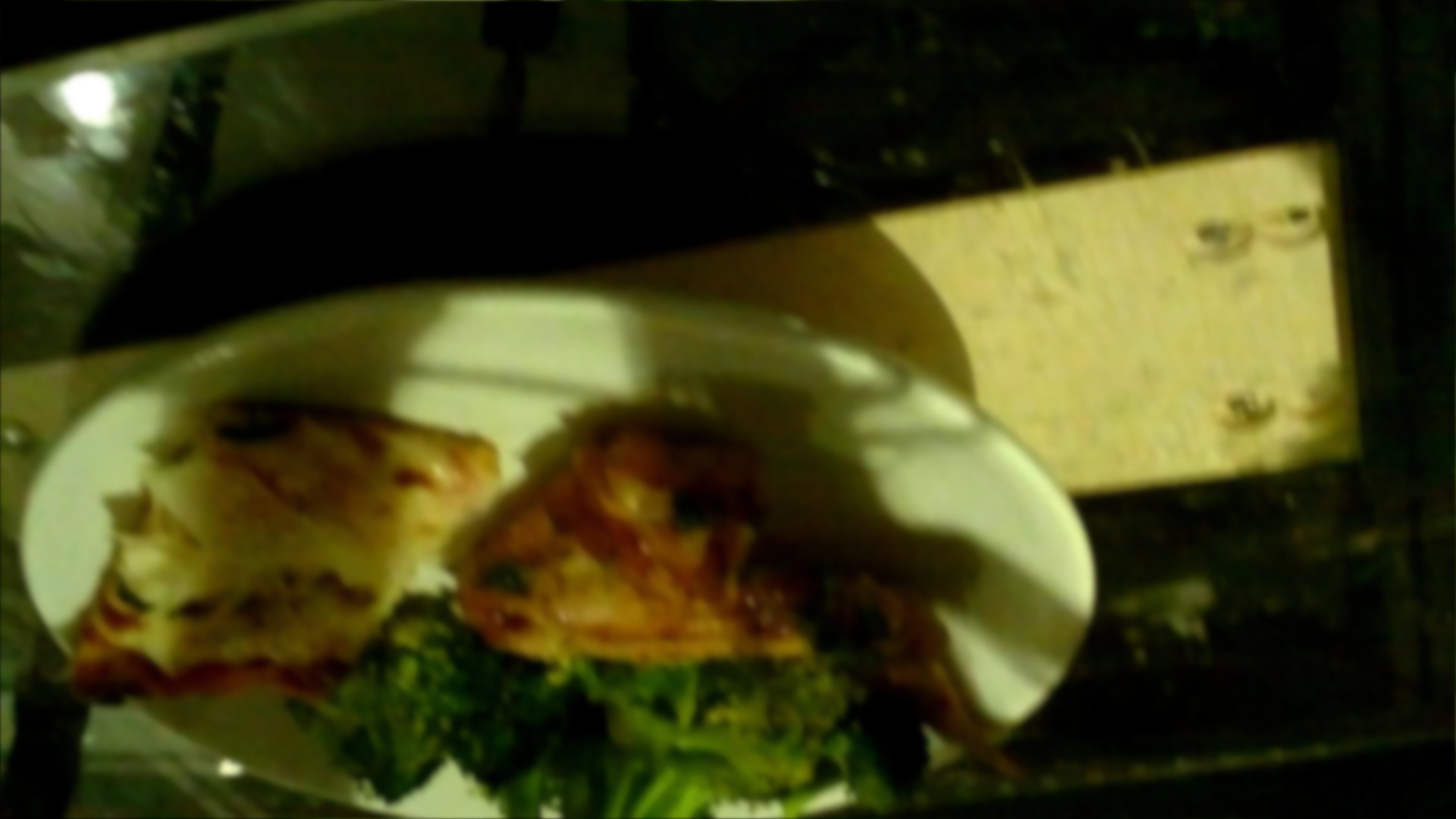} 
        & 
        \includegraphics[width=0.33\textwidth]{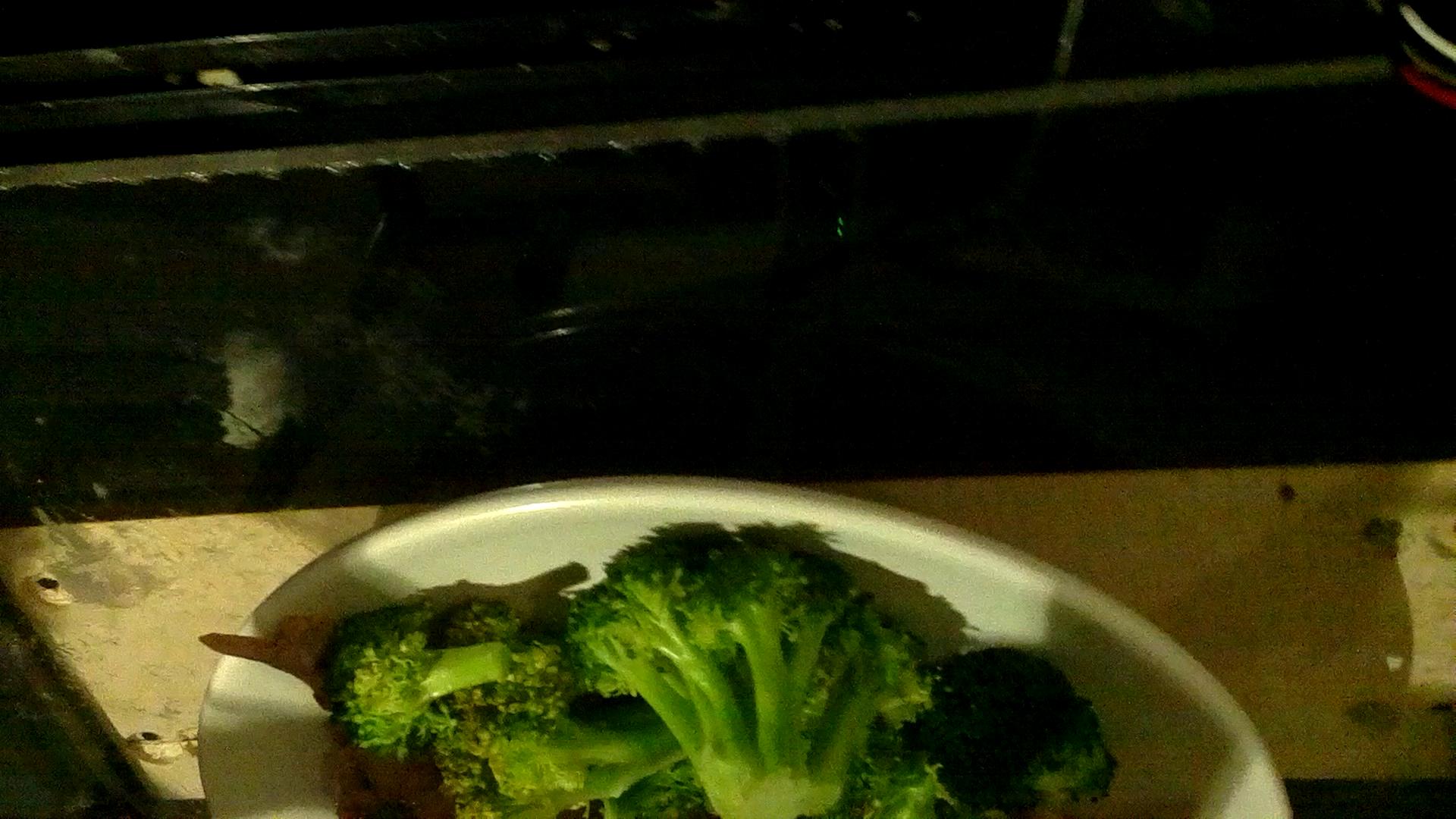} 
        \\ 
        \includegraphics[width=0.33\textwidth]{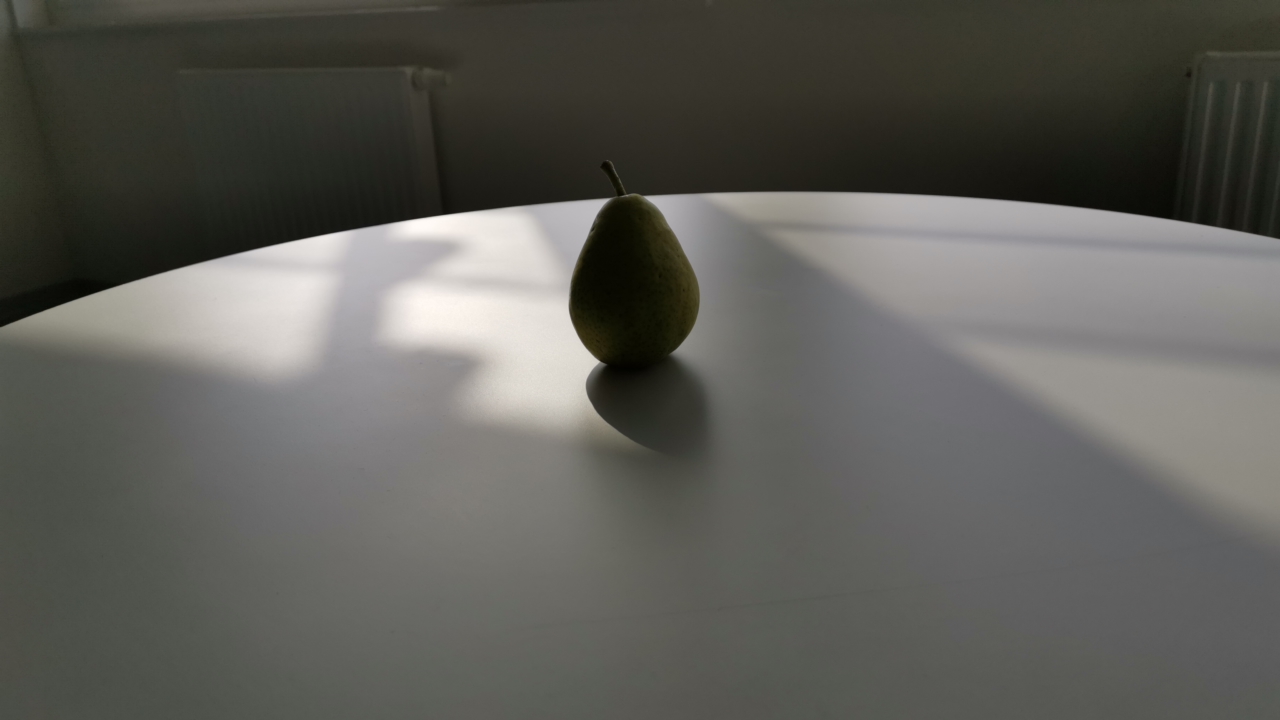} 
        & 
        \includegraphics[width=0.33\textwidth]{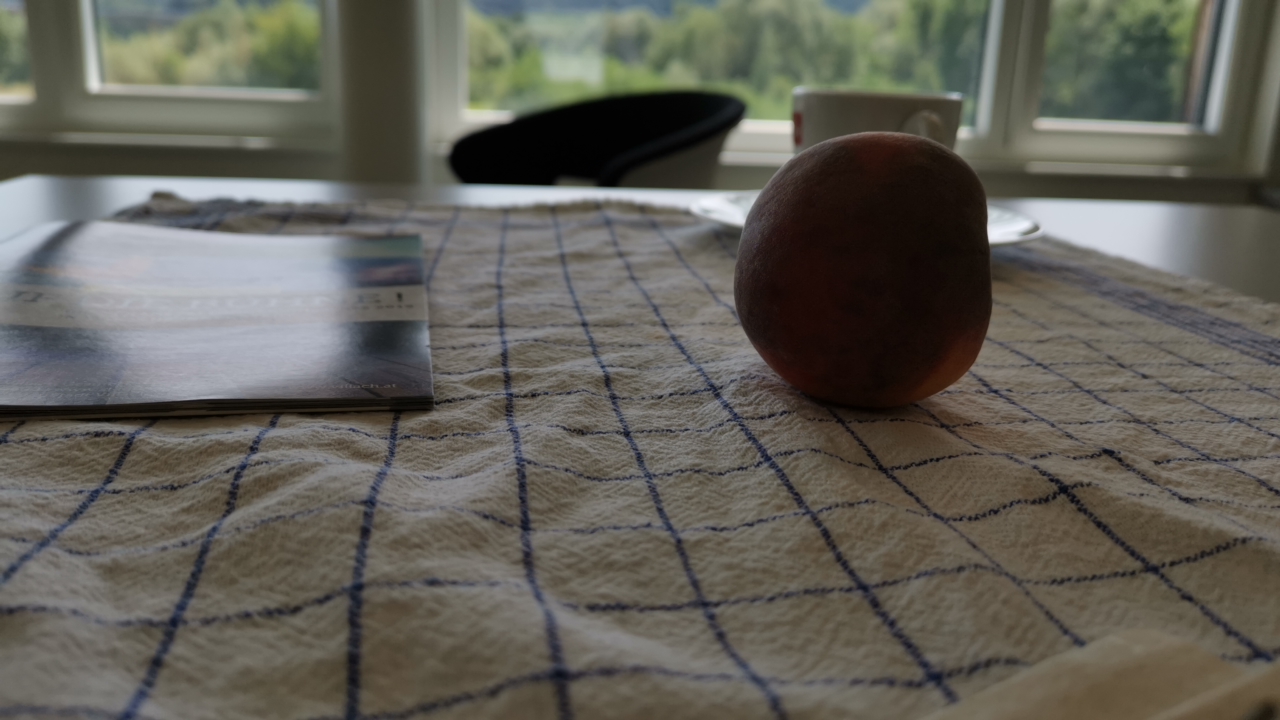} 
        & 
        \includegraphics[width=0.33\textwidth]{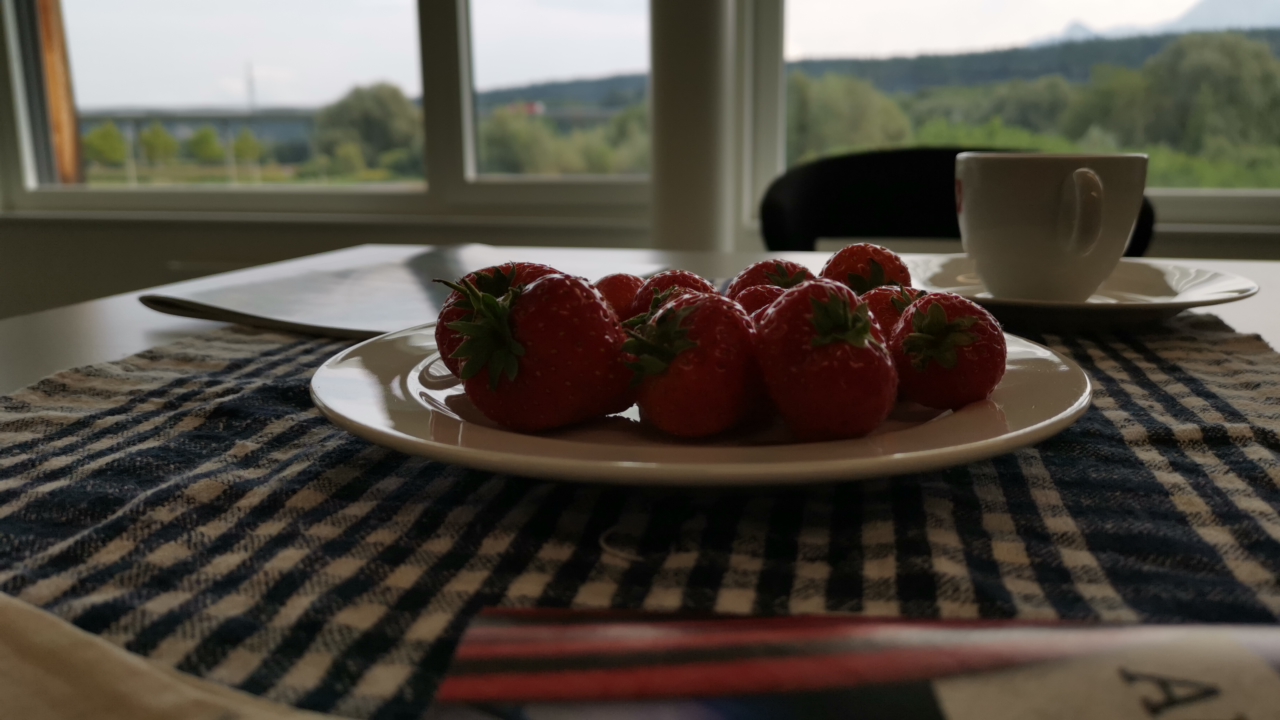} 
    \end{tabular}
    \caption{Examples of addressing the lighting challenges present in the N5k and V\&F datasets. By utilizing both natural and artificial lighting, we emphasize the reflective surfaces. Additionally, we illustrate the low-light conditions that occur in some of these scenarios.}
    \label{fig:lighting_samples}
\end{figure}

\begin{figure}[htb]
    \centering
    \setlength{\tabcolsep}{1pt}
    \begin{tabular}{ccc}
        \includegraphics[width=0.33\textwidth]{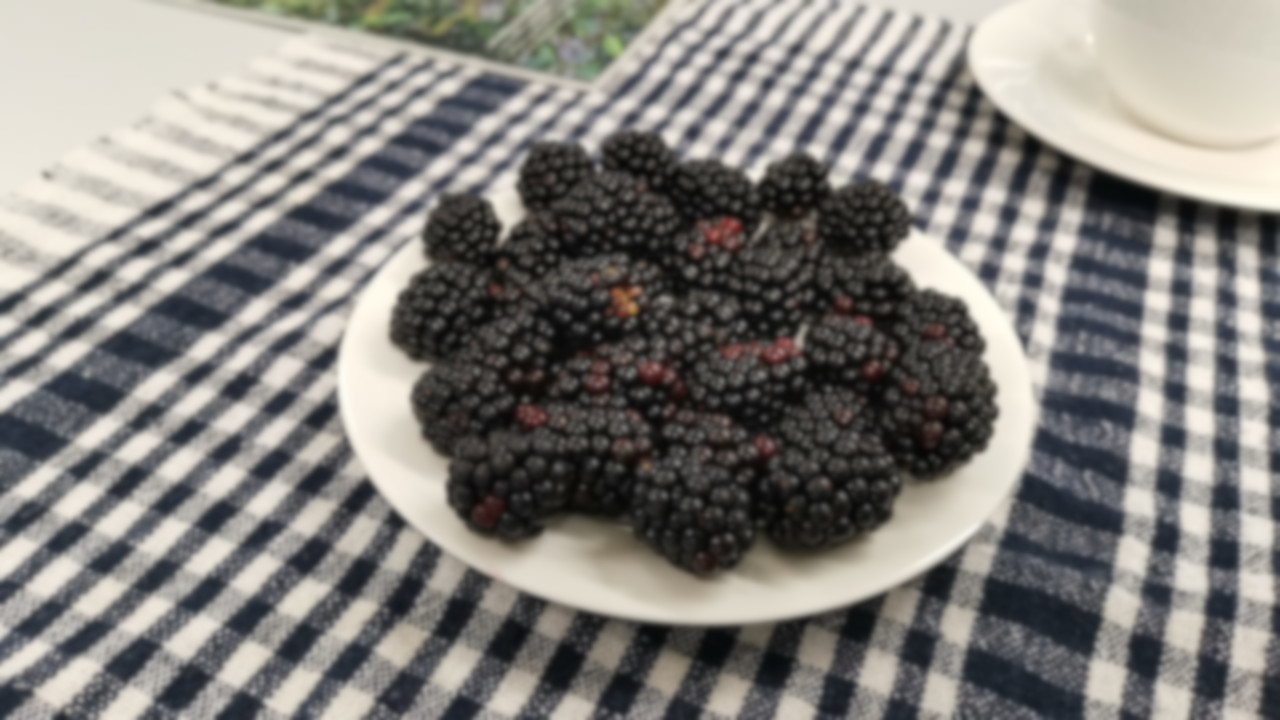}
        &
        \includegraphics[width=0.33\textwidth]{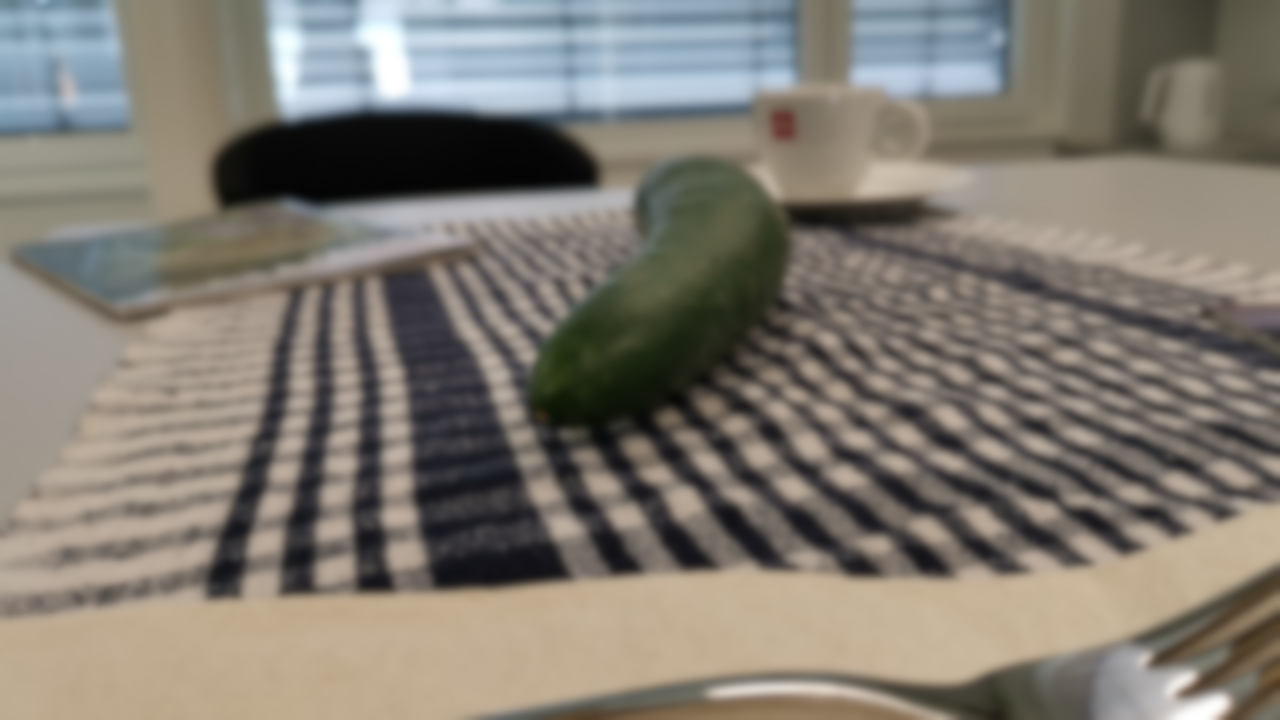}
        &
        \includegraphics[width=0.33\textwidth]{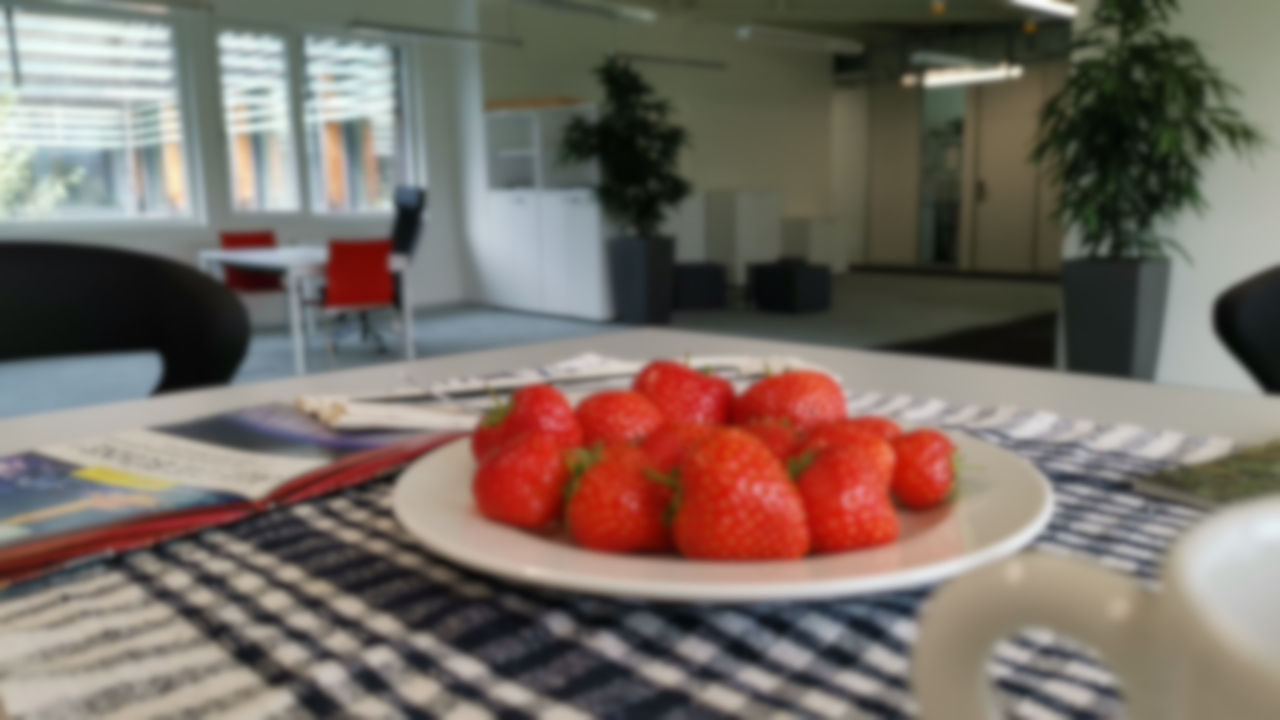}
    \end{tabular}
    \caption{Examples of tackling the challenges of motion blur found in the Vegetables and Fruits datasets. This problem frequently occurs with cameras in free motion. The samples may appear sharper because the images have been resized to fit the article page.}
    \label{fig:blur_samples}
\end{figure}

Our dataset challenges effectively handling the key challenges present in real-world datasets. The V\&F and MTF datasets introduce challenging unbounded scenarios with unrestricted camera motion, varying viewpoints, reflections, lighting fluctuations, occlusions, and motion blur—conditions illustrated in Fig.~\ref{fig:cameralocations}, Fig.~\ref{fig:occlusion_samples}, Fig.~\ref{fig:lighting_samples},  and
Fig.~\ref{fig:blur_samples}. In contrast, the bounded and controlled N5k dataset presents its own challenges, such as diverse lighting setups and multiple camera angles. Across these varied conditions, surprisingly, some methods remain robust and reliable, demonstrating strong adaptability to both uncontrolled and controlled environments.

% To quantify annotation consistency, we double-annotated a random subset of frames per source and computed per-frame mAP between annotators; inter-annotator analysis and the exact double-annotation split are reported in Table~\ref {tab:quantify_annotation_consistency}. Any substantial disagreement cases were adjudicated by a senior annotator, and the guideline was iteratively refined.

Overall, the combined BenchSeg benchmark contains 25,284 images across 55 scenes, forming one of the most structurally diverse food-segmentation testbeds to date. BenchSeg spans more than an order of magnitude in per-scene image count—from as few as 30 frames to over 1,200—capturing a broad spectrum of real-world acquisition conditions. This intentional heterogeneity provides a rigorous environment for evaluating cross-dataset generalization, model brittleness to distribution shifts, and robustness in free-motion video sequences.

\subsection{Data Availability and Licensing}
BenchSeg is constructed by curating and re-annotating frames from multiple publicly available datasets, each governed by its own license. We do not claim ownership of the original images. The licensing terms governing the original datasets used in this work are summarized in Table~\ref{tab:dataset_licenses}.

\begin{table}[htb]
\centering
\caption{Licensing terms of the source datasets used to construct BenchSeg.}
\label{tab:dataset_licenses}
\begin{tabular}{llcc}
\toprule
\textbf{Dataset} & \textbf{License} & \textbf{Redistribution} & \textbf{Commercial Use} \\
\midrule
N5k \cite{thames2021nutrition5k} & CC BY 4.0 & \checkmark & \checkmark \\
\rowcolor{gray!12}V\&F \cite{steinbrener2023learning} & CC BY 4.0 & \checkmark & \checkmark \\
FKit \cite{haroon2025vole} & CC BY 4.0 & \checkmark & \checkmark \\
\rowcolor{gray!12}MTF \cite{chen2024metafood3d} & CC BY-NC 4.0 & \checkmark &  \\
\bottomrule
\end{tabular}
\end{table}

\paragraph{Release contents}
We will release: (i) all BenchSeg binary masks in a unified format (PNG + JSON metadata), (ii) per-frame identifiers linking each annotation to its source dataset and original sequence, (iii) scripts to download/construct the corresponding RGB frames from official sources when redistribution is restricted, (iv) train/test split files, and (v) evaluation code for all spatial and temporal metrics.

\paragraph{License compliance}
For sources with non-commercial clauses (e.g., CC BY-NC), we will distribute only derived annotations and retrieval scripts, and the resulting subset remains subject to the original non-commercial terms.

\subsection{FoodSeg103: Dataset Analysis}
The \textit{FoodSeg103}\cite{wu2021large} dataset consists of 103 ingredient categories organized into 15 supercategories. We randomly partition the dataset into 70\% for training (4,983 images) and 30\% for testing. Summary statistics for these splits are provided in Table~\ref{tab:statistics}. All experiments are conducted on this dataset for in-domain training and evaluation. The total number of ingredients within each supercategory is reported in Table~\ref{table:foodseg_super_classes}.
% \begin{table}[htb]
%     \centering
%     \caption{Training and testing set statistics for FoodSeg103.}
%     \begin{tabular}{c|ccc|ccc}
%         \hline
%          & \multicolumn{3}{c|}{\#Images} & \multicolumn{3}{c}{\#Ingredients} \\
%          Dataset & Train & Test & Total & Train & Test & Total  \\
%          \hline
%          FoodSeg103 & 4,983 & 2,135 & 7,118 & 29,530 & 12,567 & 42,097 \\
%          \hline
%     \end{tabular}
%     \label{tab:statistics}
% \end{table}

\begin{table}[t]
\centering
\small
\setlength{\tabcolsep}{5pt}
\caption{Training and testing set statistics for FoodSeg103. Percentages are reported with respect to the total split.}
\label{tab:statistics}

\begin{tabular}{l
                c c c
                c c c}
\toprule
 & \multicolumn{3}{c}{\textbf{\#Images}}
 & \multicolumn{3}{c}{\textbf{\#Ingredients}} \\
\cmidrule(lr){2-4} \cmidrule(lr){5-7}

\textbf{Dataset}
 & \multicolumn{1}{c}{Train}
 & \multicolumn{1}{c}{Test}
 & \multicolumn{1}{c}{\textbf{Total}}
 & \multicolumn{1}{c}{Train}
 & \multicolumn{1}{c}{Test}
 & \multicolumn{1}{c}{\textbf{Total}} \\

\midrule
\rowcolor{gray!12}
FoodSeg103
 & 4,983 & 2,135 & \textbf{7,118}
 & 29,530 & 12,567 & \textbf{42,097} \\

\addlinespace[3pt]
 & \multicolumn{1}{c}{\footnotesize(70.0\%)}
 & \multicolumn{1}{c}{\footnotesize(30.0\%)}
 & {}
 & \multicolumn{1}{c}{\footnotesize(70.2\%)}
 & \multicolumn{1}{c}{\footnotesize(29.8\%)}
 & {} \\

\bottomrule
\end{tabular}
\end{table}

\begin{table}[t]
\centering
\small
\setlength{\tabcolsep}{6pt}
\caption{Distribution of ingredient counts across superclasses in FoodSeg103. Percentages are reported with respect to the total number of ingredients. Methods are sorted in ascending order according to count column.}
\label{table:foodseg_super_classes}

\begin{tabular}{l c c l c c}
\toprule
\textbf{Superclass} & {\raggedleft\textbf{Count$\blacktriangledown$}} & {\textbf{\%}}
& \textbf{Superclass} & {\textbf{Count$\blacktriangledown$}} & {\textbf{\%}} \\
\midrule

Vegetable & 15,719 & 37.4
& Main & 5,634 & 13.4 \\

\rowcolor{gray!12}Fruit & 6,007 & 14.3
& Meat & 4,956 & 11.8 \\

Dessert & 3,913 & 9.3
& Condiment & 1,543 & 3.7 \\

\rowcolor{gray!12}Nut & 912 & 2.2
& Seafood & 920 & 2.2 \\

Beverage & 844 & 2.0
& Fungus & 592 & 1.4 \\

\rowcolor{gray!12}Egg & 424 & 1.0
& Others & 341 & 0.8 \\

Soy & 148 & 0.4
& Soup & 121 & 0.3 \\

\rowcolor{gray!12}Salad & 23 & 0.1
& - & - & - \\
\bottomrule
\end{tabular}
\end{table}

\subsection{Implementation Settings}
\label{sec:implementation_settings}
This section details the datasets, segmentation architectures, and training hyperparameters used in our experiments.
We first describe the \hyperref[par:training_protocol]{Training Protocol}, then we describe \hyperref[par:dataset]{\textit{Dataset}}, where we outline the in-domain and out-of-domain benchmarks, data splits, and pre-training resources used for our models.
We then introduce the \hyperref[par:segmenter]{\textit{Segmenter}} and \hyperref[par:relem]{\textit{ReLeM}} configurations, specifying the backbone architectures and initialization strategies.
Finally, we summarize the optimization and regularization choices in \hyperref[par:learn-seg]{\textit{Learning Parameters for Segmenter}}, \hyperref[par:learn-relem]{\textit{ReLeM}}, and \hyperref[par:learn-yolo]{\textit{for YOLO}}, including input resolutions, data augmentation pipelines, training schedules, and hardware settings.

\paragraph{Training Protocol}
\label{par:training_protocol}
All evaluated models were trained exclusively on the FoodSeg103 dataset using their respective official training recipes. For transformer-based models, standard ImageNet or ImageNet-21k pretraining was retained where applicable. No model was fine-tuned on BenchSeg. This protocol is designed to isolate cross-dataset generalization rather than in-distribution performance. ReLeM \cite{wu2021large} denotes a multimodal pre-training strategy that injects semantic food knowledge into these segmentors, while memory-based models are combined with them to study the effect of temporal propagation under identical settings.

\paragraph{Dataset}
\label{par:dataset}
In our experiments, we use FoodSeg103, a single food image dataset, for in-domain training and evaluation, while an additional Asian food set serves as an out-of-domain test. FoodSeg103\cite{wu2021large} is randomly split into training and testing subsets with a 7:3 ratio. The training set comprises 4,983 images with 29,530 ingredient masks, and the testing set contains 2,135 images with 12,567 masks. For ReLeM training, the training split of \textit{Recipe1M+} is used to learn recipe representations, while FoodSeg103 test images are kept unseen during training. For FoodLMM \cite{yin2025foodlmm}, SegMan \cite{fu2025segman}, and FoodMem \cite{almughrabi2025foodmem}, we followed the same implementation settings and hyperparameters that were presented in their papers. 

\paragraph{Segmenter}
\label{par:segmenter}
We evaluate three segmentation architectures: CCNet \cite{huang2019ccnet}, FPN \citep{siddiqui2023panoptic}, and SeTR \cite{zheng2021rethinking}. CCNet and FPN use a ResNet-50 \citep{wu2021large} backbone pre-trained on ImageNet-1k, whereas SeTR uses a ViT-16/B \cite{zheng2021rethinking} transformer backbone pre-trained on ImageNet-21k. The ViT backbone contains 12 transformer encoder layers with 12-head self-attention, and its positional embeddings are reinitialized via bilinear interpolation. SeTR extracts features from the 12th transformer layer, followed by two convolutional layers for mask prediction. Other components follow default configurations and are randomly initialized.

\paragraph{ReLeM}
\label{par:relem}
The vision encoders in ReLeM follow the same configuration as the segmenters, using either ResNet-50 or ViT-16/B. Text inputs are preprocessed using the skip-instruction models initialized from pre-trained weights \cite{marin2021recipe1m+}.

\paragraph{Learning Parameters for Segmenter}
\label{par:learn-seg}
Input images are resized to $2049 \times 1024$ pixels with a scaling ratio between 0.5 and 2.0, from which $768 \times 768$ crops are extracted. Standard data augmentations, including random horizontal flipping and color jitter, are applied. Models are trained for 80k iterations with a batch size of 8 using SGD with momentum 0.9 and weight decay 0.0005. The initial learning rate is set to $10^{-3}$ for all architectures and decayed polynomially with a power of 0.9. No hard negative mining is applied. All segmenter experiments are conducted using the MMsegmentation platform \citep{contributors2020mmsegmentation}, except FoodLMM and Yolo. For all experiments, we used 4 Nvidia H100 GPUs (80G VRAM). For testing, using the BenchSeg dataset, we used 1 Nvidia RTX 5090 (32G VRAM) and 1 Nvidia RTX 3090 (24G VRAM). 

\paragraph{Learning Parameters for ReLeM}
\label{par:learn-relem}
Input images are resized to $256 \times 256$ and cropped to $224 \times 224$ for the vision encoder. Training is performed for 720 epochs with a batch size of 160 using the Adam optimizer \citep{adam2014method} with learning rate $10^{-4}$. A two-stage optimization strategy is adopted: first, the vision encoder is frozen, and only the text encoder is trained; once the text encoder converges, its parameters are frozen while the vision encoder is optimized.

\paragraph{Learning Parameters for YOLO}
\label{par:learn-yolo} 
The input resolution is fixed to $640\times640$. Data augmentation follows the default YOLO pipeline with the following settings: Mixup~0.15, Copy-Paste~0.3, rotation~$\pm 10^\circ$, translation~0.1, scale~0.5, horizontal flip with probability 0.5, and Mosaic augmentation enabled. We train for 300 epochs with AdamW and an early-stopping patience of 50 epochs. The batch size is automatically determined based on GPU memory availability. All other settings remain at their default values.

\subsection{Quality Metrics}
We evaluate segmentation models using standard computer vision metrics.
During training on \textit{FoodSeg103}, performance is measured using mean Intersection-over-Union (mIoU) \cite{everingham2010pascal} and mean Accuracy (mAcc) \cite{long2015fully}, which quantify class-level agreement between predictions and ground truth. 
For testing on \textit{BenchSeg}, we report Average Precision (AP; denoted mAP for consistency, but computed as single-class AUPRC for food-vs-background), Precision, Recall \cite{papadopoulos2017training}, F1-score, and IoU across mask sequences. Precision and Recall capture complementary aspects of false-positive and false-negative behavior, while the F1-score summarizes their harmonic balance. IoU provides a direct measure of spatial overlap between predicted and ground-truth masks, offering an intuitive assessment of segmentation quality under challenging variations\footnote{\label{more_det} More details are provided in the appendix.}. 

\subsubsection{Label space and evaluation mapping}
\label{sec:label_mapping}

BenchSeg provides a single binary foreground mask per frame, representing the visible edible region of the dish instance (food vs.\ background). In contrast, FoodSeg103 models predict ingredient-level logits over 103 categories plus background. To ensure fair evaluation across model families, we convert all model outputs into a binary foreground prediction $\hat m_t \in \{0,1\}^{H\times W}$ as follows.

\paragraph{Ingredient models (FoodSeg103-trained)}
Let $p_t(c \mid x_t)$ be the per-pixel probability for class $c$. We define the food foreground probability as
\begin{equation}
p^{\text{fg}}_t = 1 - p_t(\text{background} \mid x_t),
\end{equation}
and obtain the binary mask by thresholding $\hat m_t = \mathbbm{1}\{p^{\text{fg}}_t \ge \tau\}$ with $\tau=0.5$.

\paragraph{Promptable / class-agnostic models (e.g., SAM-based)}
For promptable models that output one or more candidate masks, we select the mask maximizing overlap with the previous-frame prediction (for video) or maximizing internal confidence (for single-frame), and threshold the returned probability mask if applicable. For the first frame of each sequence, we select the mask with the highest internal confidence (or the largest mask if confidences are unavailable). To limit drift, we optionally reset the selection every $K$ frames using the same first-frame rule (we use $K=30$ unless stated otherwise).

\paragraph{Multi-mask methods}
When a method produces $M$ candidate masks, we evaluate both (i) the best single mask per frame and (ii) the union mask, and report the primary protocol in the main paper and the alternative in the appendix.

\subsection{FoodSeg103 Observations}
To evaluate segmentation model performance trained on FoodSeg103, we benchmark all methods using a standardized evaluation protocol \cite{wu2021large} and report mIoU and mAcc. Since this experiment aims to compare the segmentation abilities of various architectures, we categorize all methods based on their underlying computational backbone: convolutional networks (e.g., ResNet-based models), recurrent architectures (e.g., LSTMs), and Transformer-based models. This structural grouping helps readers interpret performance differences both across models and within different architecture families.

\begin{table}[htb]
\centering
\setlength{\tabcolsep}{2pt}
\caption{Comprehensive evaluation of all segmentation methods trained on the FoodSeg103 dataset. Methods are sorted in ascending order by mIoU. Best results in \textbf{bold}, second-best in \uline{underline}, and third-best in \textit{italic}.}
\begin{tabular}{lcccc}
\toprule
\textbf{Method} & \textbf{Backbone} & \textbf{↑mIoU (\%)$\blacktriangle$} & \textbf{↑mAcc (\%)} &\textbf{ Model Size} \\
\midrule
FPN & ResNet-50 & 27.8 & 38.2 & \textit{218M} \\
\rowcolor{gray!12}ReLeM-FPN & Transformer & 28.9 & 39.7 & \textit{218M} \\
ReLeM-FPN & LSTM & 29.1 & 39.8 & \textit{218M} \\
\rowcolor{gray!12}CCNet & ResNet-50 & 35.5 & 45.3 & 381M \\
ReLeM-CCNet & Transformer & 36.0 & 46.5 & 381M \\
\rowcolor{gray!12}ReLeM-CCNet & LSTM & 36.8 & 47.4 & 381M \\
SeTR & ViT-16/B & 41.3 & 52.7 & 723M \\
\rowcolor{gray!12}ReLeM-SeTR & Transformer & 43.2 & 55.7 & 723M \\
ReLeM-SeTR & LSTM & \textit{43.9} & \textit{57.0} & 723M \\
\rowcolor{gray!12}SegMAN & Transformer & \uline{46.2} & \uline{57.2} & \uline{51.79M} \\
YOLO & YOLOv11 CNN & \textbf{71.5} & \textbf{78.39} & \textbf{10.1M} \\

\bottomrule
\end{tabular}
\label{tab:model_comparisons}
\end{table}

Table~\ref{tab:model_comparisons} compares all evaluated methods. Metrics show performance differences between convolutional, recurrent, and transformer models. Transformer models, especially SeTR variants, deliver strong segmentation across datasets, while convolutional models like CCNet perform slightly lower. YOLO-based models are lighter and faster, offering competitive recall but lower mAP, suited for quick, real-time tasks rather than detailed segmentation. The model's memory and speed show a trade-off: transformers have high cost but top accuracy; convolutional methods are balanced; YOLO variants are most efficient. Choice depends on needs—transformers suit offline tasks, while YOLO is better for real-time or limited-resource applications.

\subsection{BenchSeg Observations}
\begin{figure}
    \centering
    \includegraphics[trim={0cm 0.4cm 0.8cm 0cm},clip,width=1.0\linewidth]{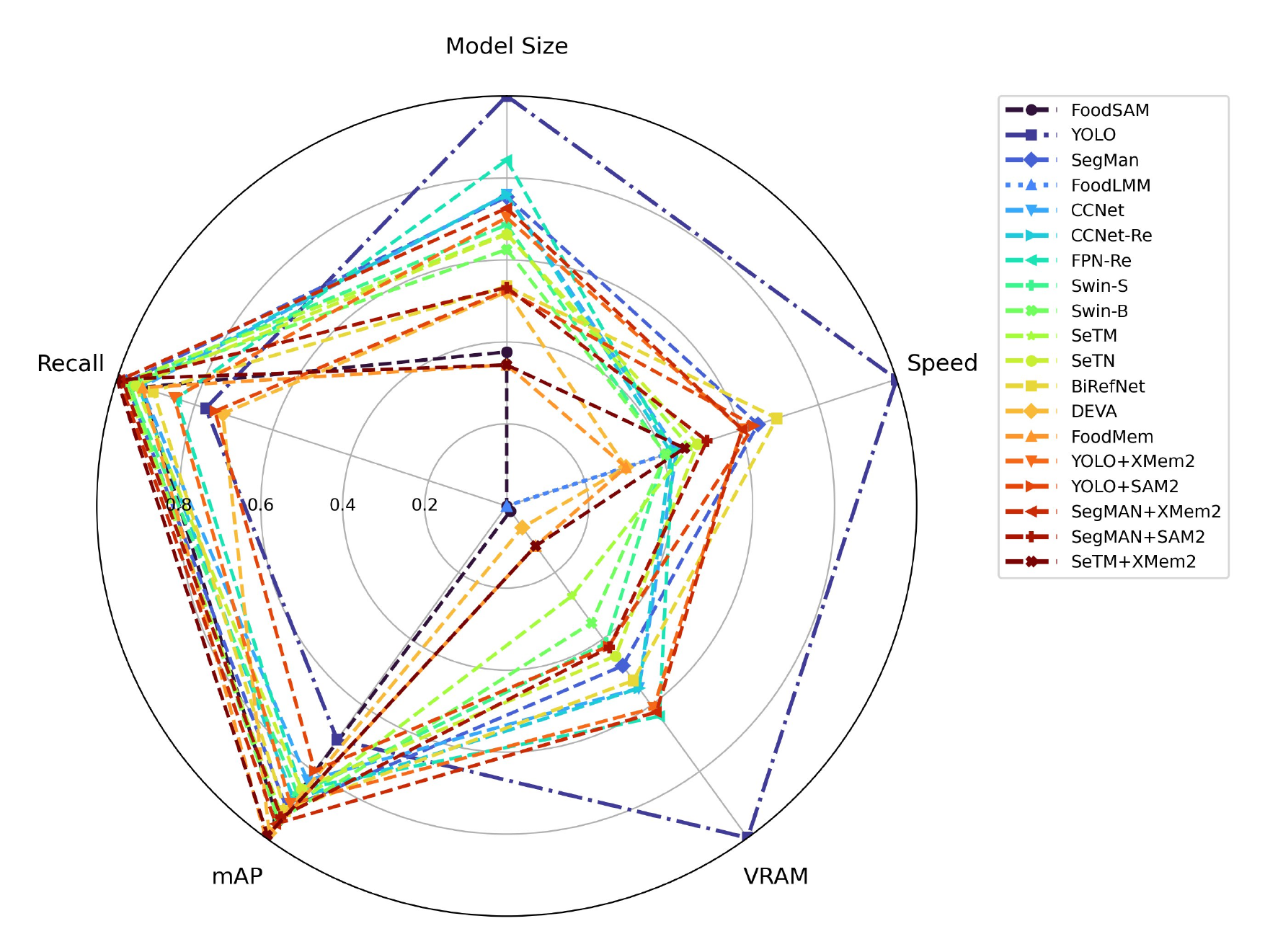}
    \caption{Illustration of the explicit trade-off between segmentation accuracy and computational cost. Configurations achieving the highest accuracy also exhibit substantially larger parameter counts, longer processing times, and increased memory usage, highlighting the balance required when selecting an optimal model setup. (mAP and recall are averaged across all datasets)}
    \label{fig:resources}
\end{figure}

\begin{figure}
    \centering
    \begin{subfigure}[t]{0.22\textwidth}
        \centering
        \includegraphics[trim={0cm 1.5cm 0.2cm 2.7cm},clip,width=1.0\textwidth]{ 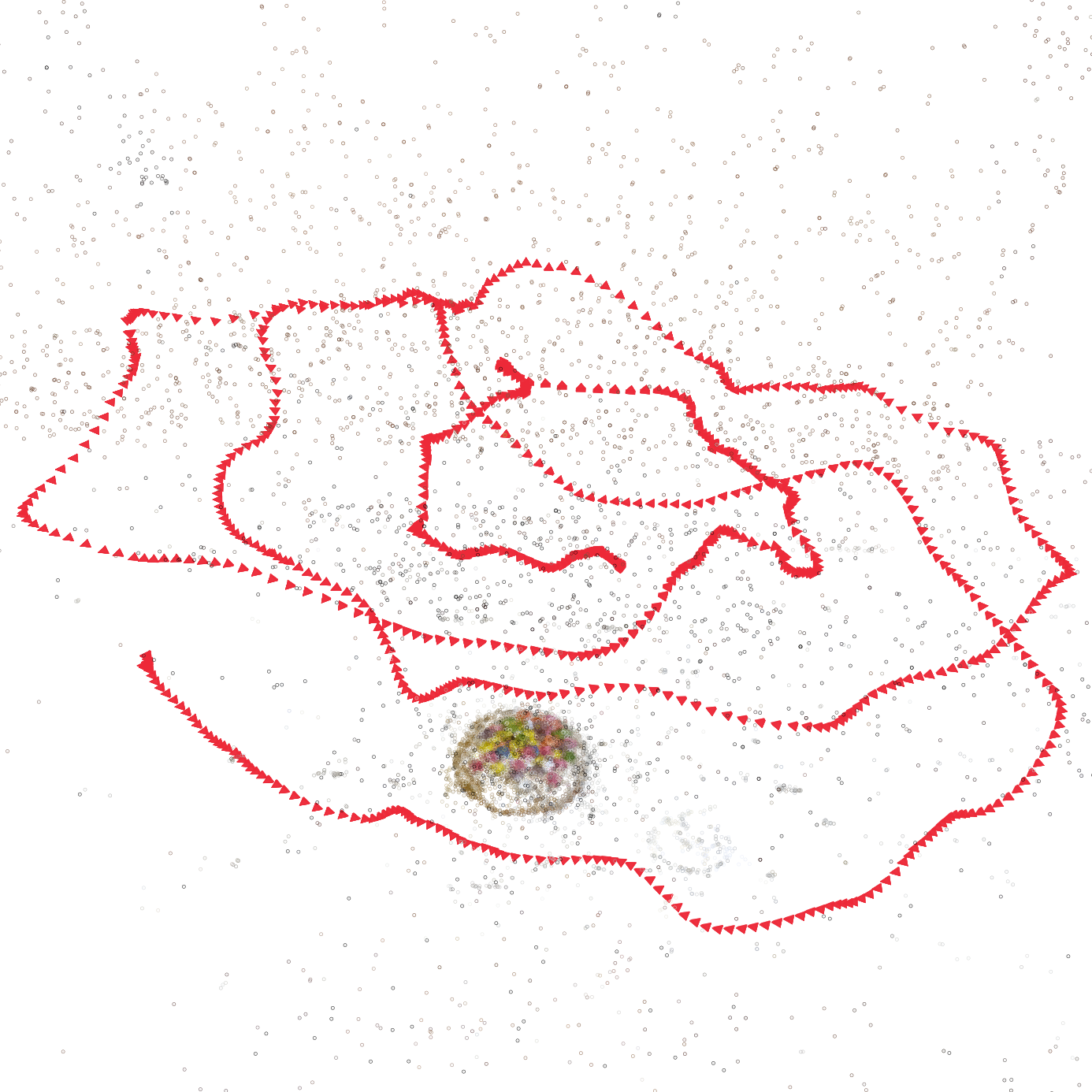}
        \caption{FoodLMM}
    \end{subfigure}
    \begin{subfigure}[t]{0.22\textwidth}
        \centering
        \includegraphics[trim={0cm 1.5cm 0.2cm 2.7cm},clip,width=1.0\textwidth]{ 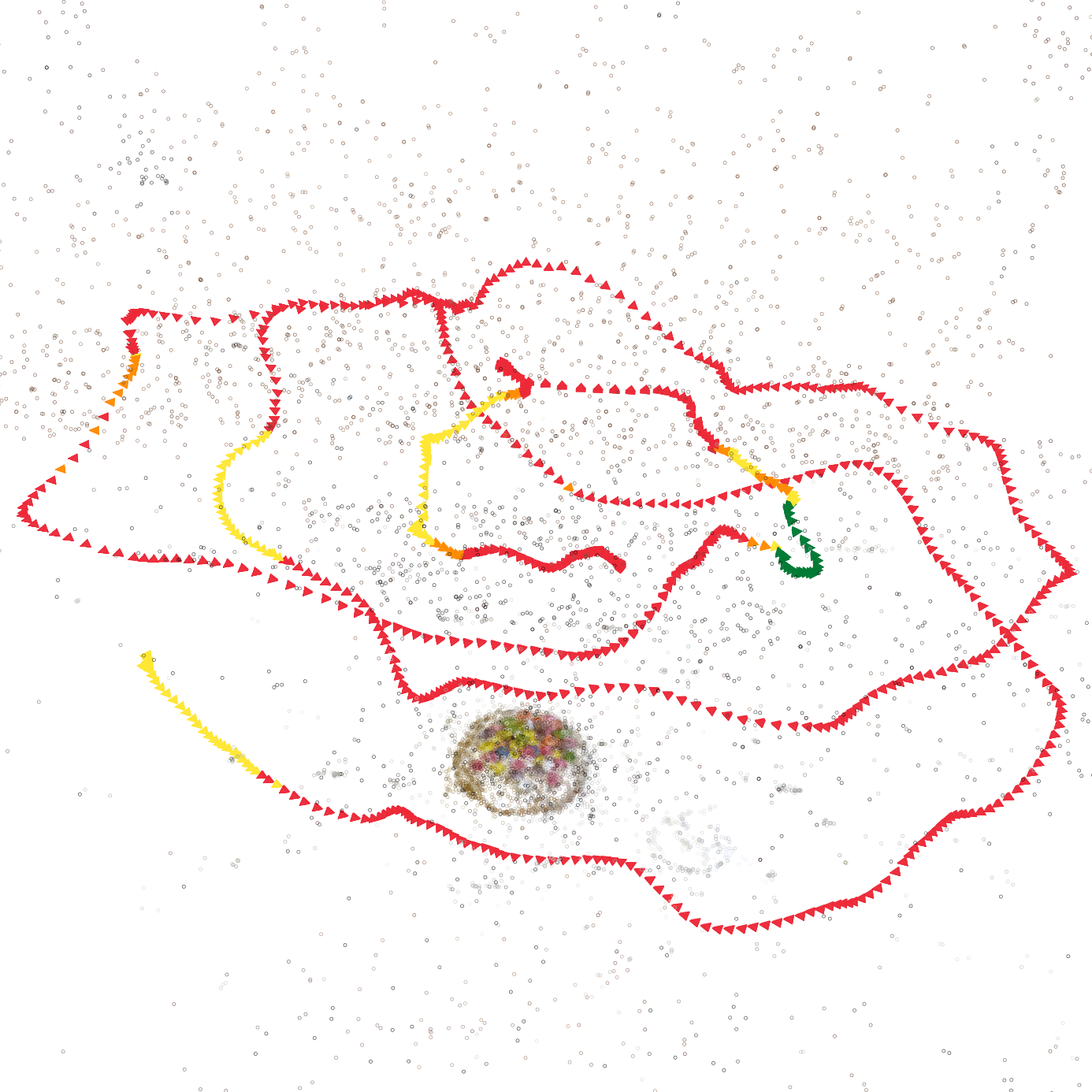}
        \caption{kMean++}
    \end{subfigure}
    \begin{subfigure}[t]{0.22\textwidth}
        \centering
        \includegraphics[trim={0cm 1.5cm 0.2cm 2.7cm},clip,width=1.0\textwidth]{ 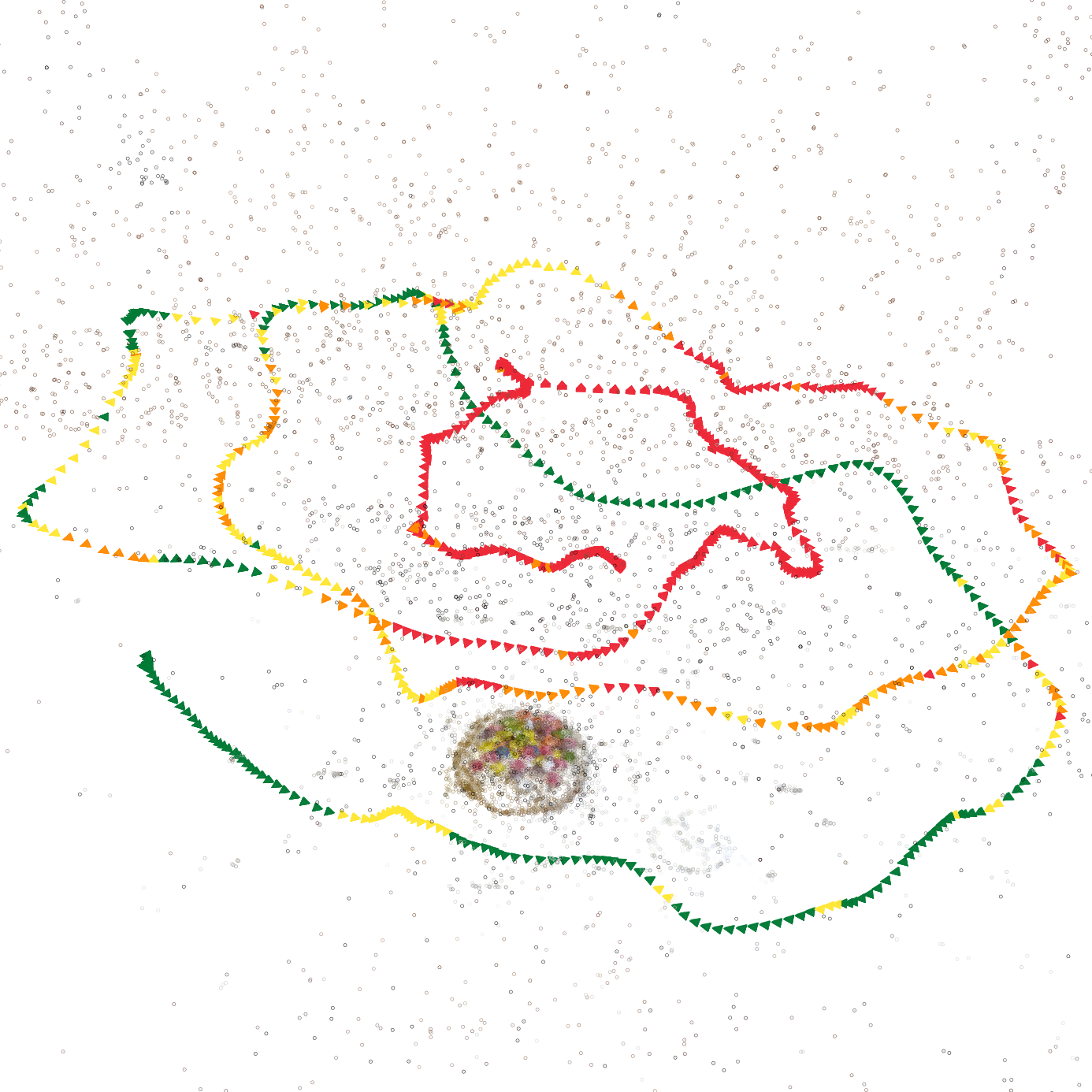}
        \caption{CCNet Relem}
    \end{subfigure}
    \begin{subfigure}[t]{0.22\textwidth}
        \centering
        \includegraphics[trim={0cm 1.5cm 0.2cm 2.7cm},clip,width=1.0\textwidth]{ 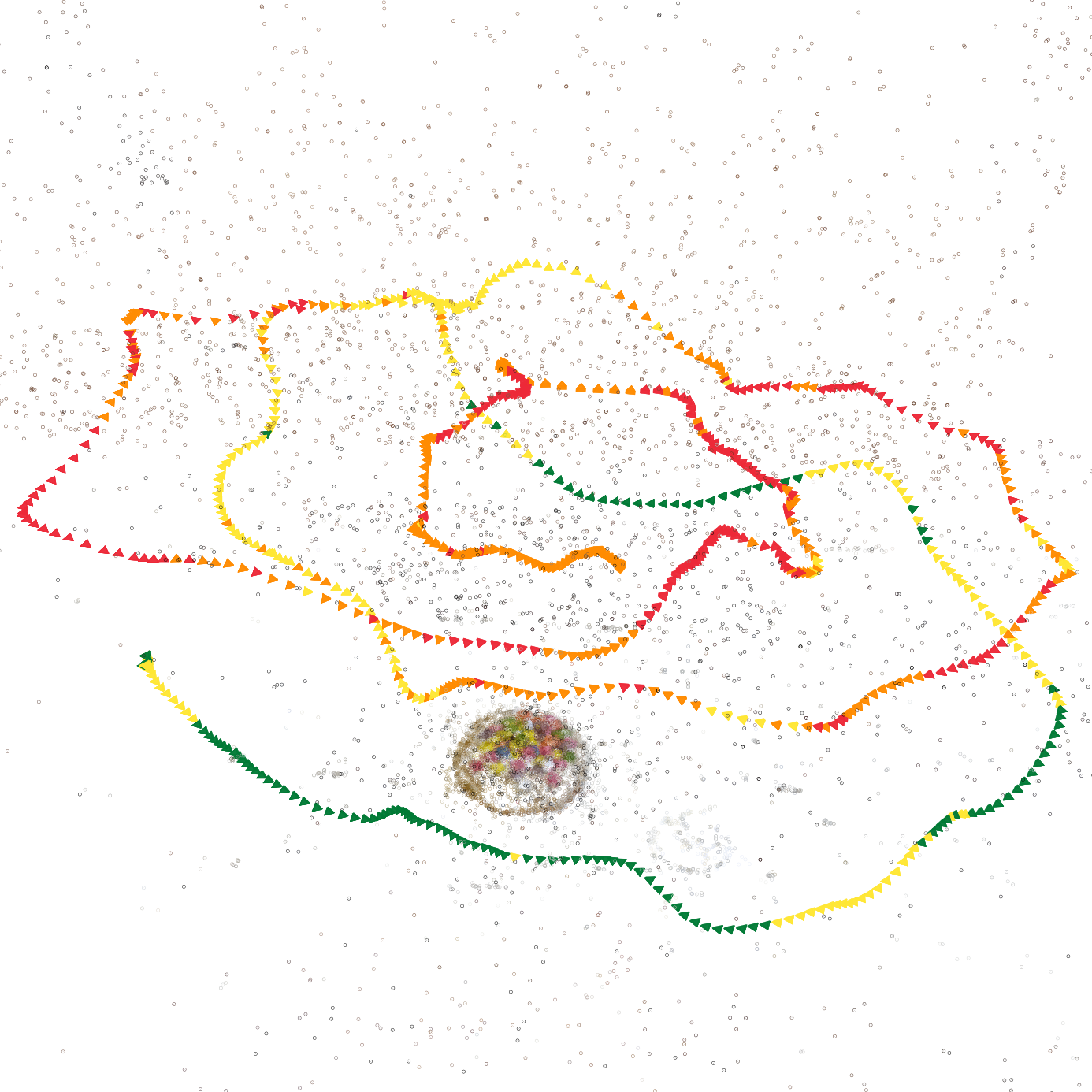} 
        \caption{CCNet}
    \end{subfigure}
    \begin{subfigure}[t]{0.22\textwidth}
        \centering
        \includegraphics[trim={0cm 1.5cm 0.2cm 2.7cm},clip,width=1.0\textwidth]{ 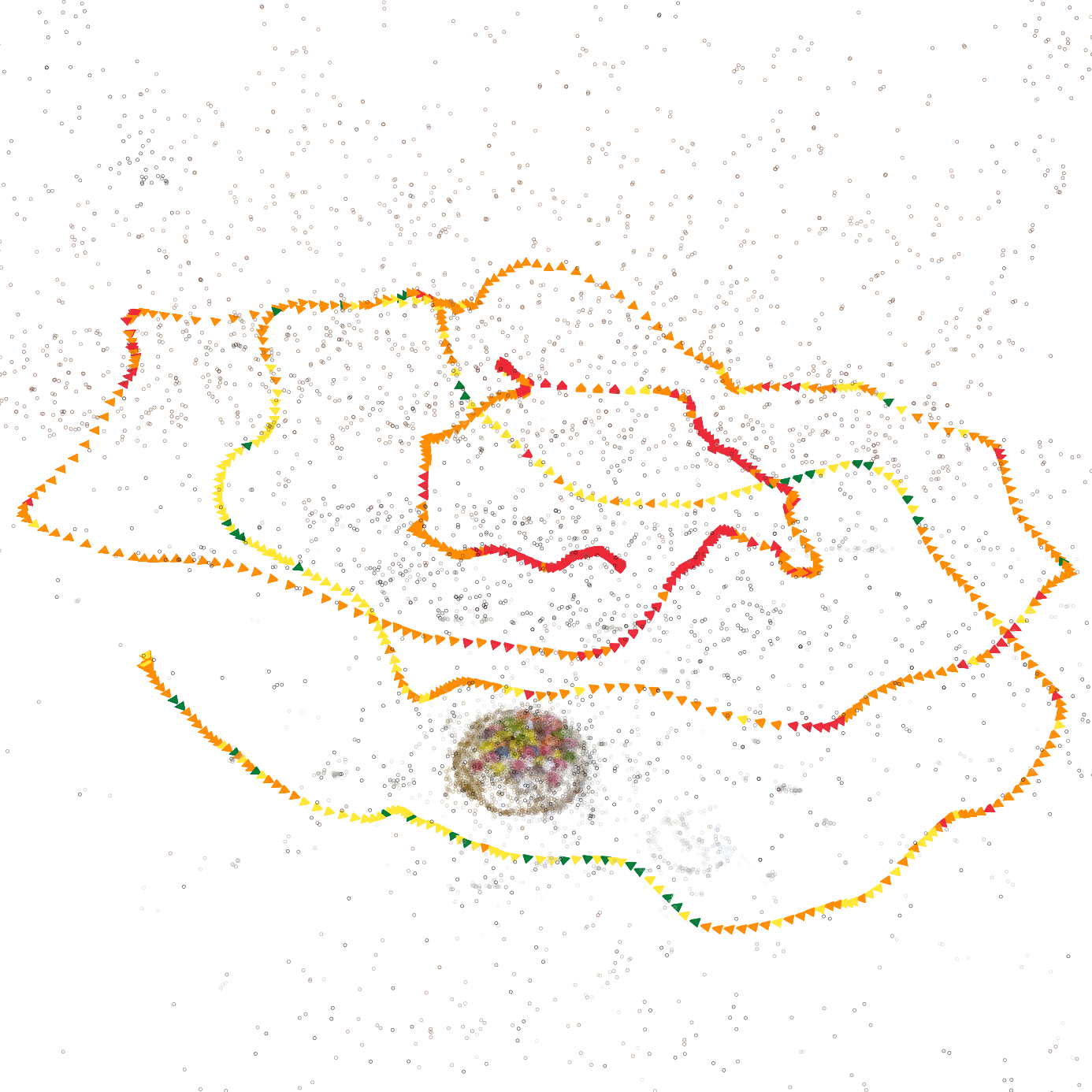}
        \caption{YOLO}
    \end{subfigure}
    \begin{subfigure}[t]{0.22\textwidth}
        \centering
        \includegraphics[trim={0cm 1.5cm 0.2cm 2.7cm},clip,width=1.0\textwidth]{ 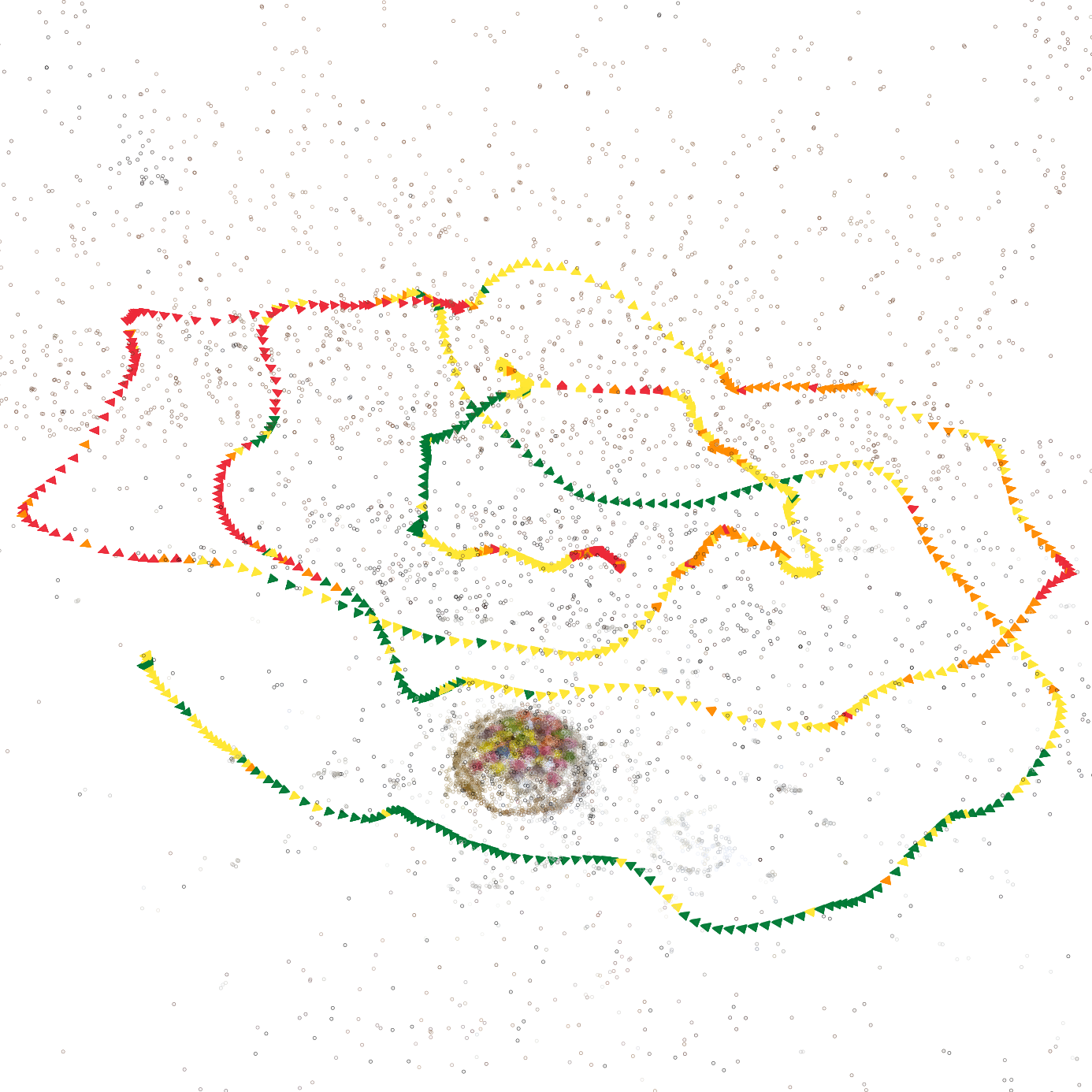}
        \caption{Swin-Small}
    \end{subfigure}
    \begin{subfigure}[t]{0.22\textwidth}
        \centering
        \includegraphics[trim={0cm 1.5cm 0.2cm 2.7cm},clip,width=1.0\textwidth]{ 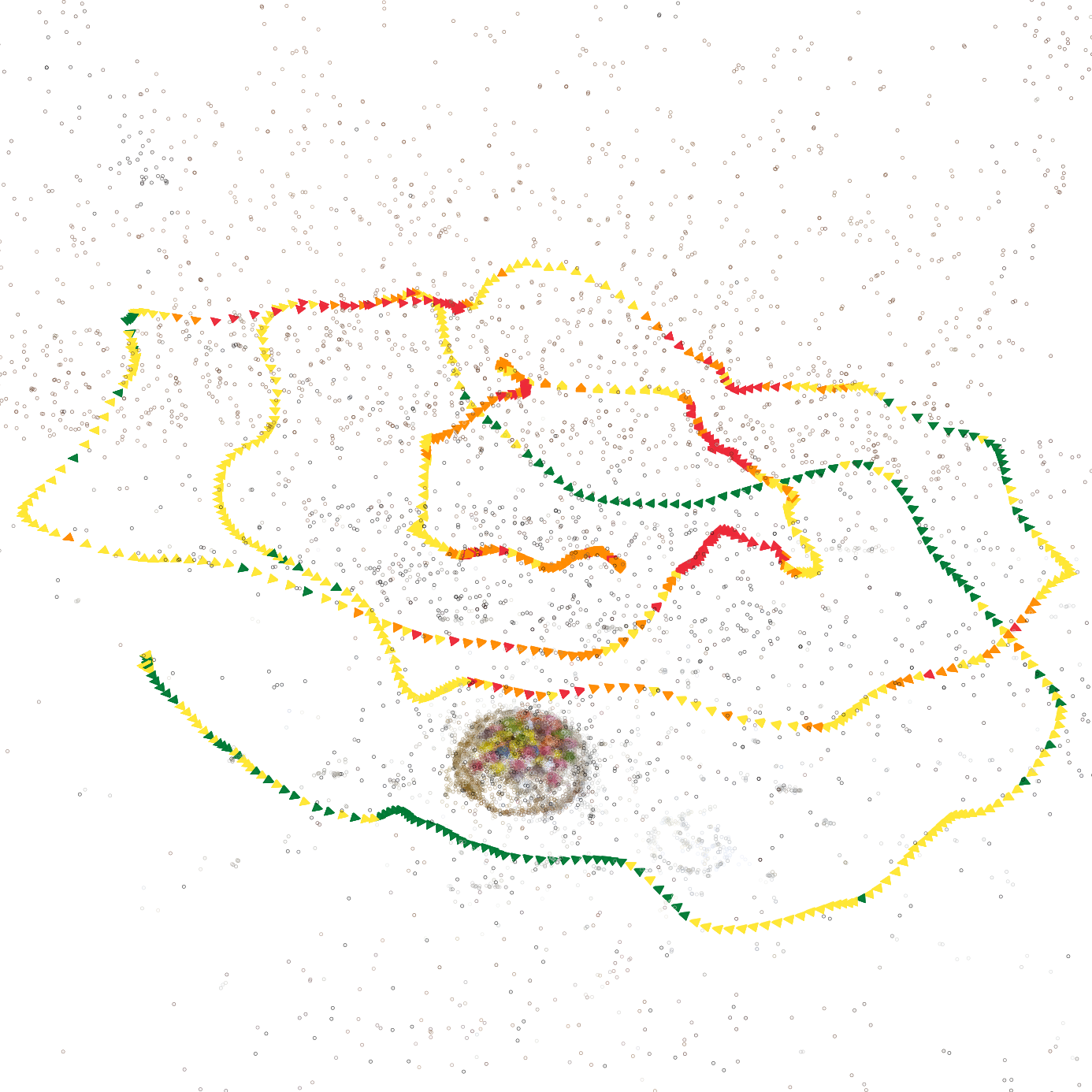}
        \caption{FPN-Relem}
    \end{subfigure}
    \begin{subfigure}[t]{0.22\textwidth}
        \centering
        \includegraphics[trim={0cm 1.5cm 0.2cm 2.7cm},clip,width=1.0\textwidth]{ 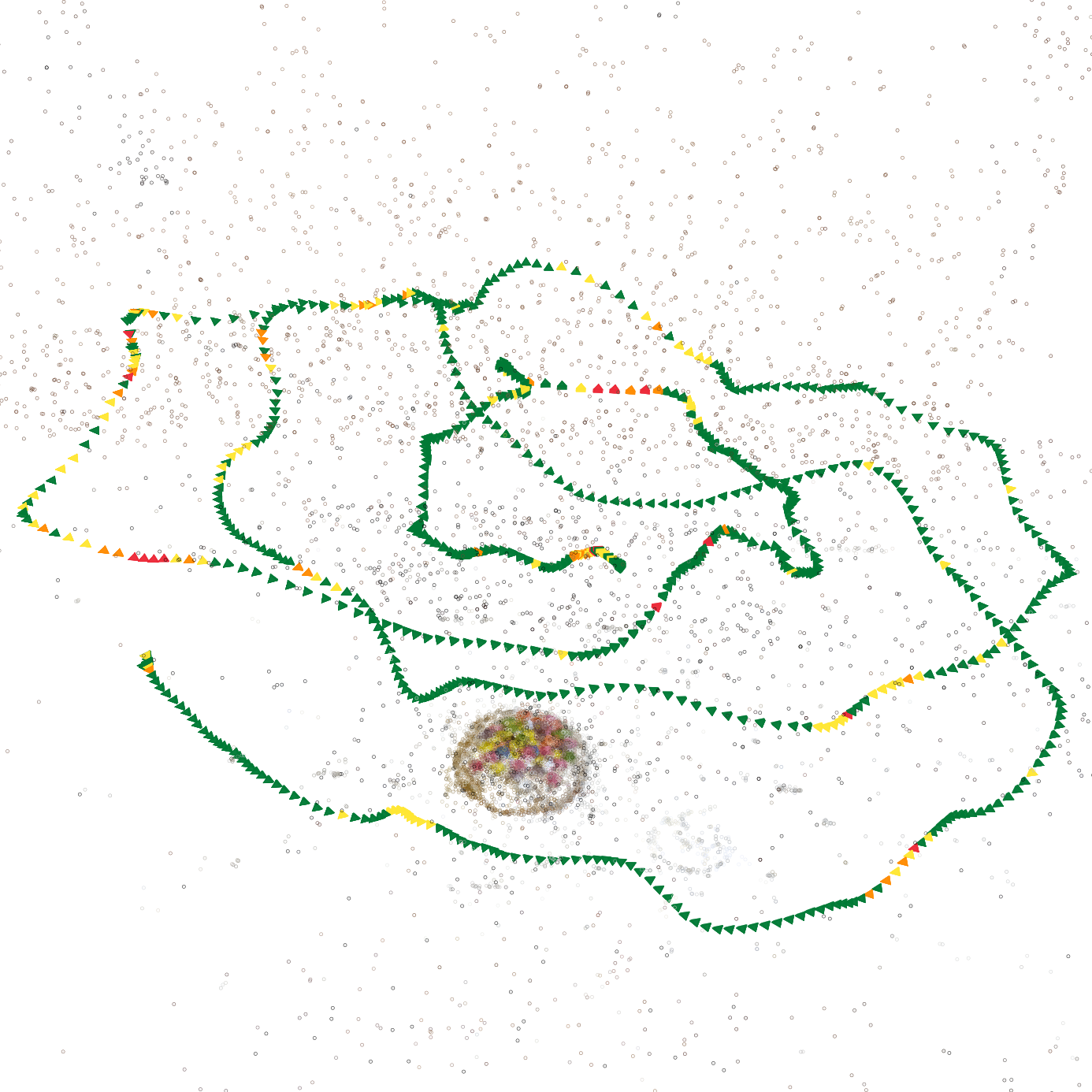}
        \caption{Swin-Base}
    \end{subfigure}
    \begin{subfigure}[t]{0.22\textwidth}
        \centering
        \includegraphics[trim={0cm 1.5cm 0.2cm 2.7cm},clip,width=1.0\textwidth]{ 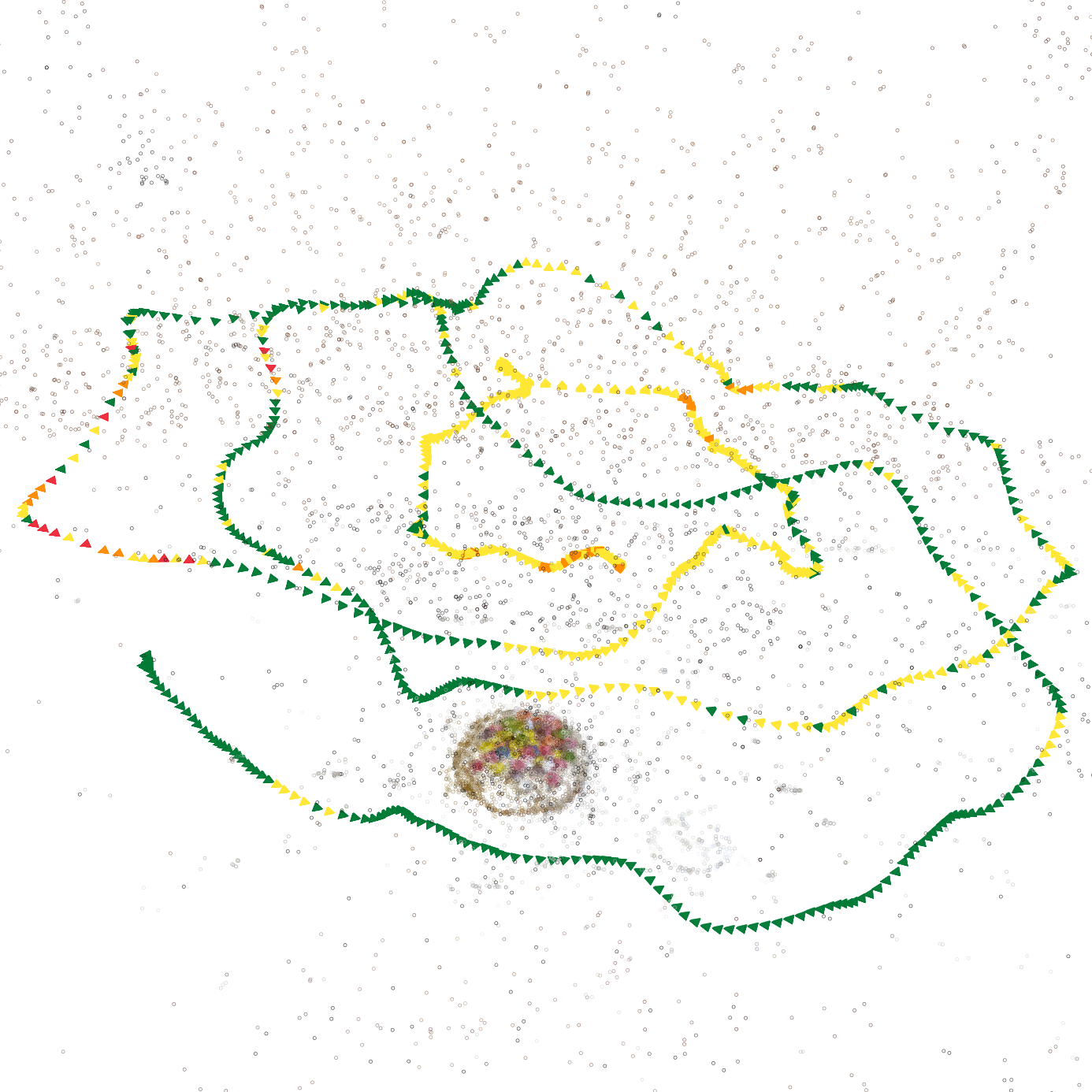}
        \caption{SeTR-Naive}
    \end{subfigure}
    \begin{subfigure}[t]{0.22\textwidth}
        \centering
        \includegraphics[trim={0cm 1.5cm 0.2cm 2.7cm},clip,width=1.0\textwidth]{ 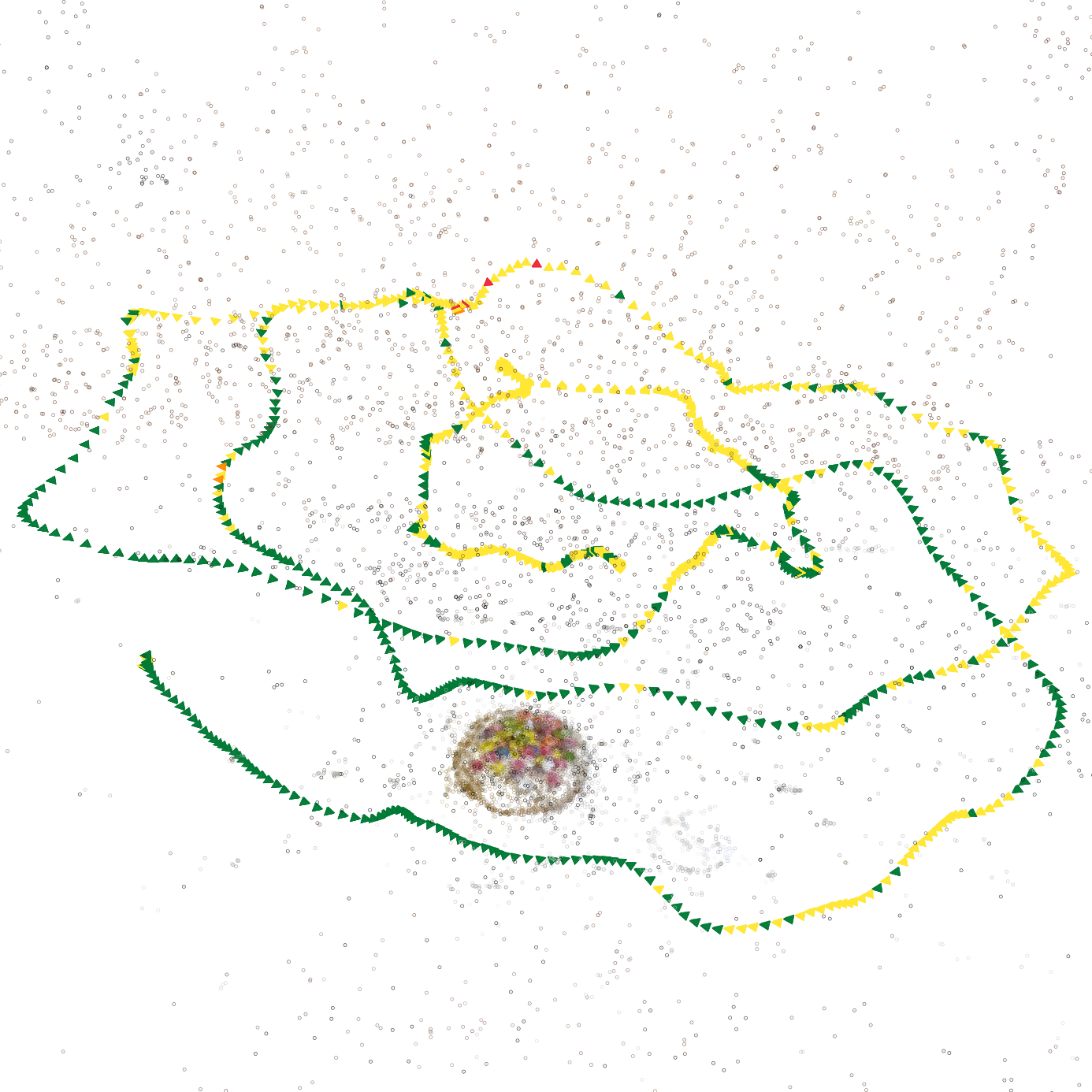}
        \caption{SegMan}
    \end{subfigure}
    \begin{subfigure}[t]{0.22\textwidth}
        \centering
        \includegraphics[trim={0cm 1.5cm 0.2cm 2.7cm},clip,width=1.0\textwidth]{ 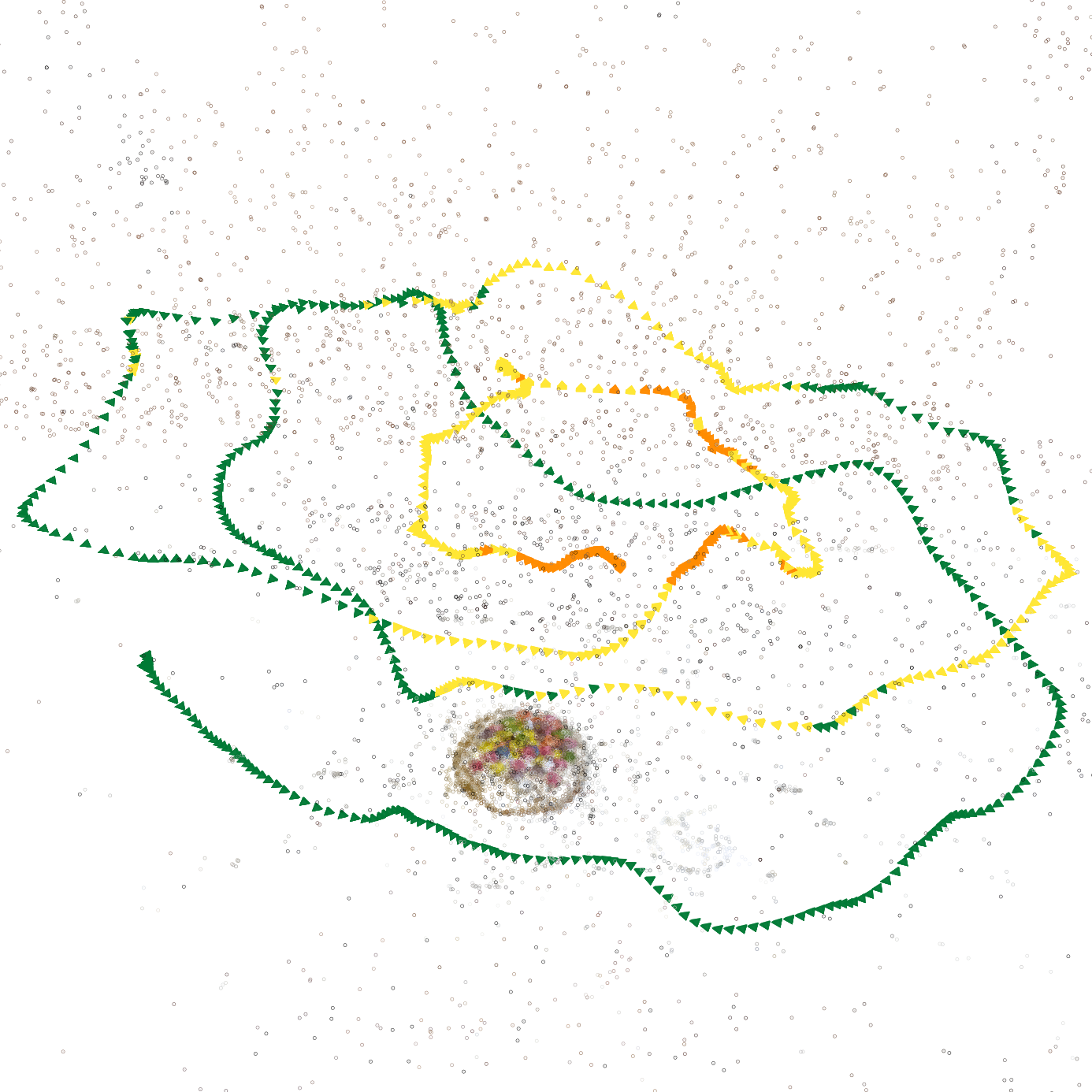}
        \caption{SeTR-MLA}
    \end{subfigure}
    \begin{subfigure}[t]{0.22\textwidth}
        \centering
        \includegraphics[trim={0cm 1.5cm 0.2cm 2.7cm},clip,width=1.0\textwidth]{ 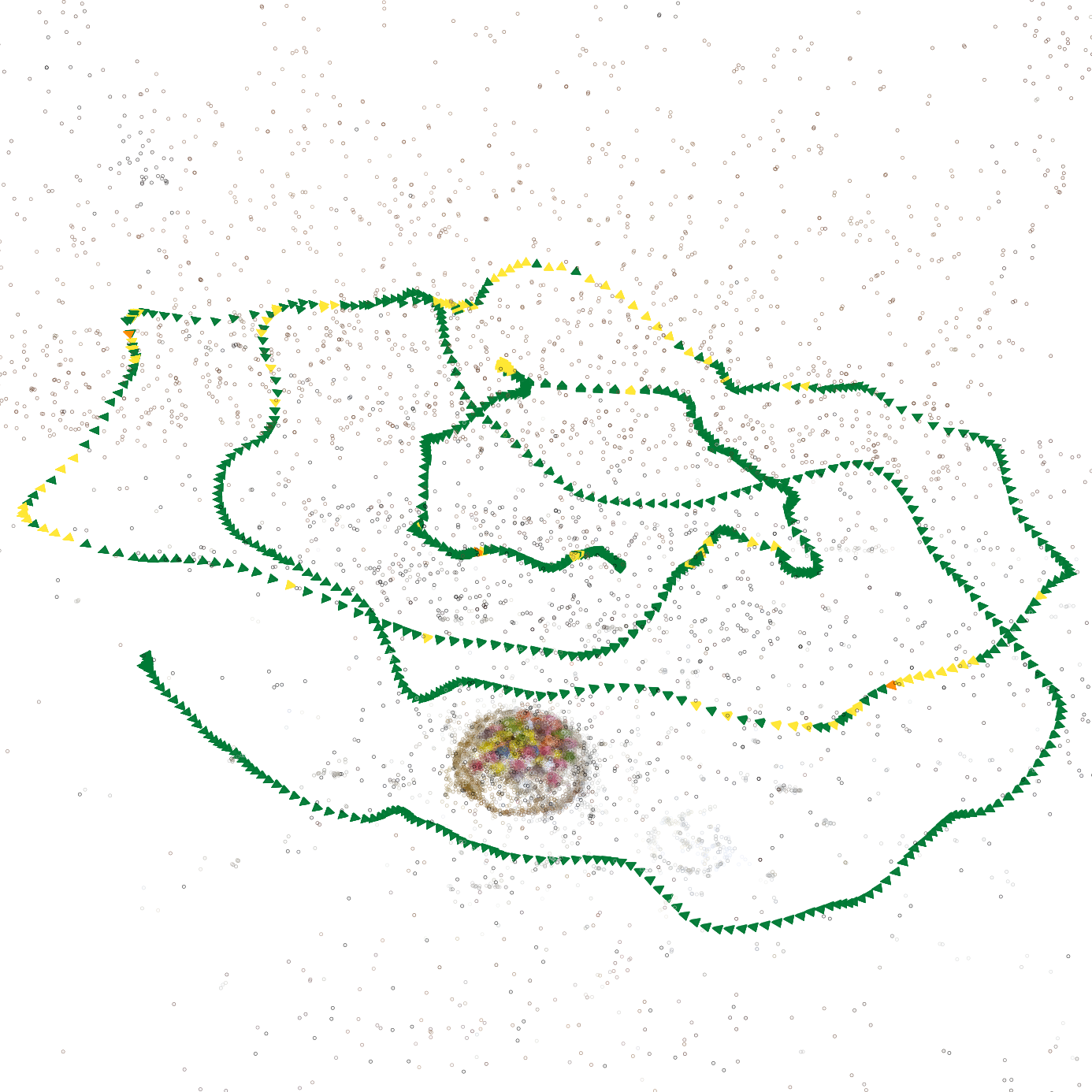}
        \caption{FoodSAM}
    \end{subfigure}
    \begin{subfigure}[t]{0.22\textwidth}
        \centering
        \includegraphics[trim={0cm 1.5cm 0.2cm 2.7cm},clip,width=1.\textwidth]{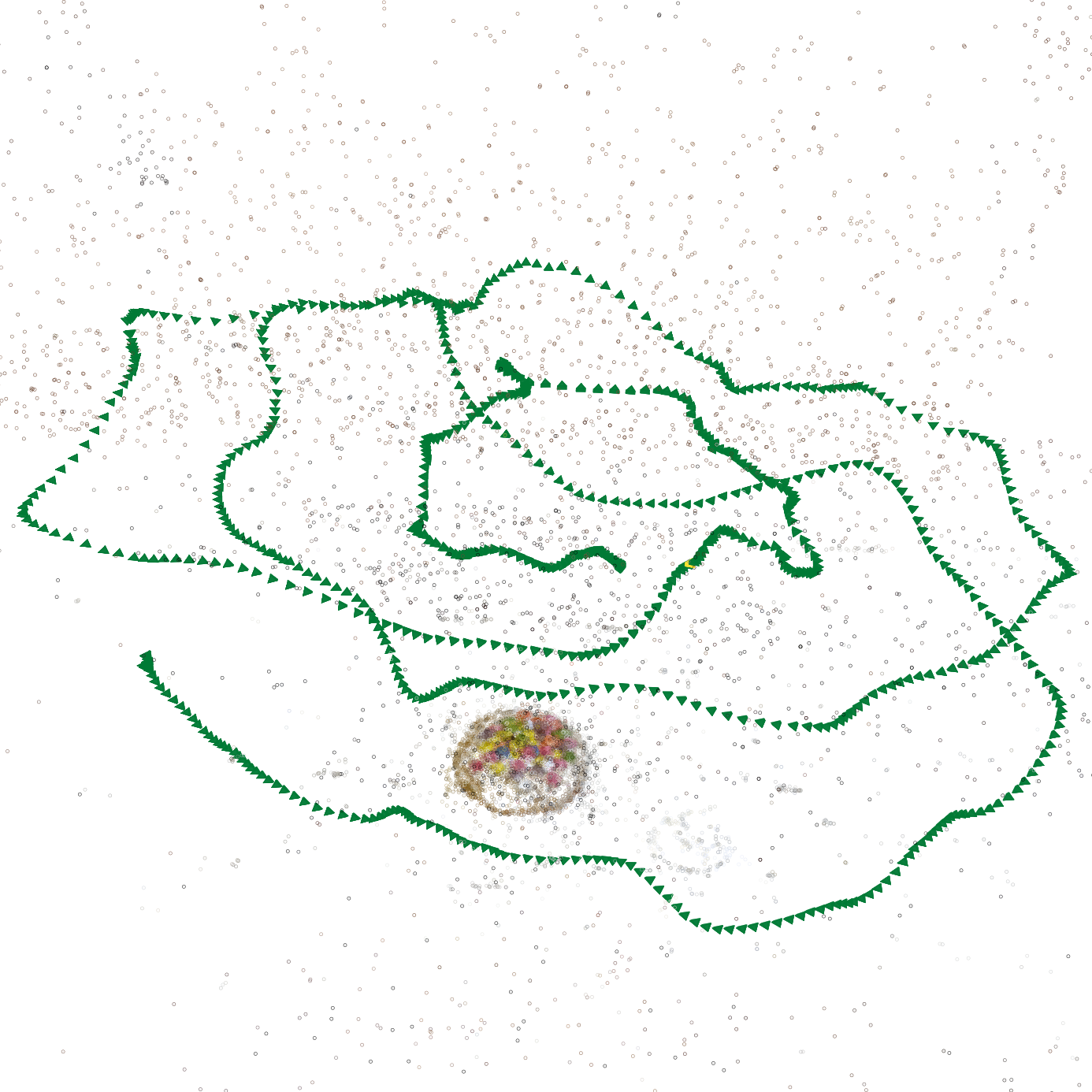}
        \caption{BiRefNet}
    \end{subfigure}
    \begin{subfigure}[t]{0.22\textwidth}
        \centering
        \includegraphics[trim={0cm 1.5cm 0.2cm 2.7cm},clip,width=1.0\textwidth]{ 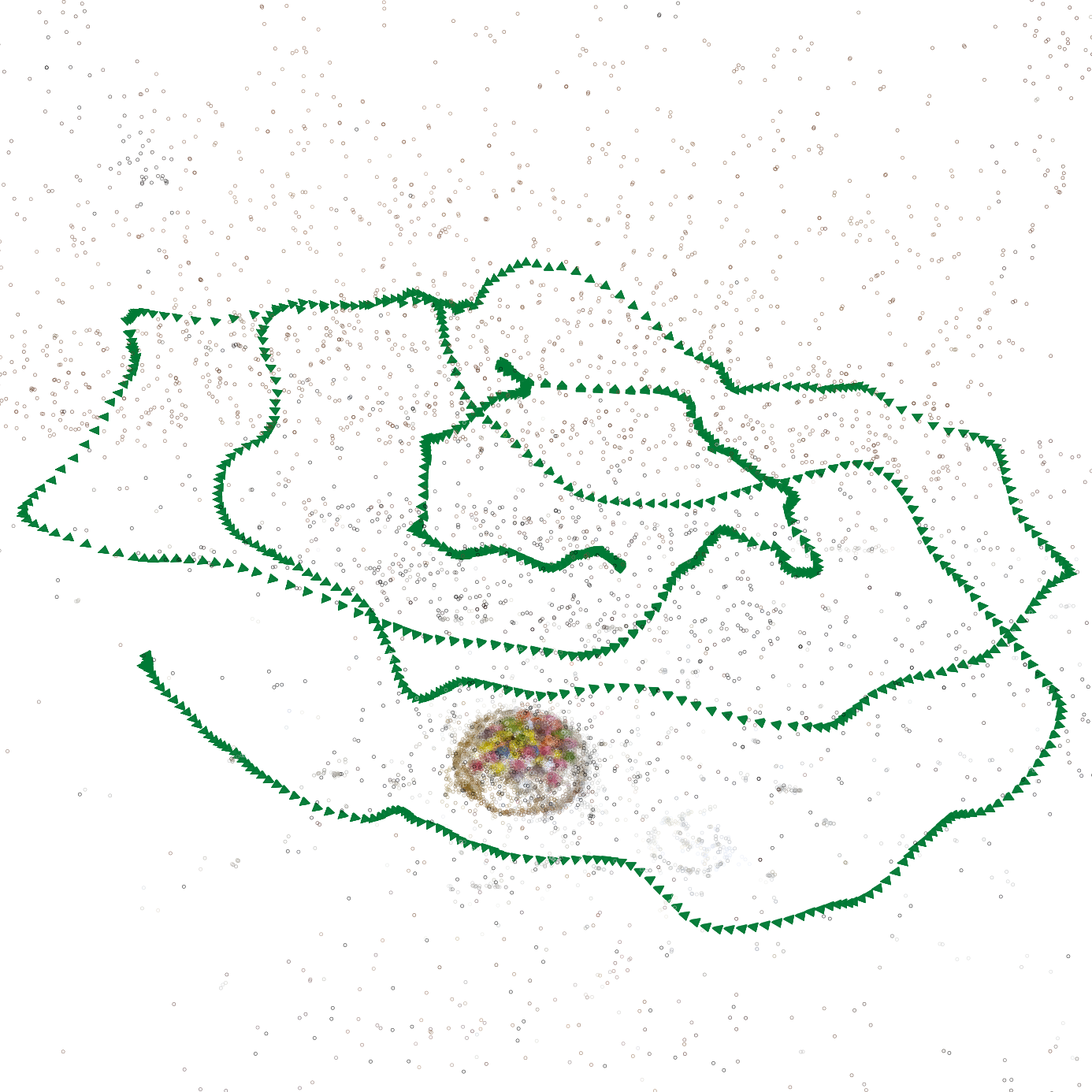}
        \caption{DEVA}
    \end{subfigure}
    \begin{subfigure}[t]{0.22\textwidth}
        \centering
        \includegraphics[trim={0cm 1.5cm 0.2cm 2.7cm},clip,width=1.0\textwidth]{ 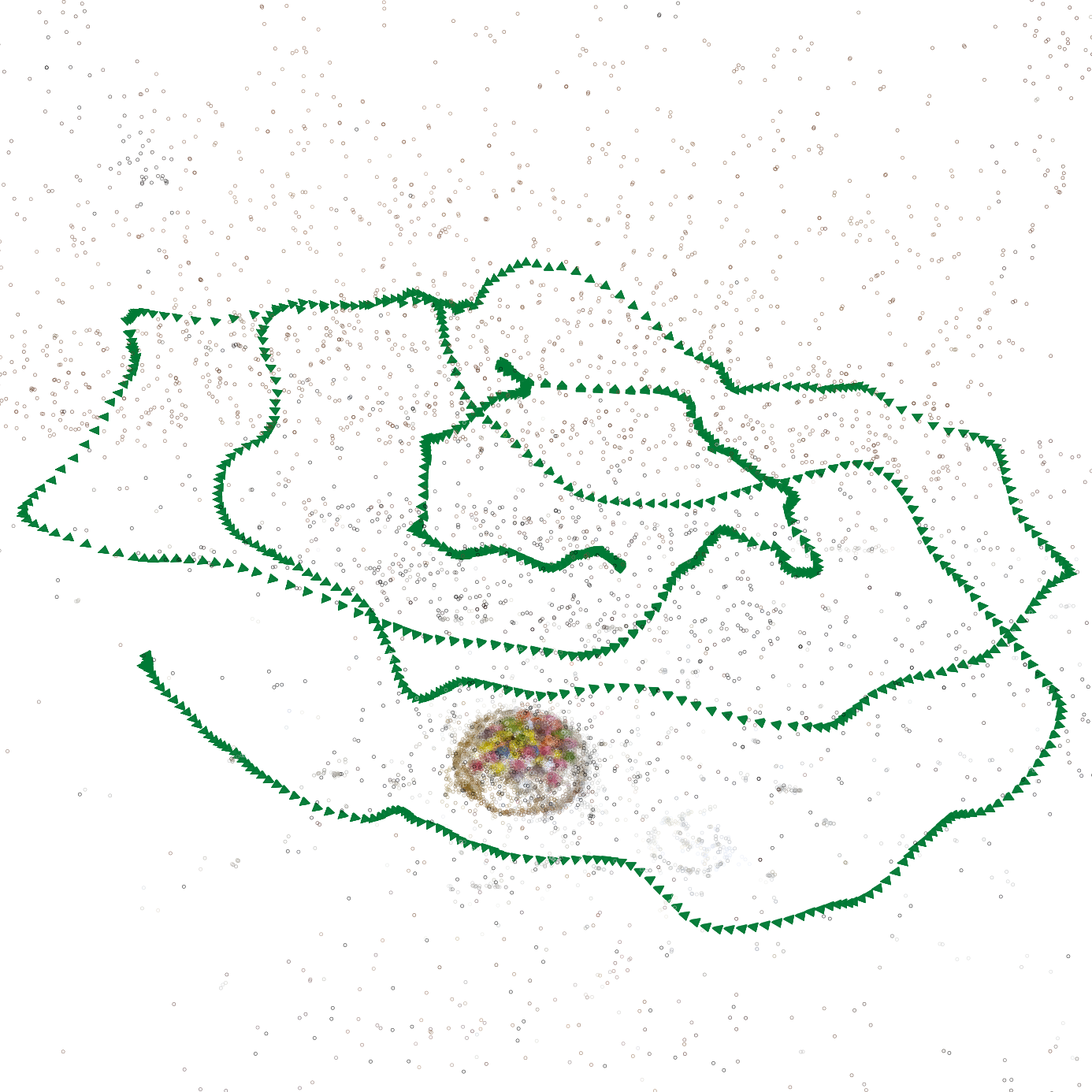}
        \caption{FoodMem}
    \end{subfigure}
    \begin{subfigure}[t]{0.22\textwidth}
        \centering
        \includegraphics[trim={0cm 1.5cm 0.2cm 2.7cm},clip,width=1.0\textwidth]{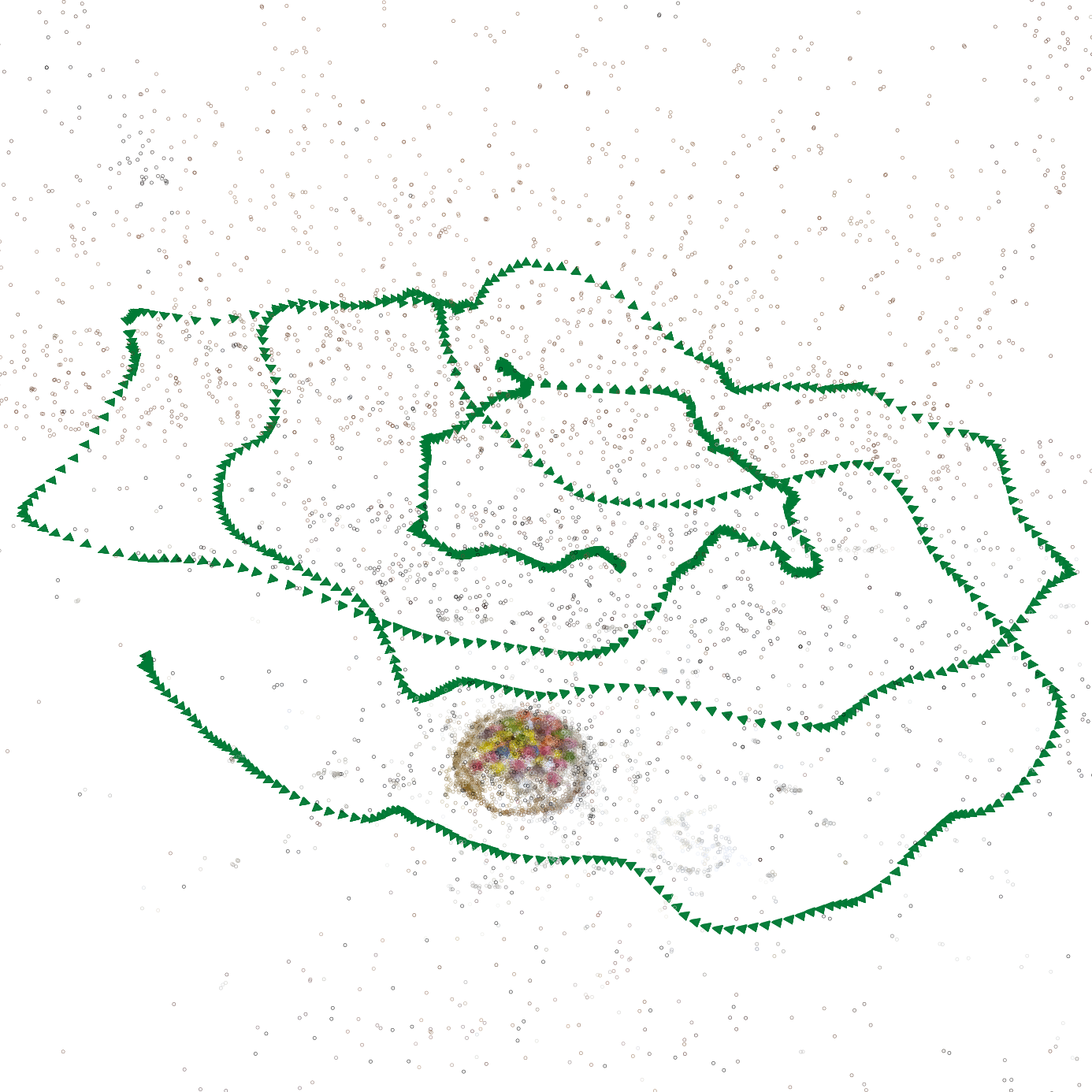}
        \caption{YOLO+XMem2}
    \end{subfigure}
    \caption{Comparison of camera locations for different methods on a Foodkit object, with cameras colored by mAP: red for $\leq$50\%, orange for 50–75\%, yellow for 75–95\%, and green for $\geq$95\%, highlighting differences in coverage and reliability across methods.}
    \label{fig:3d_comparasions}
\end{figure}

\begin{figure}
    \centering
    \begin{subfigure}[t]{0.22\textwidth}
        \centering
         \includegraphics[width=1.0\textwidth]{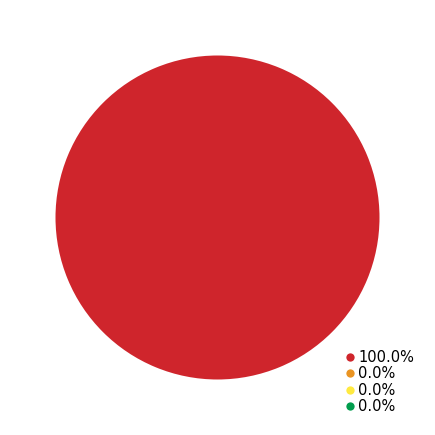}
        \caption{FoodLMM}
    \end{subfigure}
    \begin{subfigure}[t]{0.22\textwidth}
        \centering
         \includegraphics[width=1.0\textwidth]{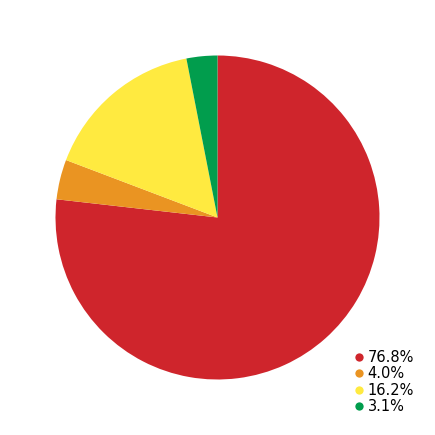}
        \caption{kMean++}
    \end{subfigure}
    \begin{subfigure}[t]{0.22\textwidth}
        \centering
         \includegraphics[width=1.0\textwidth]{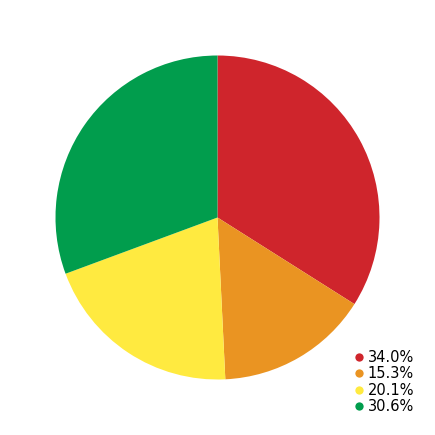}
        \caption{CCNet Relem}
    \end{subfigure}
    \begin{subfigure}[t]{0.22\textwidth}
        \centering
        \includegraphics[width=1.0\textwidth]{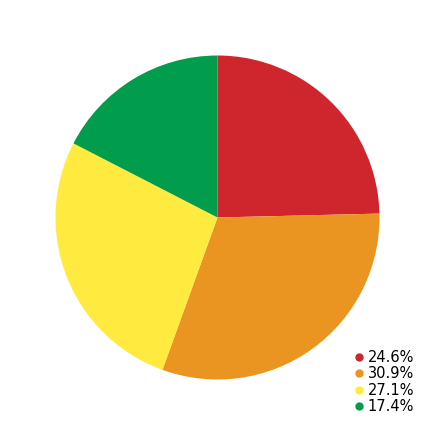}
        \caption{CCNet}
    \end{subfigure}
    \begin{subfigure}[t]{0.22\textwidth}
        \centering
         \includegraphics[width=1.0\textwidth]{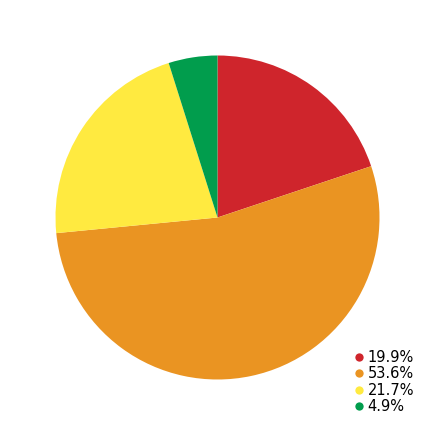}
        \caption{YOLO}
    \end{subfigure}
    \begin{subfigure}[t]{0.22\textwidth}
        \centering
         \includegraphics[width=1.0\textwidth]{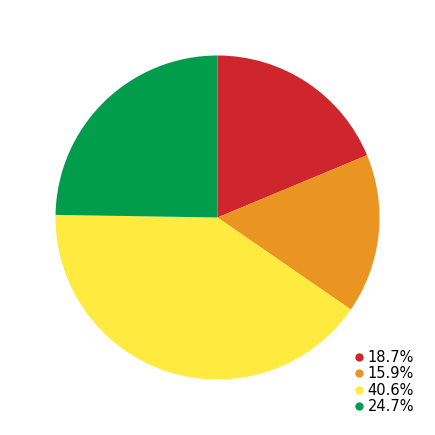}
        \caption{Swin-Small}
    \end{subfigure}
    \begin{subfigure}[t]{0.22\textwidth}
        \centering
         \includegraphics[width=1.0\textwidth]{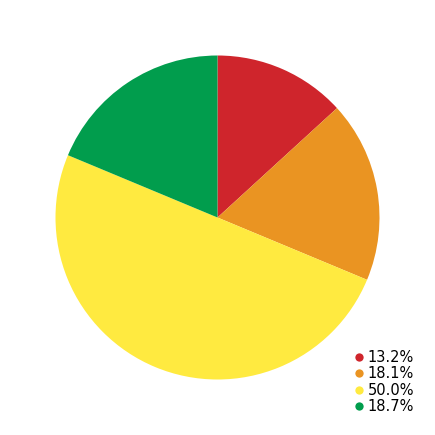}
        \caption{FPN-Relem}
    \end{subfigure}
    \begin{subfigure}[t]{0.22\textwidth}
        \centering
         \includegraphics[width=1.0\textwidth]{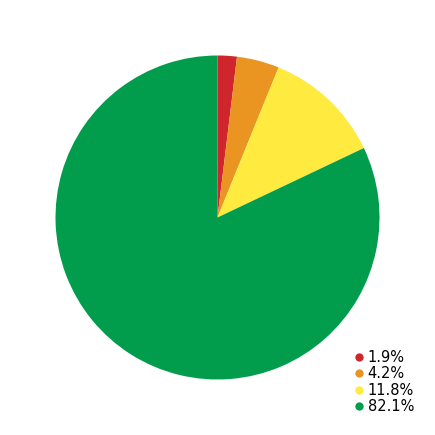}
        \caption{Swin-Base}
    \end{subfigure}
    \begin{subfigure}[t]{0.22\textwidth}
        \centering
         \includegraphics[width=1.0\textwidth]{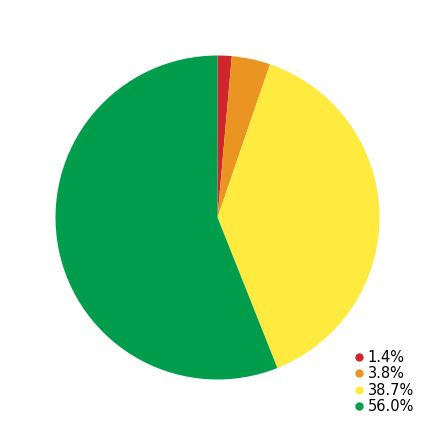}
        \caption{SeTR-Naive}
    \end{subfigure}
    \begin{subfigure}[t]{0.22\textwidth}
        \centering
         \includegraphics[width=1.0\textwidth]{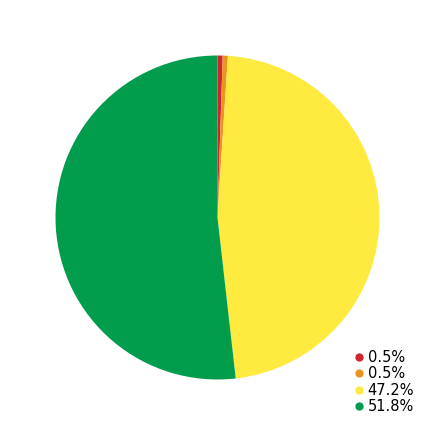}
        \caption{SegMan}
    \end{subfigure}
    \begin{subfigure}[t]{0.22\textwidth}
        \centering
         \includegraphics[width=1.0\textwidth]{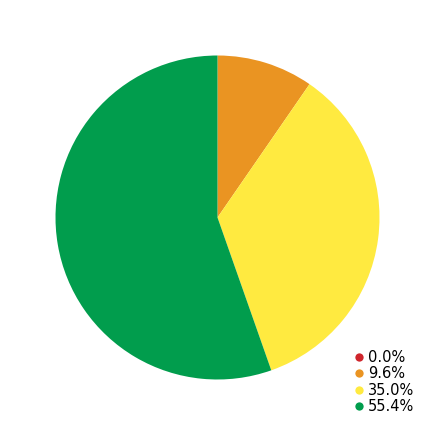}
        \caption{SeTR-MLA}
    \end{subfigure}
    \begin{subfigure}[t]{0.22\textwidth}
        \centering
         \includegraphics[width=1.0\textwidth]{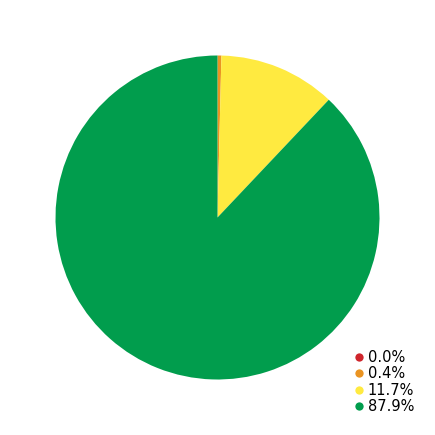}
        \caption{FoodSAM}
    \end{subfigure}
    \begin{subfigure}[t]{0.22\textwidth}
        \centering
        \includegraphics[width=1.0\textwidth]{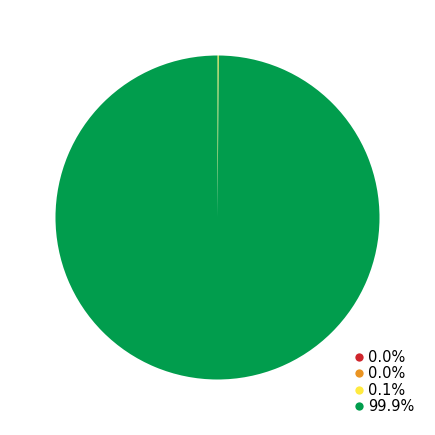}
        \caption{BiRefNet}
    \end{subfigure}
    \begin{subfigure}[t]{0.22\textwidth}
        \centering
         \includegraphics[width=1.0\textwidth]{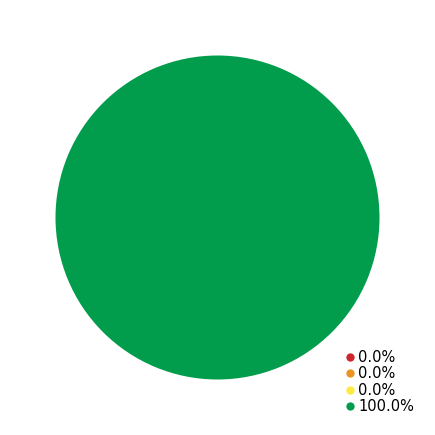}
        \caption{DEVA}
    \end{subfigure}
    \begin{subfigure}[t]{0.22\textwidth}
        \centering
         \includegraphics[width=1.0\textwidth]{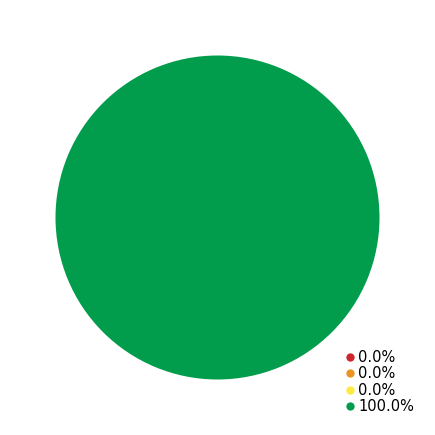}
        \caption{FoodMem}
    \end{subfigure}
    \begin{subfigure}[t]{0.22\textwidth}
        \centering
         \includegraphics[width=1.0\textwidth]{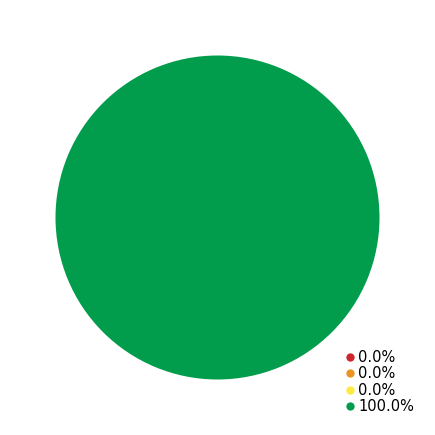}
        \caption{YOLO+XMem2}
    \end{subfigure}
    \caption{Distribution of camera views by method and mAP range for the Foodkit object, with red slices for $\le$50\%, orange for 50–75\%, yellow for 75–95\%, and green for $\geq$95\%, contrasting how reliably each method covers the object.}
    \label{fig:dist_comparasions}
\end{figure}

Two principal observations emerge from these experiments. First, hybrid pipelines that combine a strong single-frame segmenter with a temporal propagation module (e.g., XMem2) or with SAM-based refinement consistently improve mask completeness and temporal stability: these methods typically yield higher Recall and competitive or improved $\mathrm{mAP}$ relative to their single-frame counterparts. Examples include \texttt{SegMan+SAM2}, \texttt{SegMan+XMem2} and the replicated \texttt{SeTR-MLA+XMem2} entry, as shown in Table.~\ref{tab:3d_comparisons_tracker} and Table \ref{tab:3d_comparisons_tracker_2d}. Second, single-frame models vary substantially in their cross-partition generalisation: some models (e.g. \texttt{FoodSAM}, \texttt{SeTR-MLA}, \texttt{SWIN-Base}) achieve high $\mathrm{mAP}$ on N5k and MTF while degrading on V\&F or FoodKit, whereas other methods (e.g. \texttt{FoodLMM}) perform poorly across all partitions in our setup. Fig.~\ref{fig:3d_comparasions} and Fig.~\ref{fig:dist_comparasions} show examples of camera views and their distributions for all methods on the Foodkit object (donut).  Finally, there is an explicit trade-off between accuracy and computational cost: the most accurate configurations tend to incur large parameter counts, longer processing times, and greater memory consumption, as shown in Table~\ref{tab:results} and Fig.~\ref{fig:resources}.

For clarity and reproducibility, all numeric entries in Table~\ref{tab:results} are presented exactly as measured on the evaluation platform. Where measurements were not available, we indicate missing values with an em-dash (\textemdash). The remainder of this section summarizes major trends and highlights notable entries from the table. In addition, Table~\ref{tab:results_precision_f1_IoU} offers a complementary perspective by reporting Precision, F1, and IoU on the same four partitions, enabling a more fine-grained analysis of mask quality beyond $\mathrm{mAP}$ and Recall. 

To explicitly analyze temporal stability, we report continuity ($C_t$), flicker rate ($FR_{0.2}$), IoU drift ($\Delta$IoU), and IoU standard deviation ($\sigma$IoU) for all evaluated methods. As summarized in Table~\ref{tab:temporal_styled}, these metrics capture complementary aspects of temporal behavior that are not reflected by frame-wise accuracy alone. Table~\ref{tab:temporal_styled} reveals substantial differences in temporal stability between methods with comparable spatial accuracy. In particular, memory-augmented approaches consistently achieve higher continuity and lower flicker rates, indicating more stable mask propagation over time. In contrast, purely frame-based methods exhibit larger temporal fluctuations, as reflected by increased drift and standard deviation values. These results demonstrate that strong per-frame performance does not necessarily imply stable temporal behavior.

\begin{table*}[!htbp]
\centering
\scriptsize
\setlength{\tabcolsep}{3pt}
\caption{Comprehensive evaluation of all baselines on the BenchSeg benchmark.
Mean performance is reported in the first row of each method, with standard deviation shown in the second row.
Speed is measured per image; VRAM denotes peak GPU memory usage. Methods are sorted in ascending order according to Precision on FKit. Best, second-best, and third-best results are indicated by \textbf{bold}, \underline{underline}, and \textit{italic}, respectively.}
\label{tab:results}

\begin{tabular}{lccccccccccc}
\toprule
\textbf{Method} &
\multicolumn{4}{c}{\textbf{mAP (\%) ↑}} &
\multicolumn{4}{c}{\textbf{Recall (\%) ↑}} &
\multicolumn{3}{c}{\textbf{Efficiency }} \\
\cmidrule(lr){2-5} \cmidrule(lr){6-9} \cmidrule(lr){10-12}
$mean$±$std$ & FKit$\blacktriangledown$ & N5k & V\&F & MTF
 & FKit & N5k & V\&F & MTF
 & Params (M) & Speed & VRAM \\
\midrule

FLMM
 & 0.90 & 2.30 & 1.65 & 0.96
 & 1.10 & 2.10 & 1.53 & 1.37
 & 7706 & 5.5s & 13.8G \\
 & 0.15 & 0.15 & 0.12 & 0.09
 & 0.15 & 0.12 & 0.13 & 0.10
 & & & \\ 

\rowcolor{gray!12}
kMean++
 & 44.71 & 25.61 & 48.01 & 69.69
 & 57.72 & 47.29 & 57.01 & 64.78
 & -- & 15s & -- \\ \rowcolor{gray!12}
 & 39 & 19 & 34 & 21
 & 32 & 25 & 39 & 29
 & & & \\

YOLO
 & 56.91 & 76.66 & 59.68 & 69.45
 & 67.86 & 77.33 & 84.06 & 76.85
 & \textbf{10.1} & \textbf{2.9ms} & 19.6M \\
 & 43 & 25 & 38 & 39
 & 46 & 33 & 38 & 41
 & & & \\

\rowcolor{gray!12} 
Y+S2
 & 57.36 & 79.10 & \underline{90.10} & 67.63
 & 59.78 & 82.90 & \textbf{96.39} & 61.77
 & 234.6 & \uline{0.34s} & 875.7M \\ \rowcolor{gray!12}
 & 44 & 29 & 20 & 38
 & 0.47 & 36 & 11 & 40
 & & & \\

SeTN
 & 73.47 & 92.49 & 78.06 & 86.75
 & 77.78 & 96.47 & 95.27 & 93.04
 & 94.8 & 2.28s & 722.8M \\
 & 30 & 3 & 25 & 16
 & 30 & 4 & 14 & 12
 & & & \\

\rowcolor{gray!12}
CCNet
 & 75.02 & 91.81 & 74.61 & 83.97
 & 74.58 & 96.73 & 90.76 & 90.93
 & 49.9 & 4.56s & 381M \\ \rowcolor{gray!12}
 & 31 & 4 & 30 & 24
 & 32 & 4 & 27 & 25
 & & & \\

FPN-Re
 & 75.79 & 83.61 & 70.48 & 58.30
 & 84.67 & 96.95 & 94.09 & 90.03
 & \uline{28.5} & 5.21s &\textbf{217.8M} \\
 & 23 & 11 & 28 & 30
 & 21 & 4 & 17 & 22
 & & & \\

\rowcolor{gray!12}
Y+X2
 & 75.84 & 84.28 & \textit{91.02} & 74.82
 & 78.14 & 79.82 & \underline{96.58} & 92.26
 & 72.3 & 0.51s & \textit{256.7M} \\ \rowcolor{gray!12}
 & 41 & 28 & 18 & 40
 & 42 & 34 & 18 & 19
 & & & \\

Swin-S
 & 76.17 & 92.40 & 76.27 & 90.30
 & 81.71 & \textit{98.86} & 96.21 & 94.72
 & 81.3 & 6.70s & 930.4M \\
 & 29 & 3 & 26 & 12
 & 28 & 1 & 12 & 10
 & & & \\

\rowcolor{gray!12}
CCNet-Re
 & 79.17 & 92.20 & 77.67 & 86.21
 & 85.99 & 95.68 & 93.65 & 91.82
 & \textit{49.9} & 5.49s & 380M \\ \rowcolor{gray!12}
 & 26 & 4 & 26 & 22
 & 24 & 4 & 17 & 21
 & & & \\

SegMan
 & 80.57 & \underline{93.63} & 87.80 & 91.01
 & 82.91 & 98.36 & 96.00 & 95.00
 & 51.8 & \underline{0.30s} & 594.3M \\
 & 28 & 2 & 21 & 12
 & 28 & 1 & 12 & 11
 & & & \\

\rowcolor{gray!12}
FSAM
 & 88.49 & 91.92 & 90.14 & 93.48
 & 93.66 & 95.60 & 94.41 & \textbf{96.43}
 & 636 & 22m33s & 12.5G \\ \rowcolor{gray!12}
 & 28.49 & 4.74 & 18.81 & 5.13
 & 30.13 & 12.07 & 11.35 & 8.07
 & & & \\

Seg+S2
 & 85.34 & \textbf{94.33} & \underline{91.15} & 93.01
 & 83.54 & 98.10 & 96.35 & 95.60
 & 224.5 & 1.66s & 856M \\
 & 34 & 4 & 20 & 10
 & 30 & 6 & 11 & 10
 & & & \\

\rowcolor{gray!12}
SeTM
 & 87.24 & 93.29 & 84.79 & 92.28
 & 87.40 & 97.82 & 95.71 & 95.81
 & 93.2 & 3s & 2.32G \\ \rowcolor{gray!12}
 & 19 & 2 & 22 & 9
 & 20 & 2 & 11 & 9
 & & & \\

Seg+X2
 & 89.29 & 93.57 & 90.98 & \underline{94.99}
 & 87.34 & \underline{98.87} & \textbf{96.69} & 97.11
 & 62.2 & \textit{0.51s} & \underline{237M} \\
 & 28 & 3 & 18 & 4
 & 35 & 2 & 11 & 7
 & & & \\

\rowcolor{gray!12}
FoodMem
 & 92.34 & 90.98 & \textbf{94.99} & \textit{94.53}
 & 98.06 & 77.08 & 94.69 & \underline{99.41}
 & 785 & 25s & 6.25G \\ \rowcolor{gray!12}
 & 25 & 2 & 18 & 3
 & 29 & 1 & 11 & 15
 & & & \\

SeTM+X2
 & 92.34 & \textit{93.61} & 90.58 & \textbf{96.32}
 & 98.06 & \textbf{99.06} & 96.28 & \textit{98.06}
 & 785 & 3.51s & 6.25G \\
 & 25 & 2 & 18 & 3
 & 29 & 1 & 11 & 5
 & & & \\

\rowcolor{gray!12}
Swin-B
 & \textit{92.81} & 92.65 & 82.09 & 88.95
 & 89.22 & 98.44 & 96.01 & 94.65
 & 121.3 & 6.53s & 1.36G \\ \rowcolor{gray!12}
 & 18 & 5 & 23 & 14
 & 19 & 1 & 11 & 10
 & & & \\

BiRefNet
 & \underline{97.17} & 94.34 & 57.88 & 60.00
 & \textbf{98.68} & 98.54 & 95.75 & 98.50
 & 220.2 & 0.16s & 444M \\
 & 5 & 2 & 38 & 37
 & 2 & 3 & 20 & 2
 & & & \\

\rowcolor{gray!12}
DEVA
 & \textbf{97.83} & 88.25 & 85.48 & 27.37
 & \textit{98.12} & 73.01 & 93.28 & 95.11
 & 241 & 25s & 9.0G \\ \rowcolor{gray!12}
 & 4.31 & 26.54 & 30.86 & 7.76
 & 2.25 & 12.07 & 14.94 & 3.23
 & & & \\

\bottomrule
\end{tabular}

\vspace{2mm}
\scriptsize
\raggedright
Results are reported as mean ± std over scenes. 

\textbf{Abbreviations:}
Y+X2: YOLO+XMem2; Seg+X2: SegMan+XMem2;  
FSAM: FoodSAM; FLMM: FoodLMM;  
CCNet-Re: CCNet-Relem;  
Swin-S/B: Swin Small/Base;  
SeTM/SeTN: SeTR-MLA / Naive;  
S2: SAM2; X2: XMem2.

\textbf{Note:} DEVA is evaluated using a text prompt \emph{``food''} for segmentation.
\end{table*}

\begin{table}[!htbp]
\centering
\tiny
\setlength{\tabcolsep}{1.3pt} % Adjust column spacing
\caption{Evaluation on the BenchSeg benchmark using Precision, F1-score, Accuracy, and IoU.
Methods are sorted in ascending order according to Precision on FKit. 
For each metric, the best, second-best, and third-best results are highlighted in \textbf{bold}, \uline{underline}, and \textit{italics}, respectively.}
\label{tab:results_precision_f1_IoU}
\begin{tabular}{lcccccccccccccccc}
\toprule
\textbf{Method} & 
\multicolumn{4}{c}{\textbf{Precision (\%) $\uparrow$}} & 
\multicolumn{4}{c}{\textbf{F1 (\%) $\uparrow$}} & 
\multicolumn{4}{c}{\textbf{IoU (\%) $\uparrow$}} & 
\multicolumn{4}{c}{\textbf{Accuracy (\%) $\uparrow$}} \\
\cmidrule(lr){2-5} \cmidrule(lr){6-9} \cmidrule(lr){10-13} \cmidrule(lr){14-17}
 $mean±std$ & FKit$\blacktriangle$ & MTF & N5k & V\&F & FKit & MTF & N5k & V\&F & FKit & MTF & N5k & V\&F & FKit & MTF & N5k & V\&F \\
\midrule
FLMM & 0.97 & 0.96 & 2.30 & 1.65 & 1.05 & 0.98 & 1.84 & 1.71 & 2.76 & 0.94 & 1.61 & 1.58 & 97.58 & 95.69 & 81.10 & 94.98 \\
     & 9.01 & 9.48 & 14.63 & 12.05 & 0.00 & 9.64 & 12.18 & 12.33 & 9.03 & 9.32 & 11.09 & 11.57 & 1.39 & 2.35 & 8.02 & 4.20 \\ \addlinespace[3pt]

\rowcolor{gray!12}
KMeans++ & 44.71 & 69.69 & 25.61 & 48.01 & 44.24 & 64.63 & 31.21 & 49.50 & 35.22 & 51.71 & 20.40 & 39.73 & 94.60 & 97.09 & 64.04 & 95.14 \\ \rowcolor{gray!12}
          & 38.92 & 20.92 & 19.23 & 33.95 & 33.39 & 22.48 & 20.21 & 33.87 & 32.23 & 24.22 & 16.24 & 30.83 & 6.04 & 2.40 & 11.27 & 5.32 \\ \addlinespace[3pt]  
          
YOLO & 59.78 & 70.98 & 86.43 & 60.39 & 59.69 & 71.42 & 79.06 & 65.40 & 56.24 & 68.51 & 72.73 & 59.25 & 95.53 & 96.64 & 94.36 & 97.00 \\
     & 44.48 & 41.53 & 22.84 & 38.53 & 44.47 & 41.21 & 29.32 & 37.92 & 43.43 & 40.60 & 30.63 & 38.29 & 16.05 & 12.30 & 6.77 & 3.96 \\\addlinespace[3pt]

\rowcolor{gray!12}
Y+S2 & 60.29 & 72.24 & 84.87 & 91.50 & 58.93 & 69.61 & 80.12 & 93.23 & 56.44 & 66.07 & 75.99 & 90.21 & 98.96 & 81.16 & 95.74 & 99.20 \\\rowcolor{gray!12}
     & 46.77 & 42.90 & 28.59 & 17.59 & 46.19 & 41.24 & 32.72 & 16.25 & 44.94 & 40.33 & 33.79 & 19.11 & 1.38 & 33.64 & 6.24 & 2.05 \\\addlinespace[3pt]
     
Y+X2 & 75.84 & 74.82 & 84.28 & 91.02 & 76.12 & 74.49 & 78.27 & 92.94 & 75.20 & 70.10 & 72.73 & 89.72 & 99.43 & 85.86 & 95.05 & 99.18 \\
     & 41.45 & 39.83 & 28.40 & 17.76 & 41.43 & 35.79 & 31.89 & 16.41 & 41.19 & 37.88 & 32.50 & 19.10 & 1.13 & 25.76 & 5.96 & 2.05 \\\addlinespace[3pt]

\rowcolor{gray!12}
Seg+X2 & 85.44 & \uline{98.61} & 94.81 & \uline{92.60} & 84.99 & \uline{97.31} & 96.62 & \uline{93.71} & 84.24 & \uline{95.11} & 93.49 & \uline{91.17} & 99.48 & \uline{99.83} & 98.59 & \textbf{99.27} \\ \rowcolor{gray!12}
        & 34.24 & 4.09 & 2.77 & 18.08 & 34.62 & 4.60 & 1.23 & 16.58 & 34.41 & 7.87 & 2.27 & 19.44 & 1.27 & 0.25 & 0.89 & 2.04 \\\addlinespace[3pt]

FSAM & 88.49 & 81.66 & 91.87 & 90.14 & 85.63 & 81.10 & 95.25 & 87.42 & 82.64 & 76.30 & 91.05 & 88.49 & 99.48 & 98.29 & 80.72 & 94.93 \\
     & 28.49 & 28.95 & 4.74 & 18.81 & 29.61 & 30.31 & 2.81 & 11.35 & 30.28 & 32.09 & 4.80 & 19.93 & 1.47 & 3.04 & 8.01 & 4.19 \\\addlinespace[3pt]

\rowcolor{gray!12}
FPN-Re & 88.82 & 64.24 & 85.77 & 75.00 & 83.89 & 69.27 & 90.49 & 79.36 & 76.07 & 59.08 & 83.32 & 71.01 & 99.27 & 96.70 & 96.39 & 98.17 \\ \rowcolor{gray!12}
       & 19.02 & 30.46 & 11.16 & 27.07 & 19.52 & 27.10 & 7.06 & 23.53 & 23.05 & 29.76 & 10.66 & 27.19 & 1.13 & 3.61 & 2.58 & 2.38 \\\addlinespace[3pt]

SeTN & 89.36 & 95.06 & 95.18 & 82.32 & 79.62 & 91.75 & 95.76 & 85.06 & 73.23 & 86.57 & 91.92 & 78.30 & 99.20 & 99.32 & 98.29 & 98.51  \\
      &  26.00 & 13.60 & 2.81 & 24.04 & 28.79 & 12.50 & 1.80 & 20.52 & 30.07 & 15.92 & 3.24 & 24.35 & 2.09 & 2.34 & 1.07 & 2.62 \\\addlinespace[3pt]

\rowcolor{gray!12}
Swin-S & 89.42 & 96.41 & 93.27 & 78.45 & 81.97 & 94.35 & 95.94 & 83.58 & 76.02 & 90.23 & 92.24 & 76.54 & 99.22 & 99.63 & 98.36 & 98.33 \\ \rowcolor{gray!12}
      & 23.13 & 8.32 & 3.06 & 25.63 & 27.03 & 8.63 & 1.52 & 21.55 & 29.06 & 11.67 & 2.74 & 25.91 & 1.39 & 0.49 & 0.88 & 2.55 \\ \addlinespace[3pt]

CCNet-Re & 89.48 & 95.64 & \textit{95.31} & 83.60 & 84.95 & 90.42 & 95.52 & 84.61 & 79.12 & 86.12 & 91.51 & 77.97 & 99.29 & 99.51 & 98.20 & 98.65 \\
         & 21.87 & 11.37 & 2.64 & 24.10 & 23.67 & 18.45 & 2.38 & 21.58 & 26.09 & 21.68 & 4.21 & 25.16 & 1.59 & 0.78 & 1.26 & 2.23 \\\addlinespace[3pt]

\rowcolor{gray!12}
Seg+S2 & 89.59 & \textit{98.21} & \uline{95.63} & \textbf{92.74} & 87.43 & 95.98 & \uline{96.82} & \textbf{93.78} & 84.96 & \textit{92.86} & \uline{94.02} & \textbf{91.25} & 99.57 & \textit{99.75} & 98.65 & \uline{99.26} \\ \rowcolor{gray!12}
        & 27.62 & 3.95 & 2.59 & 17.84 & 28.66 & 6.09 & 3.51 & 16.37 & 28.85 & 9.92 & 5.49 & 19.36 & 1.02 & 0.37 & 1.72 & 2.04 \\\addlinespace[3pt]

SegMan & 90.73 & 96.99 & 94.97 & 90.45 & 84.81 & 94.69 & 96.56 & 91.76 & 80.34 & 90.87 & 93.37 & 87.92 & 99.41 & 99.68 & \uline{98.65} & 99.15 \\
       & 22.27 & 6.77 & 2.59 & 18.77 & 27.27 & 8.76 & 1.23 & 17.04 & 28.70 & 11.81 & 2.25 & 20.17 & 1.21 & 0.35 & 0.71 & 2.04 \\\addlinespace[3pt]

\rowcolor{gray!12}
CCNet & 91.99 & 95.11 & 93.94 & 85.65 & 79.95 & 88.14 & 95.42 & 81.03 & 74.52 & 83.71 & 91.33 & 74.69 & 99.33 & 99.48 & 98.12 & 98.74 \\ \rowcolor{gray!12}
      & 24.07 & 15.45 & 2.56 & 25.72 & 30.38 & 22.54 & 2.35 & 27.05 & 31.71 & 24.76 & 4.17 & 29.34 & 1.10 & 1.00 & 1.36 & 2.07 \\\addlinespace[3pt]

Swin-B & 92.81 & 94.89 & 93.82 & 84.44 & 89.31 & 93.43 & 96.01 & 88.00 & 84.31 & 88.87 & 92.43 & 82.30 & 99.53 & 99.54 & 98.41 & 98.77 \\
     & 17.58 & 11.19 & 4.23 & 21.88 & 19.07 & 9.66 & 2.52 & 18.71 & 21.37 & 13.54 & 4.06 & 22.54 & 0.95 & 0.79 & 0.99 & 2.25 \\\addlinespace[3pt]

\rowcolor{gray!12}
SeTM & 97.09 & 97.71 & 94.80 & 87.61 & 91.19 & \textit{95.64} & 96.33 & 89.84 & 87.00 & 92.17 & 92.94 & 84.96 & 99.62 & 99.73 & 98.53 & 98.97 \\ \rowcolor{gray!12}
      & 9.83 & 4.40 & 2.27 & 20.67 & 17.60 & 6.08 & 1.20 & 17.83 & 20.02 & 9.12 & 2.21 & 21.28 & 0.97 & 0.30 & 0.82 & 2.14 \\\addlinespace[3pt] 

SeTM+X2 & \textit{97.80} & \textbf{98.77} & 94.53 & \textit{92.59} & \textit{93.79} & \textbf{98.04} & \textit{96.68} & \textit{93.51} & \textit{92.10} & \textbf{96.31} & \textit{93.60} & 90.78 & \textit{99.75} & \textbf{99.87} & \textit{98.61} & \textit{99.24} \\
          & 13.15 & 3.31 & 2.34 & 17.87 & 19.26 & 3.04 & 1.29 & 16.48 & 21.14 & 5.41 & 2.37 & 19.37 & 0.82 & 0.21 & 0.90 & 2.04 \\\addlinespace[3pt] 

\rowcolor{gray!12}
DEVA & \uline{98.64} & 87.50 & 71.73 & 80.73 & \textbf{98.63} & 90.32 & 78.69 & 83.50 & \textbf{97.47} & 85.92 & 69.60 & 79.04 & \uline{99.93} & 99.09 & 87.30 & 97.66 \\ \rowcolor{gray!12}
     & 4.31 & 22.86 & 26.54 & 30.86 & 3.68 & 17.54 & 21.88 & 27.48 & 4.79 & 22.49 & 26.55 & 31.45 & 0.30 & 2.50 & 14.02 & 4.46 \\\addlinespace[3pt]

BiRefNet & \textbf{98.44} & 60.93 & \textbf{95.83} & 58.95 & \uline{98.47} & 67.78 & \textbf{96.95} & 65.04 & \uline{97.17} & 60.08 & \textbf{94.10} & 58.01 & \textbf{99.92} & 95.05 & \textbf{98.78} & 94.38 \\
        & 4.50 & 37.38 & 2.31 & 38.30 & 3.96 & 31.93 & 1.04 & 35.15 & 4.98 & 36.69 & 1.95 & 38.14 & 0.44 & 5.39 & 0.70 & 8.26 \\ 
\bottomrule
\end{tabular}

\vspace{2pt}
\raggedright
\tiny
Results are reported as mean ± std over scenes. 

\textbf{Abbreviations:}
Y+X2: YOLO+XMem2; Seg+X2: SegMan+XMem2; FSAM: FoodSAM; FLMM: FoodLMM;
CCNet-Re: CCNet-Relem; Swin-S/B: Swin Small/Base; SeTM/SeTN: SeTR-MLA/Naive;
S2: SAM2; X2: XMem2.

\vspace{2pt}
\textbf{Note:} DEVA is evaluated using a text prompt \emph{``food''} for segmentation.
\end{table}

\begin{table}[!htb]
\centering
\tiny
\setlength{\tabcolsep}{1pt} % Adjust column spacing
\caption{Temporal metrics across datasets and global averages. Styled for top 3 ranking per column. Higher is better (↑) for continuity; lower is better (↓) for flicker rate, IoU drift, and standard deviation.}
\label{tab:temporal_styled}
\begin{tabular}{lcccccccccccccccccccc}
\toprule
\textbf{Method} & \multicolumn{4}{c}{$C_{t}$(\%)$\uparrow$} & \multicolumn{4}{c}{$FR_{0.2}$(\%)$\downarrow$} & \multicolumn{4}{c}{$\Delta IoU$(\%)$\downarrow$} & \multicolumn{4}{c}{$\sigma IoU$(\%)$\downarrow$} & \multicolumn{4}{c}{Global$(\%)$} \\
\cmidrule(lr){2-5} \cmidrule(lr){6-9} \cmidrule(lr){10-13} \cmidrule(lr){14-17} \cmidrule(lr){18-21}
mean±std & FKIT$\blacktriangle$ & MTF & N5K & V\&F & FKIT & MTF & N5K & V\&F & FKIT & MTF & N5K & V\&F & FKIT & MTF & N5K & V\&F & $C_{t}$ & $FR$ & $\Delta IoU$ & $\sigma IoU$ \\
\midrule
FLMM & \textbf{100} & 98.1 & 97.5 & 97.1 & \textbf{0.00} & 2.10 & 3.90 & 3.50 & \textbf{0.00} & 1.90 & 2.50 & 2.90 & \textbf{0.00} & 9.30 & 11.1 & \textit{11.6} & 98.2 & 2.30 & 1.80 & \textbf{8.00} \\
& 0.00 & 13.1 & 13.6 & 15.5 & 0.00 & 0.00 & 0.00 & 0.00 & 0.00 & 13.1 & 13.6 & 15.5 & 0.00 & 9.30 & 11.1 & 11.6 & 1.30 & 1.70 & 1.30 & 5.40 \\
\addlinespace
\rowcolor{gray!12}Y+X2 & \underline{99.8} & 99.2 & \textit{98.7} & 98.4 & \textit{0.10} & 0.50 & \textit{0.50} & \textit{1.20} & \underline{0.20} & 0.80 & \textit{1.30} & 1.60 & 41.2 & 37.9 & 32.5 & 19.1 & 99.0 & \underline{0.60} & \textit{1.00} & 32.7 \\
& 1.80 & 5.60 & 4.80 & 5.10 & 0.00 & 0.00 & 0.00 & 0.00 & 1.80 & 5.60 & 4.80 & 5.10 & 41.2 & 37.9 & 32.5 & 19.1 & 0.60 & 0.50 & 0.60 & 9.70 \\
\addlinespace
FoodMem & \textit{99.7} & \textbf{99.5} & \underline{99.0} & \textit{98.5} & 0.20 & \textbf{0.10} & \textbf{0.00} & \underline{1.20} & \textit{0.30} & \textbf{0.50} & \textbf{1.00} & \textit{1.50} & 29.0 & \textbf{5.40} & \underline{2.40} & \underline{19.4} & \textbf{99.2} & \textit{0.40} & \textbf{0.80} & \textit{14.0} \\
& 2.40 & 1.50 & 1.00 & 5.10 & 0.00 & 0.00 & 0.00 & 0.00 & 2.40 & 1.50 & 1.00 & 5.10 & 29.0 & 5.40 & 2.40 & 19.4 & 0.50 & 0.50 & 0.50 & 12.4 \\
\addlinespace
\rowcolor{gray!12}S+X2 & \textit{99.7} & \textbf{99.5} & \underline{99.0} & \underline{98.6} & \textbf{0.00} & \textbf{0.10} & \textbf{0.00} & \underline{1.20} & \textit{0.30} & \textbf{0.50} & \textbf{1.00} & \underline{1.40} & 34.4 & 7.90 & \textit{2.30} & \underline{19.4} & \textbf{99.2} & \textbf{0.30} & \textbf{0.80} & 16.0 \\
& 1.70 & 1.20 & 1.10 & 5.20 & 0.00 & 0.00 & 0.00 & 0.00 & 1.70 & 1.20 & 1.10 & 5.20 & 34.4 & 7.90 & 2.30 & 19.4 & 0.50 & 0.60 & 0.50 & 14.2 \\
\addlinespace
Y+S2 & \textit{99.7} & 99.0 & 98.5 & \textit{98.5} & \textit{0.10} & 0.90 & 1.10 & \textit{1.10} & \textit{0.30} & 1.00 & 1.50 & \textit{1.50} & 44.9 & 40.3 & 33.8 & \textbf{19.1} & 98.9 & 0.80 & 1.10 & 34.5 \\
& 2.40 & 6.10 & 5.30 & 4.20 & 0.00 & 0.00 & 0.00 & 0.00 & 2.40 & 6.10 & 5.30 & 4.20 & 44.9 & 40.3 & 33.8 & 19.1 & 0.50 & 0.50 & 0.50 & 11.3 \\
\addlinespace
\rowcolor{gray!12}DEVA & 99.6 & 98.8 & 96.8 & 97.3 & \textit{0.10} & 0.80 & 3.50 & 2.50 & 0.40 & 1.20 & 3.20 & 2.70 & \textbf{4.80} & 22.5 & 26.6 & 31.4 & 98.1 & 1.70 & 1.90 & 21.3 \\
& 1.80 & 6.90 & 9.80 & 8.90 & 0.00 & 0.00 & 0.00 & 0.00 & 1.80 & 6.90 & 9.80 & 8.90 & 4.80 & 22.5 & 26.6 & 31.4 & 1.30 & 1.60 & 1.30 & 11.6 \\
\addlinespace
BiRefNet & 99.5 & 94.0 & \textbf{99.1} & 92.5 & \underline{0.20} & 10.8 & \textbf{0.00} & 12.5 & 0.50 & 6.00 & \underline{0.90} & 7.50 & \underline{5.00} & 36.7 & \textbf{1.90} & 38.1 & 96.3 & 5.90 & 3.70 & 20.4 \\
& 2.90 & 14.6 & 1.00 & 15.1 & 0.00 & 0.00 & 0.00 & 0.00 & 2.90 & 14.6 & 1.00 & 15.1 & 5.00 & 36.7 & 1.90 & 38.1 & 3.60 & 6.70 & 3.60 & 19.6 \\
\addlinespace
\rowcolor{gray!12}Seg+S2 & 99.4 & \underline{99.3} & \underline{98.8} & \textbf{98.7} & 0.50 & \underline{0.40} & 0.80 & \textbf{1.10} & 0.60 & \underline{0.70} & 1.20 & \textbf{1.30} & 28.9 & \underline{9.90} & 5.50 & \underline{19.4} & \underline{99.1} & \textit{0.70} & \underline{0.90} & \underline{15.9} \\
& 2.60 & 2.60 & 2.90 & 4.30 & 0.00 & 0.00 & 0.00 & 0.00 & 2.60 & 2.60 & 2.90 & 4.30 & 28.9 & 9.90 & 5.50 & 19.4 & 0.40 & 0.30 & 0.40 & 10.4 \\
\addlinespace
kMean++ & 97.8 & 95.7 & 97.0 & 93.9 & 1.80 & 1.70 & 2.40 & 6.00 & 2.20 & 4.30 & 3.00 & 6.10 & 32.2 & 24.2 & 16.2 & 30.8 & 96.1 & 3.00 & 3.90 & 25.9 \\
& 6.60 & 5.50 & 5.40 & 9.40 & 0.00 & 0.00 & 0.00 & 0.00 & 6.60 & 5.50 & 5.40 & 9.40 & 32.2 & 24.2 & 16.2 & 30.8 & 1.70 & 2.10 & 1.70 & 7.30 \\
\addlinespace
\rowcolor{gray!12}SegMan & 94.9 & 97.1 & \underline{98.8} & 96.0 & 7.30 & 3.10 & \textbf{0.00} & 5.20 & 5.10 & 2.90 & \underline{1.20} & 4.00 & 28.7 & 11.8 & \textit{2.30} & 20.2 & \textit{96.7} & 3.90 & 3.30 & 15.7 \\
& 10.3 & 6.70 & 1.70 & 9.20 & 0.00 & 0.00 & 0.00 & 0.00 & 10.3 & 6.70 & 1.70 & 9.20 & 28.7 & 11.8 & 2.30 & 20.2 & 1.70 & 3.10 & 1.70 & 11.3 \\
\addlinespace
CCNet & 94.8 & 94.9 & 98.6 & 90.8 & 7.10 & 7.10 & \textbf{0.00} & 14.3 & 5.20 & 5.10 & 1.40 & 9.20 & 31.7 & 24.8 & 4.20 & 29.3 & 94.8 & 7.10 & 5.20 & 22.5 \\
& 10.1 & 11.6 & 1.70 & 15.5 & 0.00 & 0.00 & 0.00 & 0.00 & 10.1 & 11.6 & 1.70 & 15.5 & 31.7 & 24.8 & 4.20 & 29.3 & 3.20 & 5.80 & 3.20 & 12.6 \\
\addlinespace
\rowcolor{gray!12}Swin-B & 94.4 & 95.7 & 98.1 & 92.5 & 8.70 & 6.60 & \underline{0.50} & 11.2 & 5.60 & 4.30 & 1.90 & 7.50 & 21.4 & 13.5 & 4.10 & 22.5 & 95.2 & 6.80 & 4.80 & 15.4 \\
& 11.9 & 10.3 & 3.60 & 12.4 & 0.00 & 0.00 & 0.00 & 0.00 & 11.9 & 10.3 & 3.60 & 12.4 & 21.4 & 13.5 & 4.10 & 22.5 & 2.40 & 4.60 & 2.40 & 8.50 \\
\addlinespace
CCNet-Re & 94.3 & 96.9 & 98.5 & 92.7 & 7.70 & \underline{2.90} & \textbf{0.00} & 10.8 & 5.70 & \textit{3.10} & \textit{1.50} & 7.30 & 26.1 & 21.7 & 4.20 & 25.2 & 95.6 & 5.40 & 4.40 & 19.3 \\
& 10.9 & 7.00 & 2.00 & 11.6 & 0.00 & 0.00 & 0.00 & 0.00 & 10.9 & 7.00 & 2.00 & 11.6 & 26.1 & 21.7 & 4.20 & 25.2 & 2.60 & 4.80 & 2.60 & 10.2 \\
\addlinespace
\rowcolor{gray!12}FSAM & 94.2 & \textit{97.3} & 98.5 & 96.6 & 7.90 & 2.50 & \textbf{0.00} & 3.30 & 5.80 & \underline{2.70} & \textit{1.50} & 3.40 & 30.3 & 32.1 & 4.80 & 19.9 & \textit{96.7} & \textit{3.40} & \textit{3.30} & 21.8 \\
& 16.1 & 7.30 & 2.00 & 9.40 & 0.00 & 0.00 & 0.00 & 0.00 & 16.1 & 7.30 & 2.00 & 9.40 & 30.3 & 32.1 & 4.80 & 19.9 & 1.80 & 3.30 & 1.80 & 12.5 \\
\addlinespace
SeTN & 93.7 & 95.9 & \underline{98.8} & 90.9 & 8.60 & \underline{2.90} & \textbf{0.00} & 15.5 & 6.30 & 4.10 & \underline{1.20} & 9.10 & 30.1 & 15.9 & 3.20 & 24.3 & 94.8 & 6.70 & 5.20 & 18.4 \\
& 10.6 & 8.30 & 1.40 & 14.1 & 0.00 & 0.00 & 0.00 & 0.00 & 10.6 & 8.30 & 1.40 & 14.1 & 30.1 & 15.9 & 3.20 & 24.3 & 3.30 & 6.90 & 3.30 & 11.7 \\
\addlinespace
\rowcolor{gray!12}FPN-Re & 92.5 & 88.5 & 94.5 & 91.3 & 10.4 & 19.4 & 3.90 & 12.3 & 7.50 & 11.5 & 5.50 & 8.70 & 23.1 & 29.8 & 10.7 & 27.2 & 91.7 & 11.5 & 8.30 & 22.7 \\
& 10.1 & 12.1 & 6.10 & 11.5 & 0.00 & 0.00 & 0.00 & 0.00 & 10.1 & 12.1 & 6.10 & 11.5 & 23.1 & 29.8 & 10.7 & 27.2 & 2.50 & 6.40 & 2.50 & 8.50 \\
\addlinespace
Swin-S & 92.5 & 96.2 & 98.6 & 91.7 & 12.1 & 4.90 & \textbf{0.00} & 13.1 & 7.50 & 3.80 & 1.40 & 8.30 & 29.1 & 11.7 & \textit{2.70} & 25.9 & 94.8 & 7.50 & 5.20 & 17.3 \\
& 13.6 & 10.0 & 2.20 & 12.7 & 0.00 & 0.00 & 0.00 & 0.00 & 13.6 & 10.0 & 2.20 & 12.7 & 29.1 & 11.7 & 2.70 & 25.9 & 3.20 & 6.20 & 3.20 & 12.3 \\
\addlinespace
\rowcolor{gray!12}YOLO & 86.0 & 80.0 & 88.9 & 80.2 & 17.6 & 23.5 & 14.7 & 29.6 & 14.0 & 20.0 & 11.1 & 19.8 & 43.4 & 40.6 & 30.6 & 38.3 & 83.8 & 21.3 & 16.2 & 38.2 \\
& 30.2 & 35.9 & 23.7 & 30.2 & 0.00 & 0.00 & 0.00 & 0.00 & 30.2 & 35.9 & 23.7 & 30.2 & 43.4 & 40.6 & 30.6 & 38.3 & 4.40 & 6.60 & 4.40 & 5.50 \\
\bottomrule
\end{tabular}
\raggedright
\tiny

\vspace{0.5em}
Results are reported as mean ± std over scenes. All temporal metrics are computed at the sequence level by evaluating consecutive frame pairs and then aggregated across the dataset. Flicker and continuity are evaluated locally between adjacent frames, whereas drift and standard deviation are computed over full sequences. We report mean and standard deviation to reflect both average behavior and variability.

\textbf{Abbreviations:}
Y+X2: YOLO+XMem2; Seg+X2: SegMan+XMem2; FSAM: FoodSAM; FLMM: FoodLMM;
CCNet-Re: CCNet-Relem; Swin-S/B: Swin Small/Base; SeTM/SeTN: SeTR-MLA/Naive;
S2: SAM2; X2: XMem2.

\vspace{2pt}
\textbf{Note:} DEVA is evaluated using a text prompt \emph{``food''} for segmentation.
\end{table}

%%%%%%%%%%%%%%%%%%%%%%%%%%%%%%%%%%%%%%%%%%%%%%%%%%%%%%%%%%%%%%%%%%%
%%%%%%%%%%%%%%%%%%%%%%%%%%%%%%%%%%%%%%%%%%%%%%%%%%%%%%%%%%%%%%%%%%%

\begin{table}[htb]
    \centering
    \caption{Qualitative comparison of segmentation outputs across all evaluated configurations, including YOLO, SegMan, YOLO+SAM2, YOLO+XMem2, SegMan+SAM2, SegMan+XMem2, the ground-truth masks, and the corresponding RGB frames. The examples illustrate differences in boundary precision, temporal consistency, and robustness under challenging food scenes. Additional qualitative results are provided in the Appendix.}
    \setlength{\tabcolsep}{1pt}
    \begin{NiceTabular}{c|ccccc}
    \hline
    Method & Segmentor & +SAM2 & +XMem2 & GT & RGB \\
    \hline
    
\Block[v-center]{}{YOLO} &
\includegraphics[width=0.17\textwidth]{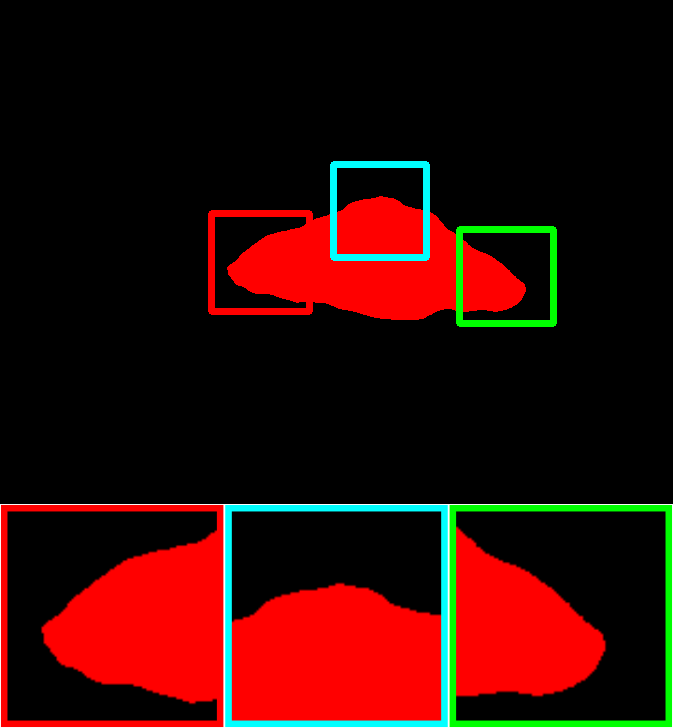} &
\includegraphics[width=0.17\textwidth]{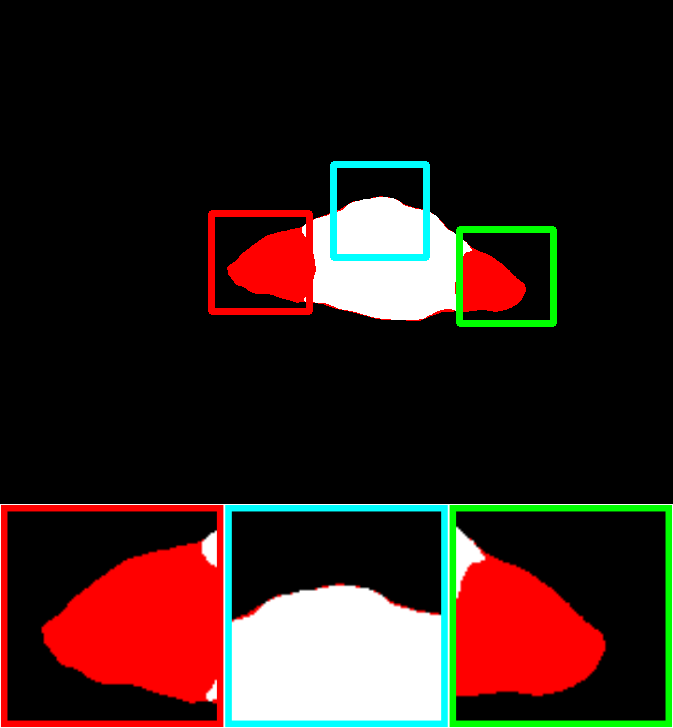} &
\includegraphics[width=0.17\textwidth]{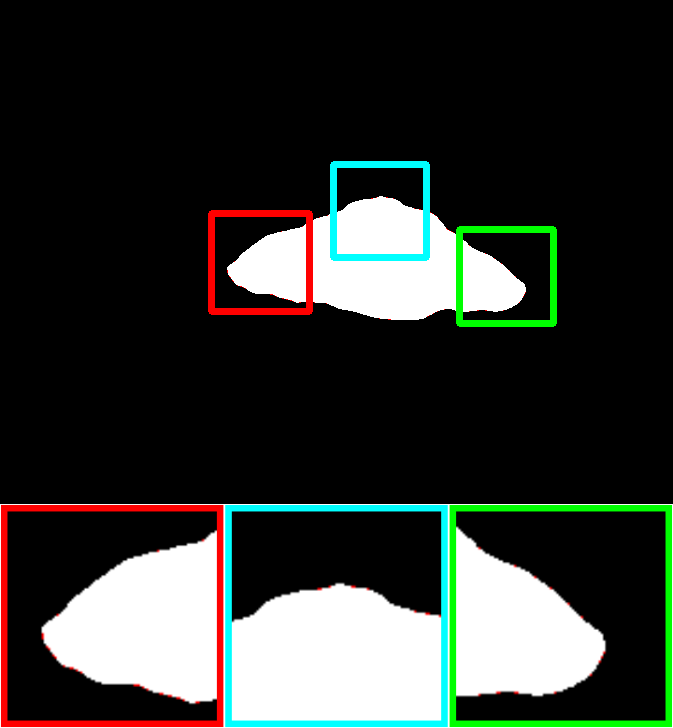} &
\includegraphics[width=0.17\textwidth]{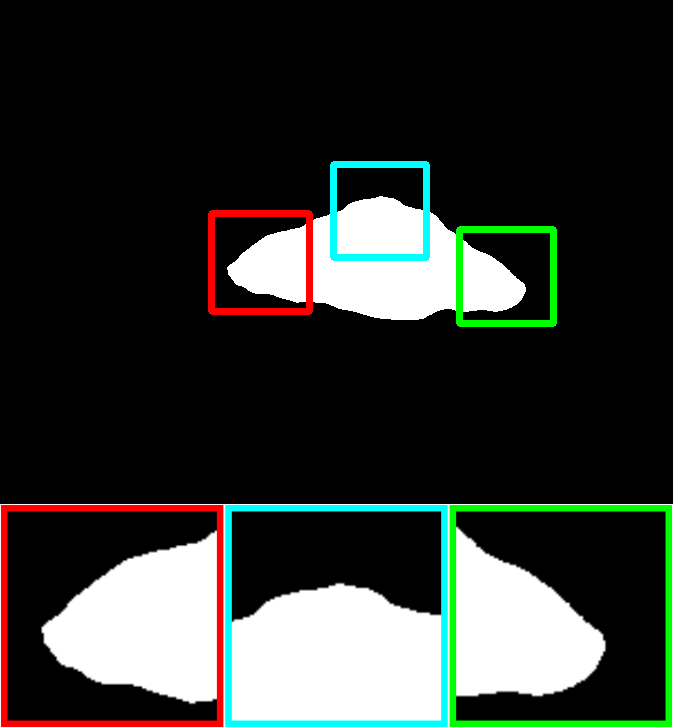} &
\includegraphics[width=0.17\textwidth]{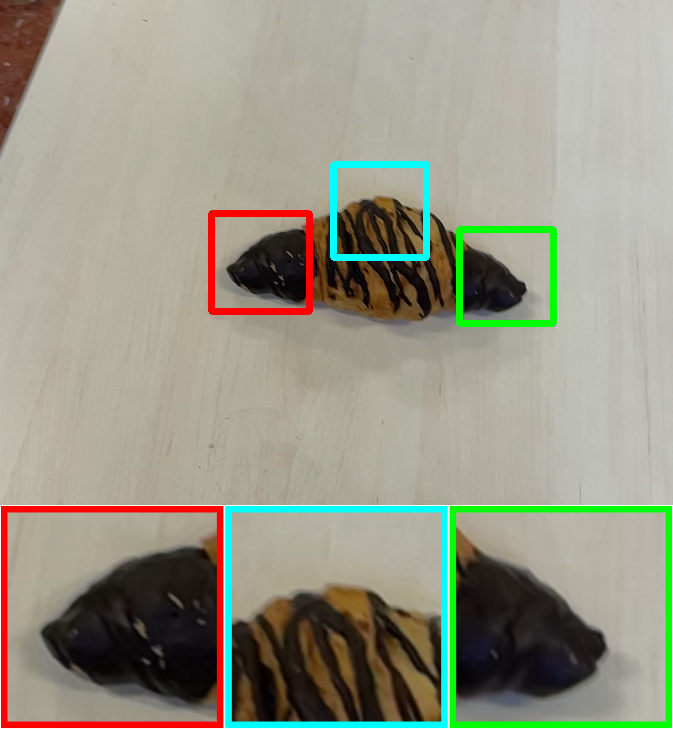} \\
\hline

\Block[v-center]{}{SegMan} & 

\includegraphics[width=0.17\textwidth]{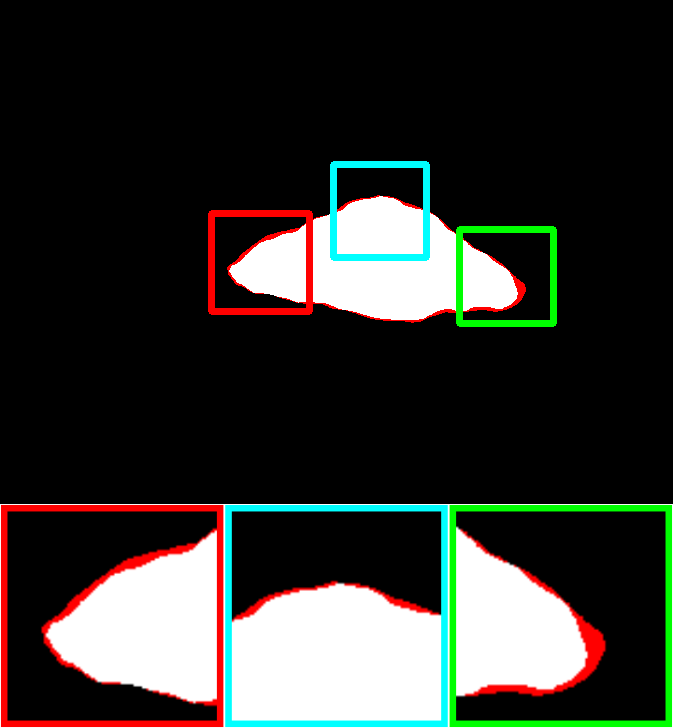} &
\includegraphics[width=0.17\textwidth]{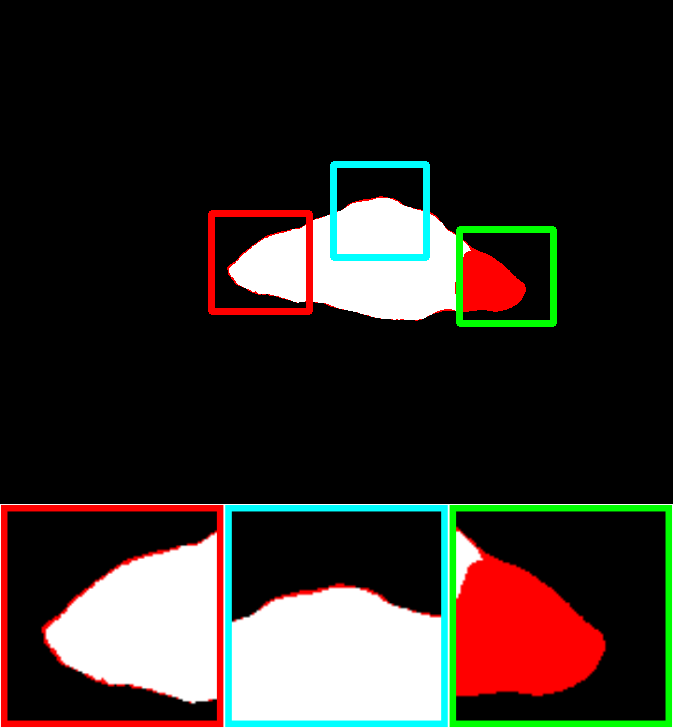} &
\includegraphics[width=0.17\textwidth]{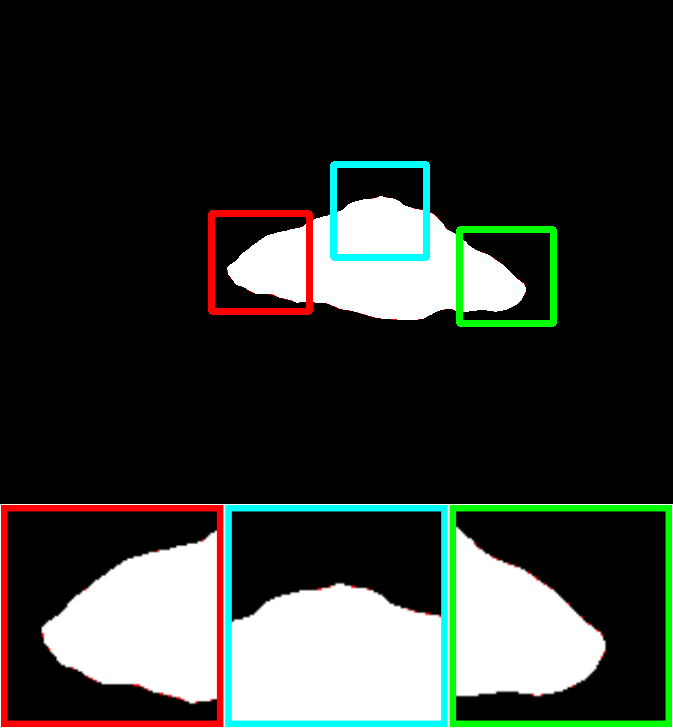} &
\includegraphics[width=0.17\textwidth]{Fig_R1/chocolate_croissant_976.png} &
\includegraphics[width=0.17\textwidth]{Fig_R1/chocolate_croissant_976_RGB.png}\\

\hline
    \end{NiceTabular}
    \label{tab:3d_comparisons_tracker_2d}

\raggedright\footnotesize{
\vspace{6px}
\tiny
Red indicates the prediction error relative to the ground truth.}
\end{table}

\begin{table}[htb]
    \centering
    \caption{Comparison of camera locations for the base segmentor, the segmentor with XMem2, and the segmentor with SAM2, with cameras colored by mAP: red for $\le$50\%, orange for 50–75\%, yellow for 75–95\%, and green for $\geq$95\%. The visualization highlights how each integration affects segmentation robustness across different viewpoints.}
    \setlength{\tabcolsep}{1pt}
    \begin{NiceTabular}{c|ccc}
    \hline
    Method & Segmentor & +XMem2 & +SAM2 \\
    \hline
    
\Block[v-center]{}{YOLO} &
\includegraphics[trim={0cm 1.5cm 0.2cm 2.7cm},clip,width=0.30\textwidth]{ YOLO.png} &
\includegraphics[trim={0cm 1.5cm 0.2cm 2.7cm},clip,width=0.30\textwidth]{ YOLO_XMEM2.png} &
\includegraphics[trim={0cm 1.5cm 0.2cm 2.7cm},clip,width=0.30\textwidth]{ 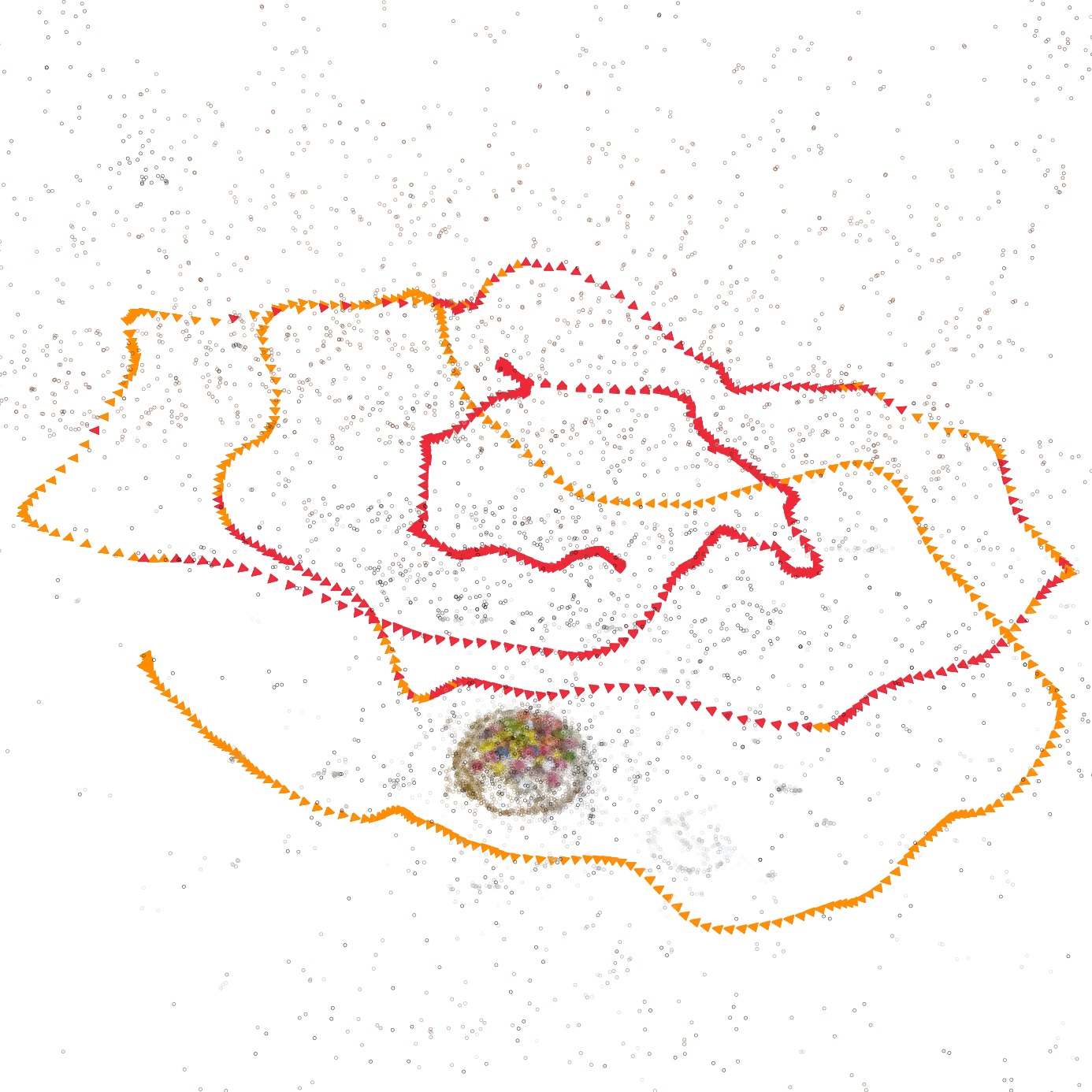} \\
\hline
\Block[v-center]{}{SegMan} & \includegraphics[trim={0cm 1.5cm 0.2cm 2.7cm},clip,width=0.30\textwidth]{ SegMAN.png} &
\includegraphics[trim={0cm 1.5cm 0.2cm 2.7cm},clip,width=0.30\textwidth]{ 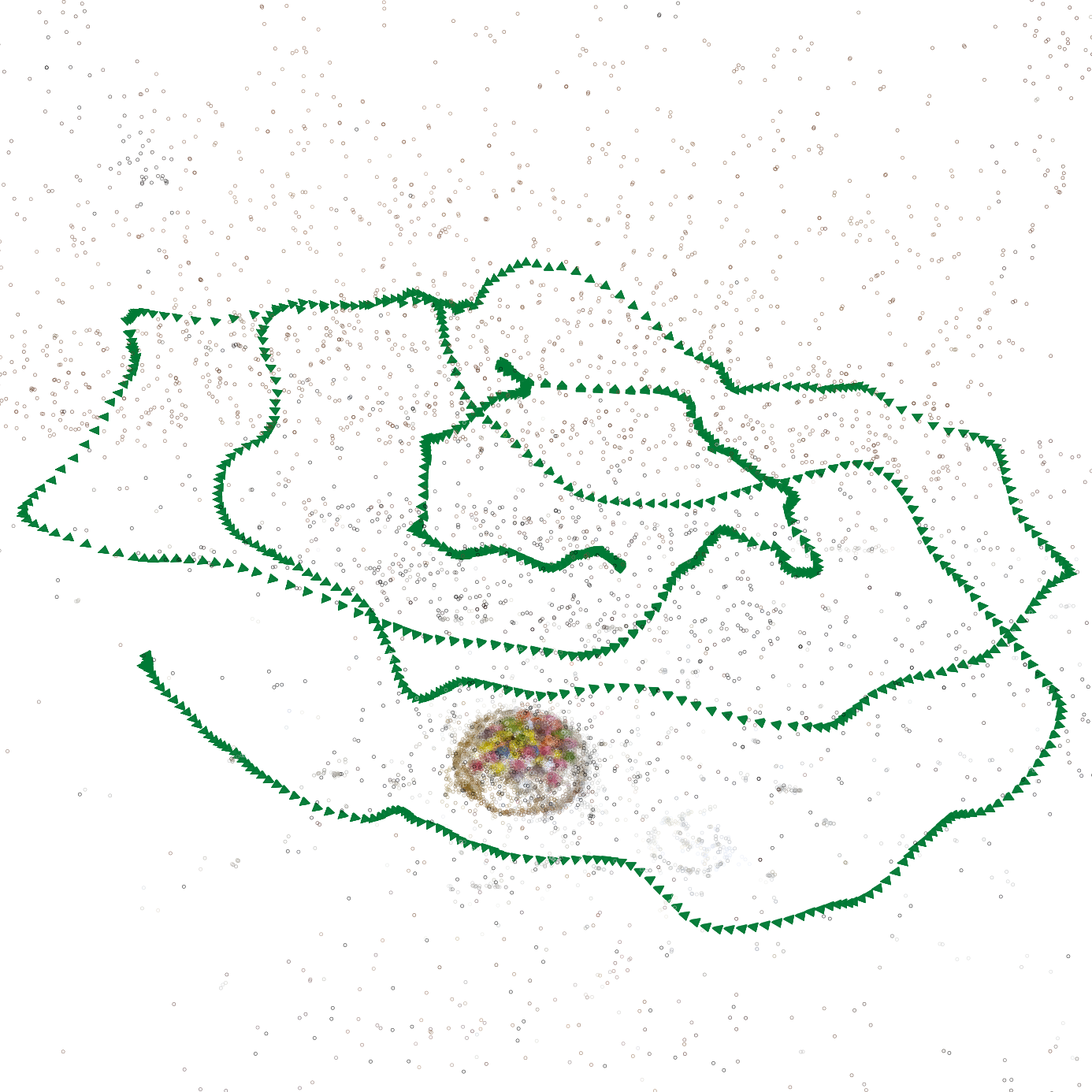} &
\includegraphics[trim={0cm 1.5cm 0.2cm 2.7cm},clip,width=0.30\textwidth]{ 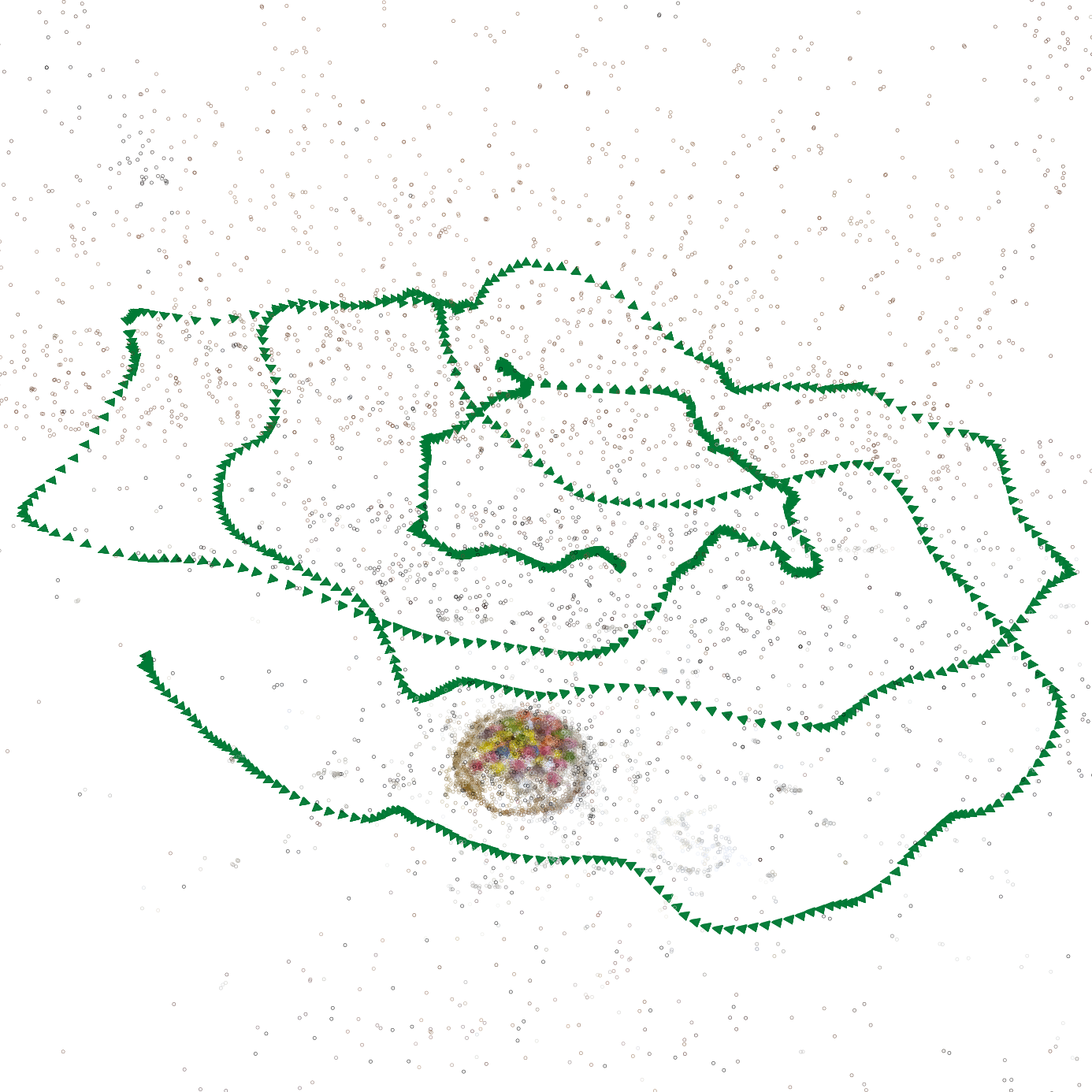} \\
\hline
% SeTR-MLA &
% \includegraphics[width=0.30\textwidth]{ SETR_MLA.png} &
% \includegraphics[width=0.30\textwidth]{ SETR_MLA_xmem2.png} &
% 
    \end{NiceTabular}
    \label{tab:3d_comparisons_tracker}
\end{table}

\paragraph{Selected observations} The highest single-frame $\mathrm{mAP}$ values on N5k are observed for \texttt{SegMan}-based variants and \texttt{BiRefNet} (0.94+), but these methods differ strongly in cross-partition behaviour: \texttt{BiRefNet} suffers on V\&F and MTF despite excellent N5k performance (see Table~\ref{tab:results}). The best overall $\mathrm{mAP}$ across multiple partitions is achieved by hybrid/ensemble methods (\texttt{SegMan+SAM2}, \texttt{SegMan+XMem2}, \texttt{SeTR-MLA+XMem2}), indicating that temporal propagation and SAM-based refinement are complementary to high-quality per-frame segmentation. Consistently, Table~\ref{tab:results_precision_f1_IoU} shows that methods with strong $\mathrm{mAP}$ on N5k and MTF (e.g. \texttt{SegMan+SAM2}, \texttt{SegMan+XMem2}, \texttt{SeTR-MLA+XMem2}) also achieve high F1 and IoU on these partitions, whereas approaches that underperform in $\mathrm{mAP}$, such as \texttt{FoodLMM}, exhibit low Precision and F1 across all splits, confirming that their weak detection scores translate directly into poor segmentation quality. 

\paragraph{Computational trade-offs} Model size correlates with resource consumption: large models (e.g. \texttt{FoodSAM}, \texttt{SeTR-MLA}, \texttt{FoodMem}) require multiple gigabytes of GPU memory and longer runtimes, whereas lightweight detectors (e.g. \texttt{YOLO}) deliver very fast per-frame inference at the cost of lower $\mathrm{mAP}$. Some hybrid pipelines add substantial parameter counts and memory overhead when combining a compact segmenter with a heavy propagation/refinement module (see \texttt{YOLO+XMem2} and \texttt{YOLO+SAM2} entries).
The metrics in Table~\ref{tab:results_precision_f1_IoU} further indicate that this cost–performance trade-off is not uniform across metrics: for instance, certain heavy models (\texttt{SeTR-MLA+XMem2} vs \texttt{SegMan+XMem2}) yield only modest gains in IoU over lighter baselines on N5k, while achieving larger improvements on challenging partitions such as FoodKit, suggesting that deployment decisions should be guided by the target operating domain rather than peak performance on a single split \footnote{Additional qualitative results are provided in the Appendix.}. 

\begin{figure}[!htbp]
    \centering
    \begin{subfigure}[t]{0.49\textwidth}
        \centering
        \includegraphics[width=1.0\textwidth]{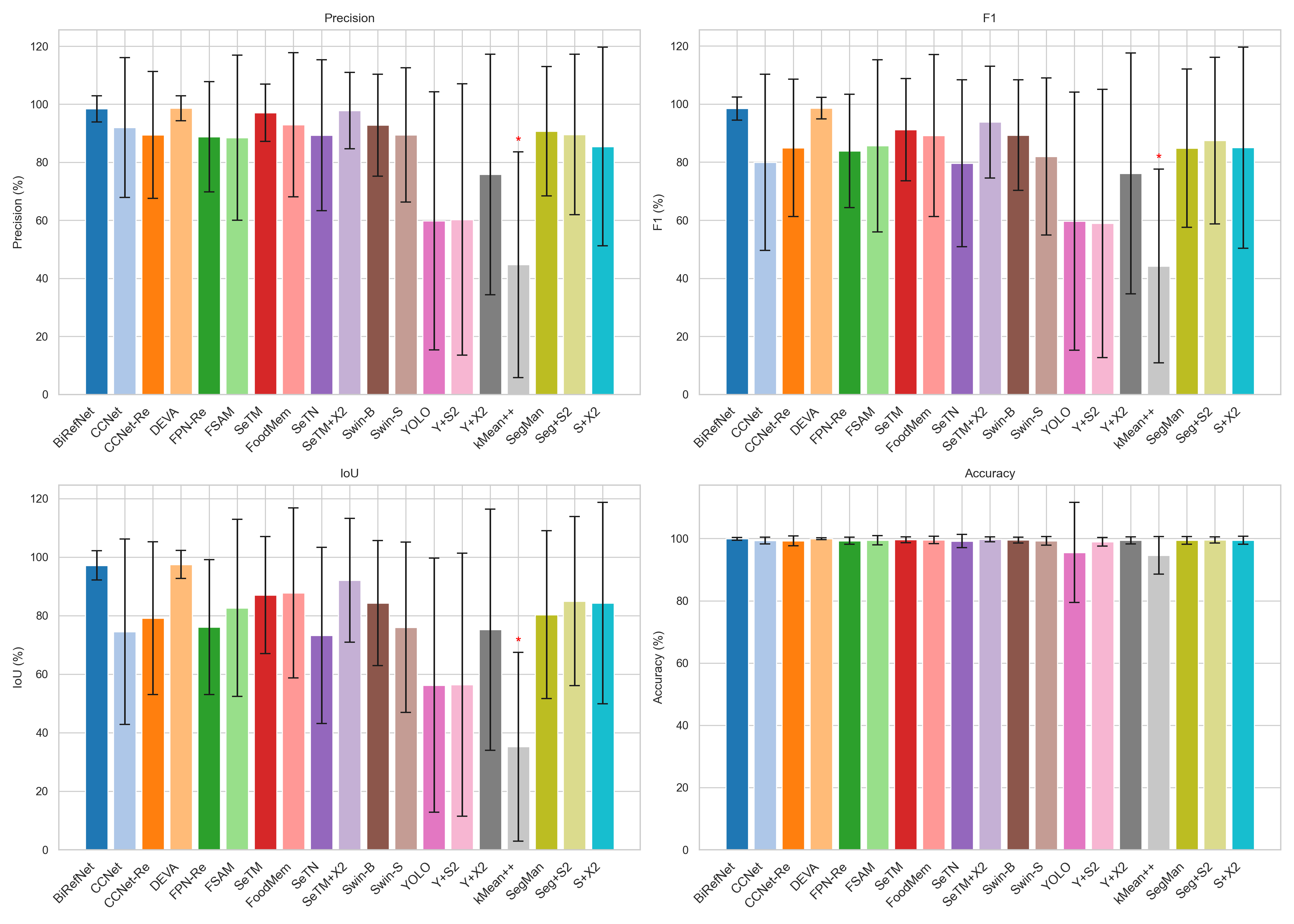}
        \caption{FKit}
    \end{subfigure}
    \begin{subfigure}[t]{0.49\textwidth}
        \centering
        \includegraphics[width=1.0\textwidth]{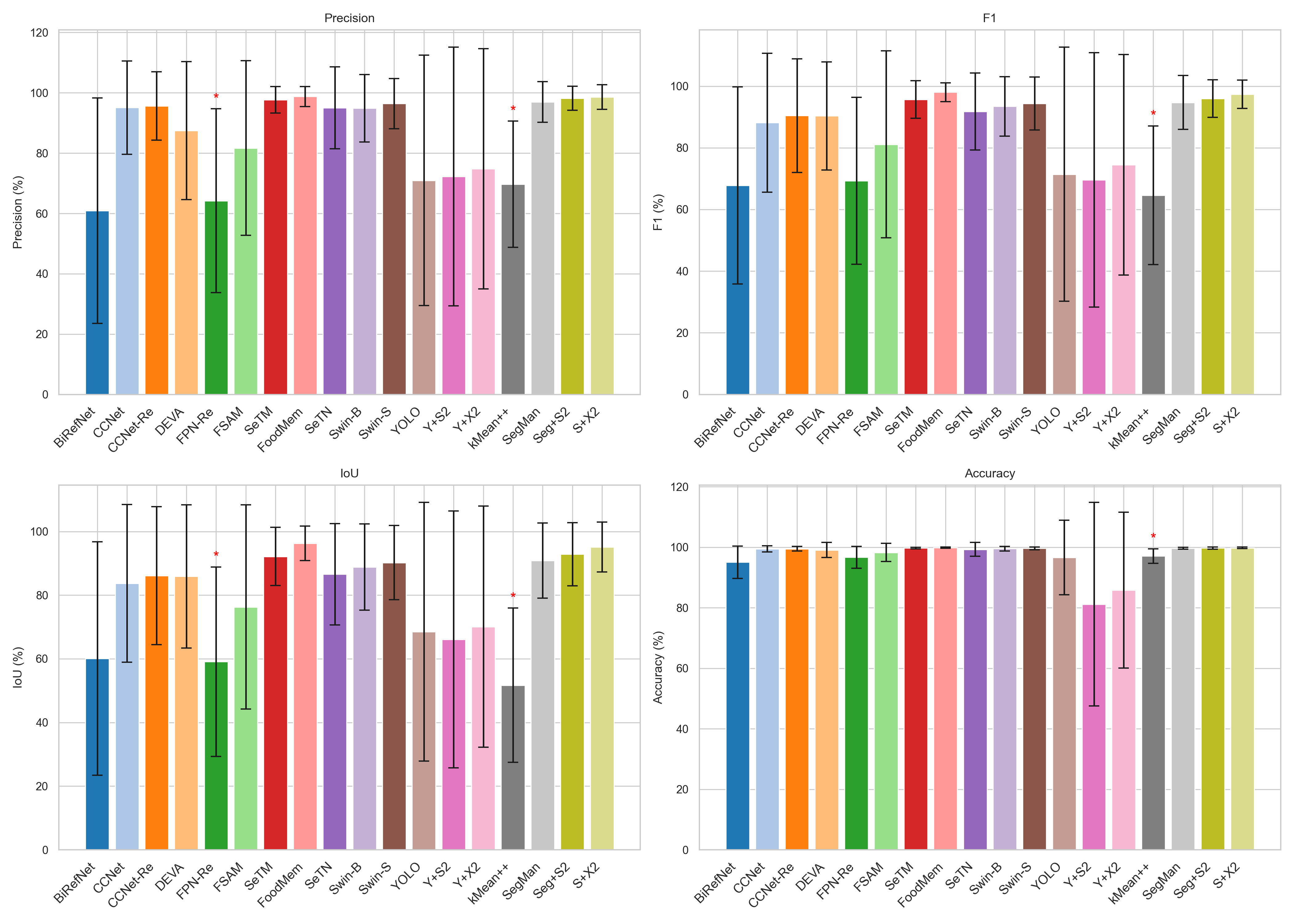}
        \caption{MTF}
    \end{subfigure}
    \begin{subfigure}[t]{0.49\textwidth}
        \centering
        \includegraphics[width=1.0\textwidth]{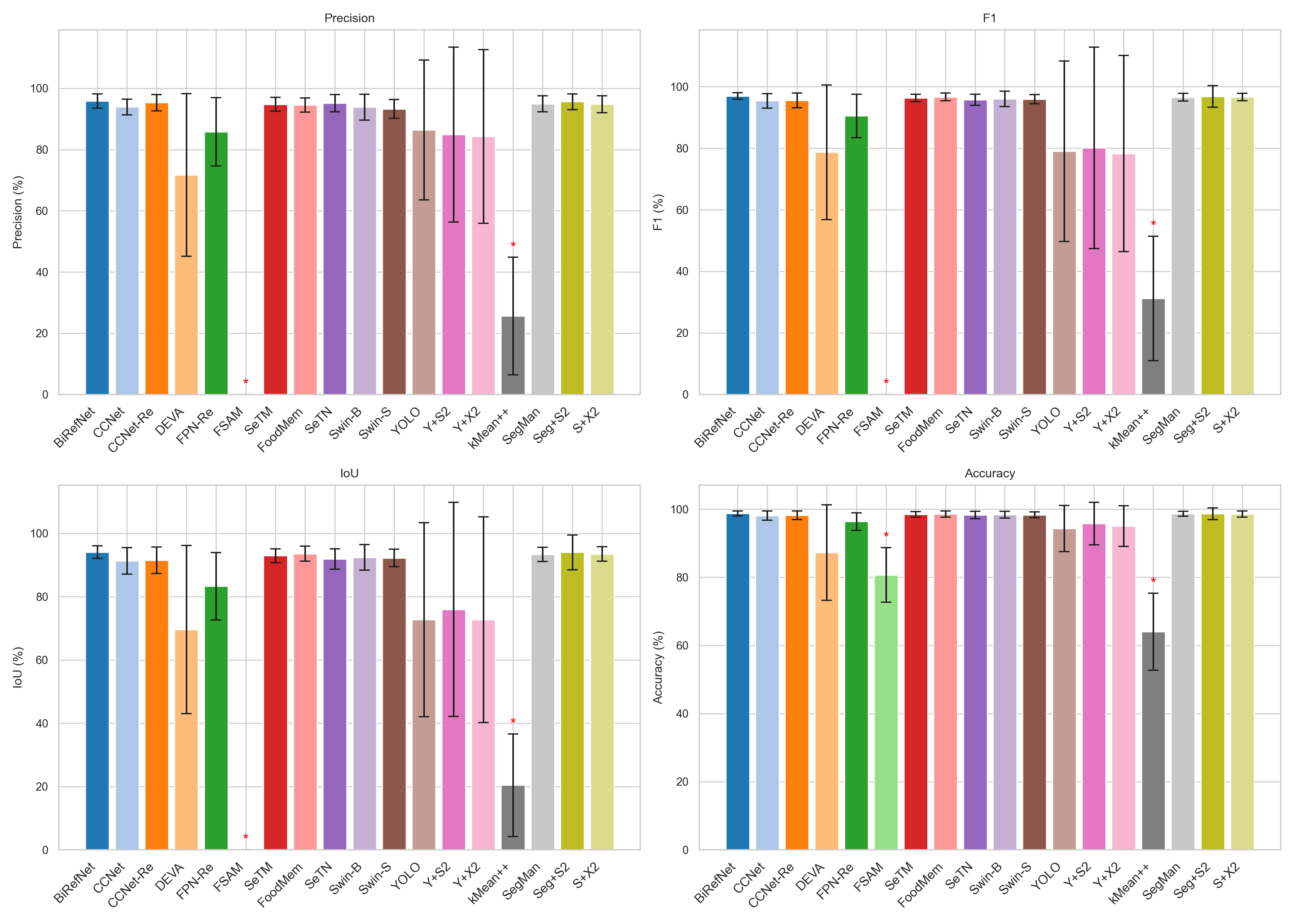}
        \caption{N5k}
    \end{subfigure}
    \begin{subfigure}[t]{0.49\textwidth}
        \centering
        \includegraphics[width=1.0\textwidth]{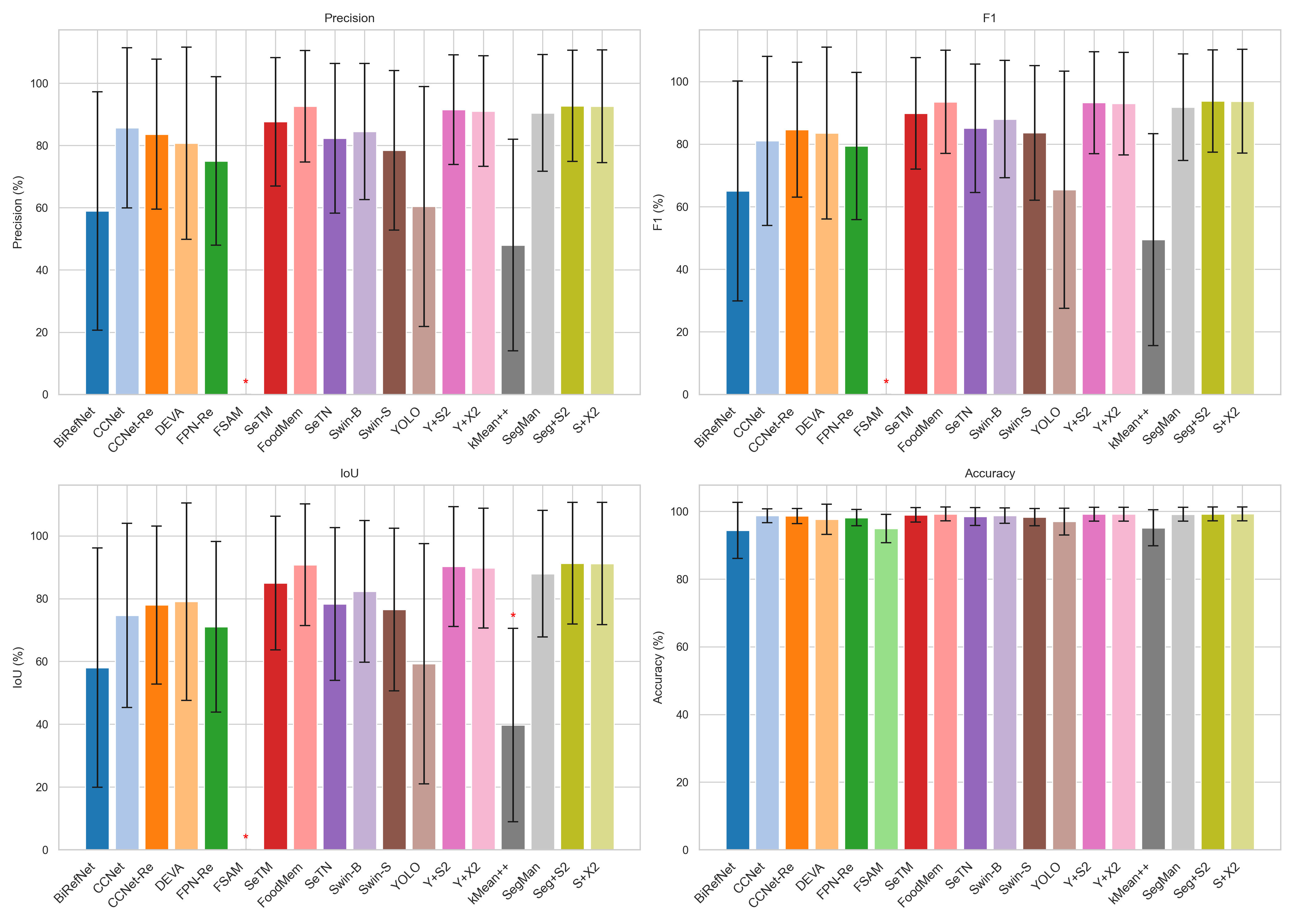}
        \caption{V\&F}
    \end{subfigure}
    \caption{Performance comparison across datasets using \textbf{Precision}, \textbf{Accuracy}, \textbf{F1 score}, and \textbf{IoU} (higher is better). Bars represent image-wise mean values, with error bars indicating standard deviation. 
    % \textbf{Red asterisks} denote cases where the best-performing method exceeds another baseline by more than the sum of their respective standard deviations, serving as a visual indicator of clear performance separation rather than a formal statistical significance test.
    }
    \label{fig:metrics_comparasions}
\end{figure}

\paragraph{Failure modes} Table \ref{tab:results} also highlights methods that fail systematically in our evaluation: \texttt{FoodLMM} shows negligible $\mathrm{mAP}$ and Recall under the current configuration, suggesting either a mismatch between model expectations and the dataset, or the need for further fine-tuning. \texttt{DEVA} exhibits strong Recall on FoodKit but a markedly low $\mathrm{mAP}$ on MTF, which points to inconsistent mask precision under certain capture conditions.
These tendencies are reflected in Table~\ref{tab:results_precision_f1_IoU}, where \texttt{FoodLMM} attains near-zero Precision, F1, and IoU on FoodKit, and \texttt{DEVA} achieves high F1/IoU on FoodKit but substantially worse scores on other partitions, reinforcing the view that some trackers and multimodal systems are highly sensitive to specific camera configurations and scene dynamics. Fig.~\ref{fig:metrics_comparasions} summarizes the performance of all baselines across datasets, reporting Precision, F1 score, IoU, and Accuracy for each method.

\paragraph{Uncertainty and Result Variability}
To characterize the stability of segmentation performance across diverse sequences and camera trajectories, we report both the mean and standard deviation of each metric over all test videos. The standard deviation reflects the variability induced by viewpoint changes, object appearance, occlusions, and motion patterns.
% \subsubsection*{Statistical testing}
% For the key pairwise comparisons reported in the text (e.g., SeTR-MLA+XMem2 vs FoodMem), we compute 95\% bootstrap confidence intervals over scenes (10,000 resamples) and perform paired two-sided Wilcoxon signed-rank tests. Differences with p < 0.05 are marked with an asterisk in the tables. The bootstrap CI quantifies uncertainty in the mean difference induced by scene-level variability; the paired test assesses whether one method consistently outperforms another across the same scenes.

\paragraph{Conclusion of results} In summary, Table~\ref{tab:results} shows that (i) combining segmentation with temporal memory or SAM-based refinement delivers the most consistent masks across all four partitions, (ii) model selection should consider the trade-off between accuracy and computational cost for deployment, and (iii) cross-dataset generalization remains a central challenge: several high-scoring models on N5k degrade substantially on V\&F or FoodKit, motivating future work on domain adaptation and multi-view training.
Taken together, Tables~\ref{tab:results} and~\ref{tab:results_precision_f1_IoU} indicate that no single configuration dominates all metrics and partitions simultaneously: methods that are Pareto-optimal in $\mathrm{mAP}$/Recall are not always optimal in IoU or F1, highlighting the importance of reporting a diverse set of metrics when benchmarking food segmentation and tracking pipelines. 
Taken together, Tables~\ref{tab:temporal_styled} and~\ref{tab:temporal_grouped_method} show that temporal stability varies substantially across methods and configurations. While some approaches achieve competitive frame-wise IoU, they exhibit pronounced flickering and drift, which are undesirable in real-world applications. These observations highlight the need to explicitly evaluate temporal behavior rather than relying solely on static metrics.

\subsection{Ablation study}
\begin{figure}[!htbp]
    \centering
    \begin{subfigure}[t]{0.49\textwidth}
        \centering
        \includegraphics[width=1.0\textwidth]{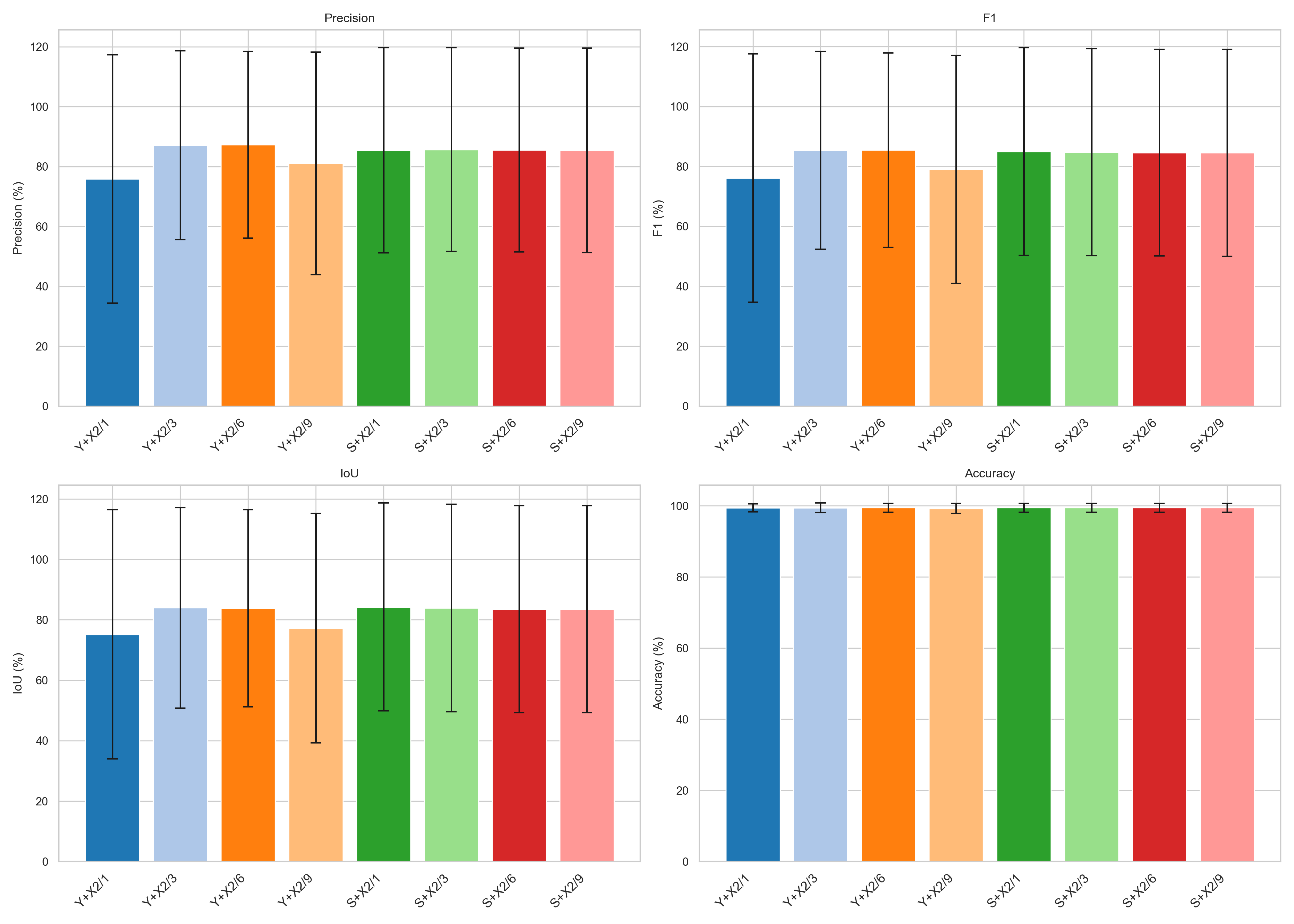}
        \caption{FKit}
    \end{subfigure}
    \begin{subfigure}[t]{0.49\textwidth}
        \centering
        \includegraphics[width=1.0\textwidth]{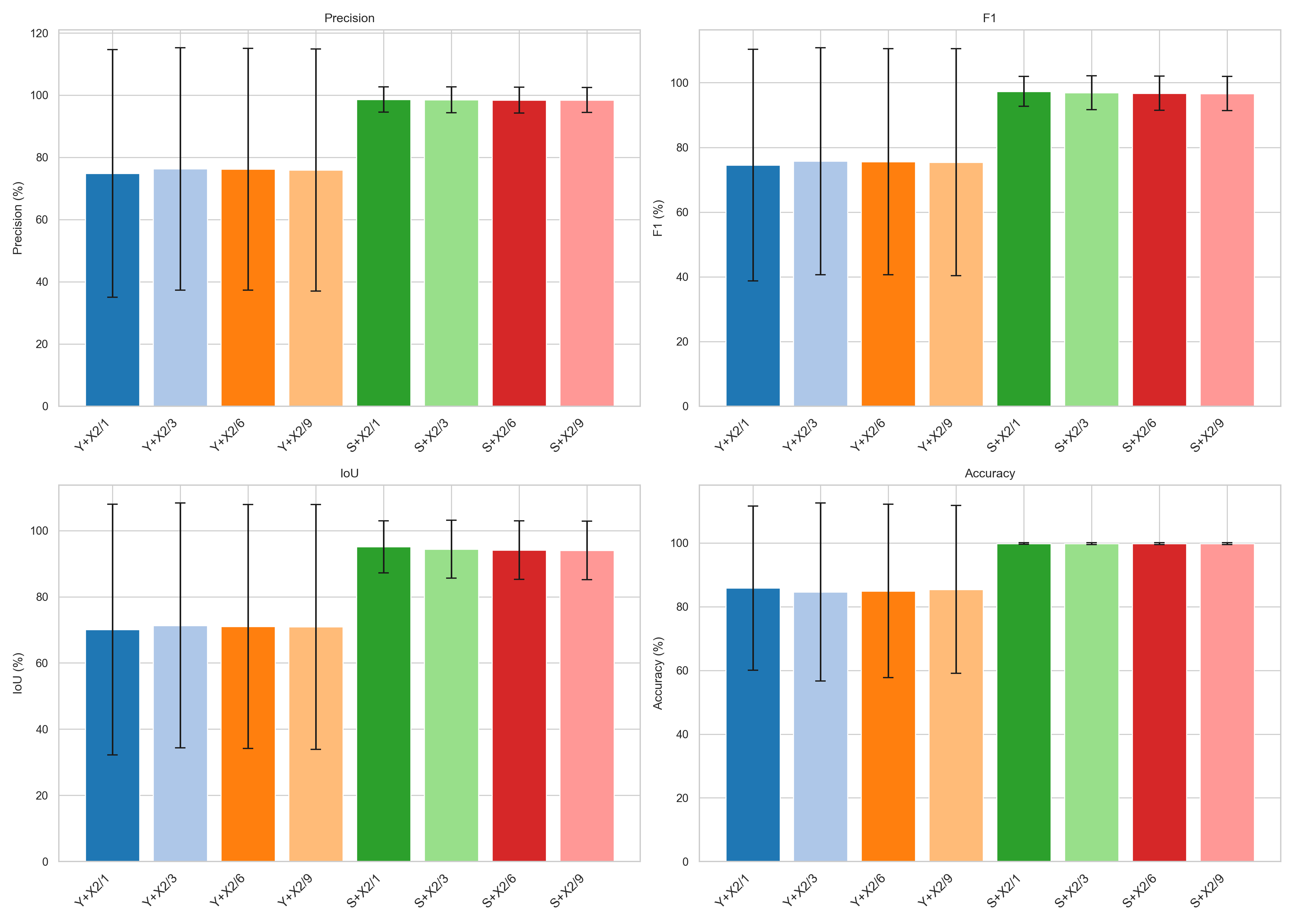}
        \caption{MTF}
    \end{subfigure}
    \begin{subfigure}[t]{0.49\textwidth}
        \centering
        \includegraphics[width=1.0\textwidth]{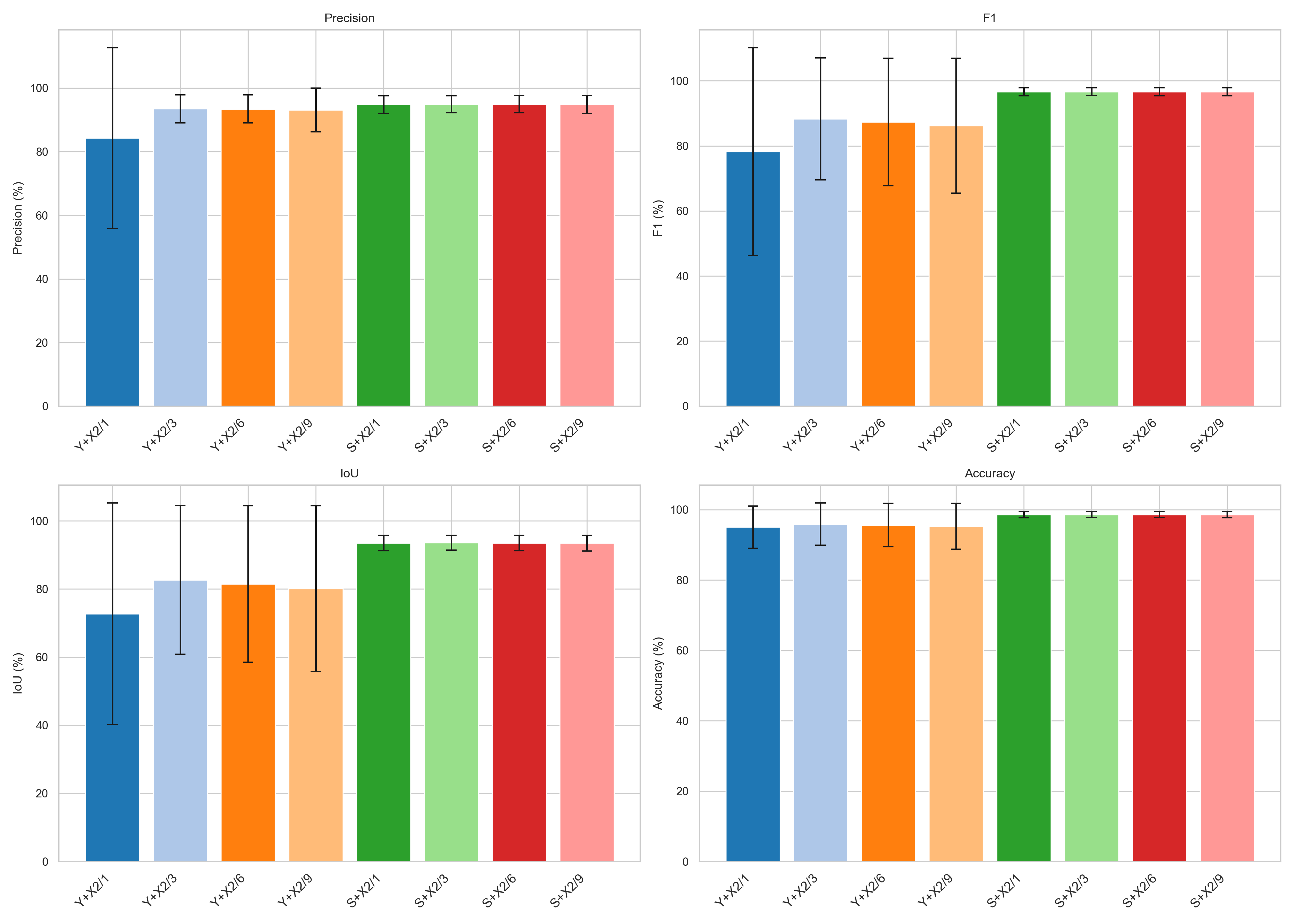}
        \caption{N5k}
    \end{subfigure}
    \begin{subfigure}[t]{0.49\textwidth}
        \centering
        \includegraphics[width=1.0\textwidth]{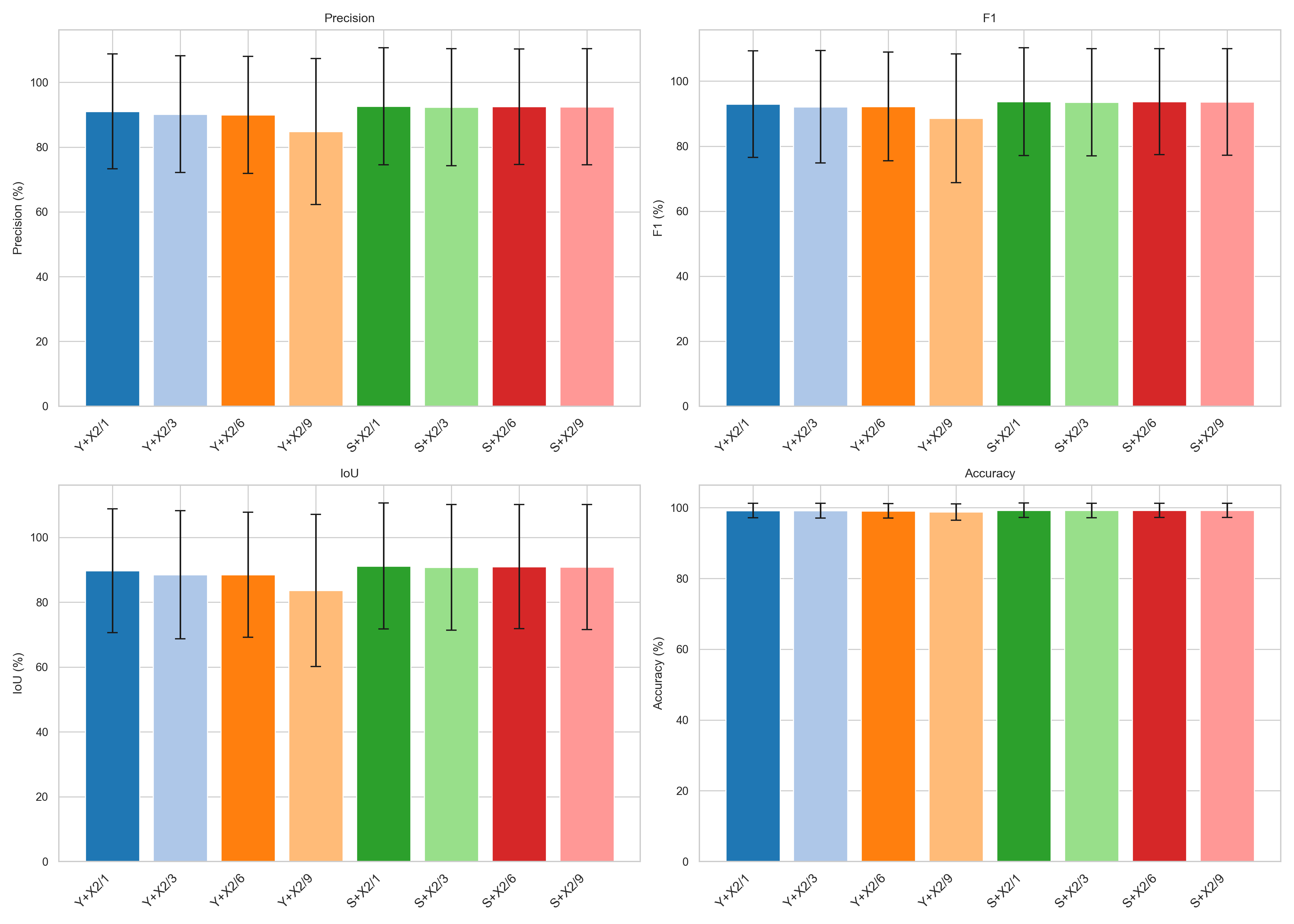}
        \caption{V\&F}
    \end{subfigure}
    \caption{Ablation performance comparison across datasets using \textbf{Precision}, \textbf{Accuracy}, \textbf{F1 score}, and \textbf{IoU} (higher is better). Bars represent image-wise mean values, with error bars indicating standard deviation. 
    % \textbf{Red asterisks} denote cases where the best-performing method exceeds another baseline by more than the sum of their respective standard deviations, serving as a visual indicator of clear performance separation rather than a formal statistical significance test.
    }
    \label{fig:metrics_comparasions_ablation}
\end{figure}

To investigate the factors underlying the efficacy of memory‑tracking segmentation models, we conducted a controlled ablation study in which we varied the number of candidate masks produced by the segmenter (e.g., SegMan), setting the mask count ($M$) to 1, 3, 6, or 9. We assessed the effect of this variation in two complementary ways. First, a qualitative evaluation provides visual comparisons of the resulting masks (see Table ~\ref{tab:3d_comparisons_tracker_2d}). Second, a quantitative evaluation reports segmentation quality using mAP and Recall (see Table ~\ref{tab:masks_ablation}), precision, F1, IoU, and Accuracy scores as shown in the Table.~\ref{tab:masks_ablation_additional}. Fig.~\ref{fig:metrics_comparasions_ablation} summarizes the ablation performance of all baselines across datasets, reporting Precision, F1 score, IoU, and Accuracy for each method and mask count ($M$).

\begin{table}[t]
\centering
\scriptsize
\setlength{\tabcolsep}{4pt}
\caption{Ablation study on the number of input masks ($M$).
Mean performance is reported in the main row, with standard deviation shown below. Methods are sorted in ascending order according to the number of masks. For each metric, the best, second-best, and third-best results are highlighted in \textbf{bold}, \uline{underline}, and \textit{italics}, respectively.}
\label{tab:masks_ablation}

\begin{tabular}{lccccccccc}
\toprule
\textbf{Method} &
&
\multicolumn{4}{c}{\textbf{mAP (\%) ↑}} &
\multicolumn{4}{c}{\textbf{Recall (\%) ↑}} \\
\cmidrule(lr){3-6} \cmidrule(lr){7-10}
$mean$±$std$  & $\boldsymbol{M}\blacktriangle$ & N5k & V\&F & MTF & FKit
 & N5k & V\&F & MTF & FKit \\
\midrule

Y+X2
 & 1
 & 84.28 & \underline{91.02} & 74.82 & 75.84
 & 79.82 & \underline{96.58} & 92.26 & 78.14 \\
 & 
 & 28.4 & 17.8 & 39.8 & 41.5
 & 34.6 & 11.1 & 19.4 & 41.6 \\

\rowcolor{gray!12}
Y+X2
 & 3
 & \textbf{93.48} & 90.14 & \textbf{76.29} & \underline{87.13}
 & \textbf{84.78} & 95.90 & \textbf{95.68} & \underline{78.88} \\ \rowcolor{gray!12}
 &
 & 4.4 & 18.0 & 39.0 & 31.5
 & 23.5 & 12.7 & 15.4 & 33.7 \\

Y+X2
 & 6
 & \underline{93.43} & \underline{89.94} & \underline{76.19} & \textbf{87.28}
 & \underline{83.35} & \underline{96.17} & \underline{95.30} & \textbf{79.13} \\
 &
 & 4.4 & 18.1 & 38.8 & 31.1
 & 24.8 & 11.7 & 15.6 & 33.2 \\

\rowcolor{gray!12}
Y+X2
 & 9
 & 93.12 & 84.80 & 75.92 & 81.07
 & 81.35 & 96.04 & 95.01 & 71.64 \\\rowcolor{gray!12}
 &
 & 6.9 & 22.6 & 39.0 & 37.2
 & 26.3 & 12.5 & 16.6 & 38.7 \\

\midrule

S+X2
 & 1
 & 94.81 & \underline{92.60} & \textbf{98.61} & 85.44
 & \textbf{98.80} & \textbf{96.65} & \textbf{97.10} & \textbf{79.28} \\
 &
 & 2.8 & 18.1 & 4.1 & 34.2
 & 2.0 & 11.1 & 7.3 & 34.7 \\

\rowcolor{gray!12}
S+X2
 & 3
 & \underline{94.88} & 92.31 & \underline{98.55} & \textbf{85.66}
 & \underline{98.78} & \underline{96.61} & \underline{96.65} & \underline{79.04} \\\rowcolor{gray!12}
 &
 & 2.7 & 18.0 & 4.2 & 34.0
 & 2.0 & 11.0 & 8.3 & 34.6 \\

S+X2
 & 6
 & \textbf{94.95} & \underline{92.49} & 98.43 & \underline{85.55}
 & 98.61 & 96.58 & 96.57 & 78.73 \\
 &
 & 2.7 & 17.8 & 4.1 & 34.1
 & 2.0 & 10.7 & 8.4 & 34.5 \\

\rowcolor{gray!12}
S+X2
 & 9
 & 94.88 & 92.42 & 98.42 & 85.43
 & 98.62 & 96.55 & 96.48 & 78.74 \\\rowcolor{gray!12}
 &
 & 2.8 & 17.9 & 4.0 & 34.1
 & 2.0 & 11.0 & 8.5 & 34.5 \\

\bottomrule
\end{tabular}

\vspace{2mm}
\scriptsize
\textbf{Abbreviations:}
Y+X2: YOLO+XMem2; S+X2: SegMan+XMem2;  
$M$: number of input masks.
\end{table}

\begin{table}[!htbp]
\centering
\tiny
\setlength{\tabcolsep}{1.1pt} % Adjust column spacing
\caption{Ablation study on the number of input masks ($M$). Mean performance is reported in the main row, with standard deviation shown below. Methods are sorted in ascending order according to the number of masks. For each metric, the best, second-best, and third-best results are highlighted in \textbf{bold}, \uline{underline}, and \textit{italics}, respectively.}
\label{tab:masks_ablation_additional}
\begin{tabular}{lccccccccccccccccc}
\toprule
\textbf{Meth.} & &
\multicolumn{4}{c}{\textbf{Precision (\%) $\uparrow$}} & 
\multicolumn{4}{c}{\textbf{F1 (\%) $\uparrow$}} & 
\multicolumn{4}{c}{\textbf{IoU (\%) $\uparrow$}} & 
\multicolumn{4}{c}{\textbf{Accuracy (\%) $\uparrow$}} \\
\cmidrule(lr){3-6} \cmidrule(lr){7-10} \cmidrule(lr){11-14} \cmidrule(lr){15-18}
 m±s &  $\boldsymbol{M}\blacktriangle$ & FKit & MTF & N5k & V\&F & FKit & MTF & N5k & V\&F & FKit & MTF & N5k & V\&F & FKit & MTF & N5k & V\&F \\
\midrule
Y+X2 & 1 & 75.84 & 74.82 & 84.28 & \textbf{91.02} & 76.12 & \textit{74.49} & 78.27 & \textbf{92.94} & 75.20 & 70.10 & 72.73 & \textbf{89.72} & \textit{99.43}  & \textbf{85.86} & 95.05 & \textbf{99.18}  \\
& & 41.45 & 39.83 & 28.40 & 17.76 & 41.43 & 35.79 & 31.89 & 16.41 & 41.19 & 37.88 & 32.50 & 19.10 & 1.13 & 25.76 & 5.96 & 2.05 \\ \addlinespace[3pt]

\rowcolor{gray!12}
Y+X2 & 3 & \uline{87.13} & \textbf{76.29} & \textbf{93.48} & \uline{90.14} & \uline{85.37} & \textbf{75.72}  & \textbf{88.30} & \textit{92.12} & \textbf{84.00} & \textbf{71.33} & \textbf{82.69} & \uline{88.52} & \uline{99.46}  & 84.62 & \textbf{95.91} & \uline{99.12} \\\rowcolor{gray!12}
& & 31.52 & 38.97 & 4.42 & 18.03 & 32.98 & 35.09 & 18.72 & 17.30 & 33.19 & 36.98 & 21.80 & 19.75 & 1.32 & 27.92 & 5.98 & 2.09 \\  \addlinespace[3pt]

Y+X2 & 6 & \textbf{87.28} & \uline{76.19} & \uline{93.43} & \textit{89.94}  & \textbf{85.44} & \uline{75.56}  & \uline{87.34} & \uline{92.20} & \uline{83.79} & \uline{71.01} & \uline{81.49} & \textit{88.50} & \textbf{99.47} & \textit{84.95} & \uline{95.65} & \textit{99.11}  \\
& & 31.11 &  38.84 & 4.38 & 18.05 & 32.43 & 34.92 & 19.59 & 16.73 & 32.61 & 36.82 & 22.99 & 19.29 & 1.27 & 27.24 & 6.20 & 2.08 \\\addlinespace[3pt]

\rowcolor{gray!12}
Y+X2 & 9 & \textit{81.07} & \textit{75.92} & \textit{93.12} & 84.80 & \textit{79.02} & 75.42 & \textit{86.17} & 88.58 & \textit{77.23} & \textit{70.91} & \textit{80.10} & 83.63 & 99.28  & \uline{85.43} & \textit{95.30} & 98.77 \\\rowcolor{gray!12}
& &  37.21 &  38.95 & 6.89 & 22.57 & 38.06 & 35.05 &  20.74 & 19.77 & 37.97 & 36.99 & 24.34 & 23.47 & 1.40 & 26.38 & 6.54 & 2.27 \\\addlinespace[3pt]

\midrule
S+X2 & 1 & \textit{85.44} & \textbf{98.61} & 94.81 & \textbf{92.60} & \textbf{84.99} & \textbf{97.31} & \textit{96.62} & \textbf{93.71} & \textbf{84.24} & \textbf{95.11} & \textit{93.49} & \textbf{91.17} & \textbf{99.48} & \textbf{99.83} & \textit{98.59} & \textbf{99.27} \\
& & 34.24 & 4.09 & 2.77 & 18.08 & 34.62 & 4.60 & 1.23 & 16.58 & 34.41 & 7.87 & 2.27 & 19.44 & 1.27 & 0.25 & 0.89 & 2.04 \\\addlinespace[3pt]

\rowcolor{gray!12}
S+X2 & 3 & \textbf{85.66} & \uline{98.55} & \uline{94.88} & 92.31 & \uline{84.79} & \uline{96.87} & \textbf{96.66} & 93.52 & \uline{83.89} & \uline{94.38} & \textbf{93.56} & 90.81 & \textit{99.47} & \uline{99.80} & \textbf{98.63} & 99.24 \\\rowcolor{gray!12}
& & 33.96 & 4.16 & 2.66 & 18.03 & 34.55 & 5.22 & 1.18 & 16.47 & 34.36 & 8.75 & 2.17 & 19.39 & 1.27 & 0.28 & 0.85 & 2.04 \\\addlinespace[3pt]

S+X2 & 6 & \uline{85.55} & \textit{98.43} & \textbf{94.95} & \uline{92.49} & \textit{84.60} & \textit{96.73} & \uline{96.63} & \uline{93.67} & \textit{83.53} & \textit{94.11} & \uline{93.50} & \uline{91.00} & \uline{99.47} & \textit{99.79} & \textbf{98.63} & \uline{99.26} \\
& & 34.06 & 4.14 & 2.74 & 17.83 & 34.49 & 5.27 & 1.21 & 16.29 & 34.27 & 8.83 & 2.22 & 19.18 & 1.26 & 0.29 & 0.85 & 2.03 \\\addlinespace[3pt]

\rowcolor{gray!12}
S+X2 & 9 & 85.43 & 98.42 & \textit{94.88} & \textit{92.42} & 84.58 & 96.65 & 96.60 & \textit{93.59} & 83.52 & 93.97 & 93.46 & \textit{90.91} & 99.47 & 99.79 & \uline{98.62} & \textit{99.25} \\\rowcolor{gray!12}
& & 34.14 & 4.01 & 2.82 & 17.93 & 34.51 & 5.29 & 1.25 & 16.42 & 34.27 & 8.84 & 2.30 & 19.29 & 1.26 & 0.29 & 0.87 & 2.03 \\

\bottomrule
\end{tabular}

\vspace{2mm}
\scriptsize
\raggedright
\textbf{Abbreviations:}
Y+X2: YOLO+XMem2; S+X2: SegMan+XMem2;  
$M$: number of input masks.
\end{table}

\begin{table}[!htb]
\centering
\tiny
\setlength{\tabcolsep}{0.9pt} % Adjust column spacing
\caption{Ablation study on the number of input masks ($M$) using temporal metrics. Mean performance is reported in the main row, with standard deviation shown below. Methods are sorted in ascending order according to the number of masks. For each metric, the best, second-best, and third-best results are highlighted in \textbf{bold}, \uline{underline}, and \textit{italics}, respectively. Higher is better (↑) for continuity; lower is better (↓) for flicker rate, IoU drift, and standard deviation.}
\label{tab:temporal_grouped_method}
\begin{tabular}{lccccccccccccccccccccc}
\toprule
\textbf{Meth.} & & \multicolumn{4}{c}{\textbf{$C_{t}$(\%)$\uparrow$}} & \multicolumn{4}{c}{\textbf{$FR_{0.2}$(\%)$\downarrow$}} & \multicolumn{4}{c}{\textbf{$\Delta IoU$(\%)$\downarrow$}} & \multicolumn{4}{c}{\textbf{$\sigma IoU$(\%)$\downarrow$}} & \multicolumn{4}{c}{\textbf{Global$(\%)$}} \\
\cmidrule(lr){3-6} \cmidrule(lr){7-10} \cmidrule(lr){11-14} \cmidrule(lr){15-18} \cmidrule(lr){19-22}
 m±s & \textbf{$M\blacktriangle$} & FKIT & MTF & N5K & V\&F & FKIT & MTF & N5K & V\&F & FKIT & MTF & N5K & V\&F & FKIT & MTF & N5K & V\&F & \textbf{$C_{t}$} & \textbf{$FR$} & \textbf{$\Delta IoU$} & \textbf{$\sigma IoU$} \\
\midrule
Y+X2 & 1 & \textbf{99.8} & \underline{99.2} & \textbf{98.7} & \textbf{98.4} & \textbf{0.1} & \textbf{0.5} & \textbf{0.5} & \textbf{1.2} & \textbf{0.2} & \underline{0.8} & \textbf{1.3} & \textbf{1.6} & 41.2 & \textit{37.9} & 32.5 & \textbf{19.1} & \textbf{99.0} & \textbf{0.6} & \textbf{1.0} & 32.7 \\
 &  & 1.8 & 5.6 & 4.8 & 5.1 & 0.0 & 0.0 & 0.0 & 0.0 & 1.8 & 5.6 & 4.8 & 5.1 & 41.2 & 37.9 & 32.5 & 19.1 & 0.6 & 0.5 & 0.6 & 9.7 \\
\addlinespace
\rowcolor{gray!12}
Y+X2 & 3 & \underline{99.7} & \textbf{99.2} & \underline{98.2} & \underline{98.2} & \textbf{0.1} & \textbf{0.5} & \underline{1.0} & \underline{1.3} & \underline{0.3} & \textbf{0.8} & \underline{1.8} & \underline{1.8} & \underline{33.2} & \underline{37.0} & \textbf{21.8} & \textit{19.8} & \underline{98.8} & \underline{0.7} & \underline{1.2} & \underline{27.9} \\
\rowcolor{gray!12} &  & 2.6 & 5.5 & 7.7 & 6.1 & 0.0 & 0.0 & 0.0 & 0.0 & 2.6 & 5.5 & 7.7 & 6.1 & 33.2 & 37.0 & 21.8 & 19.8 & 0.7 & 0.5 & 0.7 & 8.4 \\
\addlinespace
Y+X2 & 6 & \textit{99.6} & \textit{98.9} & \textit{97.5} & \textit{98.0} & \underline{0.2} & \underline{1.0} & \textit{2.4} & \textit{1.6} & \textit{0.4} & \textit{1.1} & \textit{2.5} & \textit{2.0} & \textbf{32.6} & \textbf{36.8} & \underline{23.0} & \underline{19.3} & \textit{98.5} & \textit{1.3} & \textit{1.5} & \textbf{27.9} \\
 &  & 3.4 & 6.4 & 9.9 & 7.0 & 0.0 & 0.0 & 0.0 & 0.0 & 3.4 & 6.4 & 9.9 & 7.0 & 32.6 & 36.8 & 23.0 & 19.3 & 0.9 & 1.0 & 0.9 & 8.2 \\
\addlinespace
\rowcolor{gray!12}
Y+X2 & 9 & 99.5 & 98.7 & 96.5 & 96.5 & \underline{0.2} & \underline{1.0} & 3.9 & 4.2 & 0.5 & 1.3 & 3.5 & 3.5 & \textit{38.0} & \underline{37.0} & \textit{24.3} & 23.5 & 97.8 & 2.3 & 2.2 & \textit{30.7} \\
\rowcolor{gray!12} &  & 4.1 & 7.6 & 12.3 & 9.5 & 0.0 & 0.0 & 0.0 & 0.0 & 4.1 & 7.6 & 12.3 & 9.5 & 38.0 & 37.0 & 24.3 & 23.5 & 1.5 & 2.0 & 1.5 & 7.9 \\
\midrule
S+X2 & 1 & \textbf{99.7} & \textbf{99.5} & \textbf{99.0} & \textbf{98.6} & \textbf{0.0} & \textbf{0.1} & \textbf{0.0} & \textbf{1.2} & \textbf{0.3} & \textbf{0.5} & \textbf{1.0} & \textbf{1.4} & 34.4 & \textbf{7.9} & \textit{2.3} & \textit{19.4} & \textbf{99.2} & \textbf{0.3} & \textbf{0.8} & \textbf{16.0} \\
 &  & 1.7 & 1.2 & 1.1 & 5.2 & 0.0 & 0.0 & 0.0 & 0.0 & 1.7 & 1.2 & 1.1 & 5.2 & 34.4 & 7.9 & 2.3 & 19.4 & 0.5 & 0.6 & 0.5 & 14.2 \\
\addlinespace
\rowcolor{gray!12}
S+X2 & 3 & \textbf{99.7} & \underline{99.5} & \underline{99.0} & \underline{98.5} & \underline{0.1} & \underline{0.1} & \underline{0.0} & \textit{1.3} & \textbf{0.3} & \underline{0.5} & \underline{1.0} & \underline{1.5} & 34.4 & \underline{8.7} & \textbf{2.2} & \textit{19.4} & \underline{99.2} & \underline{0.3} & \underline{0.8} & \textit{16.2} \\
\rowcolor{gray!12} &  & 1.7 & 1.2 & 1.1 & 5.0 & 0.0 & 0.0 & 0.0 & 0.0 & 1.7 & 1.2 & 1.1 & 5.0 & 34.4 & 8.7 & 2.2 & 19.4 & 0.6 & 0.6 & 0.6 & 14.1 \\
\addlinespace
S+X2 & 6 & \underline{99.7} & \textit{99.4} & \textit{99.0} & \textit{98.5} & \textit{0.1} & \textit{0.2} & \textit{0.0} & \underline{1.2} & \textit{0.3} & \textit{0.6} & \textit{1.0} & \textit{1.5} & \textbf{34.3} & \textit{8.8} & \underline{2.2} & \textbf{19.2} & \textit{99.2} & \textit{0.4} & \textit{0.8} & \underline{16.1} \\
 &  & 1.9 & 1.7 & 1.1 & 4.5 & 0.0 & 0.0 & 0.0 & 0.0 & 1.9 & 1.7 & 1.1 & 4.5 & 34.3 & 8.8 & 2.2 & 19.2 & 0.5 & 0.5 & 0.5 & 14.0 \\
\addlinespace
\rowcolor{gray!12}S+X2 & 9 & \textit{99.7} & 99.4 & 99.0 & 98.4 & 0.1 & 0.2 & 0.0 & 1.4 & \underline{0.3} & 0.6 & 1.0 & 1.6 & 34.3 & 8.8 & 2.3 & \underline{19.3} & 99.1 & 0.4 & 0.9 & 16.2 \\
\rowcolor{gray!12} &  & 1.8 & 1.7 & 1.2 & 5.1 & 0.0 & 0.0 & 0.0 & 0.0 & 1.8 & 1.7 & 1.2 & 5.1 & 34.3 & 8.8 & 2.3 & 19.3 & 0.6 & 0.7 & 0.6 & 13.9 \\
\bottomrule
\end{tabular}

\vspace{2mm}
\scriptsize
\raggedright
\textbf{Abbreviations:}
Y+X2: YOLO+XMem2; S+X2: SegMan+XMem2;  
$M$: number of input masks.
\end{table}

The ablation study reveals that the number of candidate masks provided to the memory-based tracker substantially influences segmentation stability and accuracy. Increasing the mask count generally improves recall and enriches the search space of the tracker, enabling more robust temporal consistency—particularly for challenging sequences in BenchSeg. However, this benefit is not uniform across models: lightweight architectures such as YOLO+XMem2 show notable sensitivity to the mask count ($M$), with performance varying widely across datasets, whereas SegMan+XMem2 remains comparatively stable and consistently achieves high accuracy even with a single mask. These findings indicate that stronger base segmentors depend less on mask redundancy, while weaker ones benefit from a broader mask pool, highlighting the interplay between spatial segmentation quality and temporal memory mechanisms.

We further investigate how the number of candidate masks affects temporal stability. Table~\ref{tab:temporal_grouped_method} reports the continuity and flicker metrics as a function of the number of input masks, allowing us to analyze the trade-off between redundancy and temporal robustness. As shown in Table~\ref{tab:temporal_grouped_method}, increasing the number of candidate masks generally improves temporal continuity for weaker base segmenters, while the gains saturate for stronger models. This suggests that temporal memory can compensate for spatial uncertainty to a certain extent, but cannot fully replace high-quality single-frame predictions.

\paragraph{Limitations} Although BenchSeg offers a diverse set of dishes, several important limitations remain: (i) \textbf{coverage}: the corpus does not span all cuisines or extreme imaging conditions (e.g., very low light, highly cluttered tabletops), and many scenes are relatively controlled (single-dish sweeps); (ii) \textbf{temporal assumptions}: memory-based segmentation methods assume limited inter-frame change and fairly stable illumination—abrupt occlusions, rapid motion, or highly dynamic lighting can produce fusion errors and identity switches among visually similar items; (iii) \textbf{computational cost}: transformer- and memory-augmented architectures incur substantial compute and memory demands, complicating real-time deployment on resource-constrained devices without model compression or pruning; and (iv) \textbf{deployment gap}: addressing these issues will require methods that improve robustness to occlusion and motion, more efficient architectures, and evaluation on truly unconstrained dining scenarios.

\section{Reproducibility}
To support reproducibility, we will release all training configurations, benchmark splits, and evaluation scripts. All experiments were conducted using the MMsegmentation framework with fixed random seeds. We report software versions, GPU models, batch sizes, learning rates, and inference resolutions for all experiments. Computational metrics (FPS, VRAM usage) were measured using batch size 1 on a single GPU. All scripts used to compute mAP, IoU, and related metrics will be made publicly available.

\subsection{Reproducibility checklist}
The code, model configuration files, and evaluation scripts will be released at:\footnote{https://github.com/GCVCG/benchseg} together with the exact commit hash (e.g., \texttt{commit: 80a68b1}). The release will include: (i) a Dockerfile and conda/yaml environment specifying PyTorch 1.13.1, CUDA 11.8, mmsegmentation@v0.30.0; (ii) per-method mmsegmentation config files and training commands; (iii) random seeds used for training and evaluation; (iv) per-scene metric CSVs used to compute tables; and (v) scripts to reproduce Tables \ref{tab:results}–\ref{tab:masks_ablation_additional}. Computational measurements (FPS, VRAM) were recorded on the hardware listed in Sec. \ref{sec:implementation_settings}.

\subsection{Ethical Considerations}
BenchSeg comprises images depicting food scenes that may feature human hands or personal environments. The datasets utilized as sources were gathered in accordance with their respective ethical guidelines. We do not incorporate any personally identifiable information. Users engaging with the benchmark are expected to adhere to the original dataset licenses and relevant ethical standards.

\section{Conclusions and Future Work}
We introduced BenchSeg, a large-scale dataset and benchmark for evaluating food segmentation models under realistic multi-view free-motion conditions in food videos. BenchSeg is developed using 25,284 richly annotated frames and trained strictly out-of-domain with FoodSeg103 to emphasize cross-dataset generalization and robustness. By leveraging robust per-frame segmentors combined with temporal-aware propagation mechanisms, our extensive experiments and empirical results demonstrate significant performance variations across 20 state-of-the-art segmentation architectures. We emphasized the critical importance of temporal consistency for reliable food-region prediction and observed that, when combined with temporal-aware propagation methods, low-performance food segmentors can achieve competitive results with stronger segmentors, although cross-dataset generalization remains challenging.
Beyond spatial accuracy, our analysis demonstrates that temporal stability varies significantly across methods. The proposed metrics provide a complementary diagnostic tool that exposes instability patterns invisible to frame-wise evaluations, offering more realistic insight into real-world deployment behavior.
Looking ahead, future work should focus on designing more efficient and scalable hybrid memory modules suitable for real-time inference and on expanding BenchSeg to more unconstrained dining scenarios—multi-dish scenes, clutter, and challenging illumination. These directions are essential for advancing more practical, accurate, and deployable food-understanding systems.

\section{Acknowledgments}
This work was partially funded by the EU project MUSAE (No. 01070421), 2021- SGR-01094 (AGAUR), Icrea Academia’2022 (Generalitat de Catalunya), Robo STEAM (2022-1-BG01-KA220- VET000089434, Erasmus+ EU), Deep-Sense (ACE053/22/000029, ACCIÓ), and Grants PID2022141566NB-I00 (IDEATE), PDC2022-133642-I00 (DeepFoodVol), and CNS2022-135480 (A-BMC) funded by MICIU/AEI/10.13039/501100 011033, by FEDER (UE), and by European Union NextGenerationEU/ PRTR. A. AlMughrabi acknowledges the support of FPI Becas, MICINN, Spain. U. Haroon acknowledges the support of FI-SDUR Becas, MICINN, Spain. F. Al-areqi acknowledges the support of the FI 2025 doctoral fellowship (AGAUR), Spain. The authors thankfully acknowledge the RES resources provided by the Barcelona Supercomputing Center on MareNostrum5 for IM-2025-3-0008.

\bibliographystyle{elsarticle-num} 
\bibliography{ref}

\section{Appendix}

% \subsection{Overview}
Our benchmarking includes datasets captured using mobile applications, notably Record3D for MTF, OnePose++ for FKit, and a custom app for V\&F. These datasets typically follow multi-view acquisition strategies characterized by three-ring capture topologies (more details in Sec.~\ref{sec:capturing_patterns}) with high viewpoint randomization, which are commonly adopted in practical mobile scanning scenarios. We present the notations and their definitions in Sec.~\ref{sec:notations}. Yet, we discuss representative mobile applications in Sec.~\ref{sec:mobile_computing} and common capture patterns in Sec.~\ref{sec:capturing_patterns} to contextualize the acquisition settings of existing datasets. Importantly, our benchmark itself is agnostic to the specific capture application or topology and focuses on evaluating segmentation performance under free-motion, real-world conditions. Architectural details of the baselines and additional experimental results are provided in Secs.~\ref{sec:baselines_architectural_design} and~\ref{sec:additional_experimental_results}, respectively.

\subsection{Notations}
\label{sec:notations}
Table~\ref{tab:symbols_metrics} summarizes the main symbols and evaluation metrics used in the paper for ease of reference.

\begin{table}[!htbp]
\centering
\small
\setlength{\tabcolsep}{6pt}
\caption{Summary of notations and evaluation metrics used in the paper.}
\label{tab:symbols_metrics}
\begin{tabular}{l p{8cm}}
\toprule
\textbf{Notation} & \textbf{Definition} \\
\midrule
\rowcolor{gray!12}$\mathcal{D}$ & Benchmark dataset \\
$S_i$ & i-th scene, ordered sequence of image–mask pairs \\
\rowcolor{gray!12}$n_i$ & Number of frames in scene $S_i$ \\
$x_{i,j}$ & RGB image in scene $i$, frame $j$ \\
\rowcolor{gray!12}$H, W$ & Image height and width \\
$\mathcal{M}$ & Family of segmentation models \\
\rowcolor{gray!12}$M_\theta$ & Segmentation model with parameters $\theta$ \\
$\mathcal{T}$ & Training set \\
\rowcolor{gray!12}$\mathcal{L}$ & Training loss function \\
$\theta^\star$ & Optimized parameters \\
\rowcolor{gray!12}$\hat m$ & Binary prediction mask \\
$\tau$ & Threshold for binarization \\
\rowcolor{gray!12}$\mathcal{S}$ & Scene sequence \\
$\mathcal{K}$ & Set of keyframe indices \\
\rowcolor{gray!12}$\mathcal{K}^\star$ & Selected keyframes \\
$K_{\max}$ & Maximum number of keyframes \\
\rowcolor{gray!12}$\mathcal{C}$ & Perceptual diversity objective for keyframe selection \\
$\hat m_k$ & Predicted mask on keyframe $k$ \\
\rowcolor{gray!12}$\mathcal{P}$ & Propagation operator \\
$\tilde m_j$ & Propagated mask for frame $j$ \\
\rowcolor{gray!12}$\Phi$ & Fusion function \\
$m_j^{\mathrm{final}}$ & Final fused mask for frame $j$ \\
\rowcolor{gray!12}TP, FP, FN, TN & True/false positives/negatives \\
Precision & Ratio of true positives to predicted positives \\
\rowcolor{gray!12}Recall & Ratio of true positives to actual positives \\
F1-score & Harmonic mean of precision and recall \\
\rowcolor{gray!12}IoU & Ratio of intersection to union of prediction and ground truth \\
Accuracy & Ratio of correctly classified pixels to total pixels \\
\rowcolor{gray!12}mIoU & Mean IoU across classes \\
mAcc & Mean accuracy across classes \\
\rowcolor{gray!12}mAP & Mean average precision across classes \\
\bottomrule
\end{tabular}
\end{table}

\subsection{Mobile Computing}
\label{sec:mobile_computing}
Modern mobile solutions provide powerful options for capturing volumetric data and 3D models using consumer-grade devices. Scaniverse\footnote{https://scaniverse.com/} offers seamless LiDAR and photogrammetry-based Gaussian splatting with real-time processing directly on iOS/Android, along with flexible export and editing tools using an intuitive interface. Polycam\footnote{https://poly.cam/} supports LiDAR, photogrammetry, and splat-based capture across iOS, Android, and web platforms, with features such as guided capture, AI-assisted model generation, scene editing, and measurement reporting—making it suitable for architectural and creative workflows. Luma AI \footnote{https://lumalabs.ai/} uses NeRF-inspired neural reconstruction to produce highly detailed volumetric models without requiring depth sensors, achieving impressive results on reflective surfaces. KIRI Engine \footnote{https://www.kiriengine.app/} combines photogrammetry, LiDAR, neural surface reconstruction, and Gaussian splatting in a single app, offering masking, on-device cleanup, quad-mesh retopology, and export to standard 3D formats. NeRFCapture \footnote{https://github.com/jc211/NeRFCapture} enables users to stream ARKit-posed RGB and depth frames to reconstruction frameworks like Instant-NGP \cite{muller2022instant}, facilitating high-fidelity capture workflows. OnePose++\cite{he2022onepose++} and  Record3D utilize iPhone LiDAR sensors to record live RGB-D videos and export point clouds and mesh formats, with streaming capabilities over USB or Wi-Fi. Solaya\footnote{https://www.solaya.ai/} leverages generative AI and neural reconstruction to generate 3D object scans on iOS devices. Lastly, LogMeal\footnote{https://logmeal.com/es/} focuses on dietary imaging by using deep learning to recognize dishes, estimate ingredients and nutrition, and support portion analysis from smartphone images. Table \ref{tab:3d_capture_apps} shows a comparative overview for 3D capturing applications for Mobile devices.   

% \begin{table*}[ht]
% \tiny
% \setlength{\tabcolsep}{0.1em}
% \caption{Comparative Overview of Mobile Apps for 3D Capture and Volumetric Reconstruction}
% \begin{tabular}{lllll}
% \toprule
% App & Capture Technology & Export Files & Mobility & Editing \& Masking \\ \midrule
% Polycam    & LiDAR, photogrammetry             & Mesh & iOS, Android, Web & Guided capture, measurement tools \\ 
% Scaniverse & LiDAR, photogrammetry             & Mesh & iOS, Android     & Trimming, smoothing \\ 
% KIRI Engine & Photogrammetry, NeRF, 3DGS, LiDAR & Mesh & iOS, Android, Web & Masking, mesh retopology \\ 
% Luma AI   & NeRF         & Mesh & iOS, Android     & Some AR widgets \\ 
% NeRFCapture & ARKit RGB (LiDAR optional)       & Poses/RGBD (streamed) & iOS              & - \\ 
% Record3D  & LiDAR / TrueDepth RGB-D           & Mesh, Alembic (streamed)       & iOS              & - \\ 
% LogMeal & IMU RGB & Poses, RGB & iOS, Android & - \\
% Solaya & IMU RGB & Poses, RGBD & iOS & App is in progress \\
% \bottomrule
% \end{tabular}
% \label{tab:3d_capture_apps}
% \end{table*}

\begin{table}[!htbp]
\centering
\small
\setlength{\tabcolsep}{1pt}
\caption{Comparison of mobile applications for 3D capture and reconstruction.
Checkmarks indicate supported functionality.
Photogrammetry refers to multi-view mesh reconstruction, NeRF to neural radiance fields,
and 3DGS to 3D Gaussian Splatting.}
\label{tab:3d_capture_apps}

\begin{tabular}{@{}p{2cm}cccccp{2.5cm}p{3.5cm}@{}}
\toprule
\textbf{App} & \multicolumn{2}{c}{\textbf{Sensor}} & \multicolumn{3}{c}{\textbf{Method}} & \textbf{Export}  & \textbf{Notes} \\
\cmidrule(lr){2-3} \cmidrule(lr){4-6} 
& Image & LiDAR & Photo. & NeRF & 3DGS &
& \\
\midrule

Polycam
& \checkmark & \checkmark & \checkmark &  & \checkmark
& Mesh, splats
& Guided capture, object masking \\

Scaniverse
& \checkmark & \checkmark & \checkmark &  & \checkmark
& Mesh, splats
& Editing: Trimming, smoothing \\

KIRI Engine
& \checkmark & \checkmark & \checkmark & \checkmark & \checkmark
& Mesh, splats
& Masking, mesh retopology \\

Luma AI
& \checkmark &  & & \checkmark & \checkmark
& Mesh; splats
& Scene capture, limited editing \\

RealityScan
& \checkmark &  & \checkmark &  & 
& Mesh; splats
& object masking \\

NeRFCapture
& \checkmark & \checkmark &  & \checkmark & 
& Images, poses
& Designed for NeRF dataset acquisition \\

Record3D
&  & \checkmark &  &  & 
& RGBD, Mesh
& Depth-video capture, no reconstruction \\

LogMeal
& \checkmark & \checkmark &  &  & 
& Metadata, poses
& Food-oriented visual-inertial capture \\

Solaya
& \checkmark & \checkmark & \checkmark & \checkmark & \checkmark
& Mesh, Images, poses, splats
&  Guided capture, object masking \\

OnePose++
&  & \checkmark &  &  & 
& RGBD, Mesh
& Depth-video capture, no reconstruction \\

V\&F App
&  & \checkmark &  &  & 
& RGB, IMU
& video capture, no reconstruction \\

\bottomrule
\end{tabular}
\end{table}

\subsection{Capturing Patterns}
\label{sec:capturing_patterns}
A foundational pattern in view-dependent reconstruction is the LLFF \cite{mildenhall2019local} grid-like sampling strategy. In this method, cameras capture a structured 2D grid of overlapping views spanning a scene’s frontal plane, adhering to a stipulation that disparity between adjacent images remains below 64 pixels to prevent geometric inconsistencies and aliasing \cite{mildenhall2021nerf}. The dense overlap ensures robust multiperspective stereo constraints, enabling multiplane image reconstruction with high-fidelity appearance. Practical research (e.g., Zip-NeRF \cite{barron2023zip}, LLFF-based 3DGS initializations) further leverages grid-like input for efficient volumetric rendering \cite{wu2024local}. When transferred to Gaussian Splatting, this grid framework supports placing Gaussians within an implicit frustum, thereby improving optimization of positional and opacity parameters, reducing training time to minutes, and facilitating real-time rendering.
\begin{figure}[htb]
    \centering
    \includegraphics[trim={7cm 3cm 8cm 3cm},clip,width=0.24\linewidth]{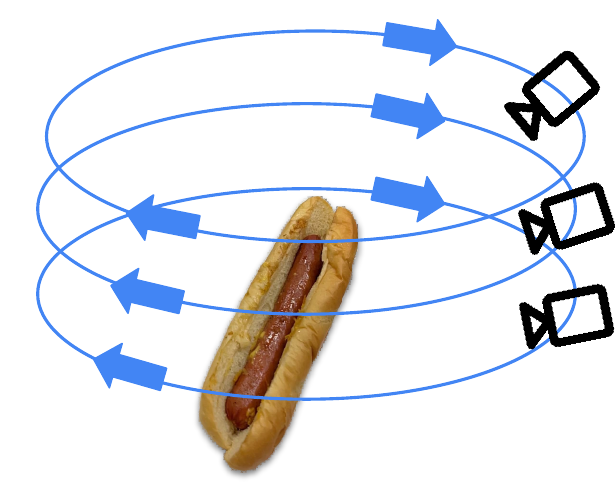}
    \includegraphics[trim={7cm 3cm 8cm 3cm},clip,width=0.24\linewidth]{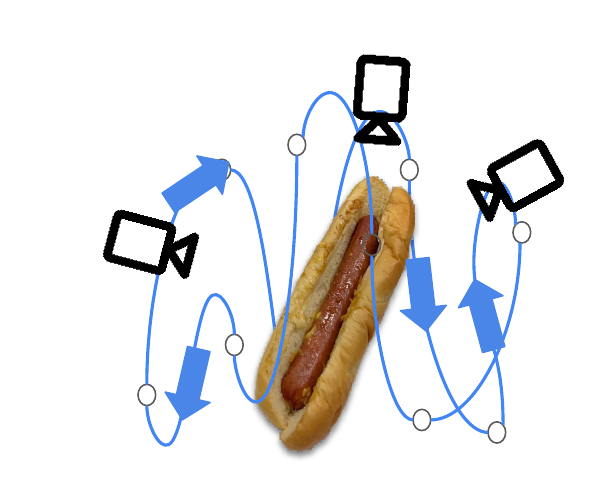}
    \includegraphics[trim={6cm 2cm 6cm 2cm},clip,width=0.24\linewidth]{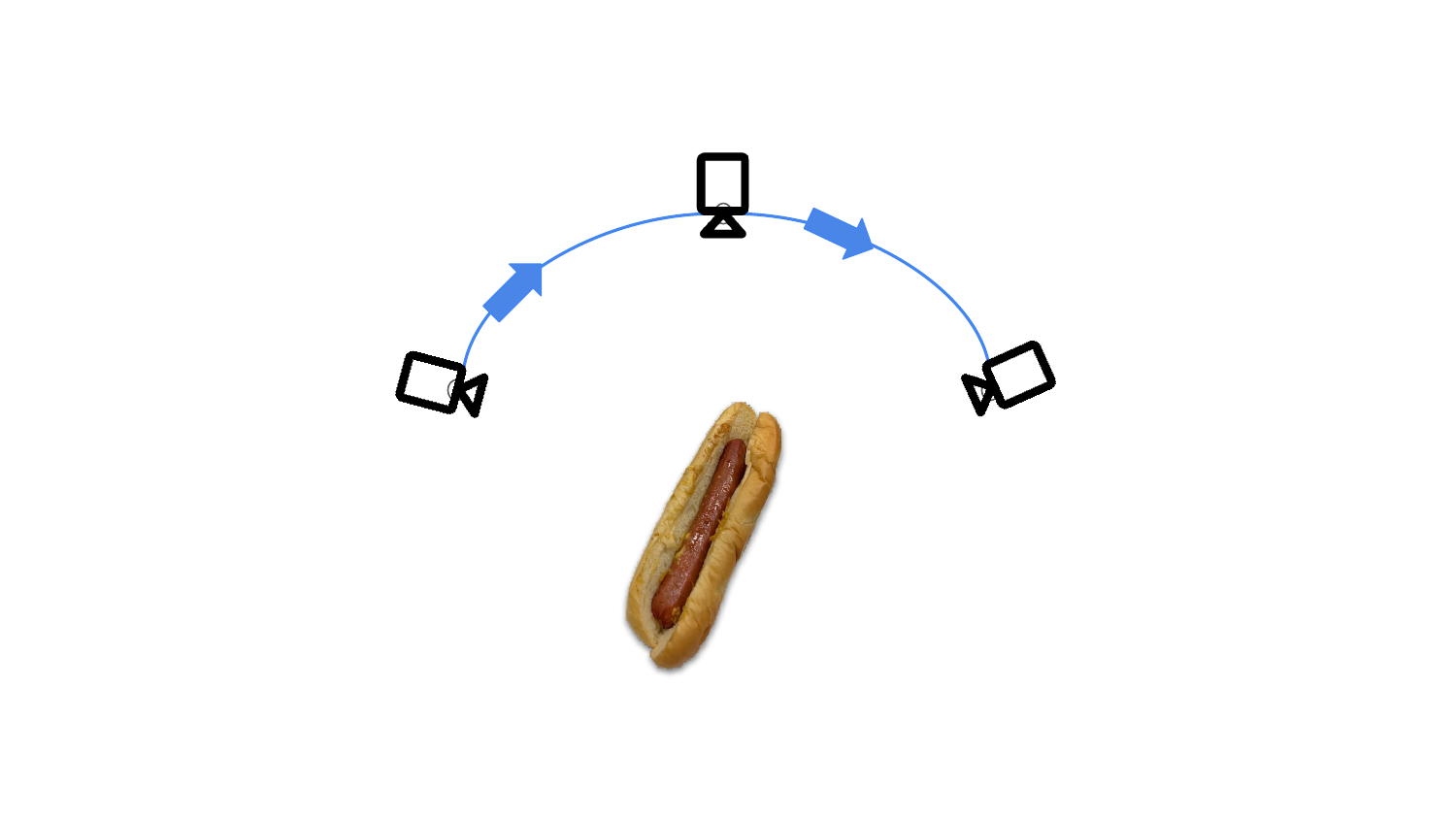}
    \includegraphics[trim={8cm 4cm 8cm 4cm},clip,width=0.24\linewidth]{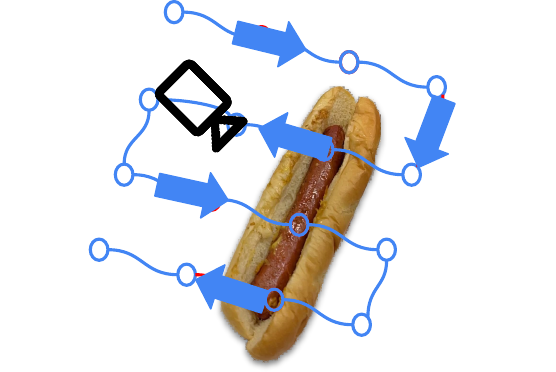}
    \caption{To the left, there are three donuts, followed by the sinusoidal pattern illustrating capturing trends, while in the third pattern, there is a circular arc pattern. The final pattern is grid-like in design. }
    \label{fig:capturing-patterns}
\end{figure}
For capturing individual objects—such as food items or smaller artifacts—a prevalent approach is the object-centric hemisphere (or circular arc) trajectory. Here, the camera circumnavigates the object along a fixed-radius arc, varying its elevation to achieve complete and symmetric coverage, particularly over surfaces that are occluded in planar capture. By sampling camera views across a semi-spherical shell, researchers ensure consistent angular variation and a uniform parallax distribution, thereby promoting accurate surface reconstruction. This approach is widely used in controlled photogrammetry and NeRF setups (i.e., Blender dataset and Zip-NeRF \cite{barron2023zip} dataset), especially in isolated-object scenarios where dense surface sampling is desirable and view diversity is paramount.

Hybrid trajectories, such as the “three-donut” and sine-function orbits, combine horizontal circular motion with vertical displacement. In the three-donut scheme, the camera performs three circular laps at distinct elevations, providing enhanced coverage of both top and side views and minimizing occlusion-related reconstruction gaps. In contrast, sinusoidal patterns path the camera along a helical path defined by a sinusoid, providing continuous elevation variation during rotation. These 3D camera paths enforce rich view overlap across all spatial dimensions, thereby improving structural consistency—particularly for complex curved or textured surfaces in both indoor and outdoor environments.

The spiral (or “rainbow”) pattern, introduced in ScanNeRF \cite{de2023scannerf} and later adapted in Gaussian-splatting-based pipelines for immersive storytelling, describes a smooth, arching trajectory resembling a slime or rainbow. This pattern typically swirls from ground level towards zenith while circumnavigating the scene center. By sweeping through both radial and vertical axes, spiral captures enable dense volumetric sampling with minimal view redundancy, which is effective in initializing splat positions in latent volumetric space. The result is a balanced distribution of viewpoint data that supports high-resolution geometric fidelity and stable view synthesis without the need for structured rigs or a checkerboard. 

In Fig.~\ref{fig:capturing-patterns}, each pattern presents distinct advantages: grid sampling simplifies stereo registration and optimization, hemisphere offers object-focused coverage, donut/sinusoid paths allow elevation-varied multi-view overlap, and spiral trajectories sustain continuous, volumetric sampling. In practical pipeline design, these patterns are often combined parametrically to balance reconstruction fidelity, training speed, and capture convenience in NeRF and Gaussian-Splatting systems.

\subsection{Baselines architectural design}
\label{sec:baselines_architectural_design}
In this section, we briefly discuss the 20 state-of-the-art models and their architectures. All of the baseline models were trained on the FoodSeg103 dataset and evaluated on the BenchSeg dataset. 

\begin{figure}[htb]
    \centering
    \includegraphics[width=1.0\linewidth]{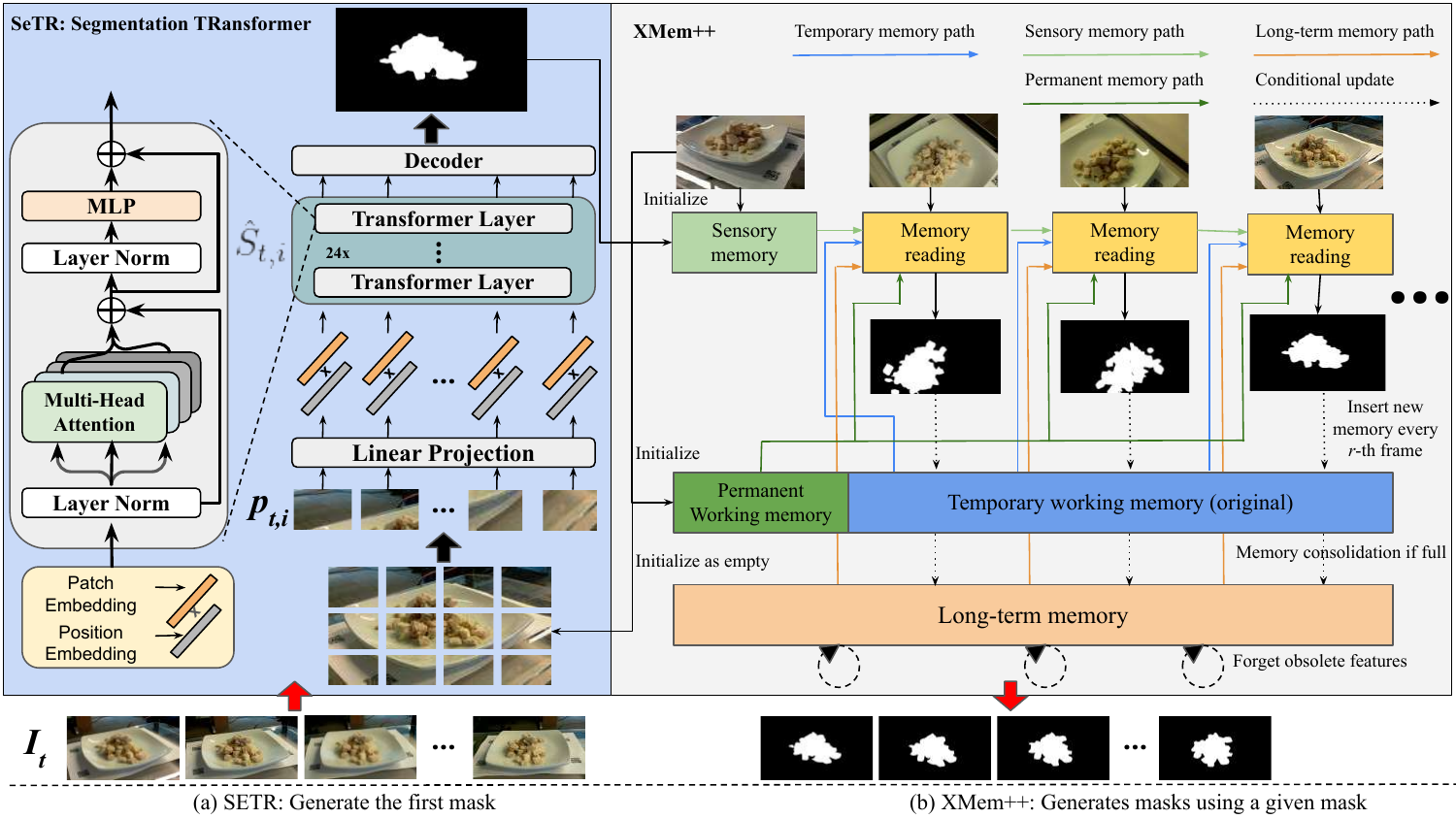}
    \caption{Overview of the FoodMem video segmentation framework, comprising a transformer‑based per‑frame segmentation module and a memory‑augmented temporal tracking module that stores and retrieves feature memories to ensure temporally coherent object masks across video frames}
    \label{fig:baseline_foodmem}
\end{figure}

\paragraph{FoodMem \cite{he2024metafood}} is a two-phase deep learning framework designed for accurate and near-real-time segmentation and tracking of food items in video sequences. The architecture decouples the problem into (1) a transformer-based segmentation module that generates initial high-quality mask predictions for food regions on a per-frame basis using a segmentation transformer backbone, and (2) a memory-augmented tracking module that propagates and refines these masks temporally across subsequent video frames by maintaining a learned spatio-temporal memory of object appearances and positions, thereby ensuring temporal coherence and minimizing flicker artifacts common in frame-by-frame segmentation. The memory component stores intermediate mask features and leverages them to guide consistent mask tracking under variations in camera motion, lighting, and scene complexity, effectively bridging the gap between single-image segmentation and continuous video segmentation. By combining transformer-based representation learning with a memory-based temporal association mechanism, FoodMem achieves substantial improvements in segmentation stability, accuracy, and inference speed compared to existing state-of-the-art models, while requiring minimal hardware resources for real-time operation as shown in Fig.~\ref{fig:baseline_foodmem}.

\begin{figure}[htb]
    \centering
    \includegraphics[trim={0.5cm 2cm 1cm 2cm},clip,width=1.0\linewidth]{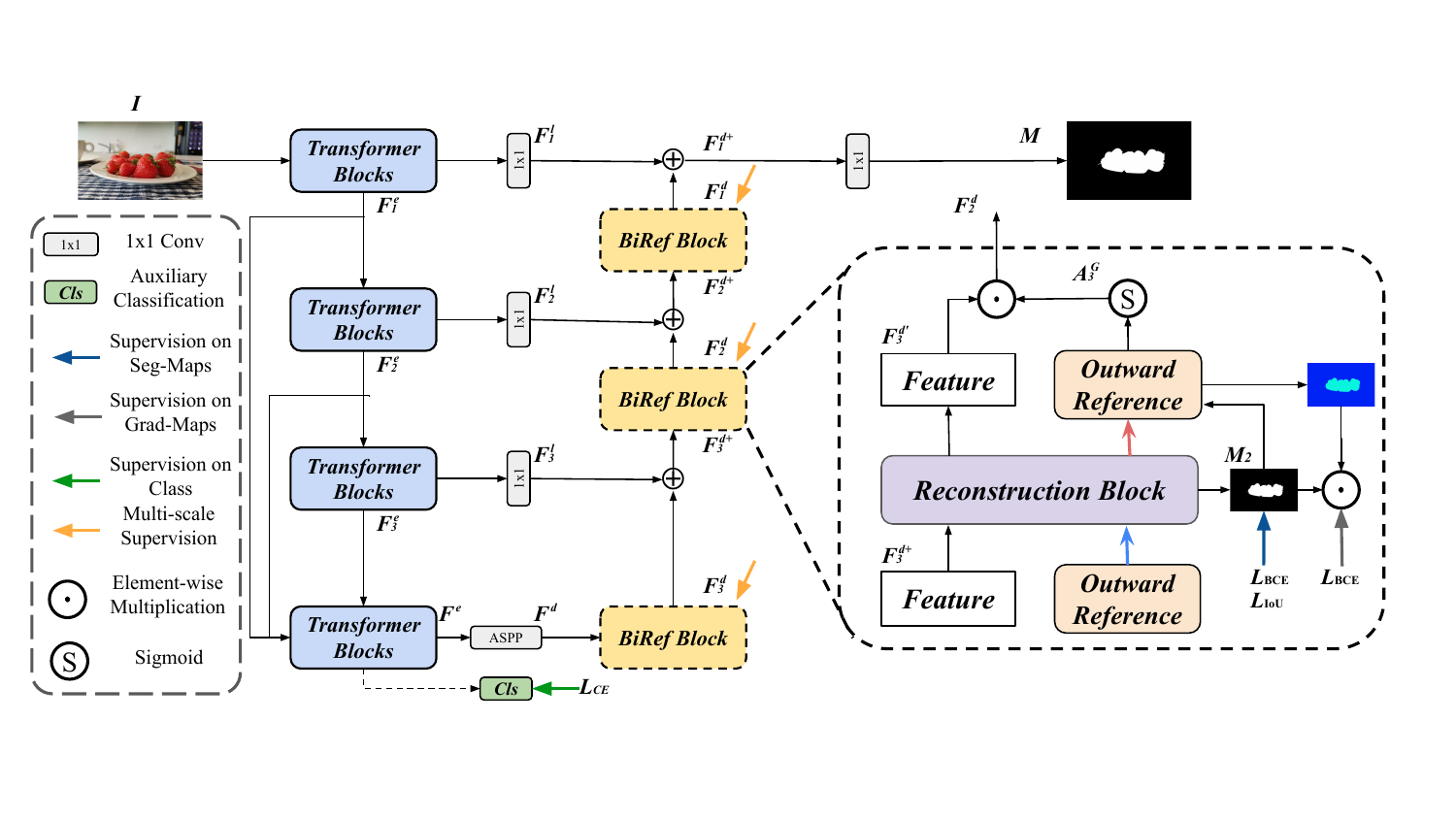}
    \caption{Diagram of BiRefNet’s bilateral reference segmentation network, highlighting the localization and reconstruction modules connected via bilateral reference paths that fuse high‑resolution inward image cues with outward gradient‑based structural features to improve boundary precision.}
    \label{fig:baseline_birefnet}
\end{figure}

\paragraph{BiRefNet \cite{zheng2024bilateral}} is a high-resolution, dichotomous image segmentation network that integrates a bilateral reference framework to jointly leverage global semantic cues and fine details for accurate binary mask prediction. The architecture is organized into two principal components: a localization module (LM) and a reconstruction module (RM) interconnected via the proposed Bilateral Reference (BiRef) mechanism. The localization module employs a hierarchical feature extractor (e.g., a transformer-based backbone) to capture multi-scale semantic representations and to facilitate object localization in high-resolution inputs. Features from the LM are propagated to corresponding stages in the RM through lateral connections. At the same time, the BiRef mechanism operates at each reconstruction stage to fuse inward references — original image patches that preserve high-resolution detail — with outward references derived from auxiliary gradient maps that emphasize fine structural information. This bilateral interaction enhances the model’s ability to maintain both global context and boundary precision during segmentation. Auxiliary gradient supervision and multi-stage hierarchical supervision are incorporated to refine the level of detail further and improve convergence. The combined effect of LM, RM, and BiRef achieves state-of-the-art performance on high-resolution binary segmentation benchmarks, with strong capabilities in salient object detection, camouflaged object extraction, and background removal, as shown in Fig.~\ref{fig:baseline_birefnet}.

\begin{figure}[htb]
    \centering
    \includegraphics[trim={4cm 0cm 4cm 0cm},clip,width=1.0\linewidth]{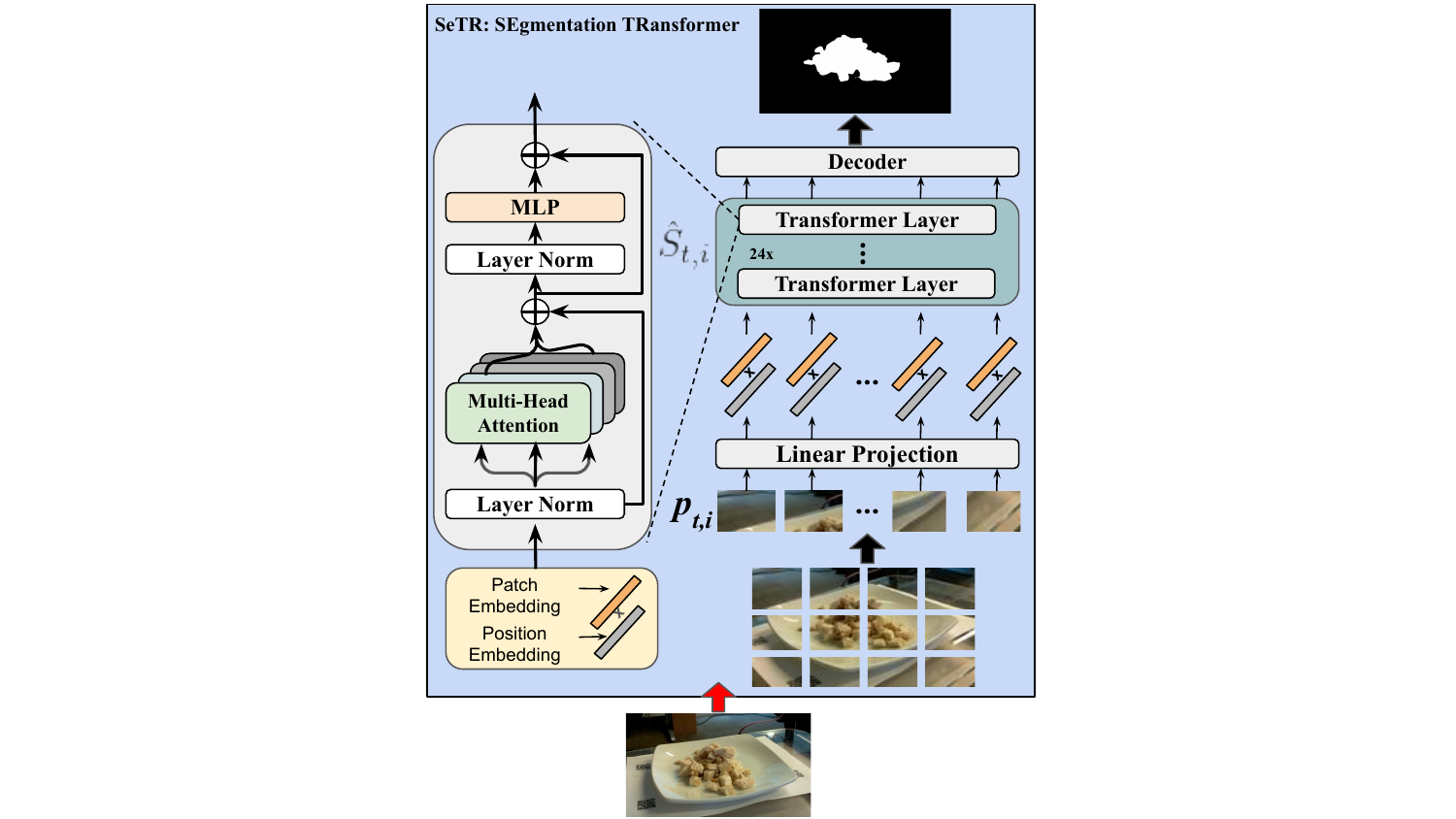}
    \caption{SeTR architecture diagram illustrating the reformulation of semantic segmentation as a sequence‑to‑sequence task, where an input image is split into patches, processed by a pure transformer encoder modeling global context, and decoded to dense pixel‑level predictions via appropriate upsampling decoders.}
    \label{fig:baseline_setr}
\end{figure}

\paragraph{SeTR \cite{zheng2021rethinking}} reformulates semantic segmentation as a sequence‑to‑sequence prediction problem by replacing the conventional convolutional encoder with a pure transformer that operates on a sequence of image patches. An input image is first decomposed into fixed‑size patches, and each patch is flattened and projected into a high‑dimensional embedding space with learned positional encodings; this sequence of patch embeddings is then processed by a multi‑layer transformer encoder that models global context at every layer through multi‑head self‑attention, eliminating the need for progressive spatial downsampling inherent in FCN‑based backbones. A lightweight decoder subsequently reconstructs high‑resolution semantic predictions from the transformer outputs, with variants including naive upsampling, progressive upsampling, and multi‑level feature aggregation to balance detail recovery and contextual integration. This architecture captures long‑range dependencies more effectively than traditional FCNs, leading to state‑of‑the‑art performance on major benchmarks such as ADE20K, Pascal Context, and Cityscapes, as shown in Fig~\ref{fig:baseline_setr}.

\begin{figure}[htb]
    \centering
    \includegraphics[trim={2cm 0.5cm 0cm 0cm},clip,width=1.0\linewidth]{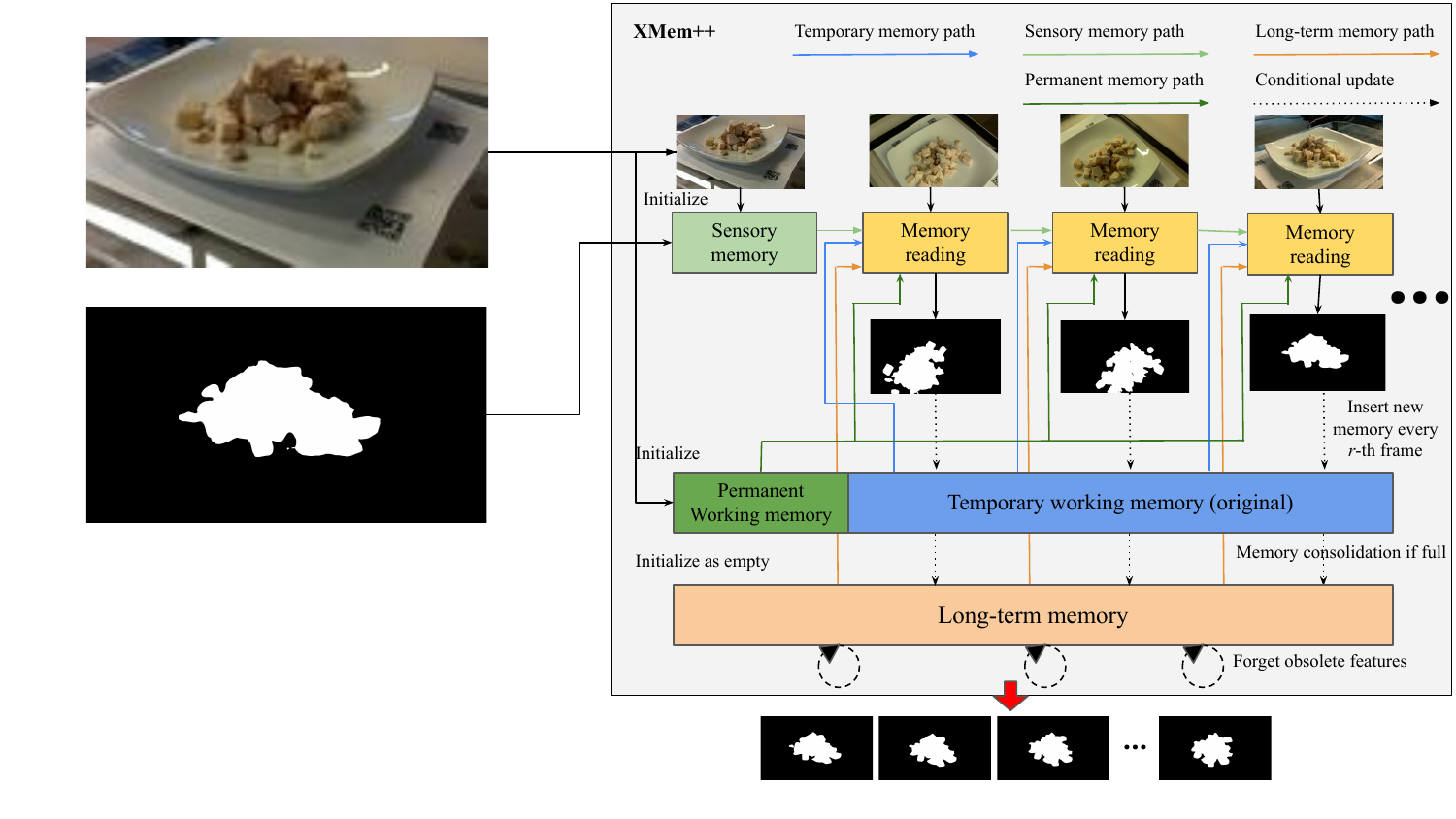}
    \caption{Illustration of XMem2’s memory‑augmented video object segmentation architecture, featuring tiered memory stores (working, long‑term, and permanent memory) and an attention‑based frame suggestion mechanism to maintain consistent segmentation across long sequences.}
    \label{fig:baseline_xmem2}
\end{figure}

\paragraph{XMem2 \cite{bekuzarov2023xmem++}} is a memory‑augmented semi‑supervised video object segmentation architecture that builds on the foundational XMem model by incorporating enhanced memory mechanisms for robust long‑range temporal segmentation. The core architecture maintains and dynamically updates feature memories that store deep representations of annotated reference frames and their masks, enabling the model to propagate object segmentation information across subsequent frames based on feature similarity rather than per‑frame learning. The original XMem \cite{cheng2022xmem} framework draws inspiration from the human Atkinson–Shiffrin memory model, employing multiple interconnected memory stores—such as short‑term and long‑term representations—to balance detailed recent features with compact long‑range context, and uses a memory-reading mechanism to retrieve relevant information for mask-prediction in query frames. XMem2 (also known as XMem++) extends this design by introducing a permanent memory module that retains multiple user‑provided annotations to improve segmentation consistency, and an attention‑based frame suggestion mechanism that selects the next best frames for annotation to optimize performance. This architecture operates in real time without retraining between inputs, delivers highly consistent masks with minimal annotated frames, and achieves production‑level performance on challenging video segmentation scenarios, as shown in Fig.~\ref{fig:baseline_xmem2}.

\begin{figure}[htb]
    \centering
    \includegraphics[trim={0cm 0cm 0cm 0cm},clip,width=1.0\linewidth]{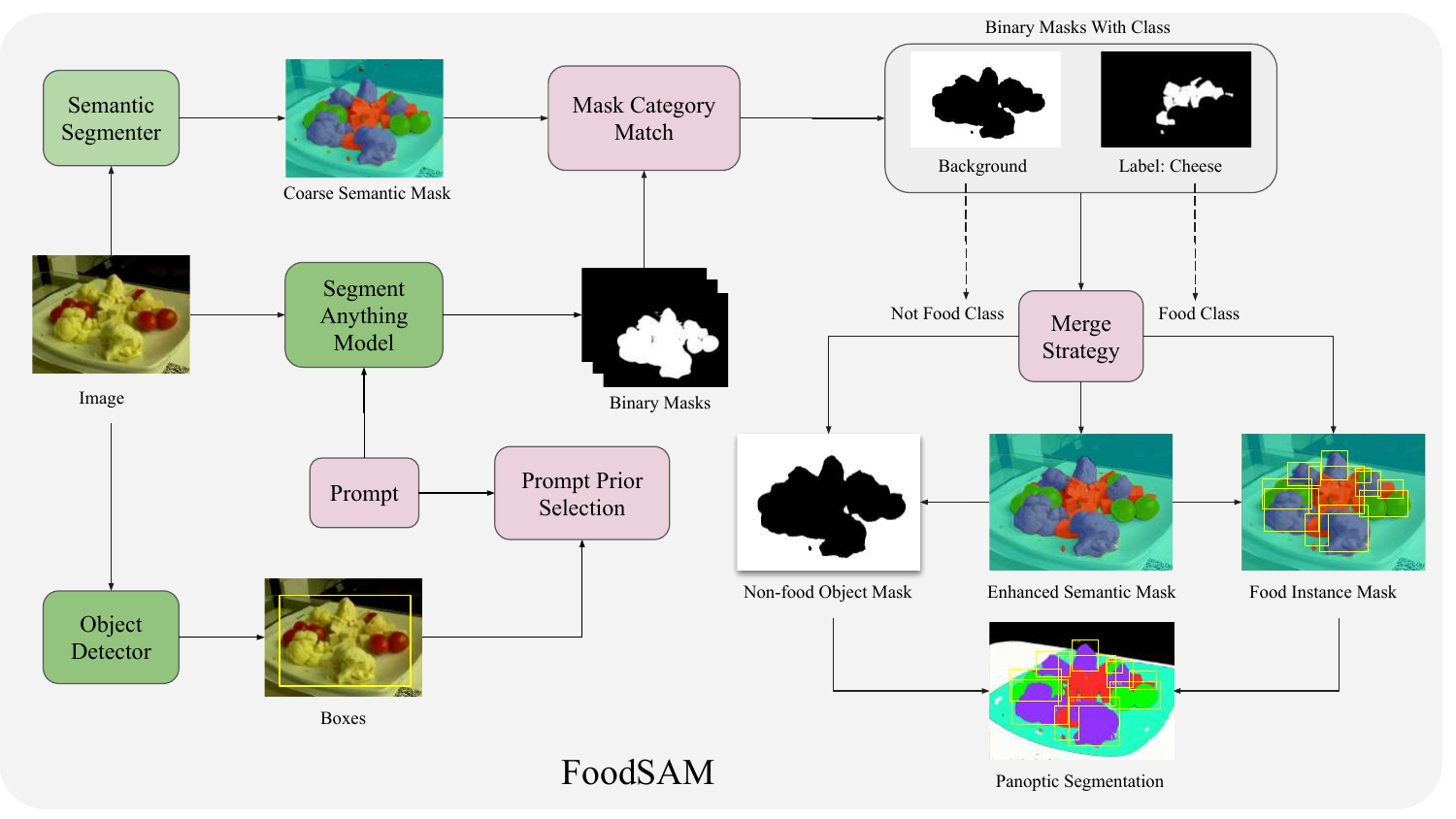}
    \caption{Diagram of the FoodSAM segmentation framework where class‑agnostic masks generated by SAM are fused with semantic segmentation outputs and integrated with an object detection branch to realize zero‑shot semantic, instance, and panoptic segmentation.}
    \label{fig:baseline_foodsam}
\end{figure}

\paragraph{FoodSAM \cite{lan2023foodsam}} is a modular, zero-shot food image segmentation framework that leverages the foundational Segment Anything Model (SAM) and integrates it with task-specific semantic and detection components to achieve multi-granularity segmentation. At its core, SAM generates high-quality class-agnostic binary masks from food images, providing dense region proposals without category labels. These masks are then fused with coarse semantic segmentation outputs produced by a dedicated semantic segmenter to inject class-specific information, thereby enhancing the fidelity and interpretability of the segmented food regions. To handle fine-grained scene understanding beyond semantic maps, FoodSAM incorporates an object detection module that identifies non-food elements, enabling unified panoptic segmentation of both food and background objects. The architecture also supports instance-level segmentation by treating each food ingredient as an independent entity and associating SAM’s mask instances with semantic categories via a mask–category matching strategy. Additionally, inspired by recent advances in promptable segmentation, FoodSAM integrates a prompt-prior selection mechanism that allows flexible, interactive segmentation using various prompt types (e.g., points, bounding boxes), thereby extending its zero-shot capabilities across semantic, instance, panoptic, and promptable tasks. This hybrid design balances the generalization strength of SAM with domain-specific segmentation quality through synergistic fusion and multi-module coordination, as shown in Fig.~\ref{fig:baseline_foodsam}.

\begin{figure}[htb]
    \centering
    \includegraphics[trim={0cm 1cm 0cm 1cm},clip,width=1.0\linewidth]{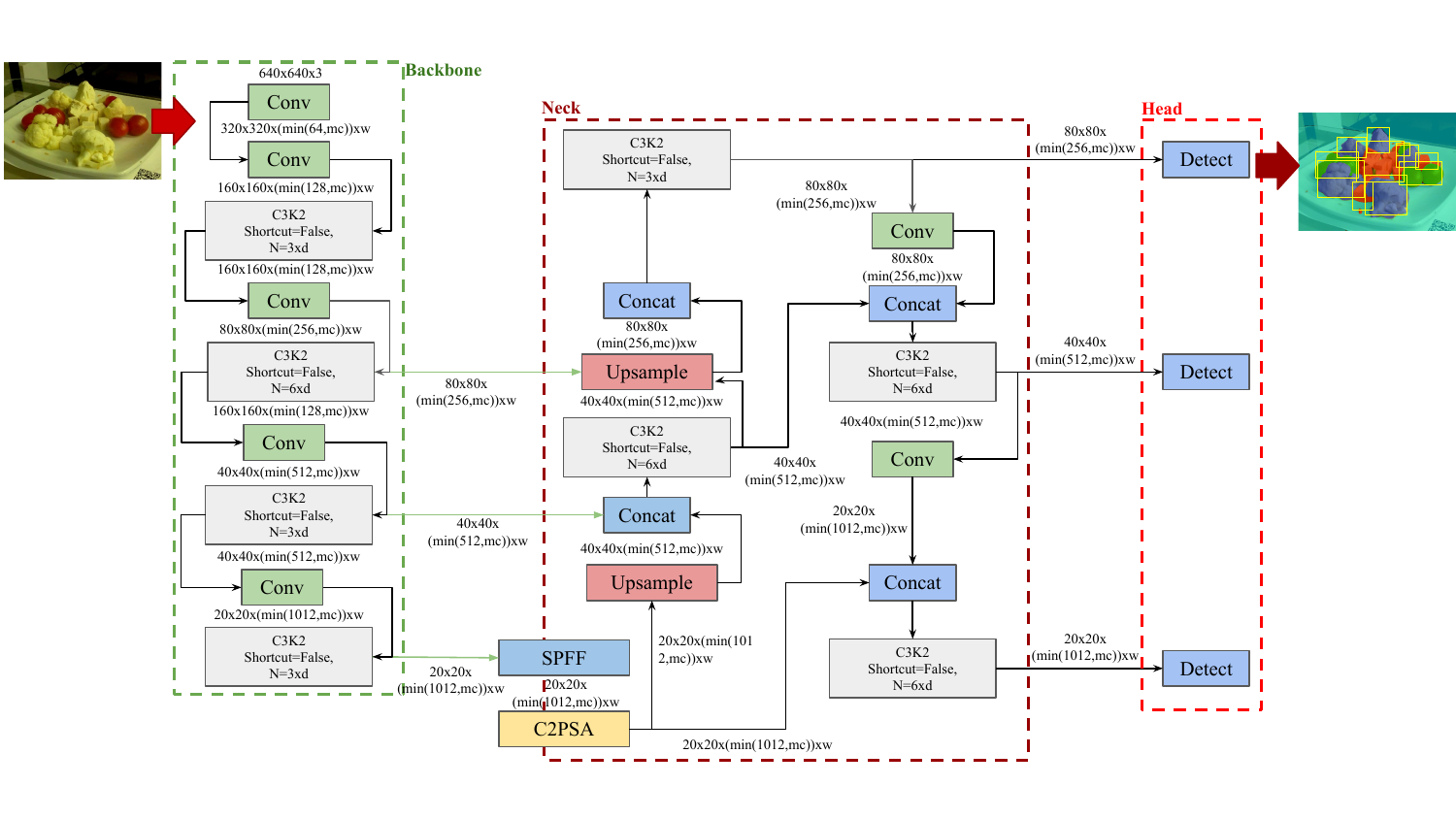}
    \caption{Overview of YOLOv11’s architecture with a shared backbone and multi‑scale neck feeding both detection and lightweight segmentation heads, where segmentation masks are predicted in real time alongside object bounding boxes for efficient instance segmentation.}
    \label{fig:basline_yolov11}
\end{figure}

\paragraph{YOLOv11 \cite{sapkota2025yolo}} achieves segmentation by extending the base object-detection architecture with a dedicated segmentation head that generates high‑resolution, pixel‑level instance masks in addition to bounding boxes. The model retains its standard backbone and neck (multi‑scale feature extractor and feature fusion modules) to produce rich hierarchical feature maps, which are then fed into a specialized segmentation head trained jointly with the detection head to predict class‑specific masks for each detected object. This design enables YOLOv11‑seg to perform real‑time instance segmentation by leveraging shared features for both tasks, with effective multi‑scale feature aggregation and lightweight attention mechanisms enhancing mask precision—especially for small and overlapping objects—without significantly increasing inference cost. Variants such as YOLO11n‑seg through YOLO11x‑seg balance computational efficiency and segmentation accuracy across model scales, making the architecture suitable for real‑time applications requiring both spatial localization and detailed mask delineation, as shown in Fig.~\ref{fig:basline_yolov11}.

\begin{figure}[htb]
    \centering
    \includegraphics[trim={0cm 3cm 0cm 3cm},clip,width=1.0\linewidth]{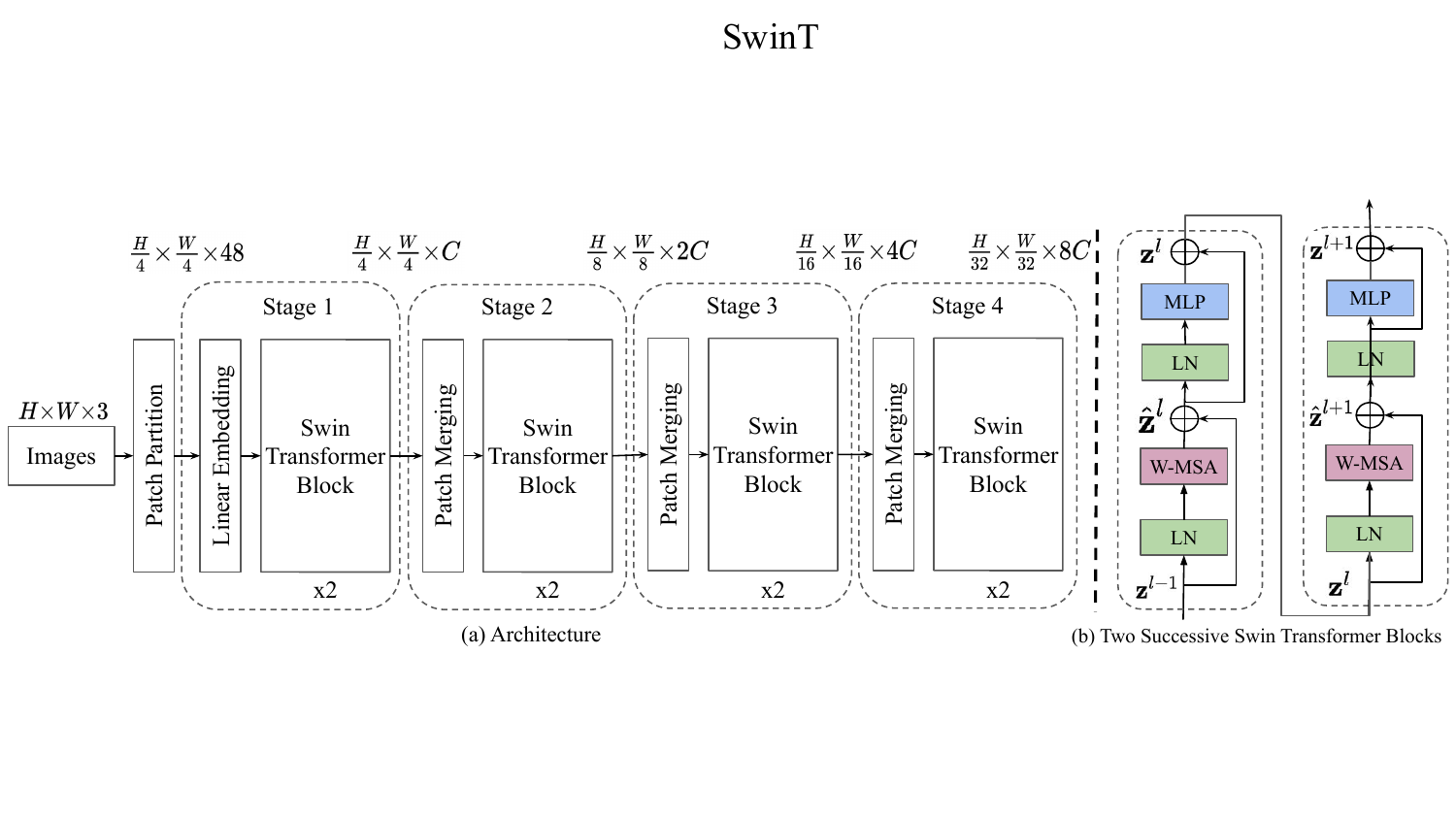}
    \caption{Hierarchical Swin Transformer architecture with patch partitioning, multi‑stage window‑based self‑attention (W‑MSA/SW‑MSA), and patch‑merging layers that progressively build a feature pyramid for dense vision tasks.}
    \label{fig:baseline_swin}
\end{figure}

\paragraph{Swin \cite{liu2021swin}} is a hierarchical vision Transformer that introduces shifted window-based self-attention to efficiently capture both local and global image features across multiple scales. Starting from an input image, it partitions the image into non-overlapping patches, which are linearly embedded into token representations, analogous to tokens in language models. The model is organized into multiple stages, each comprising a sequence of Swin Transformer blocks and patch-merging layers that progressively reduce spatial resolution while increasing feature dimensionality, thereby forming a hierarchical feature pyramid similar to convolutional backbones but learned via self-attention. Within each block, window-based multi-head self-attention (W-MSA) computes self-attention locally within fixed-size windows, reducing computational complexity to linear in the input size, and shifted window multi-head self-attention (SW-MSA) alternates window partitioning between consecutive layers to enable cross-window connections and facilitate information exchange across adjacent regions. Residual connections, layer normalization, and intermediate MLPs are interleaved with these attention modules, yielding a scalable backbone that supports high-resolution dense-prediction tasks such as object detection and semantic segmentation, achieving state-of-the-art performance, as shown in Fig. ~\ref{fig:baseline_swin}.

\begin{figure}[htb]
    \centering
    \includegraphics[trim={0cm 0cm 0cm 0cm},clip,width=1.0\linewidth]{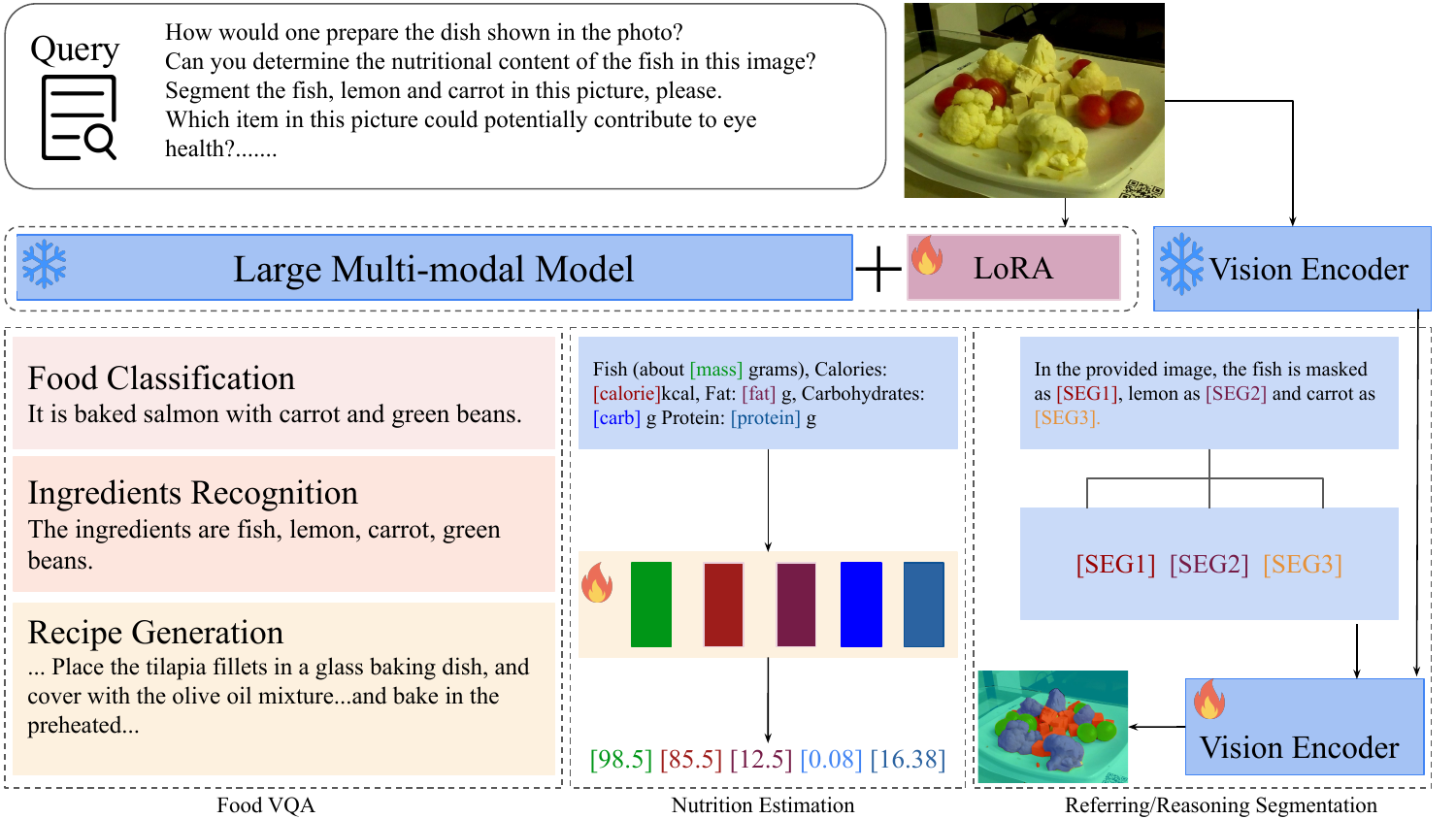}
    \caption{Illustration of FoodLMM’s multimodal architecture, showing the integration of a shared vision–language backbone with task‑specific tokens and output heads for segmentation, recipe generation, ingredient recognition, and nutritional estimation, where visual and textual features are fused and routed to dedicated decoders for structured outputs.}
    \label{fig:baseline_foodlmm}
\end{figure}

\paragraph{FoodLMM \cite{yin2025foodlmm}} Fig.~\ref{fig:baseline_foodlmm} extends the architecture of a foundation Large Multi-Modal Model (LMM) with domain-specific adaptations to address the unique challenges of food understanding. At its core, FoodLMM builds on a pre-trained multimodal backbone (e.g., a vision-encoder + language model fusion such as LLaVA) that jointly processes visual and textual inputs. To enable food-centric tasks beyond pure text output, the architecture introduces novel task-specific tokens and heads: segmentation tokens that feed into a segmentation decoder to generate one or more food instance masks, and regression heads attached to dedicated nutritional tokens that predict quantitative nutritional values. For standard vision–language tasks (e.g., food classification, ingredient recognition, recipe generation), FoodLMM uses text-generation heads operating on the shared multimodal latent representations; for segmentation and nutrition estimation, the additional tokens’ hidden states are routed to specialized decoders that produce structured outputs. This multi-head design enables the model to handle heterogeneous outputs (text, masks, numerical estimates) in a unified architecture. Training involves a two-stage strategy: first, multi-task learning on diverse food benchmarks using an instruction-following paradigm to imbue broad domain competence; second, fine-tuning on curated multi-round dialogue and reasoning segmentation datasets to enhance conversational reasoning and complex segmentation capabilities

\begin{figure}[htb]
    \centering
    \includegraphics[trim={5cm 0cm 5cm 0cm},clip,width=1.0\linewidth]{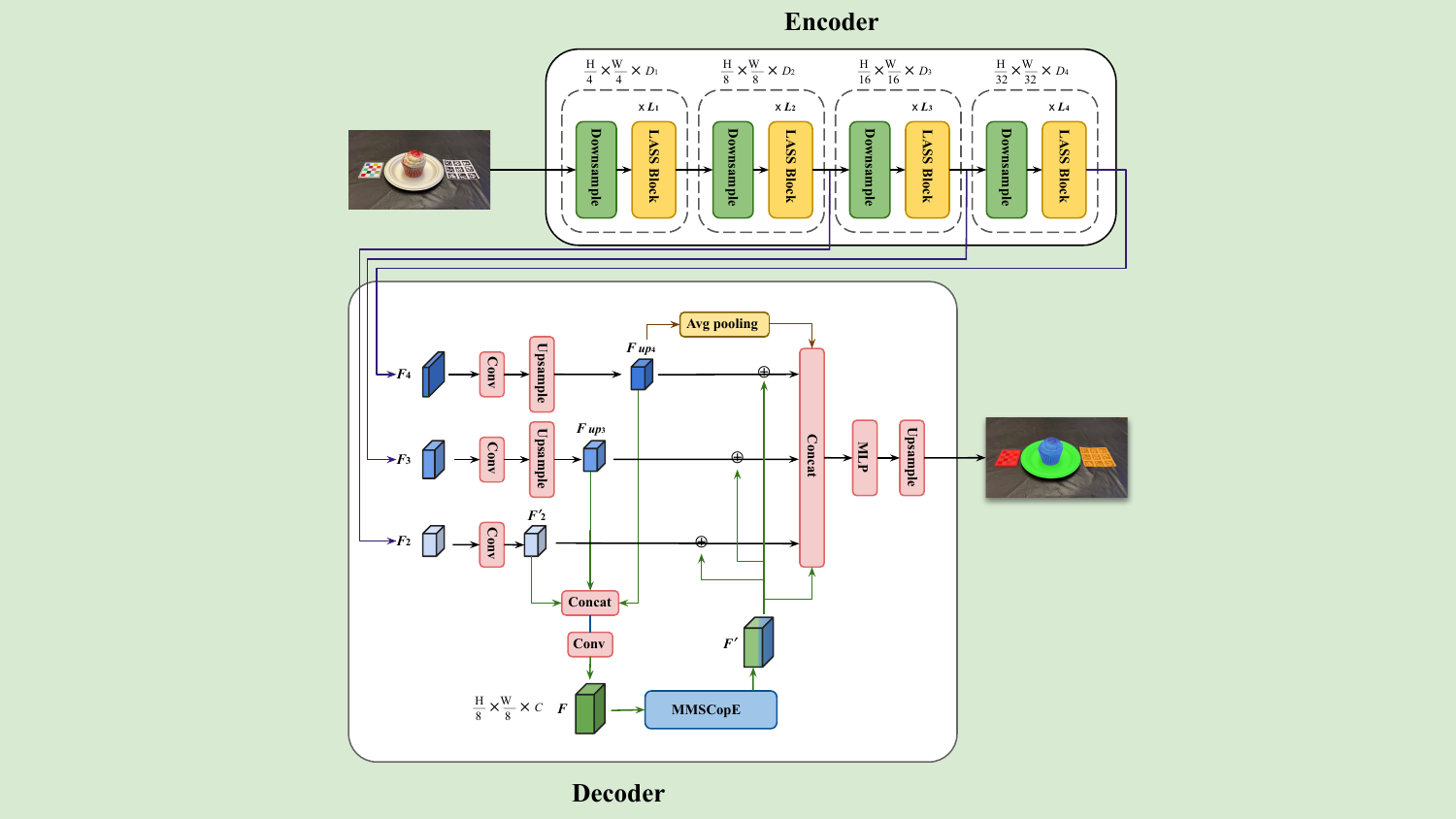}
    \caption{SegMAN schematic illustrating its hybrid encoder with sliding local attention and state‑space blocks for omni‑scale feature encoding, coupled with an MMSCopE decoder that adaptively aggregates multi‑scale contextual features for enhanced semantic segmentation.}
    \label{fig:baseline_segman}
\end{figure}

\paragraph{SegMan \cite{fu2025segman}} is a novel semantic segmentation architecture that synergistically combines local attention mechanisms with dynamic state-space modeling to achieve omni-scale context representation while preserving fine-grained detail. The core of the model is a hybrid SegMAN Encoder composed of a hierarchical, multi-stage feature extractor in which each stage integrates sliding local attention (e.g., neighborhood attention) and a 2D-Selective-Scan state-space block to efficiently model long-range global context with linear computational complexity, while also capturing local dependencies critical for precise boundary delineation. Feature maps from successive stages form a resolution pyramid that balances spatial detail and abstract semantics. In the SegMAN Decoder, the MMSCopE (Mamba-based Multi-Scale Context Extraction) module adaptively aggregates multi-scale contextual features from the encoder outputs, enabling rich cross-resolution representation that scales with input size and enhances segmentation quality. The combined design achieves efficient global context fusion, high-fidelity local encoding, and adaptive multi-scale decoding, leading to improved segmentation accuracy and lower computational cost than prior approaches on benchmarks such as ADE20K, Cityscapes, and COCO-Stuff, as shown in Fig.~\ref{fig:baseline_segman}.

\begin{figure}[!htbp]
    \centering
    \includegraphics[trim={0cm 0.5cm 0.5cm 0cm},clip,width=1.0\linewidth]{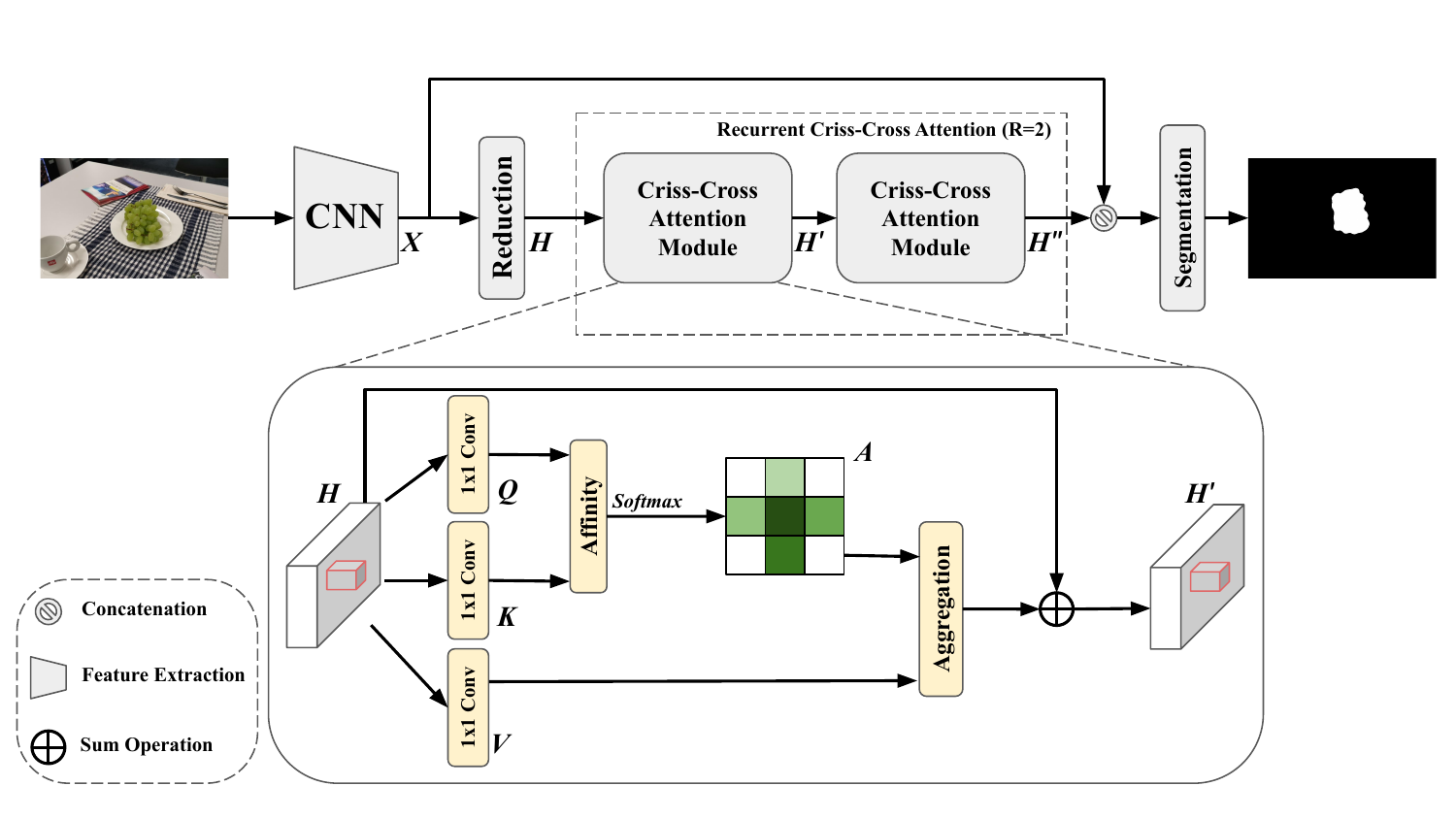}
    \caption{Schematic of the Criss‑Cross Network showing the recurrent criss‑cross attention module that captures long‑range contextual dependencies by attending along horizontal and vertical directions with linear complexity, integrated atop a backbone feature extractor.}
    \label{fig:baseline_ccent}
\end{figure}

\paragraph{CCNet \cite{huang2019ccnet}}is a semantic segmentation architecture that introduces an efficient Criss-Cross Attention mechanism to capture dense, long-range contextual information across the entire input image. Rather than computing full non-local self-attention over all pixel pairs with quadratic complexity, CCNet’s criss-cross attention module computes attention along horizontal and vertical directions for each pixel, aggregating context from all positions along these paths with linear complexity in the spatial dimensions. A recurrent application of this module enables each pixel to iteratively aggregate dependencies across the whole image, effectively expanding the receptive field without incurring high computational cost. This recurrent criss-cross attention is integrated on top of a standard backbone feature extractor and paired with a category-consistent loss to enforce discriminative feature learning. Compared to conventional non-local blocks, CCNet significantly reduces GPU memory usage and FLOPs while maintaining or improving segmentation accuracy, achieving state-of-the-art performance on benchmarks such as Cityscapes and ADE20K, as shown in Fig.~\ref{fig:baseline_ccent}.

\begin{figure}[!htbp]
    \centering
    \includegraphics[trim={2.cm 1cm 6cm 1cm},clip,width=1.0\linewidth]{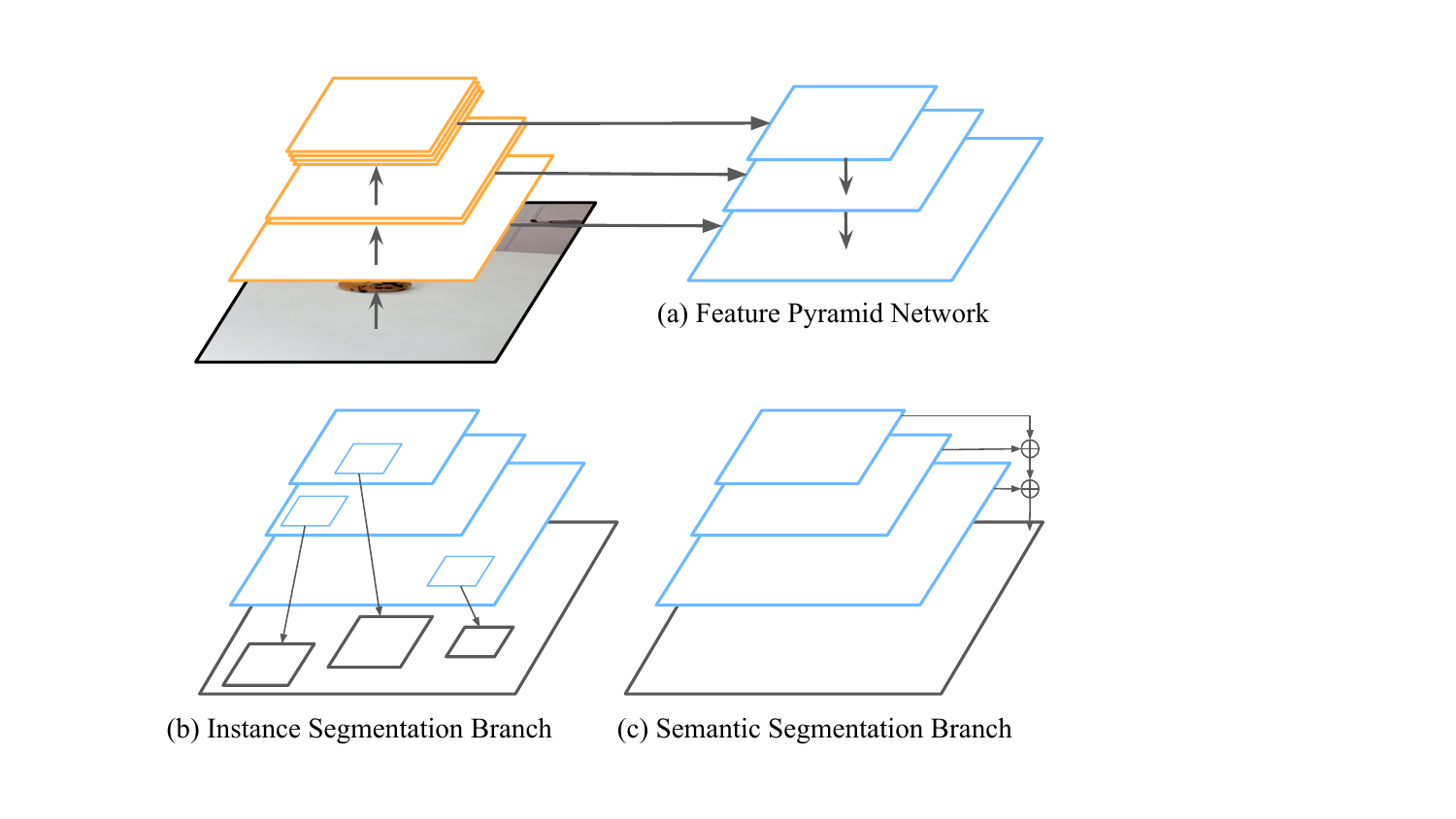}
    \caption{which extends a shared multi‑scale Feature Pyramid Network (FPN) backbone with parallel instance and semantic segmentation branches to jointly predict instance masks and dense per‑pixel class labels for unified panoptic segmentation.}
    \label{fig:baseline_fpn}
\end{figure}

\paragraph{FPN \cite{Kirillov_2019_CVPR}} extends the classical Feature Pyramid Network backbone to jointly address instance and semantic segmentation within a unified architecture, enabling efficient panoptic segmentation. Building on a shared FPN backbone, which constructs multi‑scale feature maps by combining bottom‑up backbone features with top‑down pathways via lateral connections, Panoptic FPN incorporates a lightweight semantic segmentation branch alongside the existing region‑based instance segmentation branch (e.g., Mask R‑CNN) to produce dense pixel‑level class predictions for “stuff” categories and instance masks for “thing” categories in parallel. The semantic branch upsamples and merges multi‑scale FPN features into a high‑resolution representation, typically using stacked convolution, normalization, and bilinear upsampling layers, before generating per‑pixel class scores. The instance branch operates via standard region proposal and mask heads. Joint training with balanced instance and semantic losses enables the model to produce coherent panoptic outputs with improved efficiency and competitive accuracy relative to separate single‑task networks, as shown in Fig.~\ref{fig:baseline_fpn}.

\subsection{Quality Metrics Details}
We detailed the quality metric to enhance the reproducibility. All metrics are computed from the standard confusion-matrix parameters: true positives (TP), false positives (FP), false negatives (FN), and true negatives (TN). For a predicted segmentation $\mathrm{pred}$ and ground truth $\mathrm{gt}$,
\begin{equation}
\mathrm{TP} = \sum(\mathrm{pred} = 1 \wedge \mathrm{gt} = 1)
\label{eq:tp}
\end{equation}
\begin{equation}
\mathrm{FP} = \sum(\mathrm{pred} = 1 \wedge \mathrm{gt} = 0)
\label{eq:fp}
\end{equation}
\begin{equation}
\mathrm{FN} = \sum(\mathrm{pred} = 0 \wedge \mathrm{gt} = 1)
\label{eq:fn}
\end{equation}
\begin{equation}
\mathrm{TN} = \sum(\mathrm{pred} = 0 \wedge \mathrm{gt} = 0)
\label{eq:tn}
\end{equation}

Using \eqref{eq:tp}–\eqref{eq:tn}, we define Precision and Recall as
\begin{equation}
\mathrm{Precision} = \frac{\mathrm{TP}}{\mathrm{TP} + \mathrm{FP}}
\label{eq:precision}
\end{equation}
\begin{equation}
\mathrm{Recall} = \frac{\mathrm{TP}}{\mathrm{TP} + \mathrm{FN}}
\label{eq:recall}
\end{equation}
which quantify robustness to false positives and false negatives, respectively \cite{papadopoulos2017training}. Based on \eqref{eq:precision} and \eqref{eq:recall}, the F1-score is given by
\begin{equation}
\mathrm{F1} = \frac{2 \times \mathrm{Precision} \times \mathrm{Recall}}{\mathrm{Precision} + \mathrm{Recall}}
\label{eq:f1}
\end{equation}
while IoU and Accuracy are computed as
\begin{equation}
\mathrm{IoU} = \frac{\mathrm{TP}}{\mathrm{TP} + \mathrm{FP} + \mathrm{FN}}
\label{eq:iou}
\end{equation}
\begin{equation}
\mathrm{Accuracy} = \frac{\mathrm{TP} + \mathrm{TN}}{\mathrm{TP} + \mathrm{TN} + \mathrm{FP} + \mathrm{FN}}
\label{eq:accuracy}
\end{equation}

In semantic segmentation, mean Intersection-over-Union (mIoU) is obtained by first computing the per-class IoU using \eqref{eq:iou} and then averaging these values across all semantic classes, providing a fair per-class measure regardless of class imbalance \cite{everingham2010pascal}. Similarly, mean Accuracy (mAcc) is computed by averaging the per-class accuracies derived from \eqref{eq:accuracy}, rather than relying on global pixel accuracy, which would otherwise be dominated by highly frequent classes \cite{long2015fully}.

\paragraph{Average Precision (AP)}
Because BenchSeg uses a single foreground class (food vs.\ background), we report AP as the area under the precision--recall curve obtained by sweeping the foreground threshold $\tau \in [0,1]$ on the predicted probability map. This corresponds to the standard AUPRC for binary segmentation. We report AP for the foreground class (food) and denote it as mAP for consistency with prior food-segmentation papers, although it reduces to a single-class AP in our setting.

Each metric emphasizes a different aspect of model behaviour. Mean IoU (mIoU) measures spatial overlap between prediction and ground-truth masks on a class-wise basis and is relatively robust to class imbalance \cite{everingham2010pascal}. Mean Accuracy (mAcc) measures pixel classification accuracy within each class, balancing the influence of rare and frequent categories \cite{long2015fully}. In our binary setting, AP (denoted as mAP for consistency with prior food-segmentation work) summarizes the precision--recall trade-off over all thresholds, i.e., the area under the precision—recall curve for the foreground (food) class. Precision \eqref{eq:precision} focuses on the reliability of positive predictions, whereas Recall \eqref{eq:recall} reflects the completeness of recovered target regions \cite{papadopoulos2017training}. The F1-score \eqref{eq:f1} summarizes the trade-off between Precision and Recall as a single scalar, and the IoU \eqref{eq:iou} offers an intuitive measure of spatial alignment quality between predicted and ground-truth masks under challenging conditions.

\subsection{Additional Experimental Results}
\label{sec:additional_experimental_results}
Table \ref{tab:3d_comparisons_tracker}, Table \ref{tab:FoodKit_2d_comparisons}, Table \ref{tab:MTF_2d_comparisons}, and Table \ref{tab:N5K_2d_comparisons} show additional qualitative results on FKit, MTF, and N5k datasets. Table~\ref{tab:efficiency} summarizes computational efficiency across baselines. We report mean and 95th-percentile GPU utilization to capture efficiency and sustained peak load, together with mean and maximum GPU memory usage to reflect resource footprint. Runtime is measured as the average time per sample, derived from system timestamps, and provides a practical indicator of deployability.

\begin{table}[!htbp]
\centering
\caption{Computational efficiency across baselines.
We report GPU utilization, memory footprint, and runtime.
Methods are sorted in descending order by mean GPU. For each metric, the best, second-best, and third-best results are highlighted in \textbf{bold}, \uline{underline}, and \textit{italics}, respectively. The efficiency results are reported for a representative N5k scene as a microbenchmark; full-benchmark averages are provided in the supplementary material.}
\label{tab:efficiency}
\begin{tabular}{lccccc}
\toprule
\textbf{Method} & 
\multicolumn{2}{c}{\textbf{GPU (\%)}} & 
\multicolumn{2}{c}{\textbf{Memory (MiB)}} &  \textbf{Time (ms)} \\
\cmidrule(lr){2-3} \cmidrule(lr){4-5} \cmidrule(lr){6-6}
 & $\blacktriangle$Mean↓ & 95th↓ & Mean↓ & Max↓ & Avg↓ \\
\midrule
Y+S2 & 51.87 & 100.0 & 2441.83 & \textit{3665.0} & 106.63 \\
\rowcolor{gray!12}
Swin-S & 45.62 & 100.0 & 5010.27 & 9291.0 & \uline{101.56} \\
Swin-B & 45.19 & 100.0 & 5386.42 & 10386.0 & \uline{101.56} \\
\rowcolor{gray!12}
CCNet-Re & 34.20 & 100.0 & 3598.93 & 8440.0 & 101.59 \\
FPN-Re & 29.17 & 99.0 & 3682.03 & 11893.0 & \textit{101.6} \\
\rowcolor{gray!12}
CCNet & 26.50 & 100.0 & 3427.04 & 9578.0 & \textbf{101.5} \\
SeTM & 14.84 & 100.0 & 3362.07 & 9708.0 & 106.21 \\
Y+X2 & \textit{8.53} & \uline{28.0} & \uline{1263.69} & \textbf{1980.0} & 107.56 \\
\rowcolor{gray!12}
SeTM+X2 & \uline{5.63} & \textit{30.25} & \textit{1329.62} & 8870.0 & 105.56 \\
YOLO & \textbf{2.44} & \textbf{10.9} & \textbf{950.62} & \uline{2729.0} & 103.14 \\
\bottomrule
\end{tabular}

\vspace{2pt}
\raggedright
\tiny
\textbf{Abbreviations:}
Y+X2: YOLO+XMem2; CCNet-Re: CCNet-Relem; Swin-S/B: Swin Small/Base; SeTM/SeTN: SeTR-MLA/Naive; S2: SAM2; X2: XMem2.

\textbf{Note:} the runtime includes the model load time. 
\end{table}

\begin{table}[!htbp]
    \centering
    \caption{Qualitative comparison of segmentation outputs across all evaluated configurations, including YOLO, SegMan, YOLO+SAM2, YOLO+XMem2, SegMan+SAM2, SegMan+XMem2, the ground-truth masks, and the corresponding RGB frames. The examples illustrate differences in boundary precision, temporal consistency, and robustness under challenging food scenes. }
    \setlength{\tabcolsep}{1pt}
    \begin{tabular}{c|ccccc}
    \hline
    Method & Segmentor & +SAM2 & +XMem2 & GT & RGB \\
    \hline
    
\multirow{3}{*}{Yolo} &

\includegraphics[width=0.17\textwidth]{Fig_R1/chocolate_croissant_976_YOLO_binary.png} &
\includegraphics[width=0.17\textwidth]{Fig_R1/chocolate_croissant_976_YOLO_SAM2.png} &
\includegraphics[width=0.17\textwidth]{Fig_R1/chocolate_croissant_976_YOLO_XMEM2_n1_binary.png} &
\includegraphics[width=0.17\textwidth]{Fig_R1/chocolate_croissant_976.png} &
\includegraphics[width=0.17\textwidth]{Fig_R1/chocolate_croissant_976_RGB.png} \\

& \includegraphics[width=0.17\textwidth]{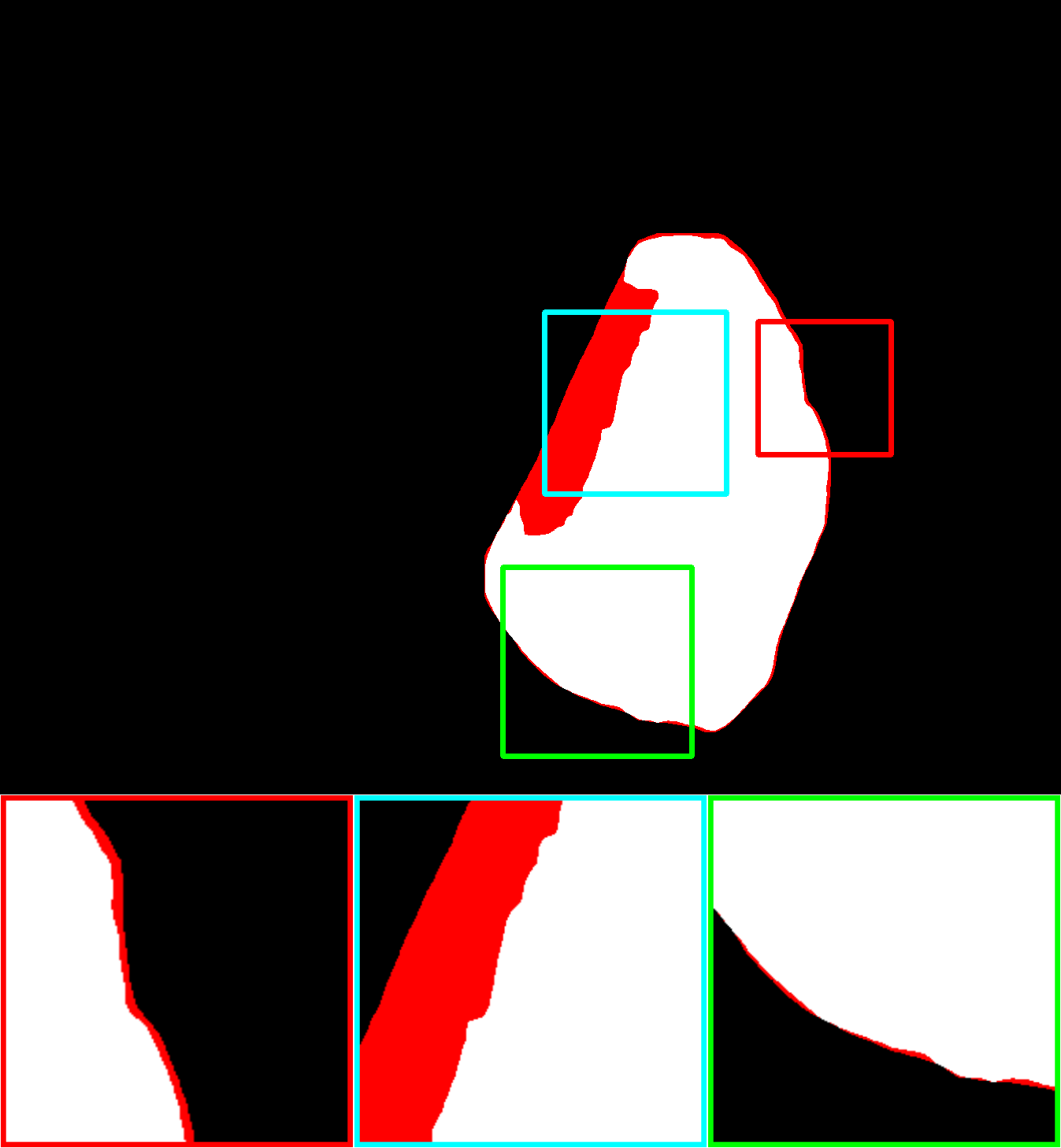} &
\includegraphics[width=0.17\textwidth]{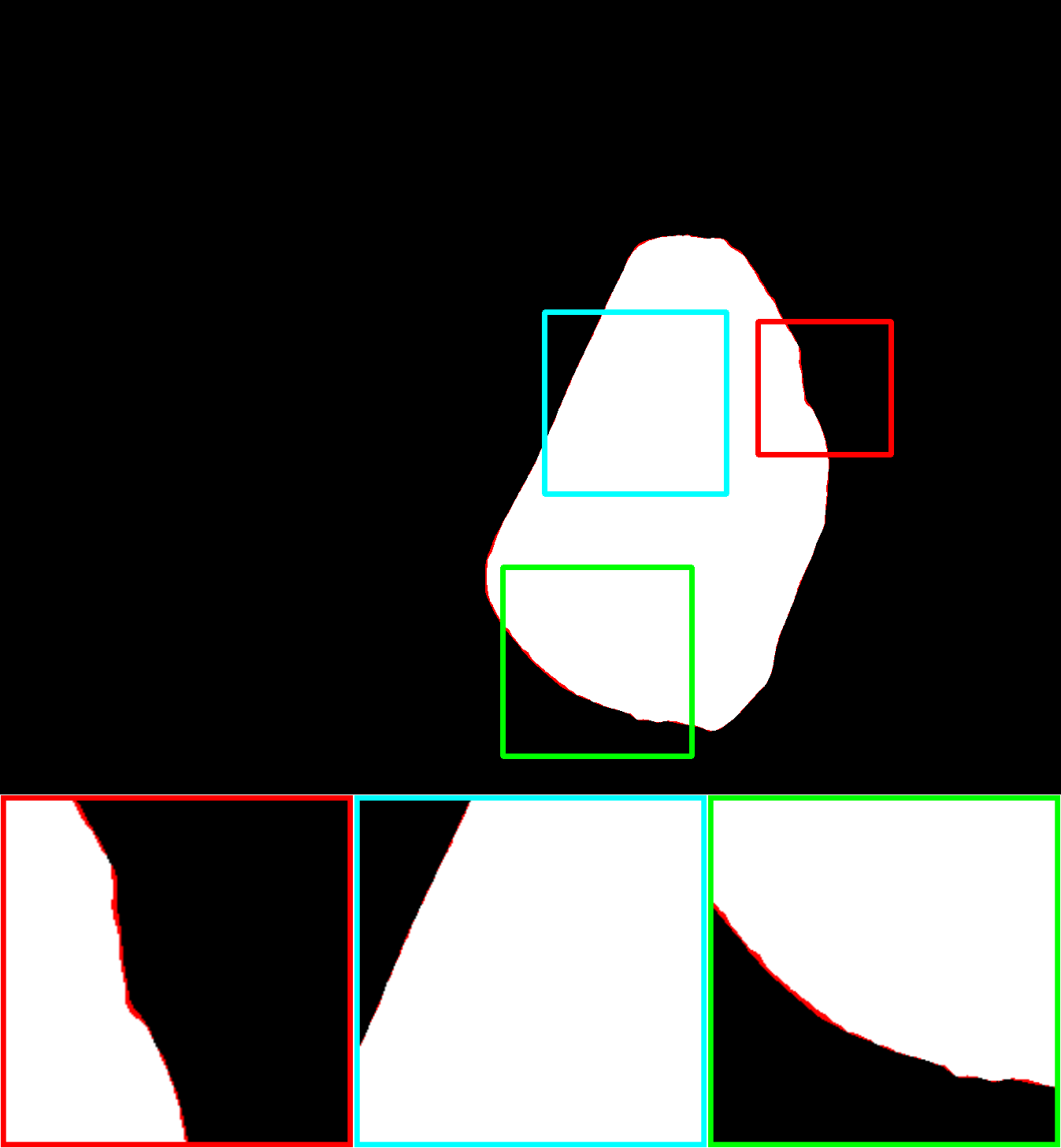} &
\includegraphics[width=0.17\textwidth]{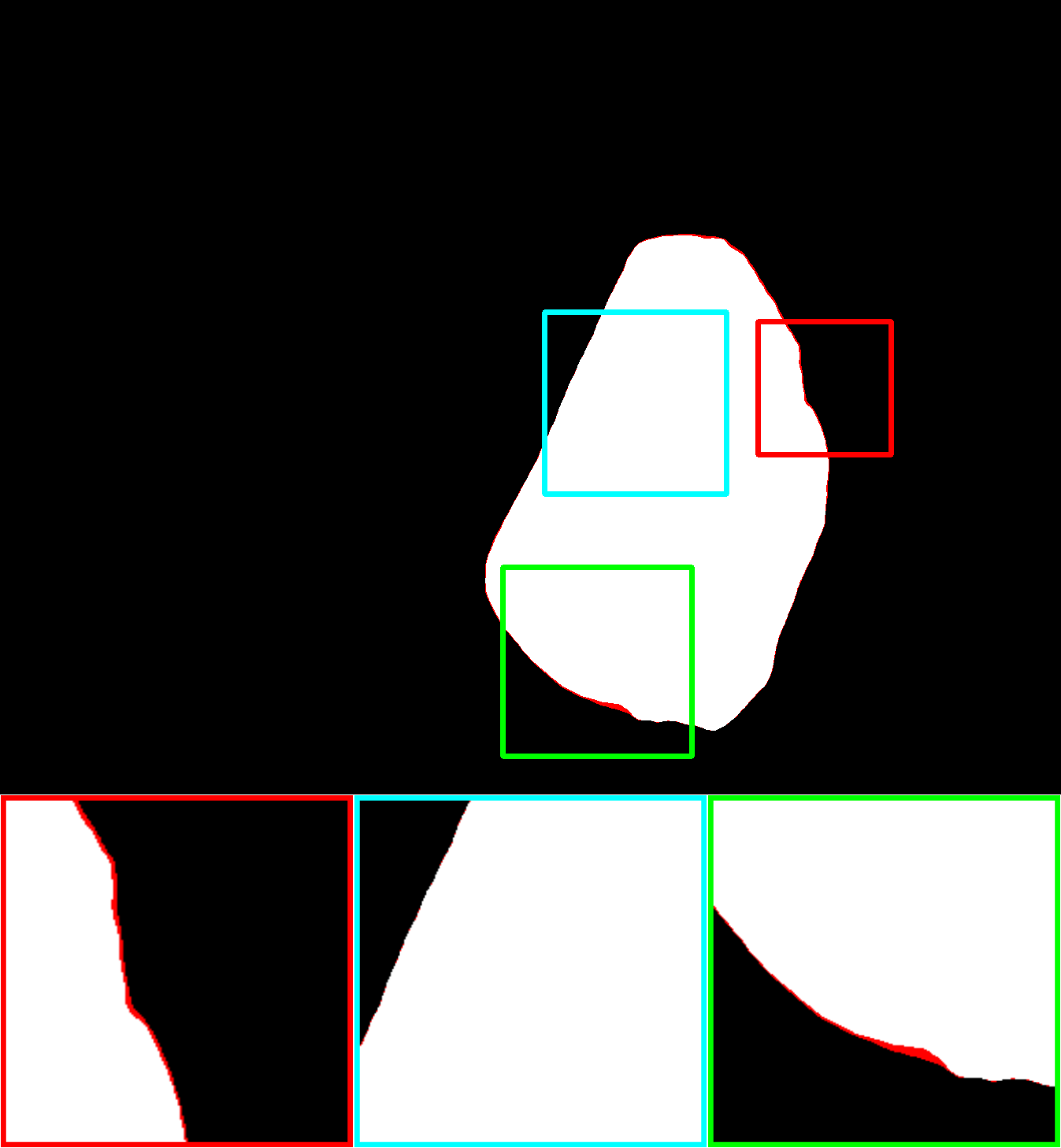} &
\includegraphics[width=0.17\textwidth]{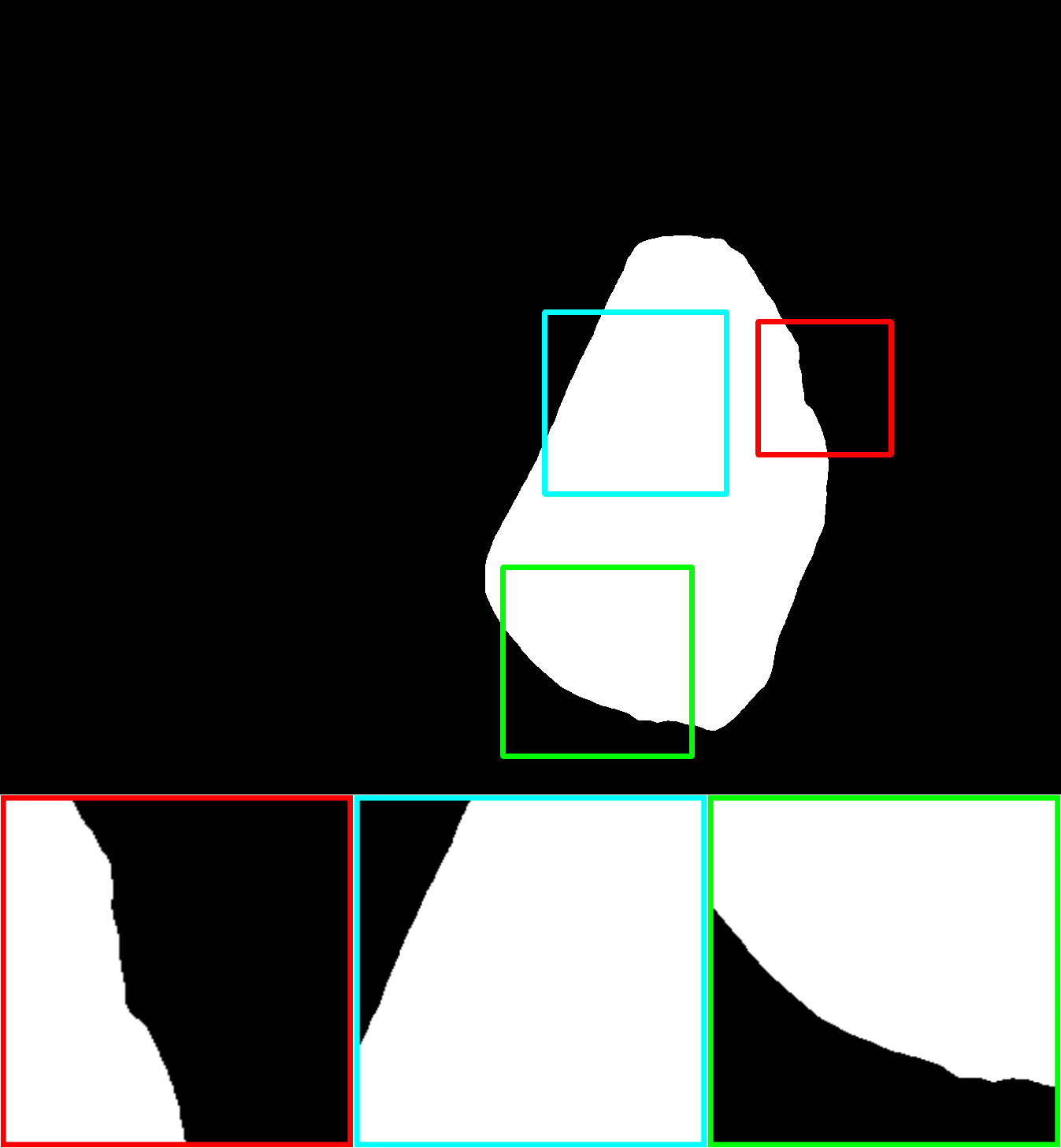} &
\includegraphics[width=0.17\textwidth]{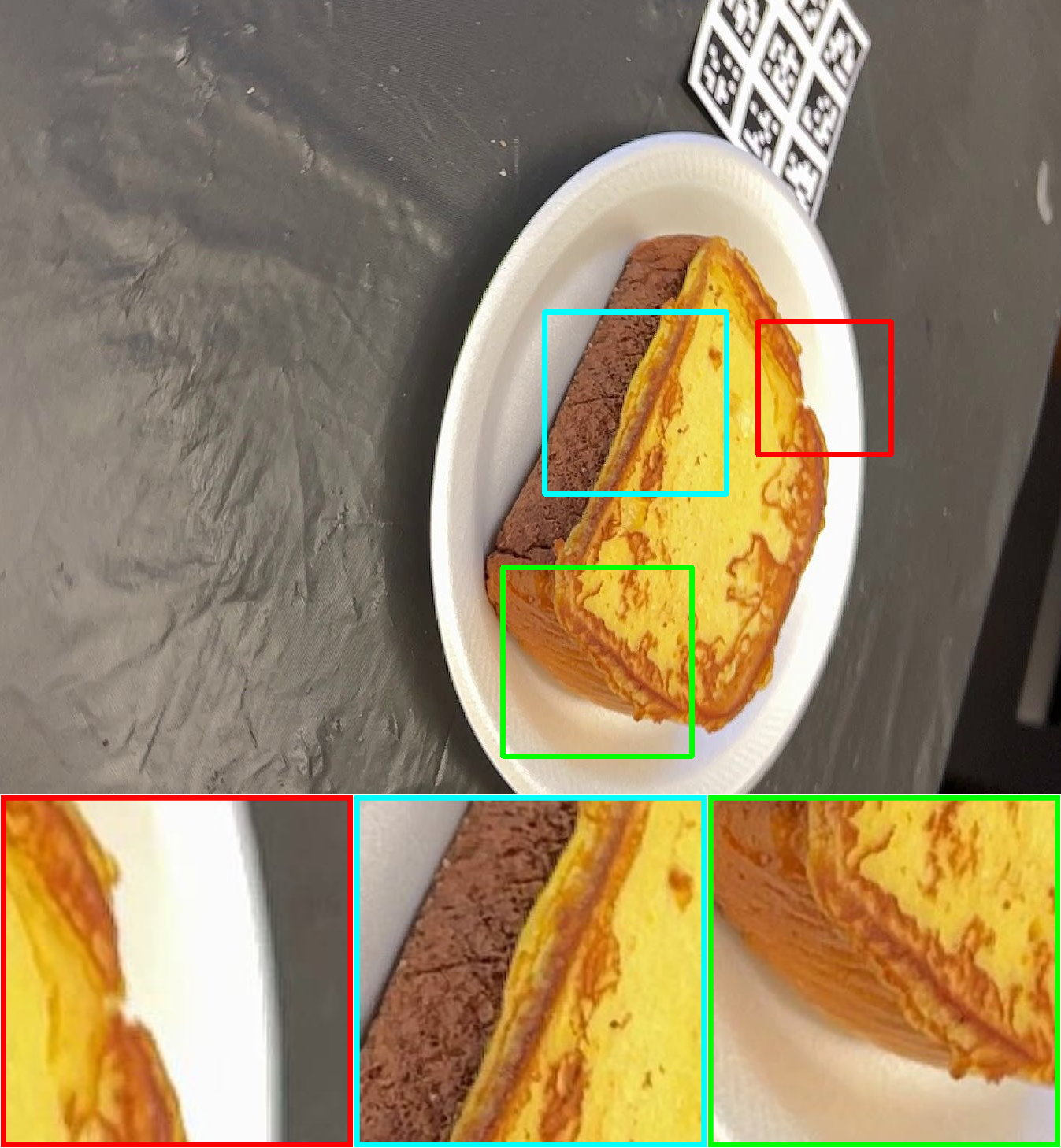} \\

& \includegraphics[width=0.17\textwidth]{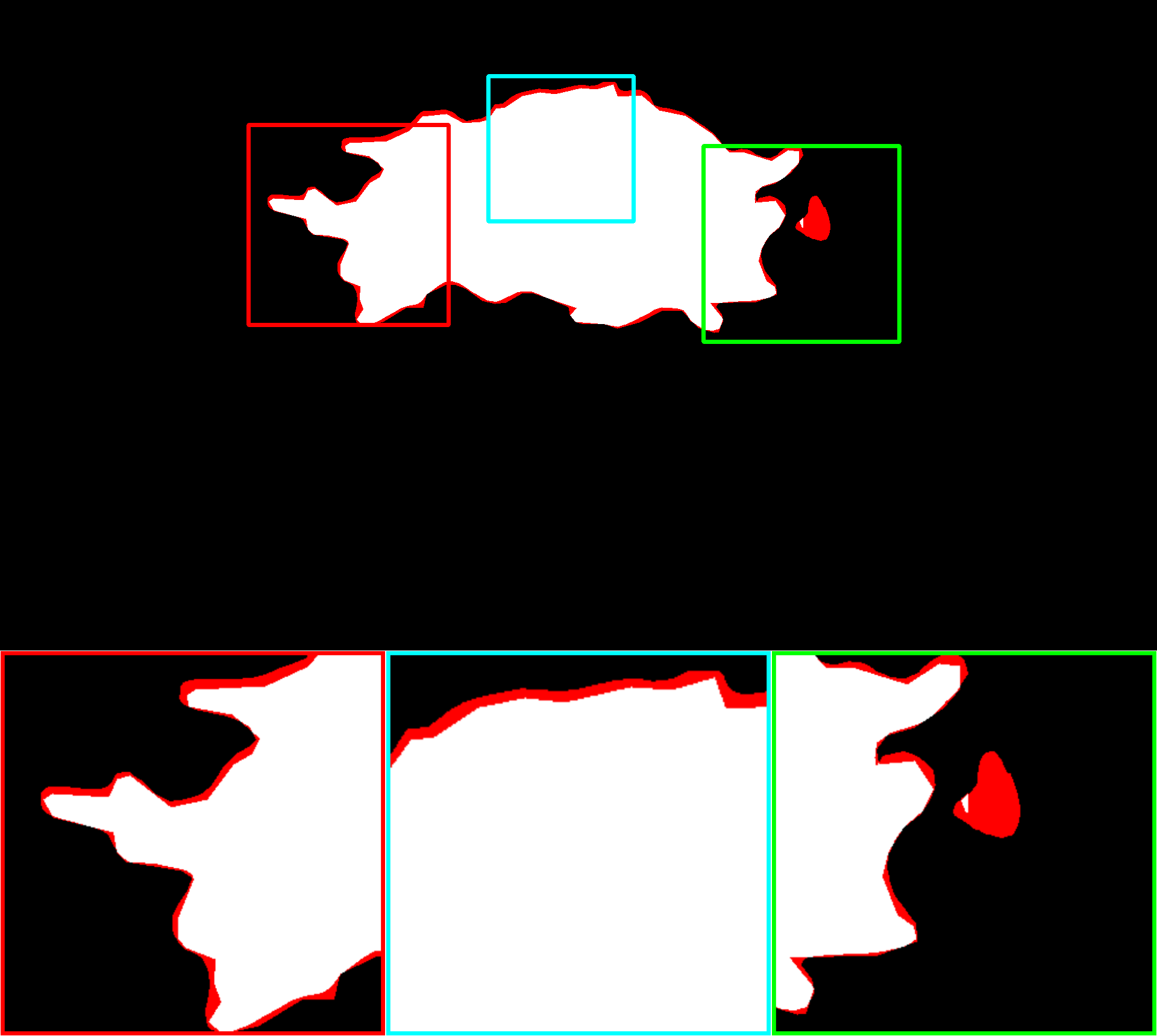} &
\includegraphics[width=0.17\textwidth]{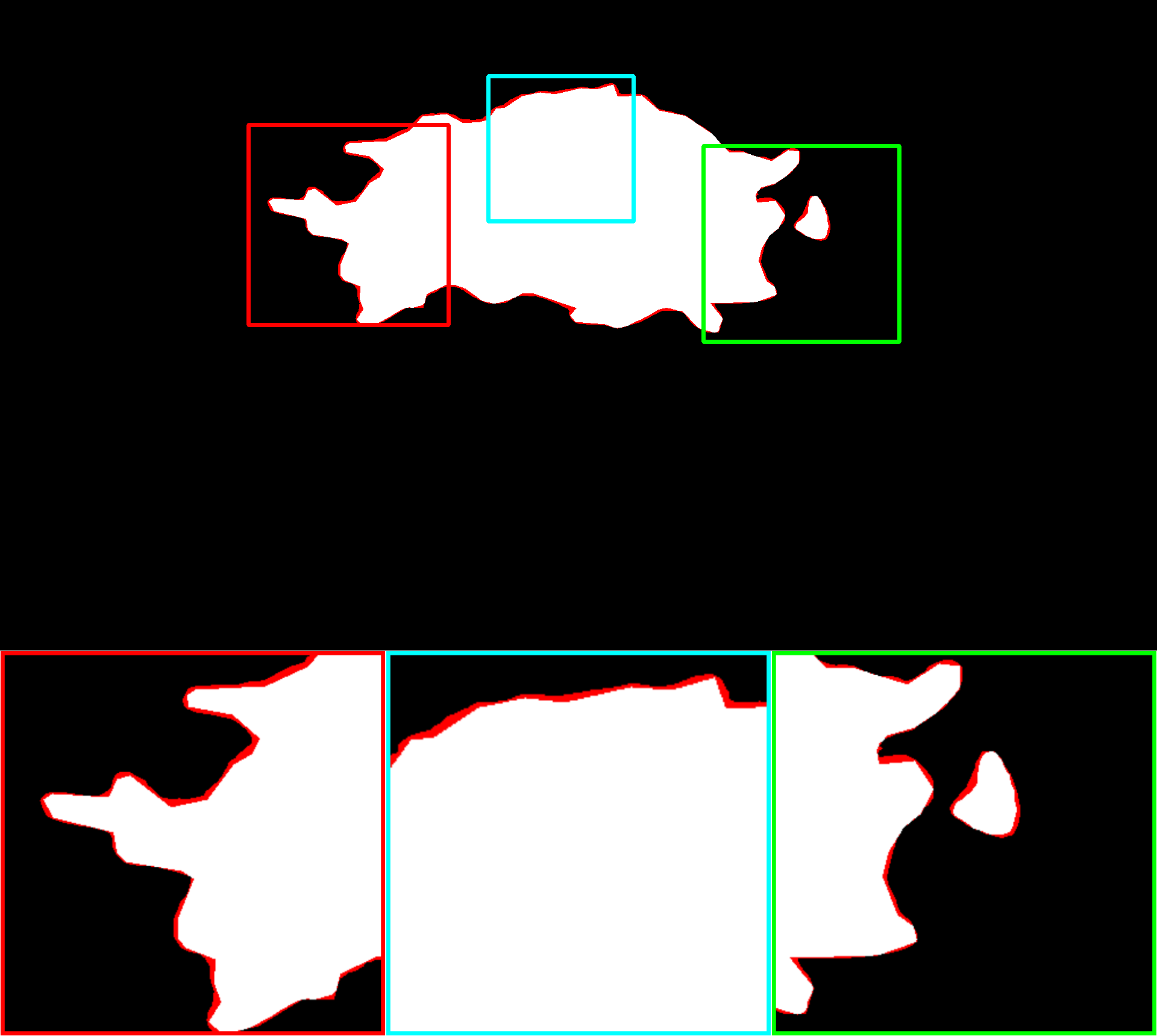} &
\includegraphics[width=0.17\textwidth]{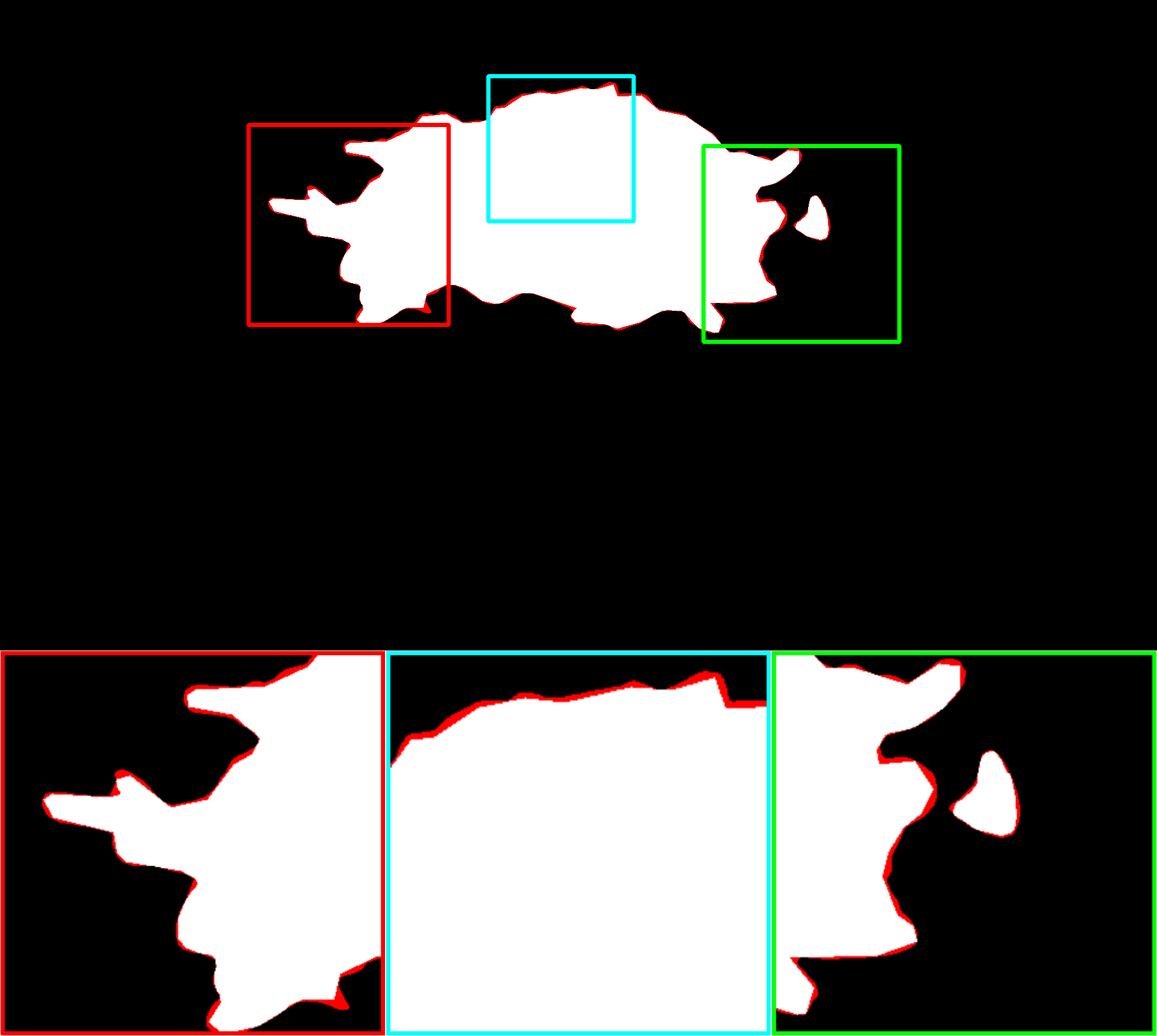} &
\includegraphics[width=0.17\textwidth]{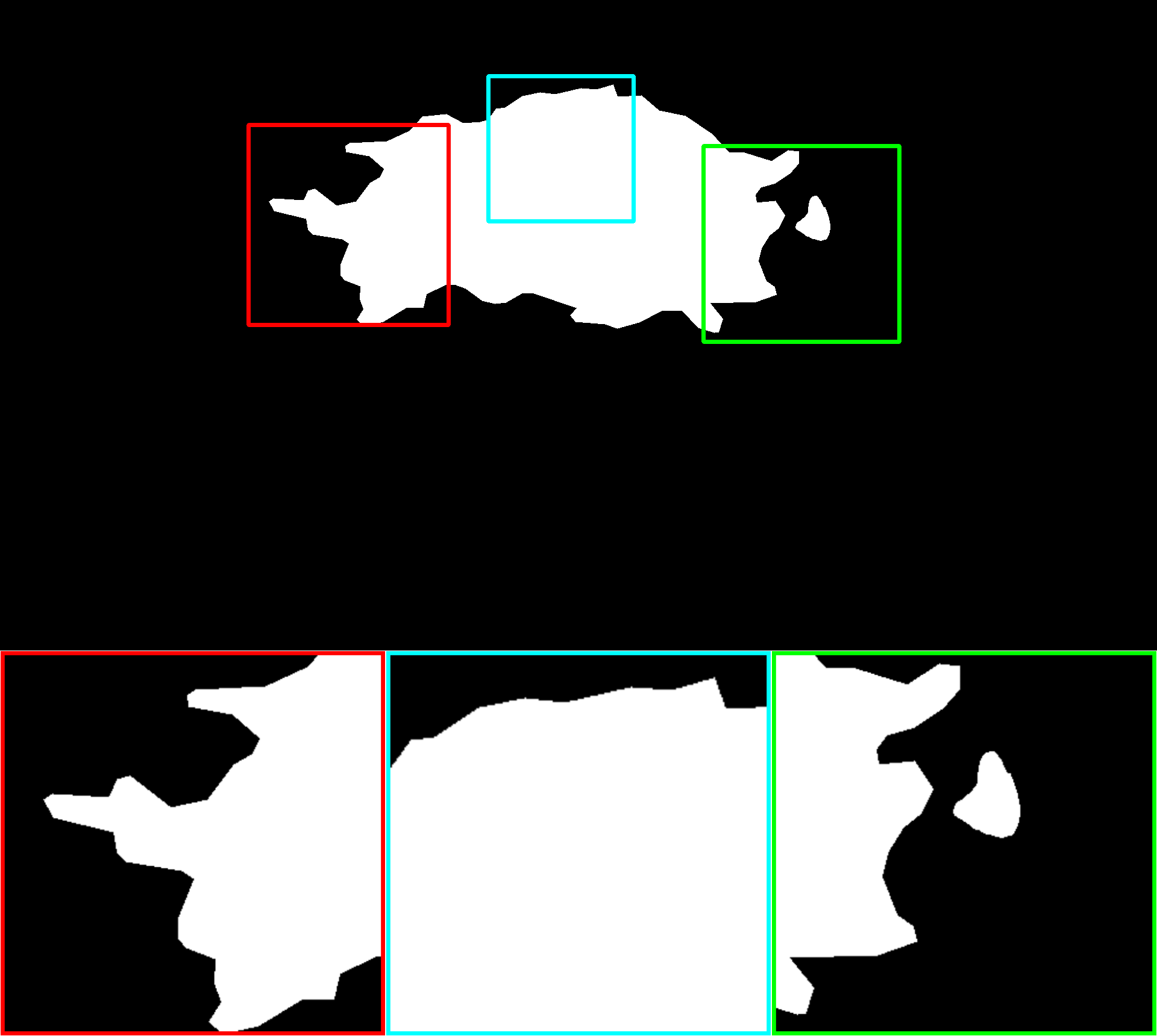} &
\includegraphics[width=0.17\textwidth]{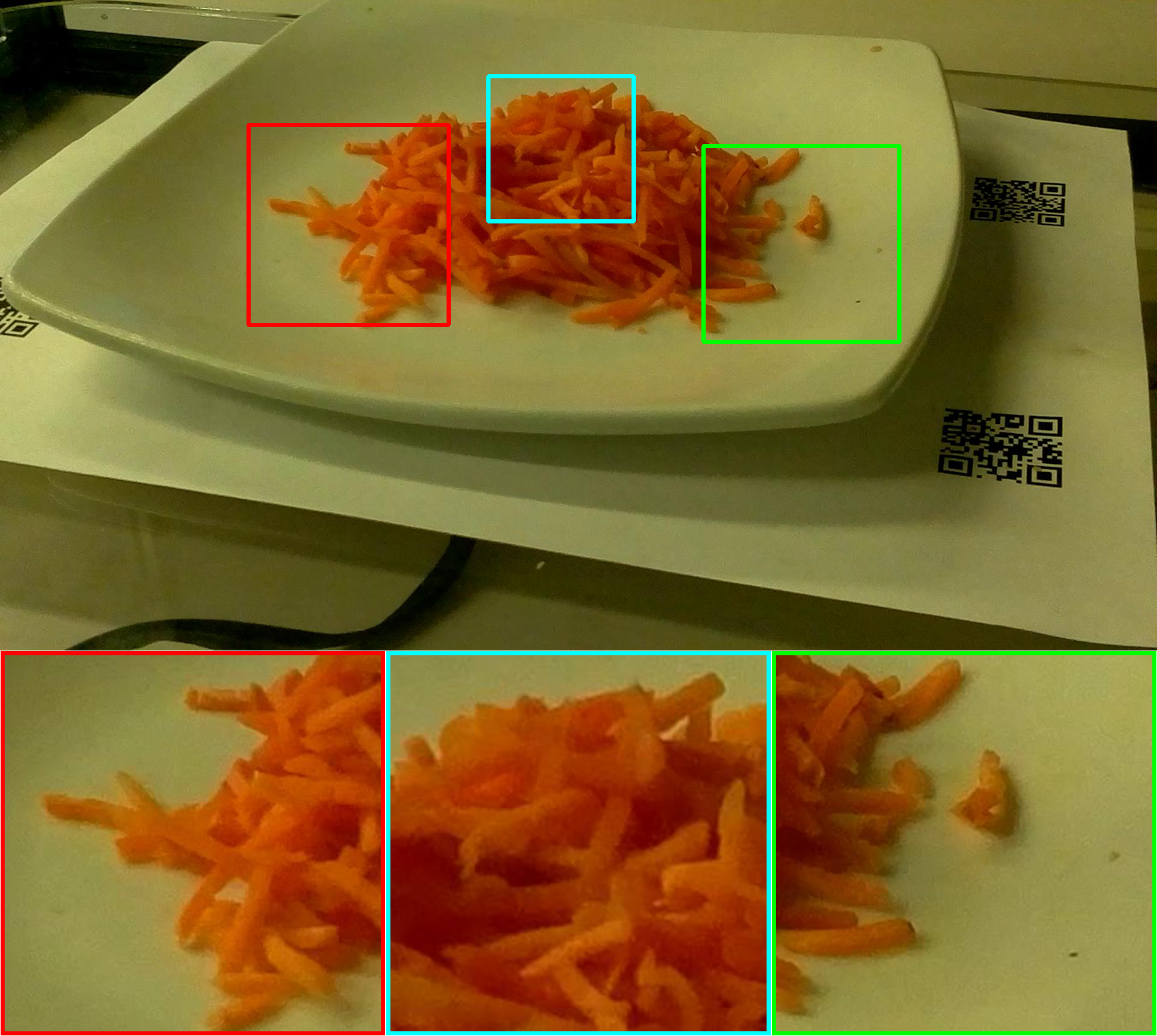} \\

& \includegraphics[width=0.17\textwidth]{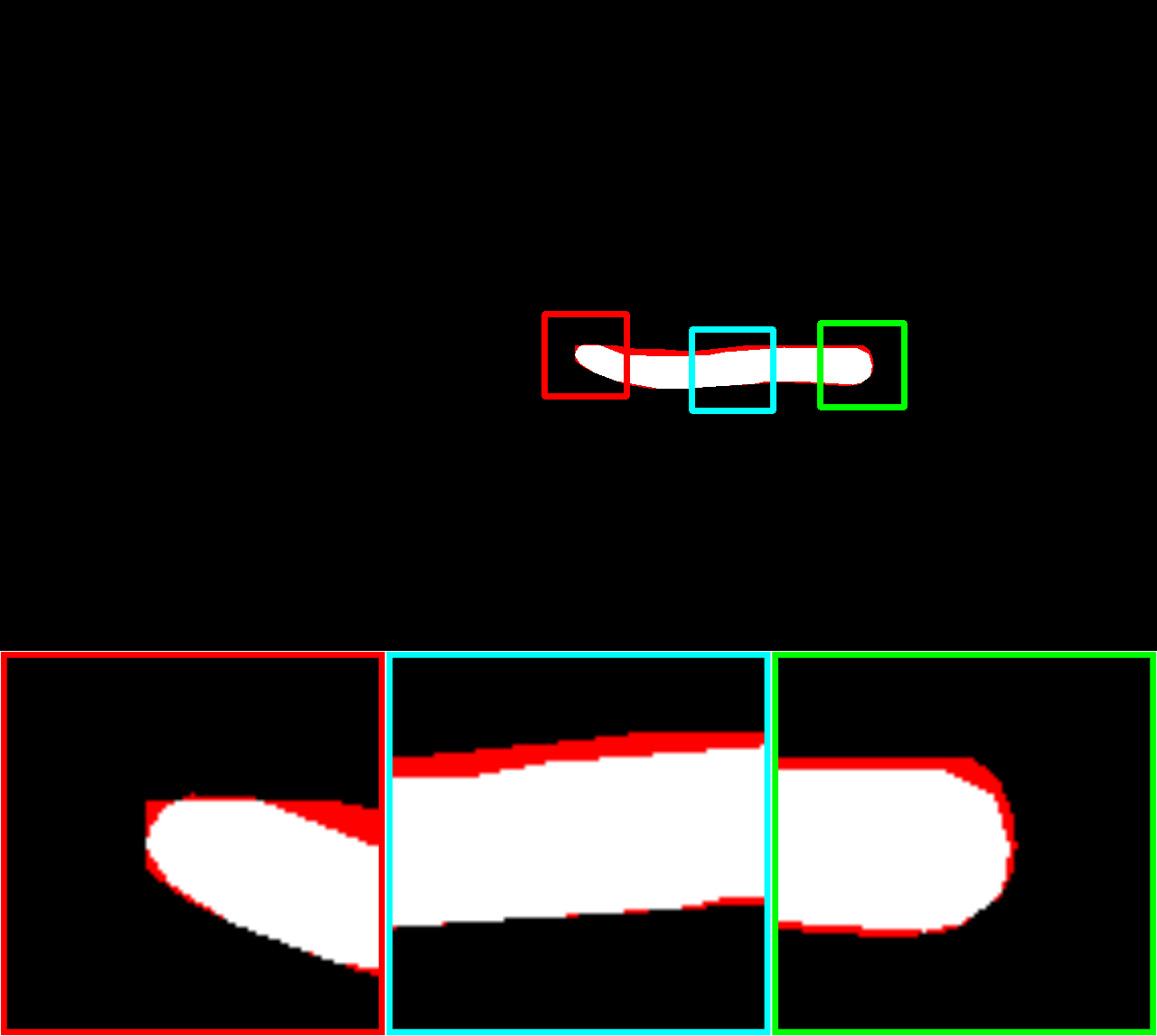} &
\includegraphics[width=0.17\textwidth]{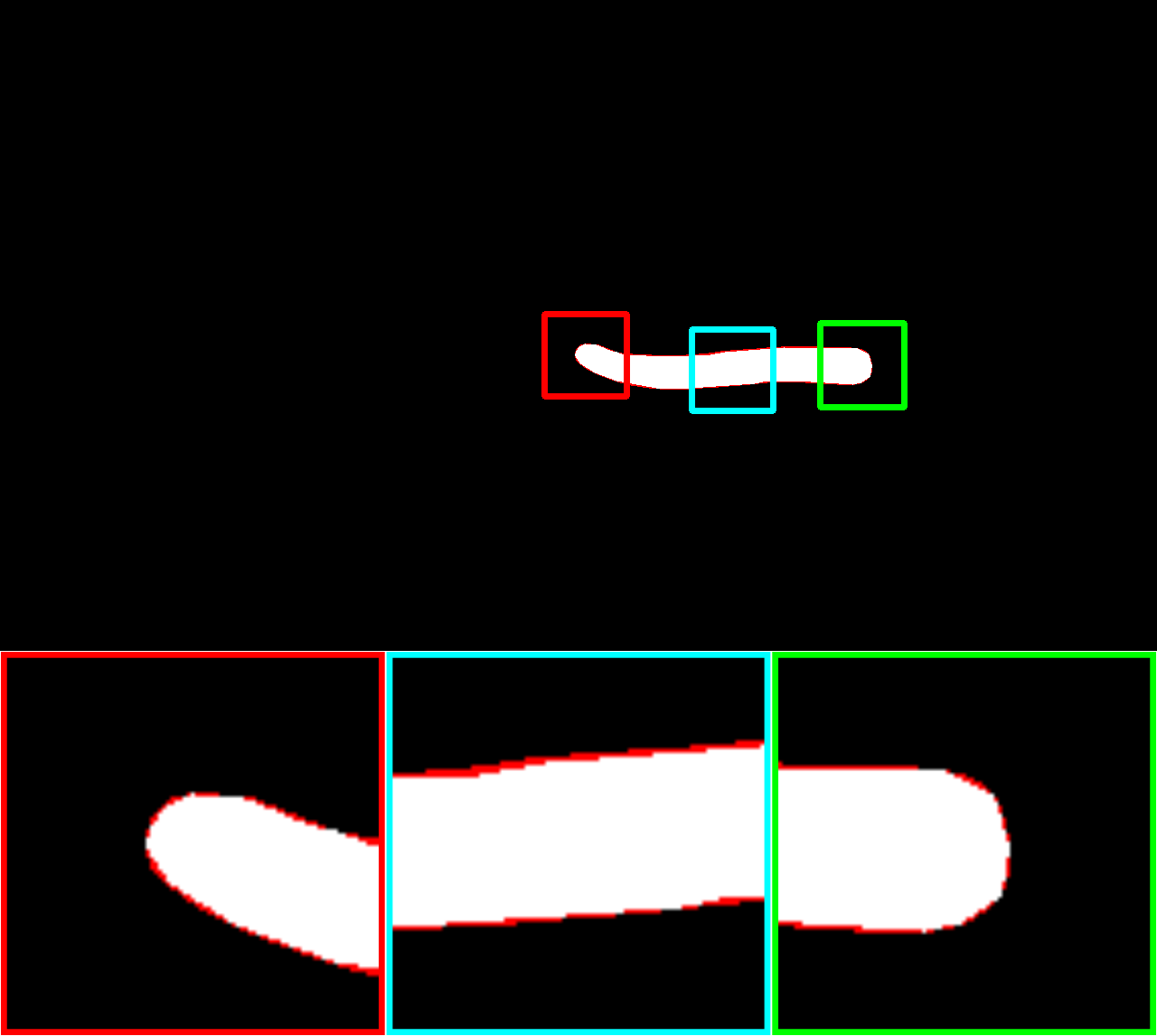} &
\includegraphics[width=0.17\textwidth]{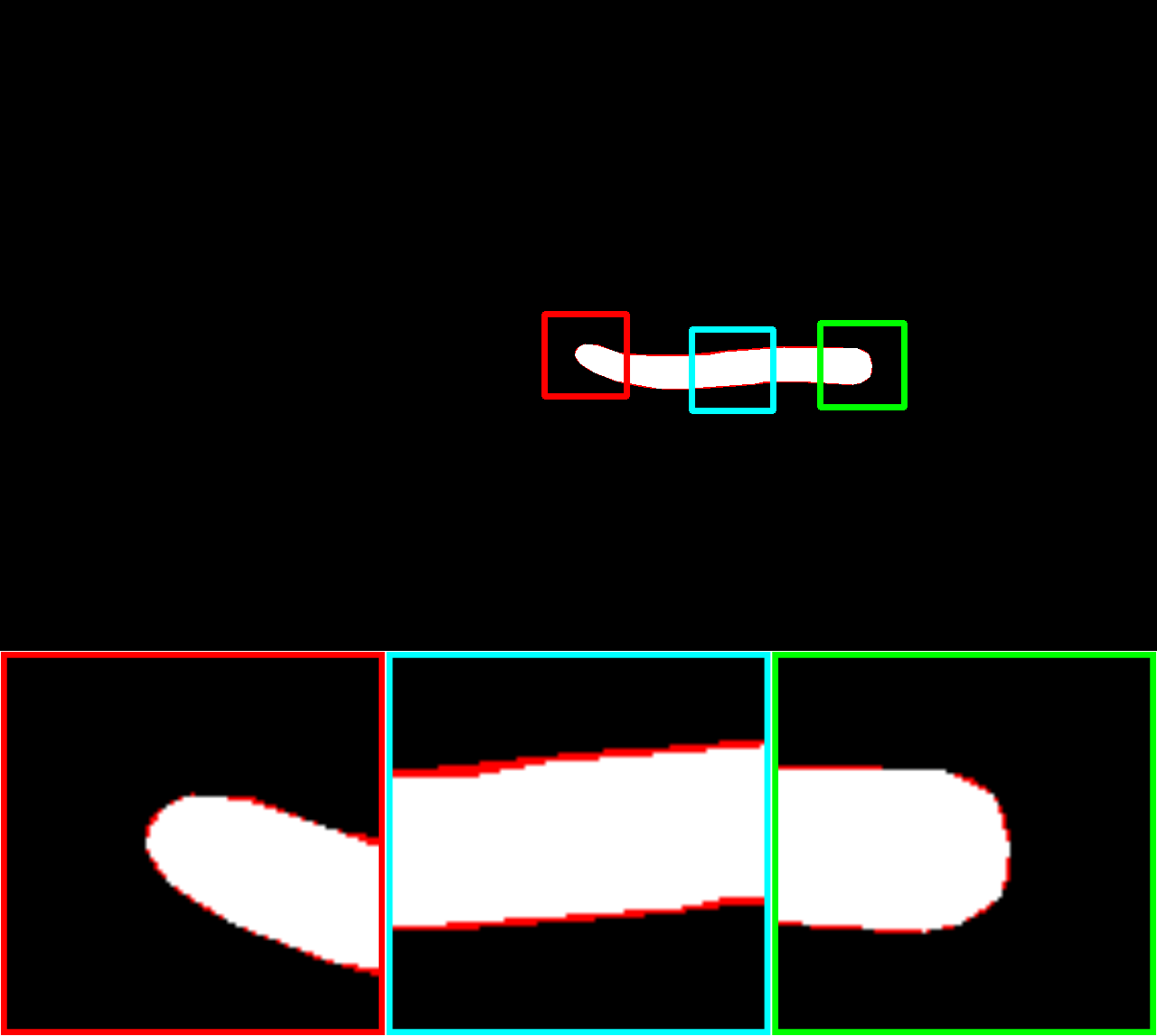} &
\includegraphics[width=0.17\textwidth]{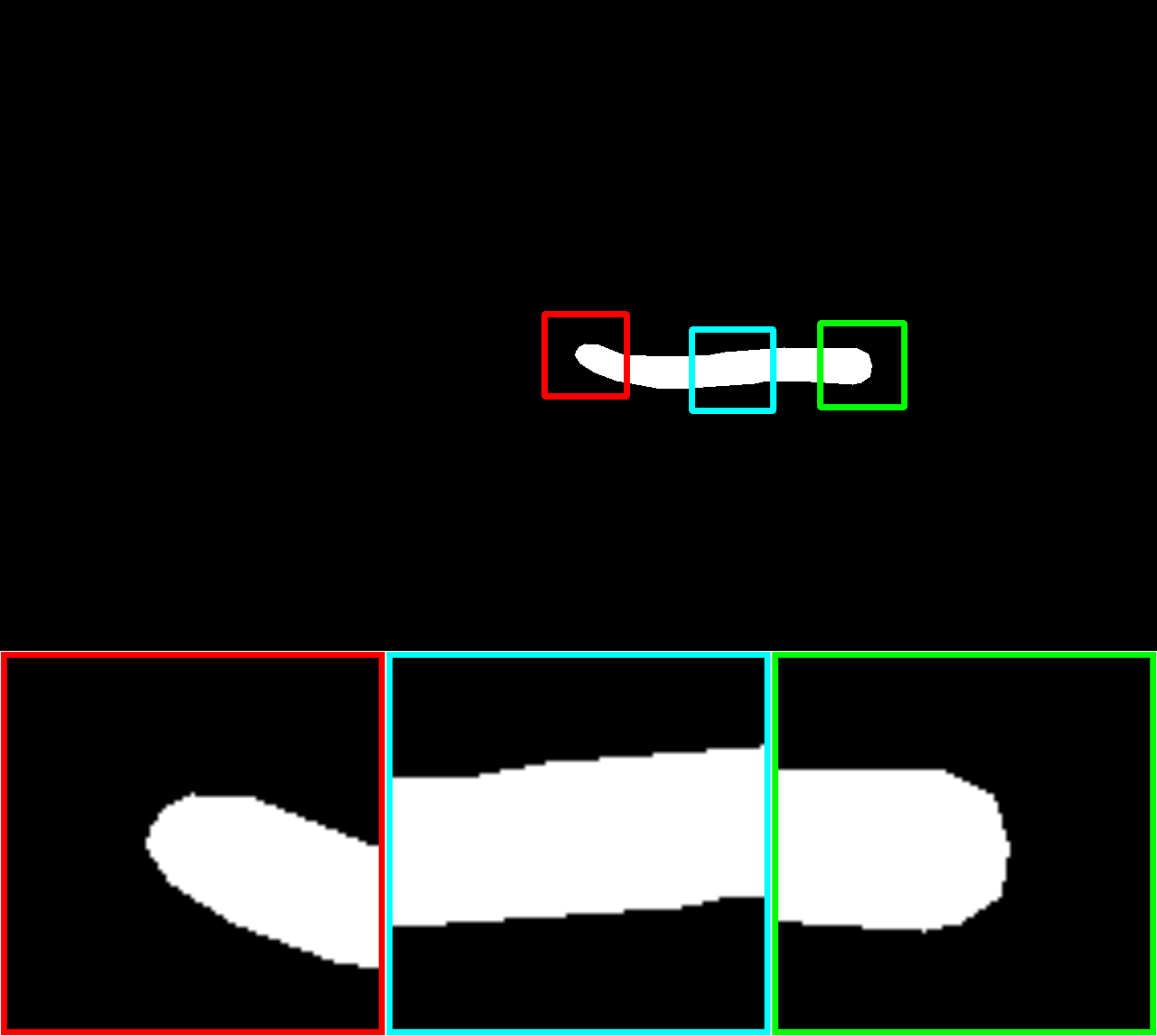} &
\includegraphics[width=0.17\textwidth]{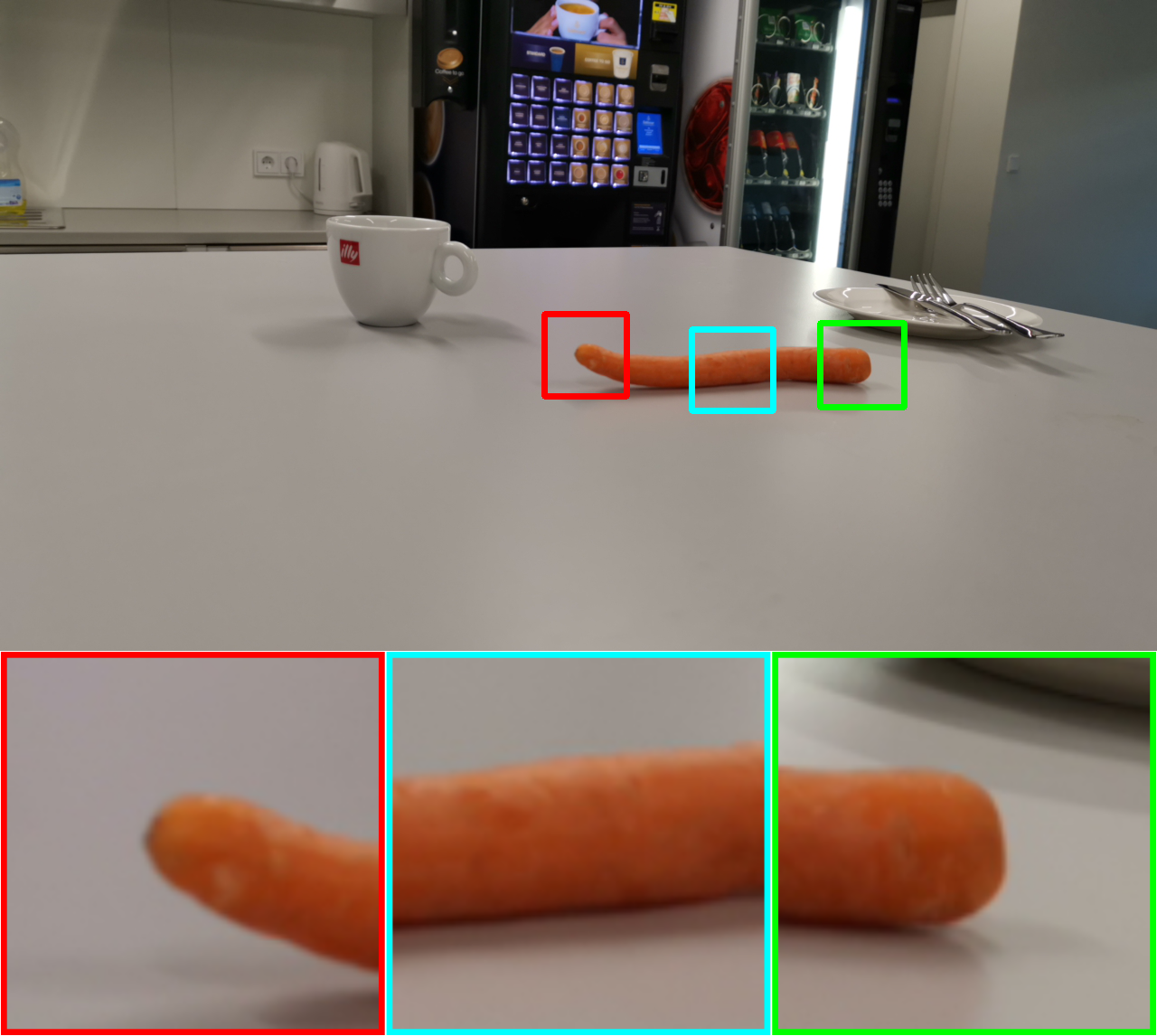} \\
\hline

\multirow{3}{*}{SegMan} & 

\includegraphics[width=0.17\textwidth]{Fig_R1/chocolate_croissant_976_segMAN.png} &
\includegraphics[width=0.17\textwidth]{Fig_R1/chocolate_croissant_976_segMAN_SAM2.png} &
\includegraphics[width=0.17\textwidth]{Fig_R1/chocolate_croissant_976_segMAN_XMEM2_n1.png} &
\includegraphics[width=0.17\textwidth]{Fig_R1/chocolate_croissant_976.png} &
\includegraphics[width=0.17\textwidth]{Fig_R1/chocolate_croissant_976_RGB.png}\\

& \includegraphics[width=0.17\textwidth]{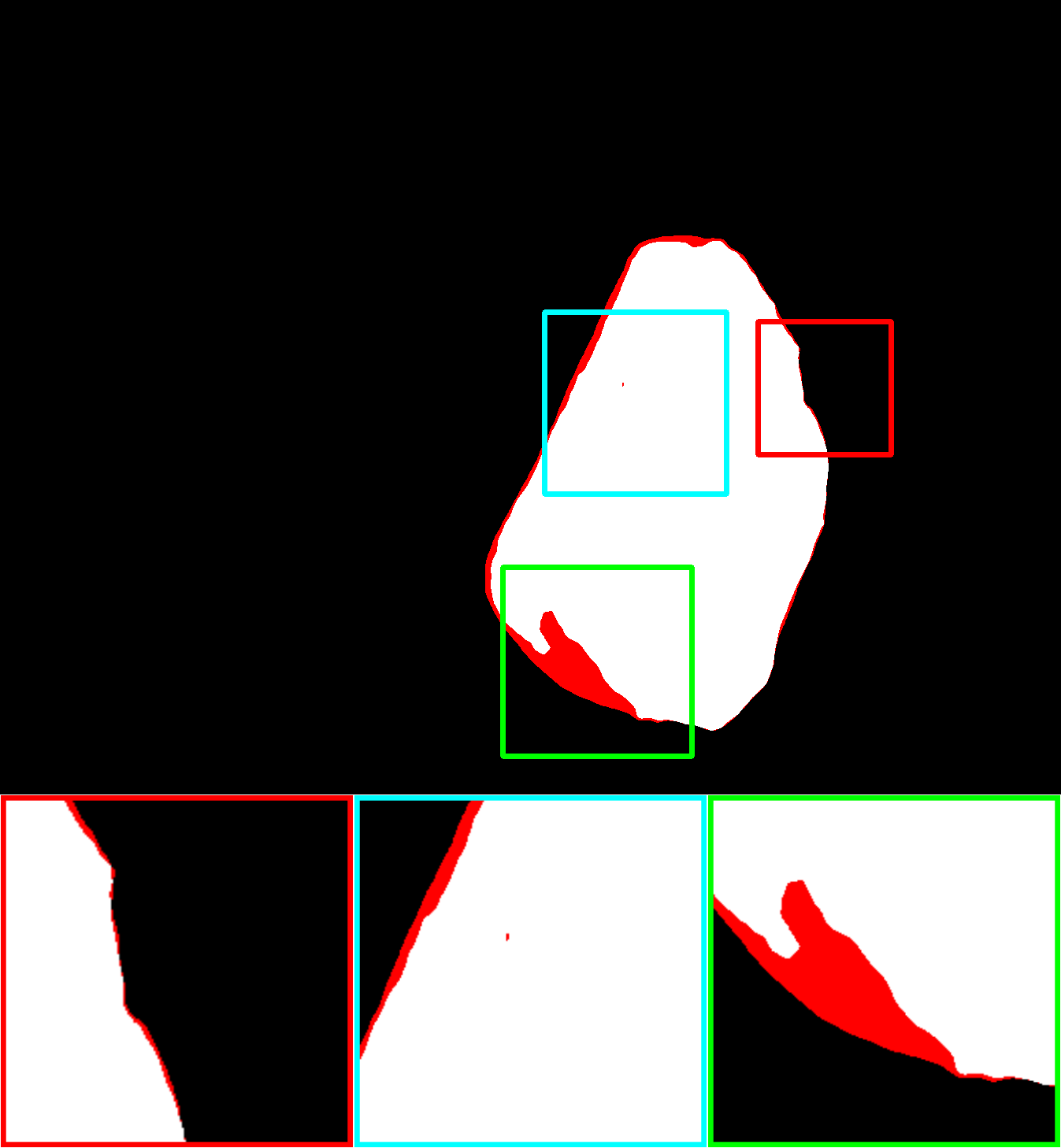} &
\includegraphics[width=0.17\textwidth]{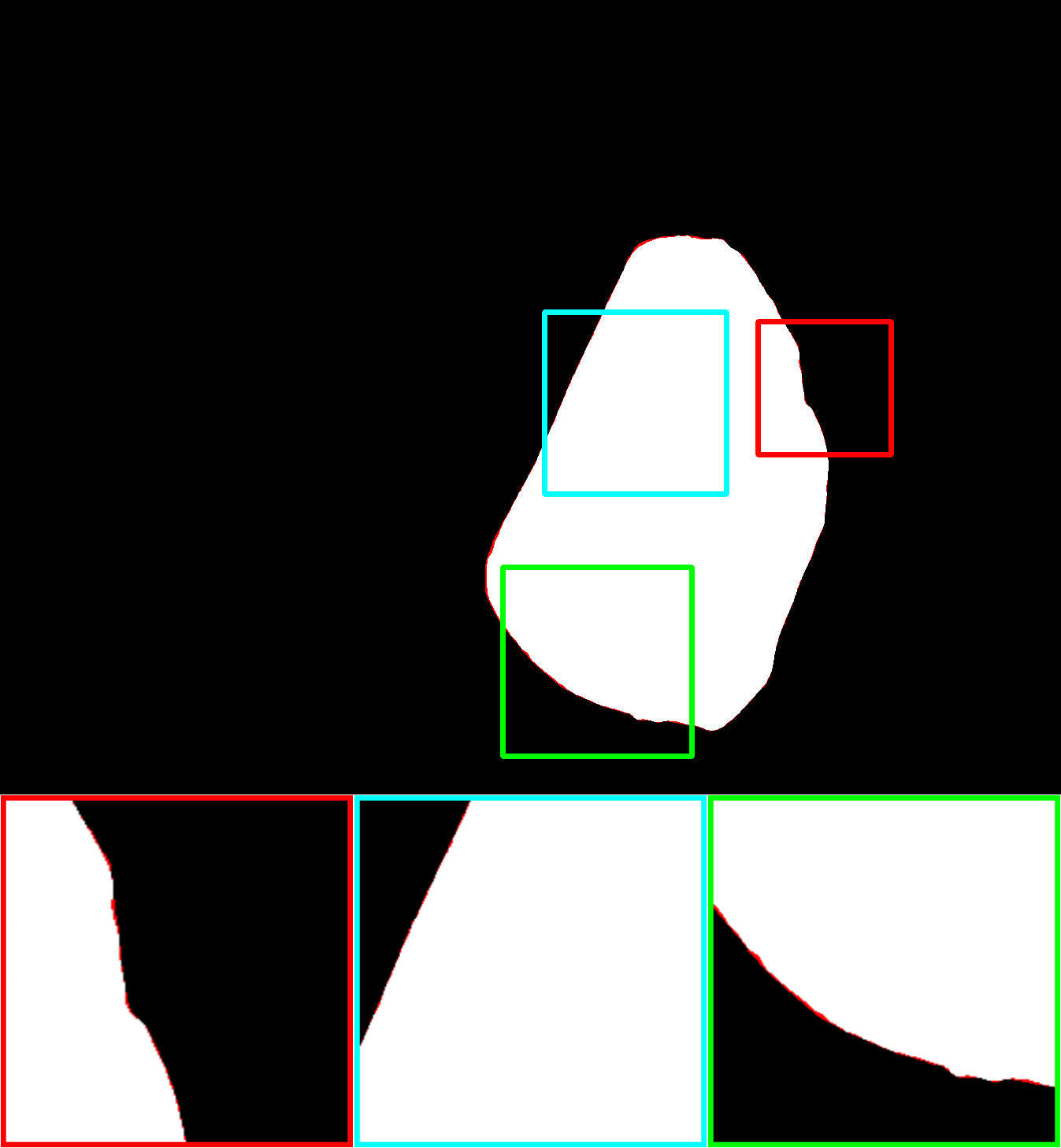} &
\includegraphics[width=0.17\textwidth]{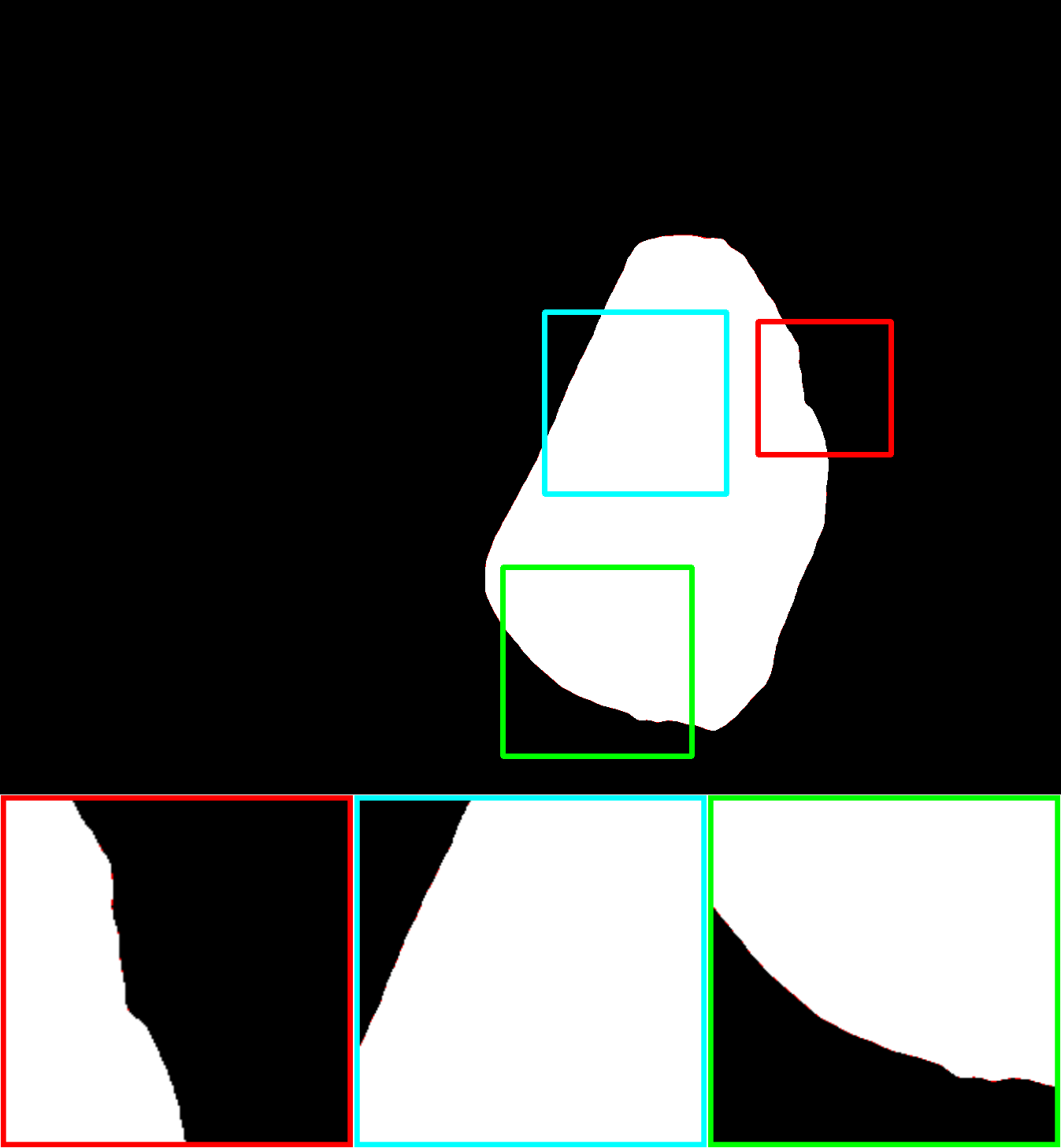} &
\includegraphics[width=0.17\textwidth]{Fig_R1/5_065_mask.png} &
\includegraphics[width=0.17\textwidth]{Fig_R1/5_065.png}\\

& \includegraphics[width=0.17\textwidth]{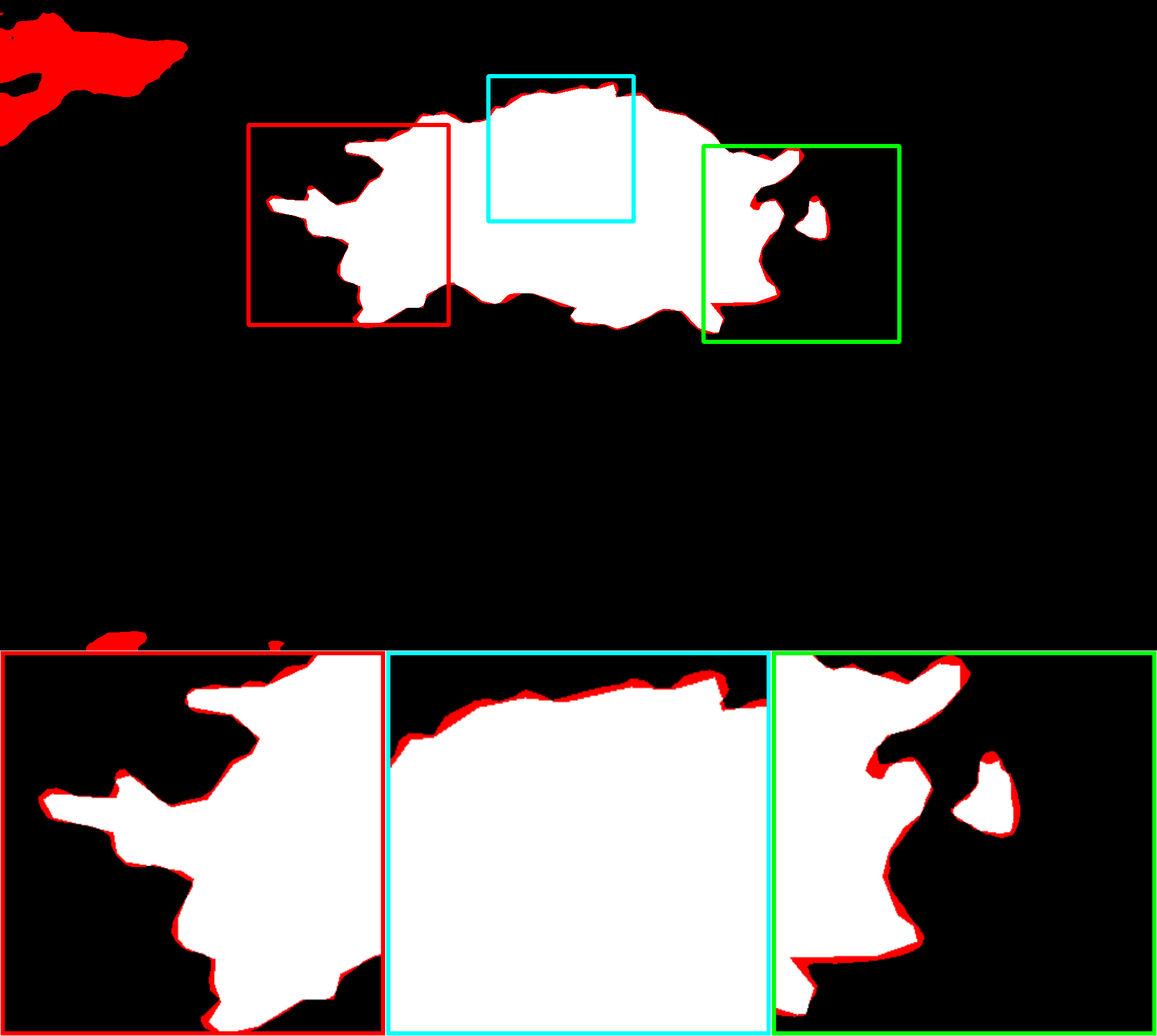} &
\includegraphics[width=0.17\textwidth]{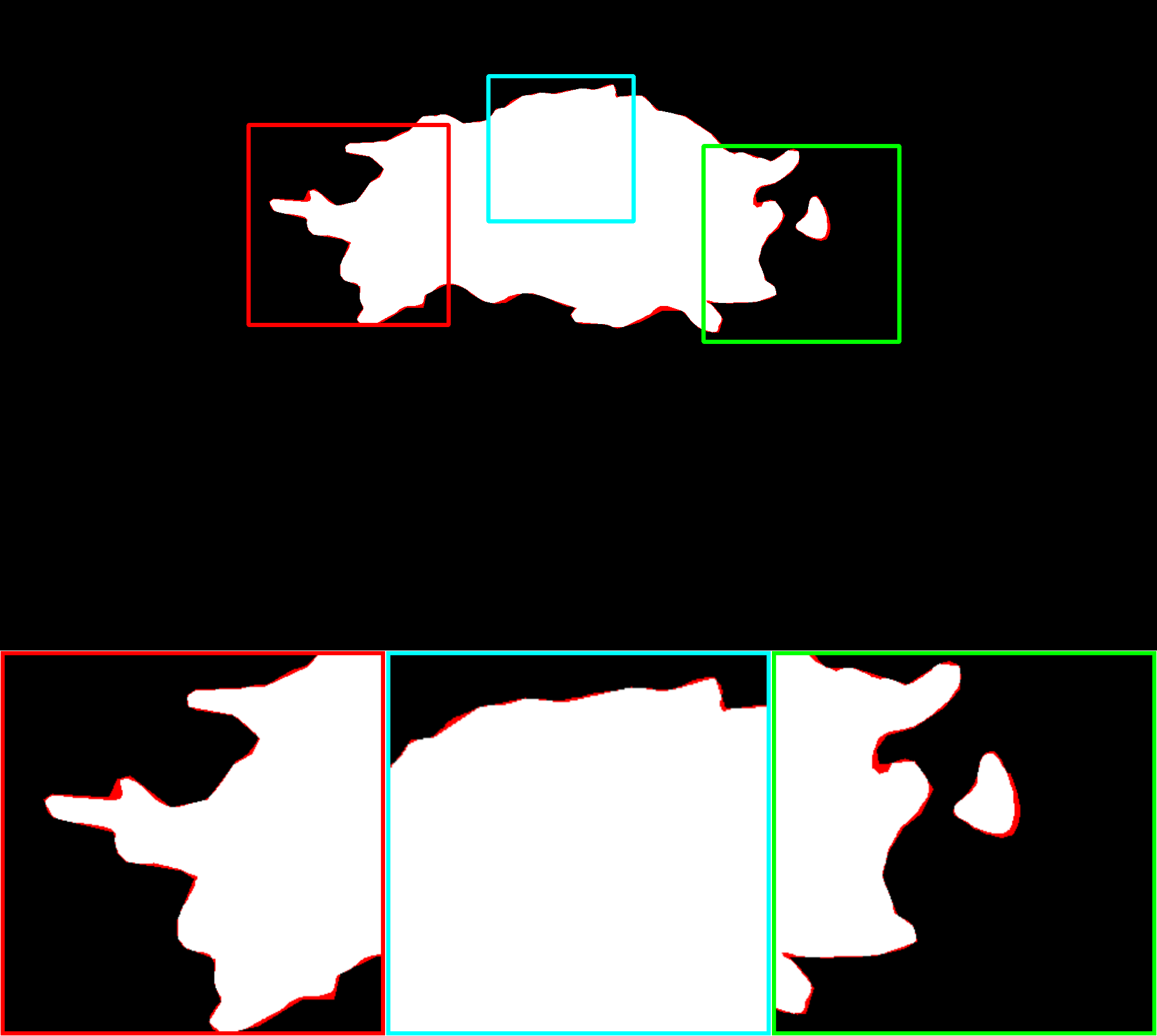} &
\includegraphics[width=0.17\textwidth]{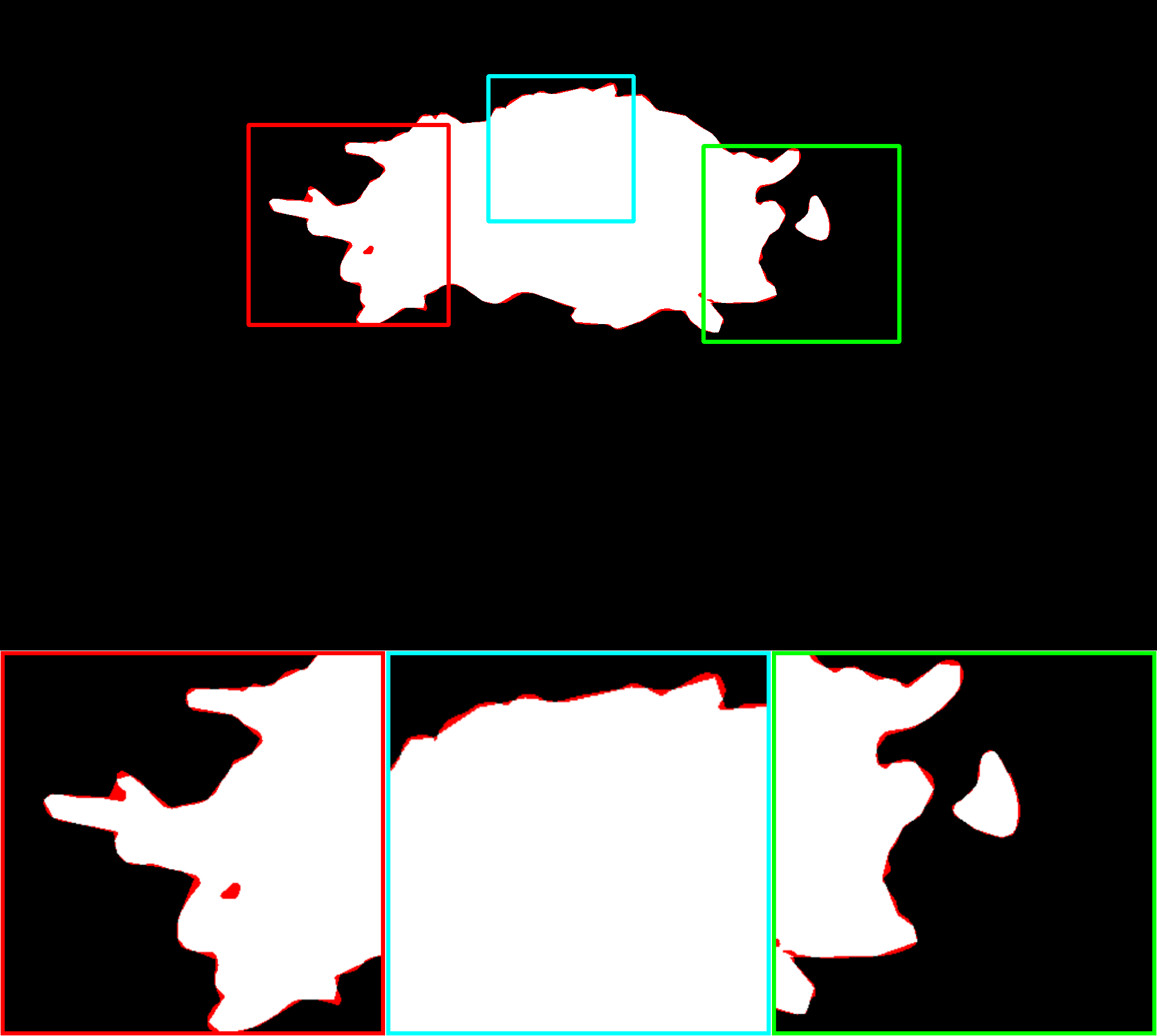} &
\includegraphics[width=0.17\textwidth]{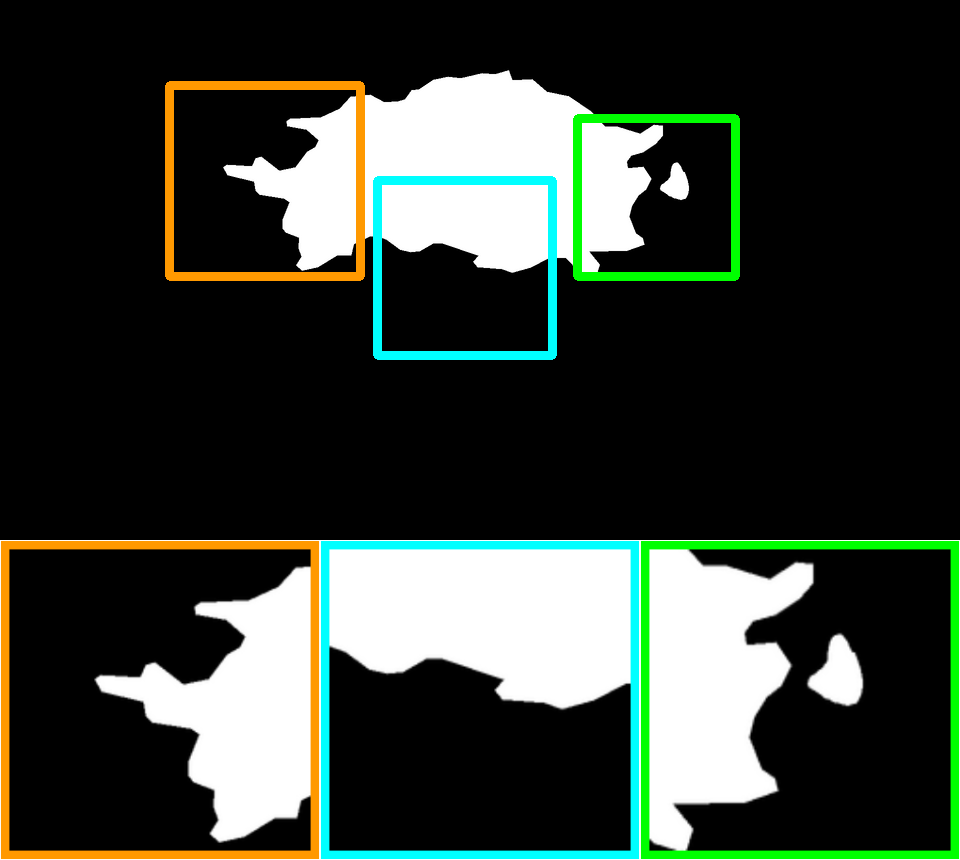} &
\includegraphics[width=0.17\textwidth]{Fig_R1/2_056_RGB.png}\\

& \includegraphics[width=0.17\textwidth]{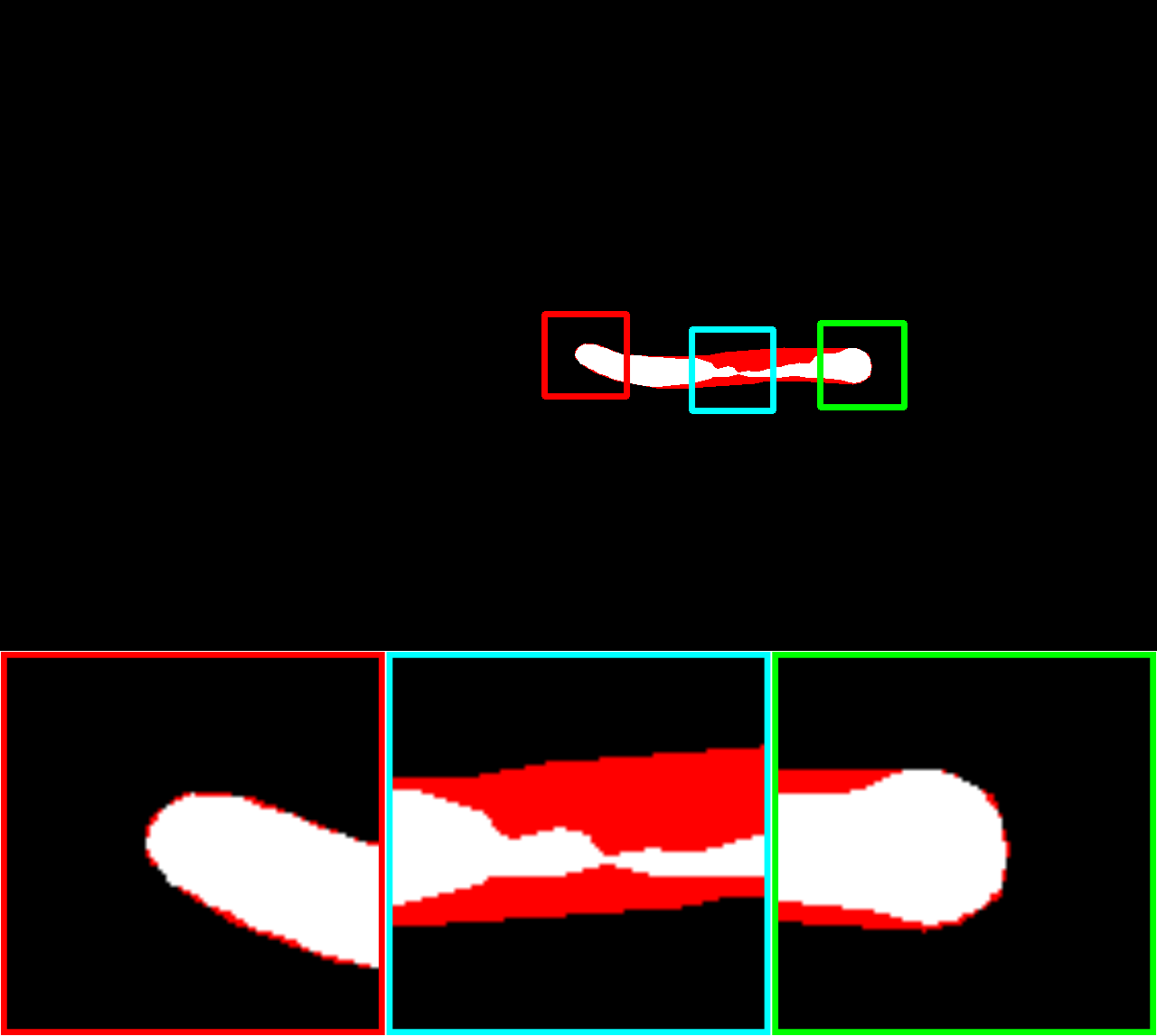} &
\includegraphics[width=0.17\textwidth]{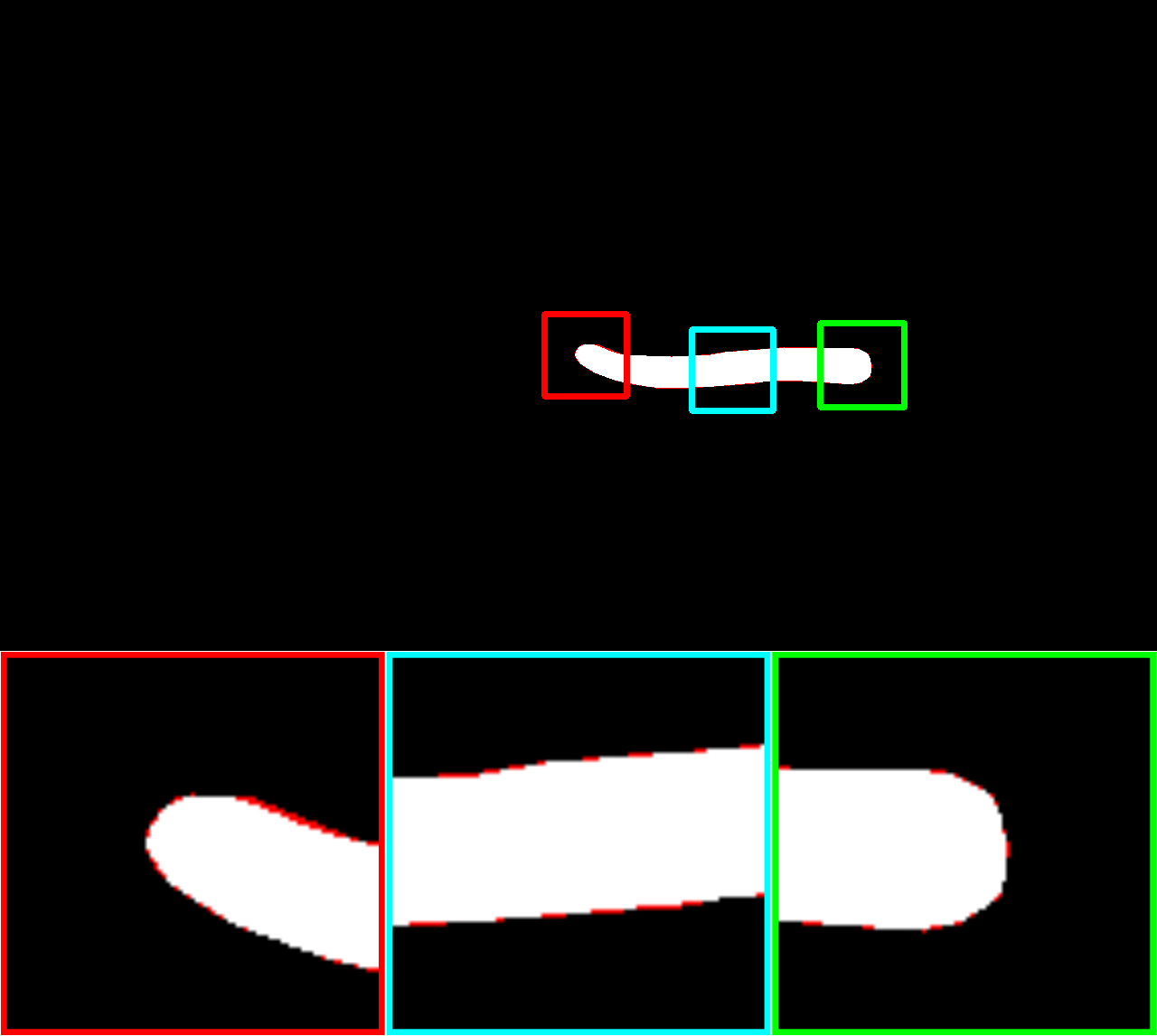} &
\includegraphics[width=0.17\textwidth]{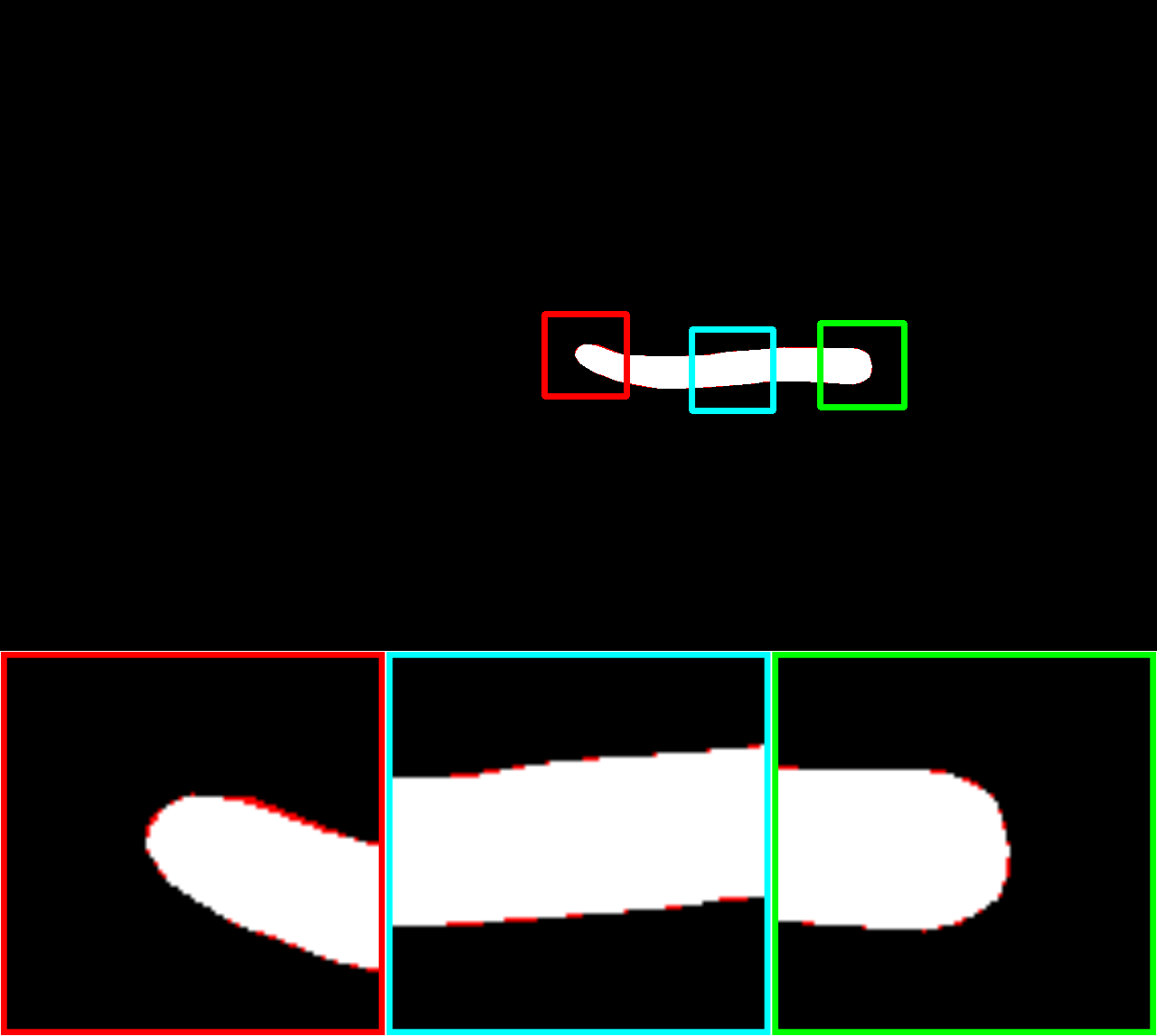} &
\includegraphics[width=0.17\textwidth]{Fig_R1/5_203_GT.png} &
\includegraphics[width=0.17\textwidth]{Fig_R1/5_203_RGB.png} \\
\hline

    \end{tabular}
    \label{tab:3d_comparisons_tracker}
\end{table}

\begin{table}[!htbp]
    \centering
    \small
    \caption{Qualitative comparison of 2D Methods on FoodKit Dataset}
    \setlength{\tabcolsep}{1pt}
    \begin{tabular}{c|ccccccc}
    \hline
    %Datasets & BiRefNet & CCNET RELEM & FPN RELEM & SETR MLA & SWIN BASE & GT \\
    FoodKit & BiRefNet & CCNET & FPN & SeTR & SWIN & GT & RGB \\
    \hline
    % \hline
    
\raisebox{+1.5cm}{\centering{Aguacate}} &

\includegraphics[width=0.12\textwidth]{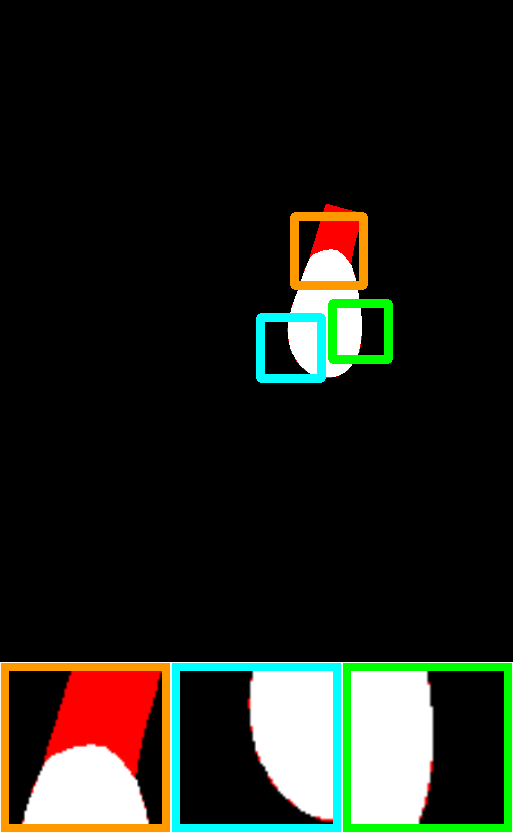} &
\includegraphics[width=0.12\textwidth]{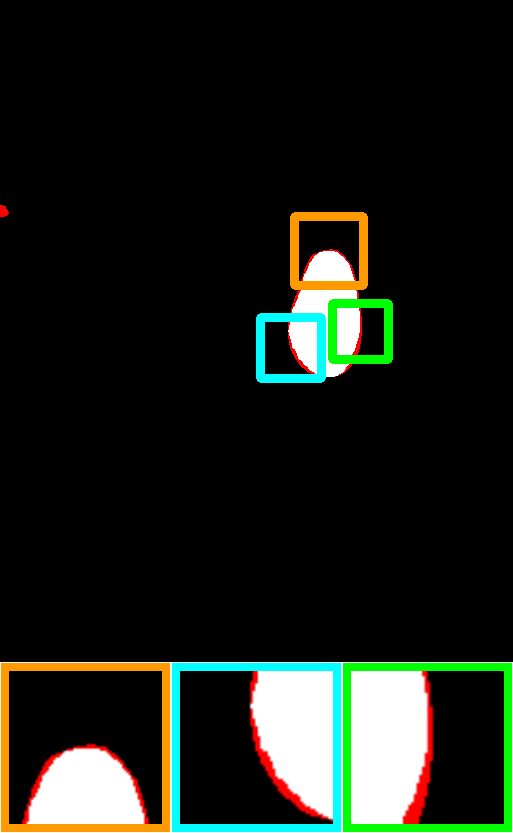} &
\includegraphics[width=0.12\textwidth]{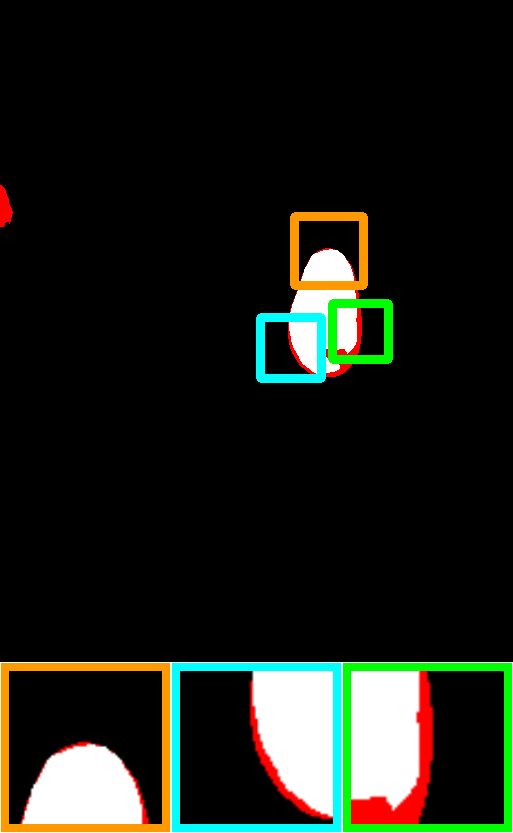} &
\includegraphics[width=0.12\textwidth]{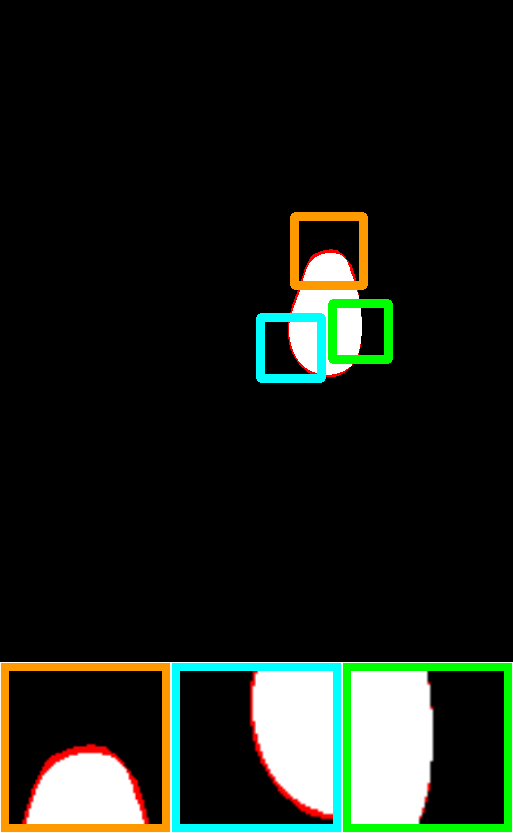} &
\includegraphics[width=0.12\textwidth]{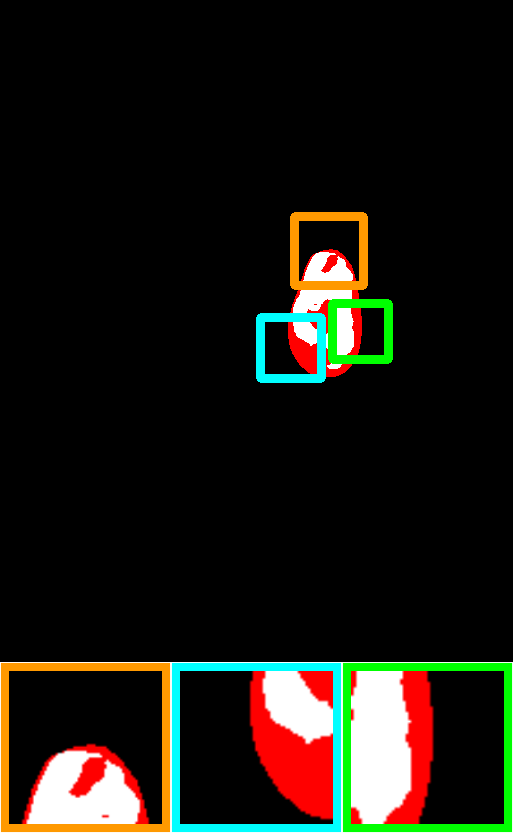} &
\includegraphics[width=0.12\textwidth]{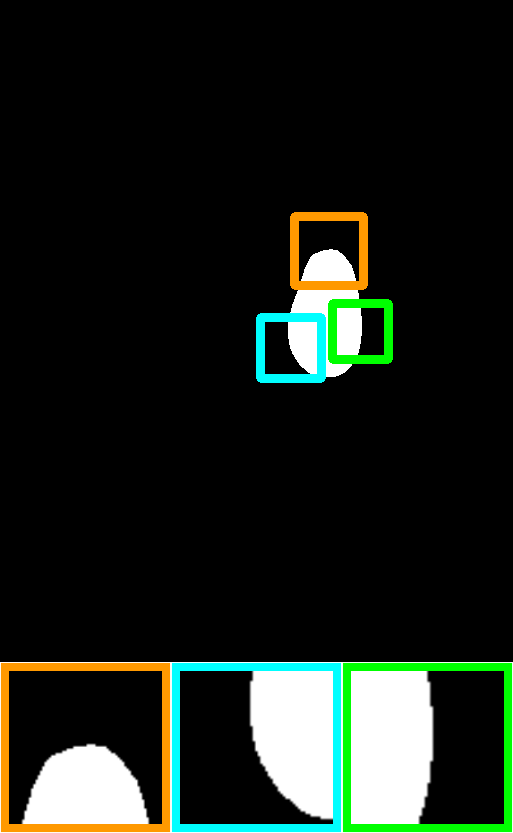} &
\includegraphics[width=0.12\textwidth]{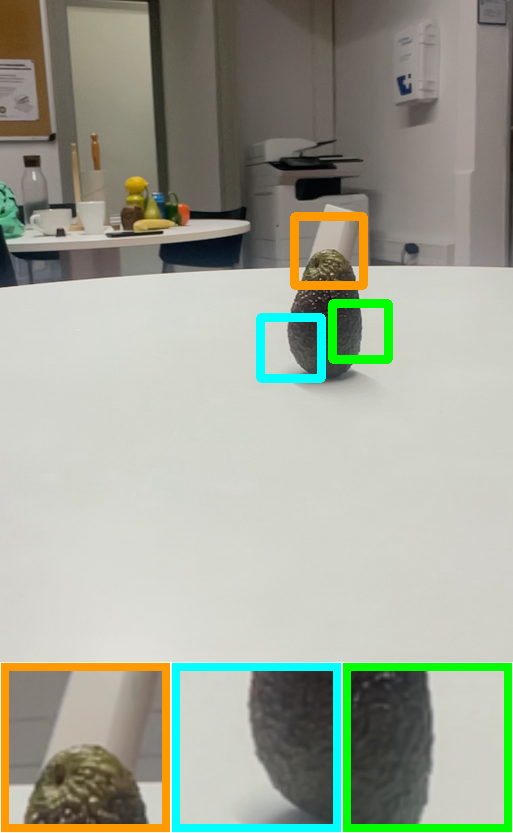} \\

\hline

\raisebox{+1.5cm}{\centering{Apple Pie}} & 

\includegraphics[width=0.12\textwidth]{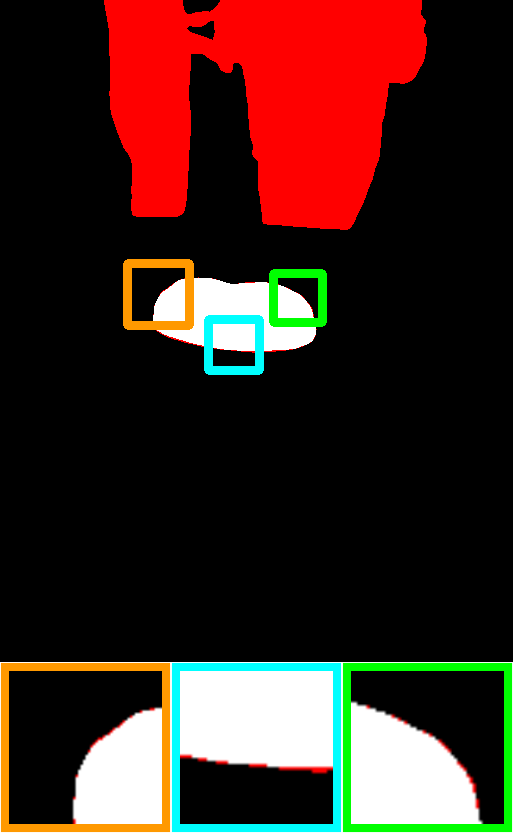} &
\includegraphics[width=0.12\textwidth]{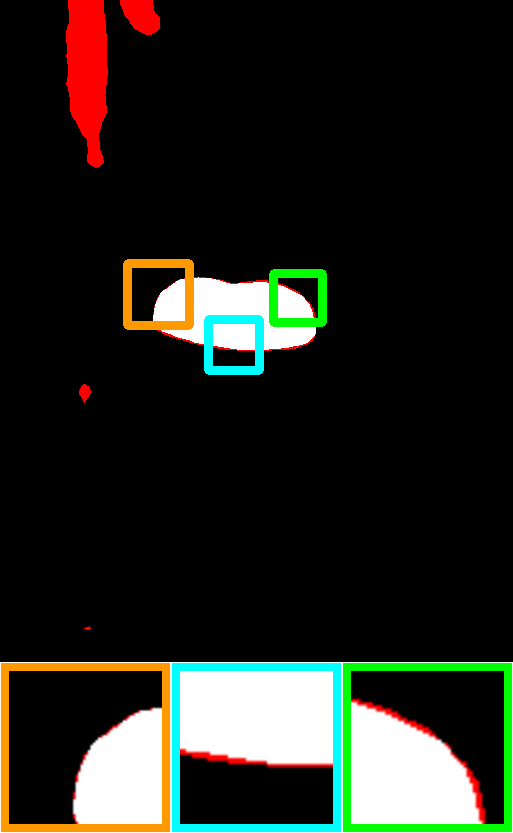} &
\includegraphics[width=0.12\textwidth]{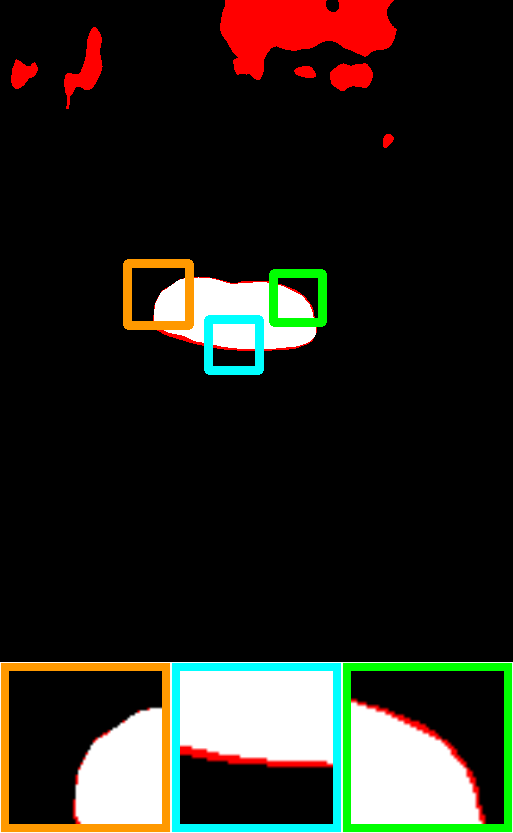} &
\includegraphics[width=0.12\textwidth]{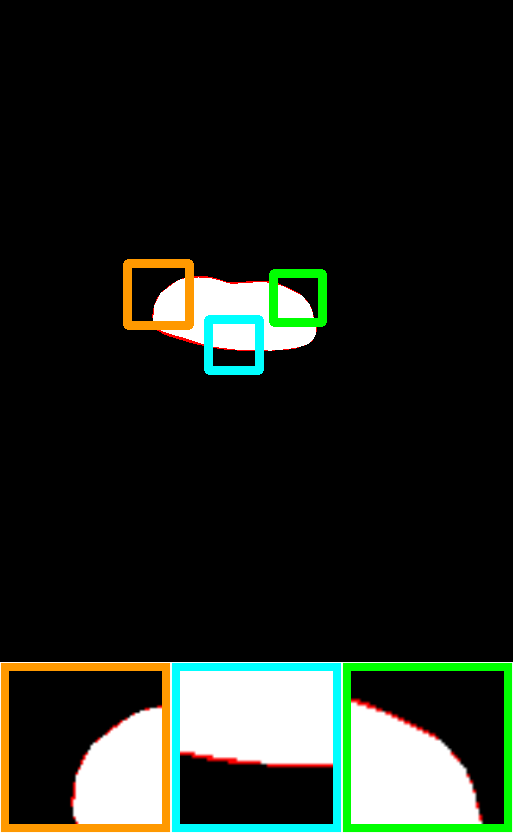} &
\includegraphics[width=0.12\textwidth]{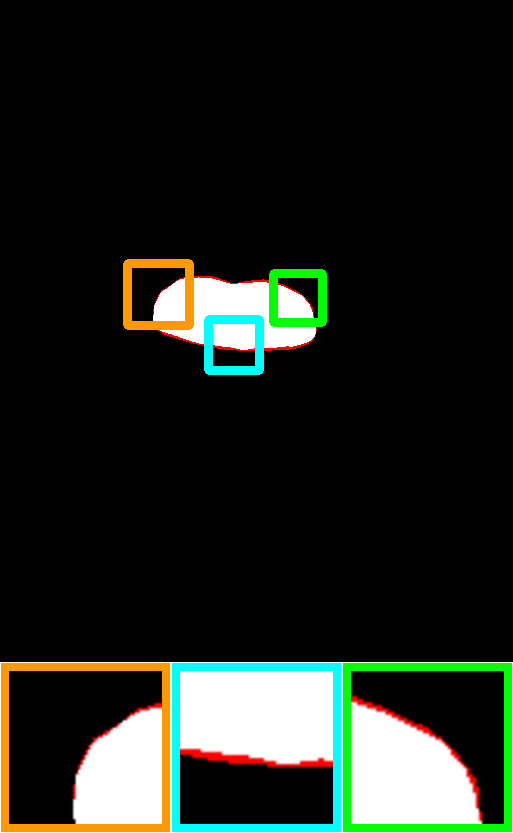} &
\includegraphics[width=0.12\textwidth]{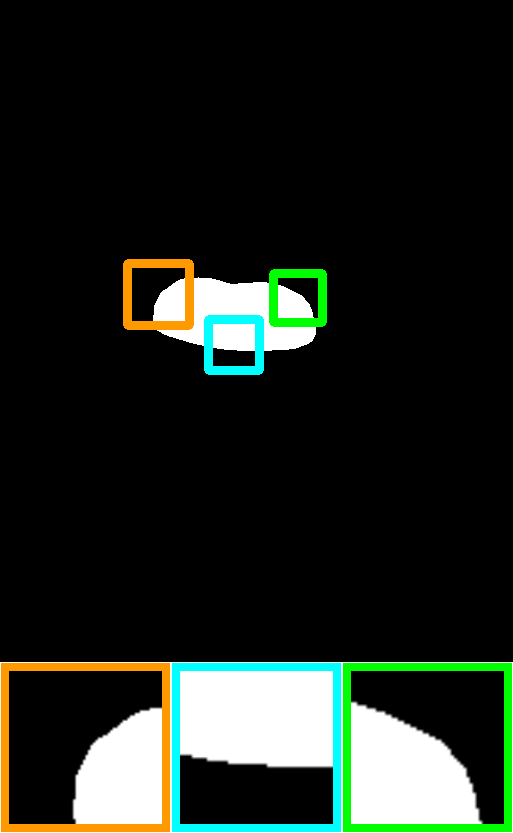} &
\includegraphics[width=0.12\textwidth]{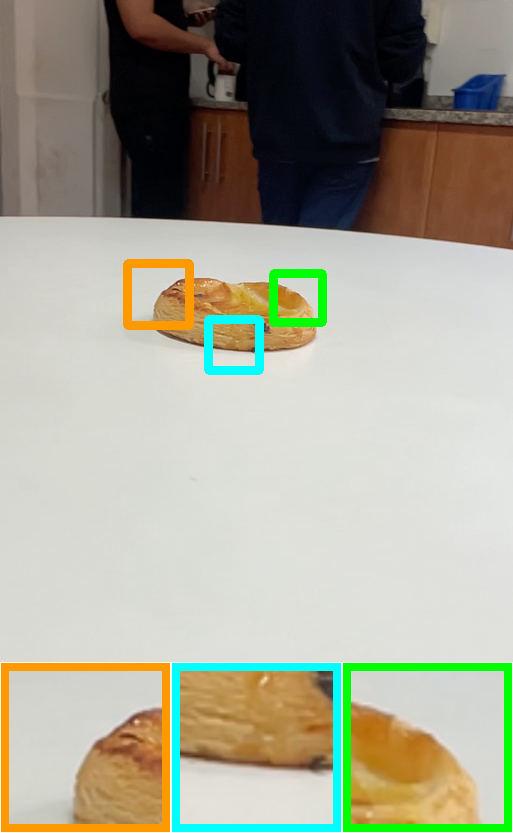} \\

\hline

\raisebox{+1.5cm}{\centering{Choc. Bomb}} &

\includegraphics[width=0.12\textwidth]{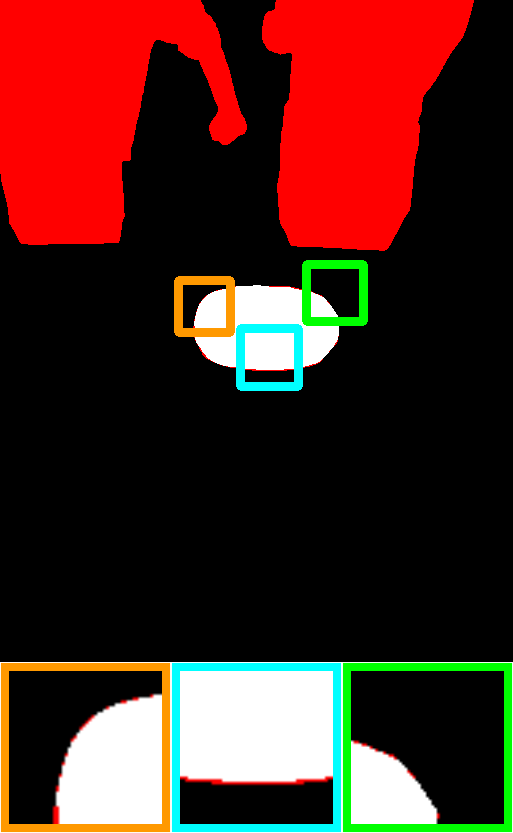} &
\includegraphics[width=0.12\textwidth]{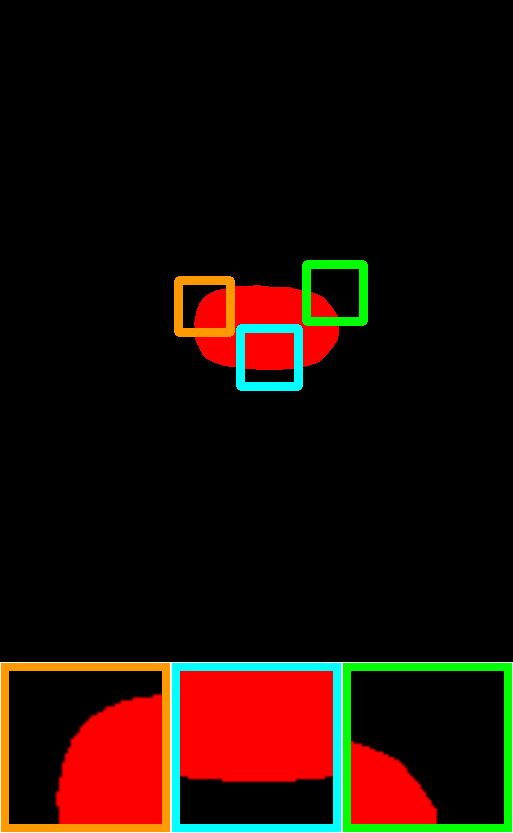} &
\includegraphics[width=0.12\textwidth]{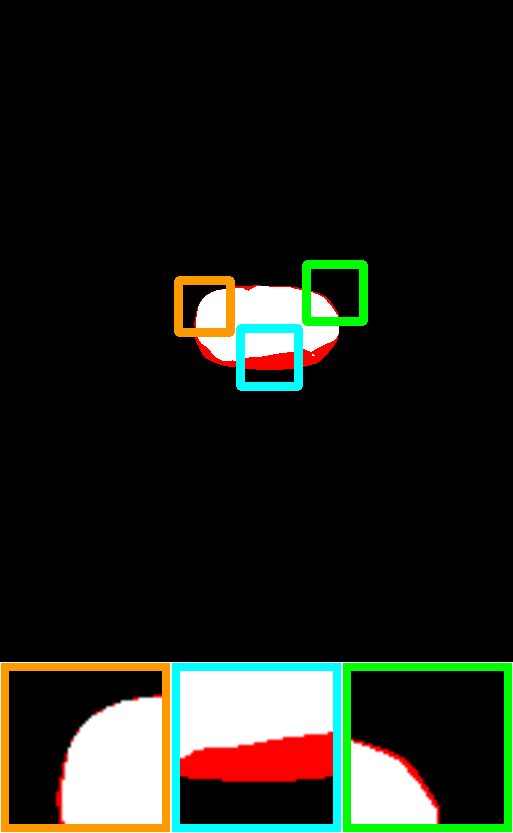} &
\includegraphics[width=0.12\textwidth]{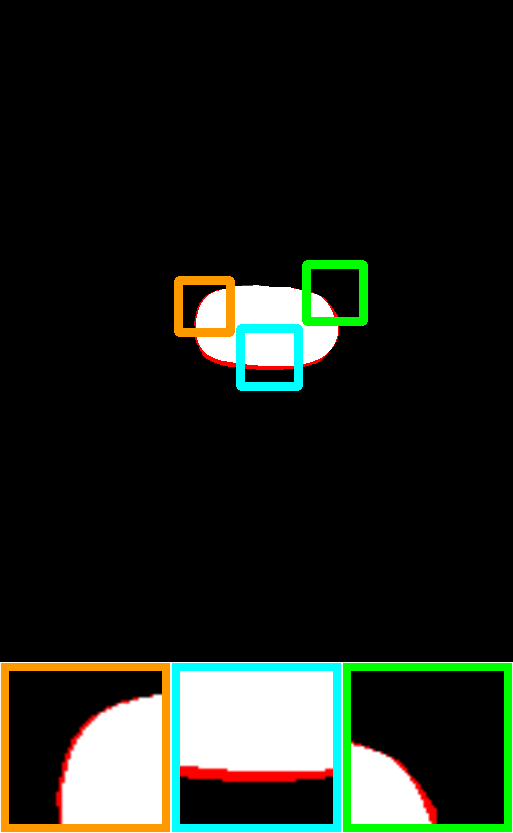} &
\includegraphics[width=0.12\textwidth]{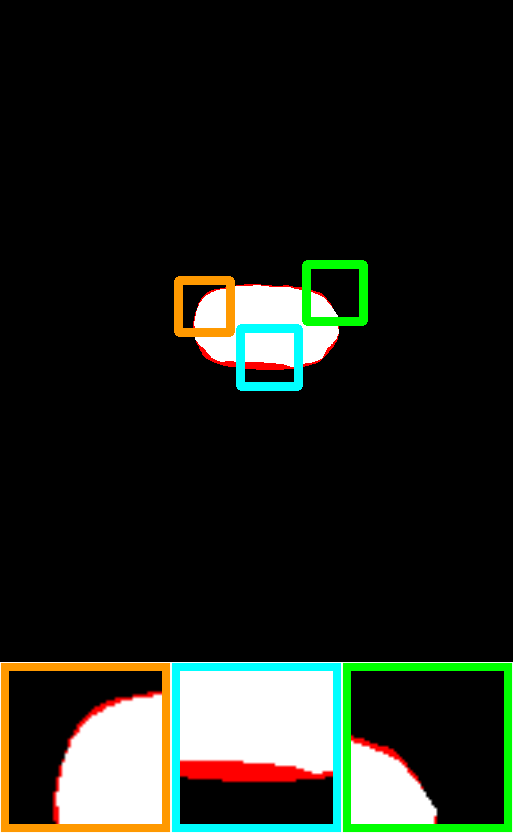} &
\includegraphics[width=0.12\textwidth]{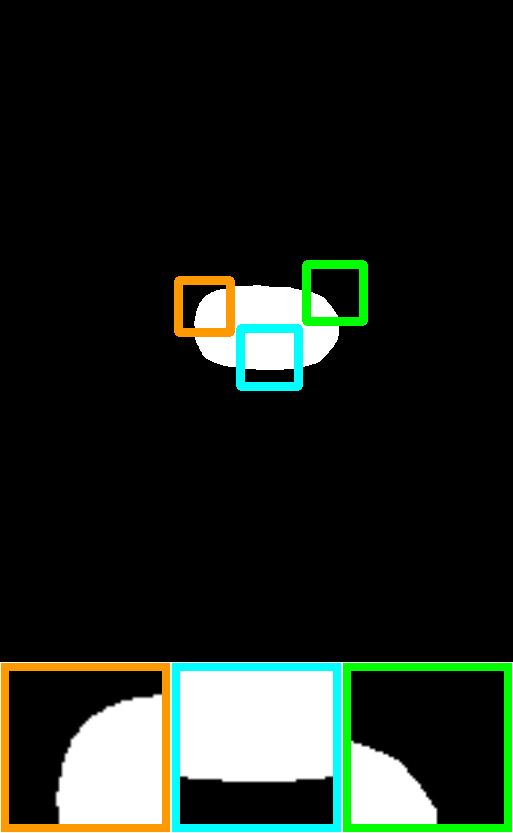} &
\includegraphics[width=0.12\textwidth]{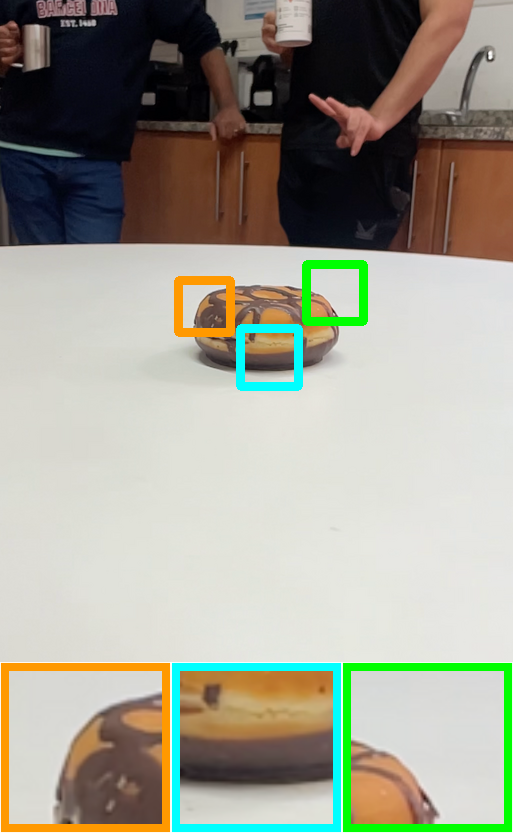} \\

\hline

\raisebox{+1.5cm}{\centering{Samosa}} &

\includegraphics[width=0.12\textwidth]{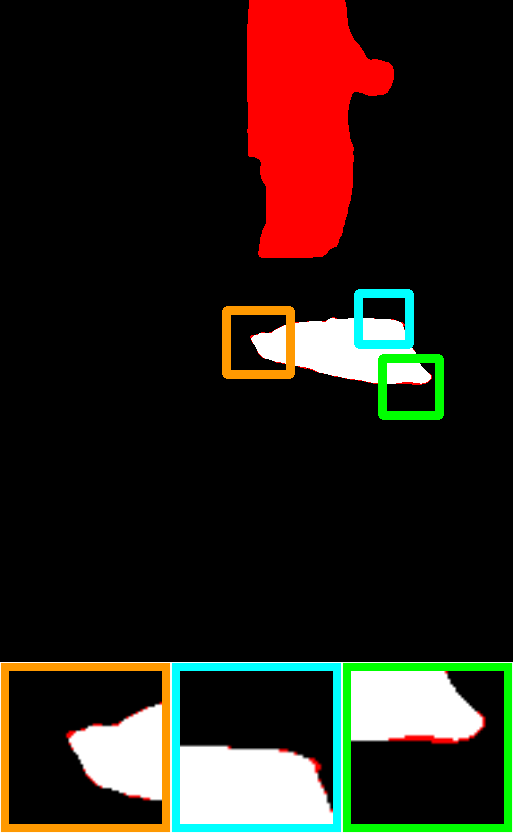} &
\includegraphics[width=0.12\textwidth]{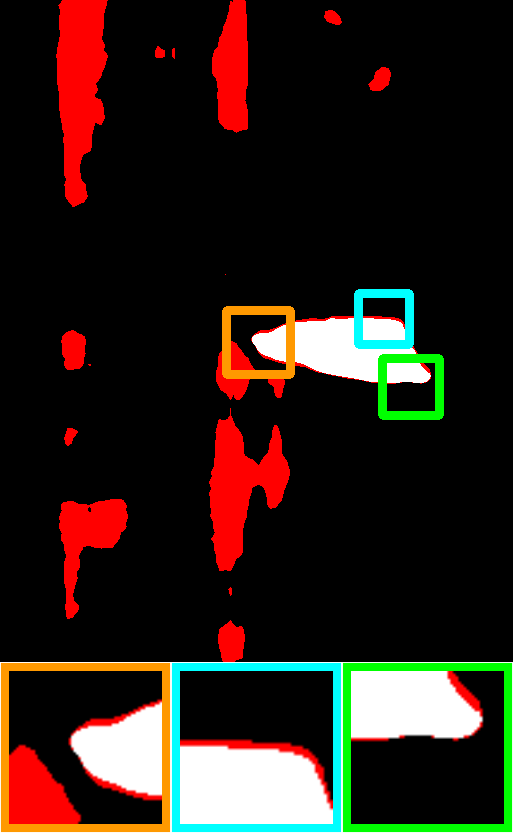} &
\includegraphics[width=0.12\textwidth]{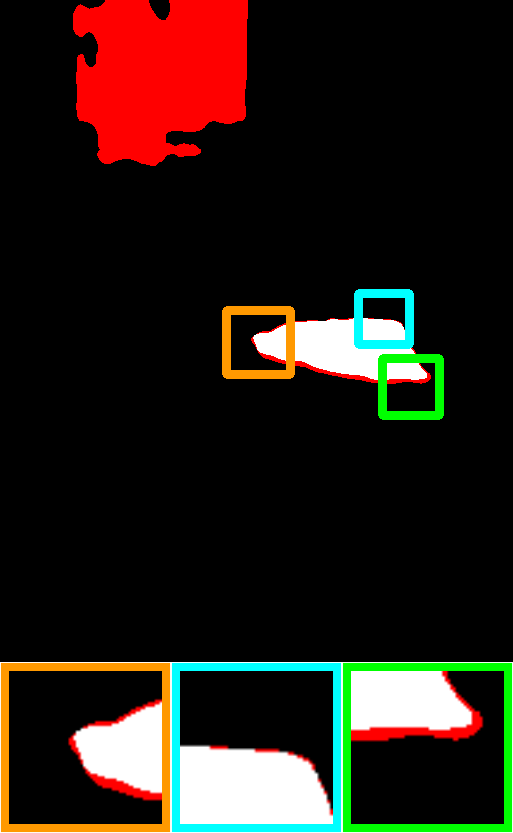} &
\includegraphics[width=0.12\textwidth]{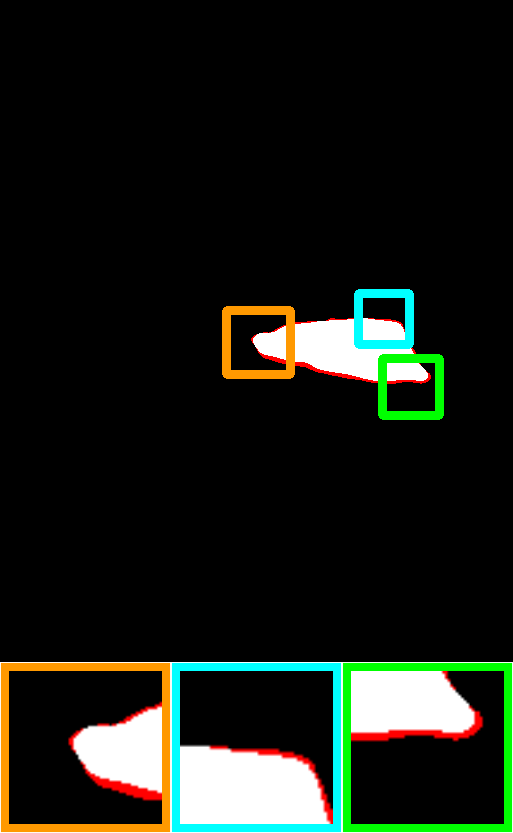} &
\includegraphics[width=0.12\textwidth]{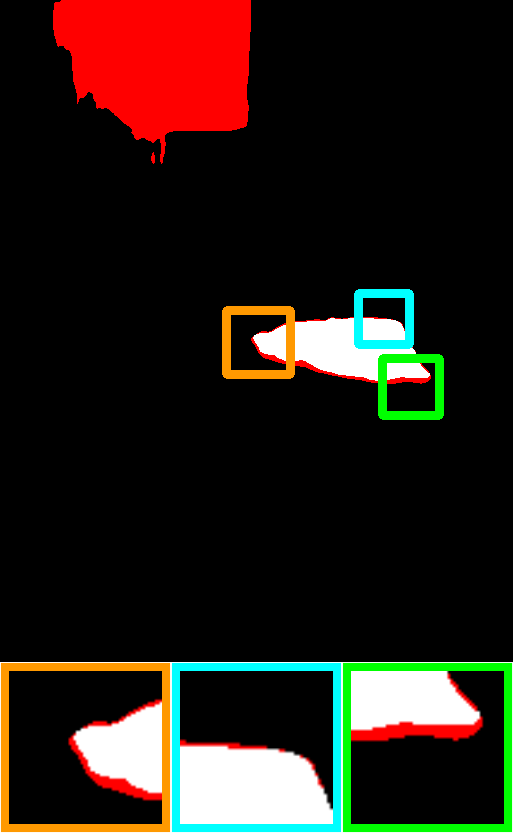} &
\includegraphics[width=0.12\textwidth]{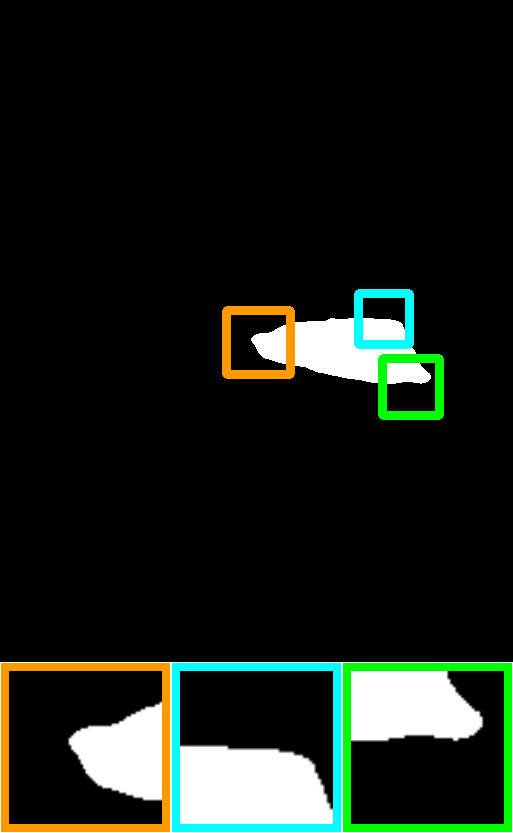} &
\includegraphics[width=0.12\textwidth]{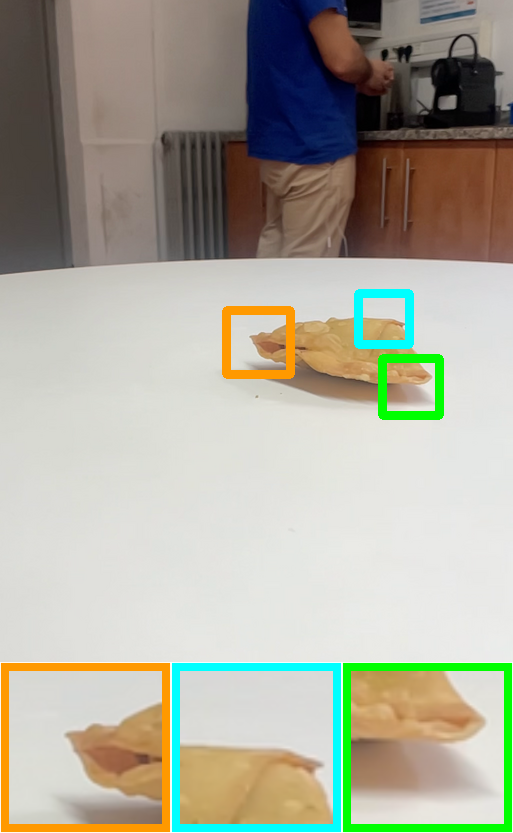} \\

\hline

\raisebox{+1.5cm}{\centering{French Bread}} & 

\includegraphics[width=0.12\textwidth]{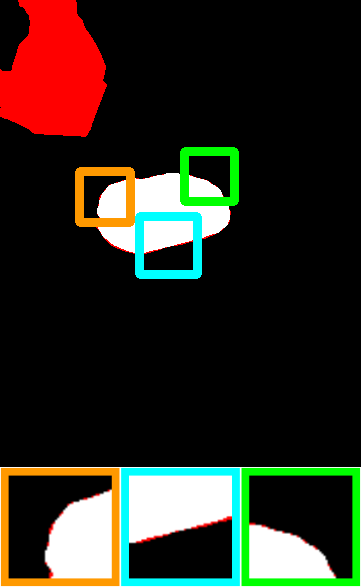} &
\includegraphics[width=0.12\textwidth]{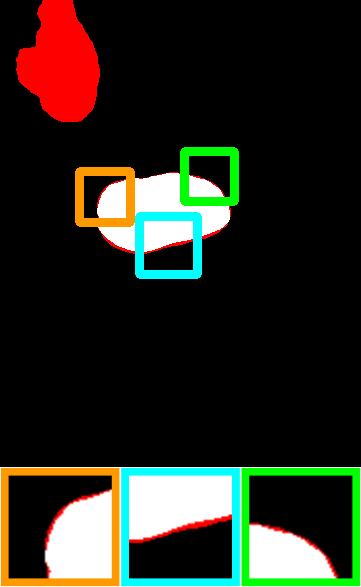} &
\includegraphics[width=0.12\textwidth]{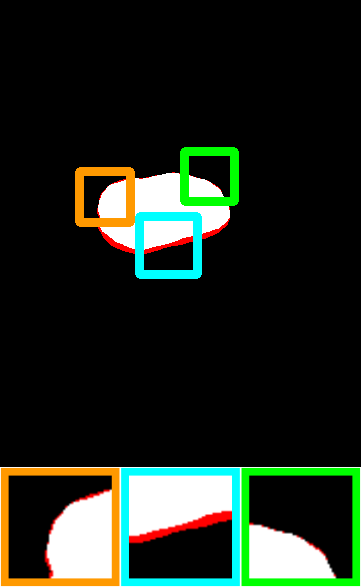} &
\includegraphics[width=0.12\textwidth]{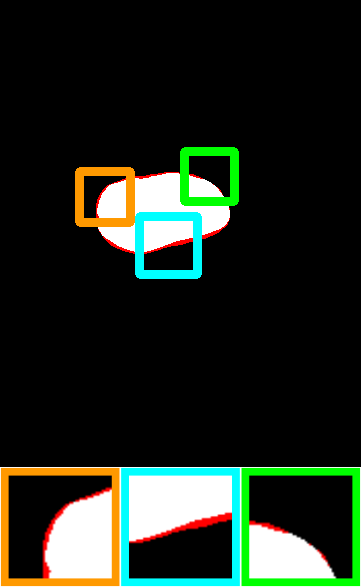} &
\includegraphics[width=0.12\textwidth]{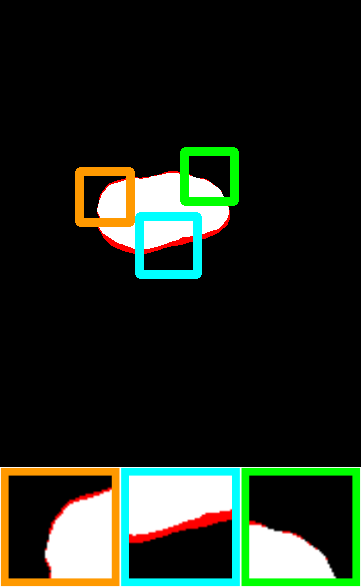} &
\includegraphics[width=0.12\textwidth]{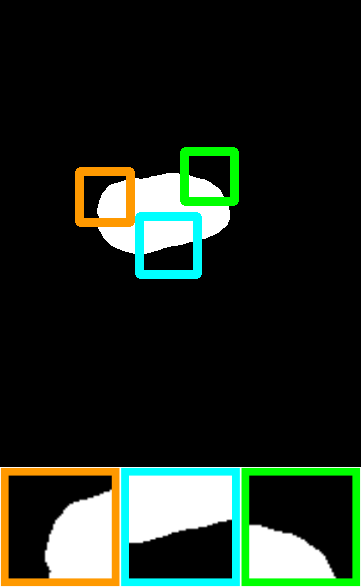} &
\includegraphics[width=0.12\textwidth]{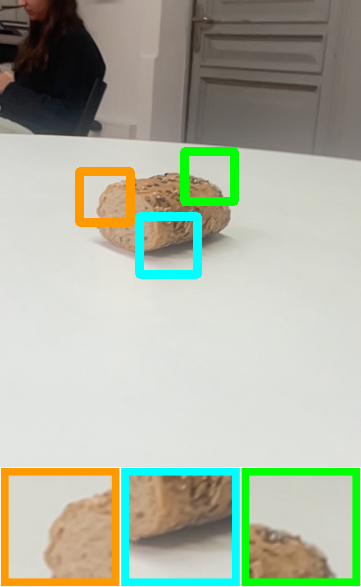} \\

\hline

    \end{tabular}
    \label{tab:FoodKit_2d_comparisons}
\end{table}

%%%%%%%%%%%%%%%%%%%%%%%%%%%%%%%%%%%%%%%%%%%%%%%%%%%%%

\begin{table}[!htbp]
    \centering
    \small
    \caption{Qualitative comparison of 2D Methods on MTF Dataset}
    \setlength{\tabcolsep}{1pt}
    \begin{tabular}{c|ccccccc}
    \hline
    %Datasets & BiRefNet & CCNET RELEM & FPN RELEM & SETR MLA & SWIN BASE & GT \\
    MTF & BiRefNet & CCNET & FPN & SeTR & SWIN & GT & RGB \\
    \hline
    \hline
    
\raisebox{+1.5cm}{\centering{5}} &

\includegraphics[width=0.12\textwidth]{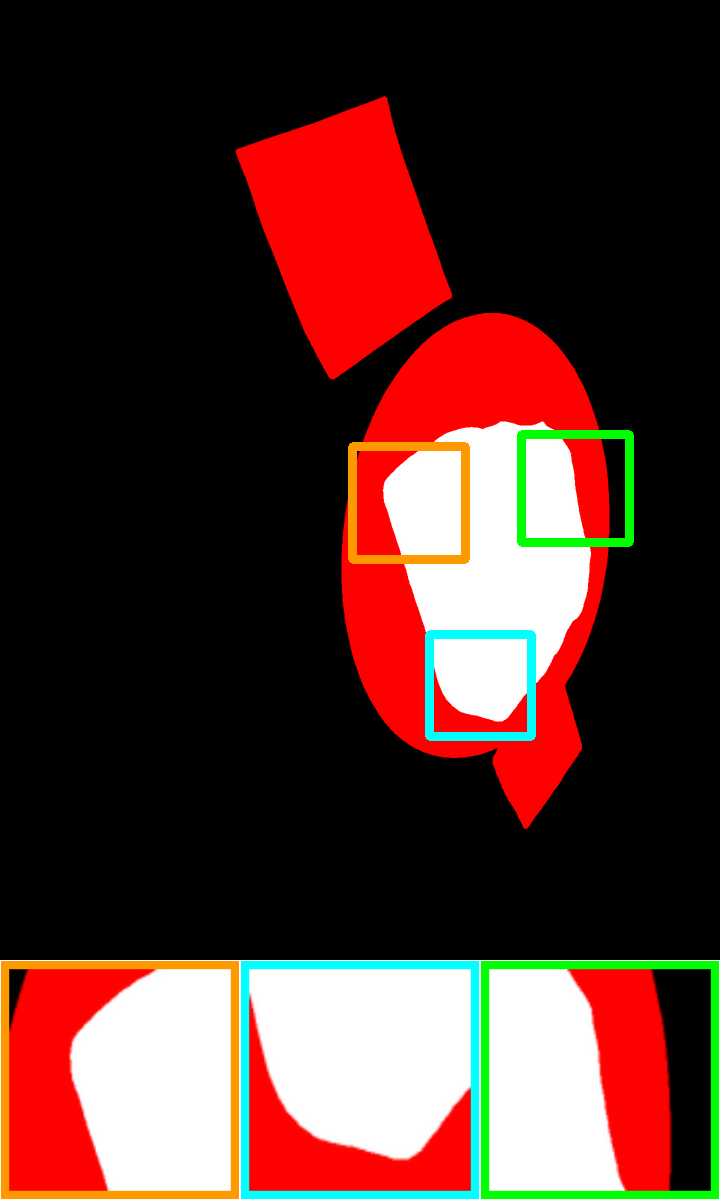} &
\includegraphics[width=0.12\textwidth]{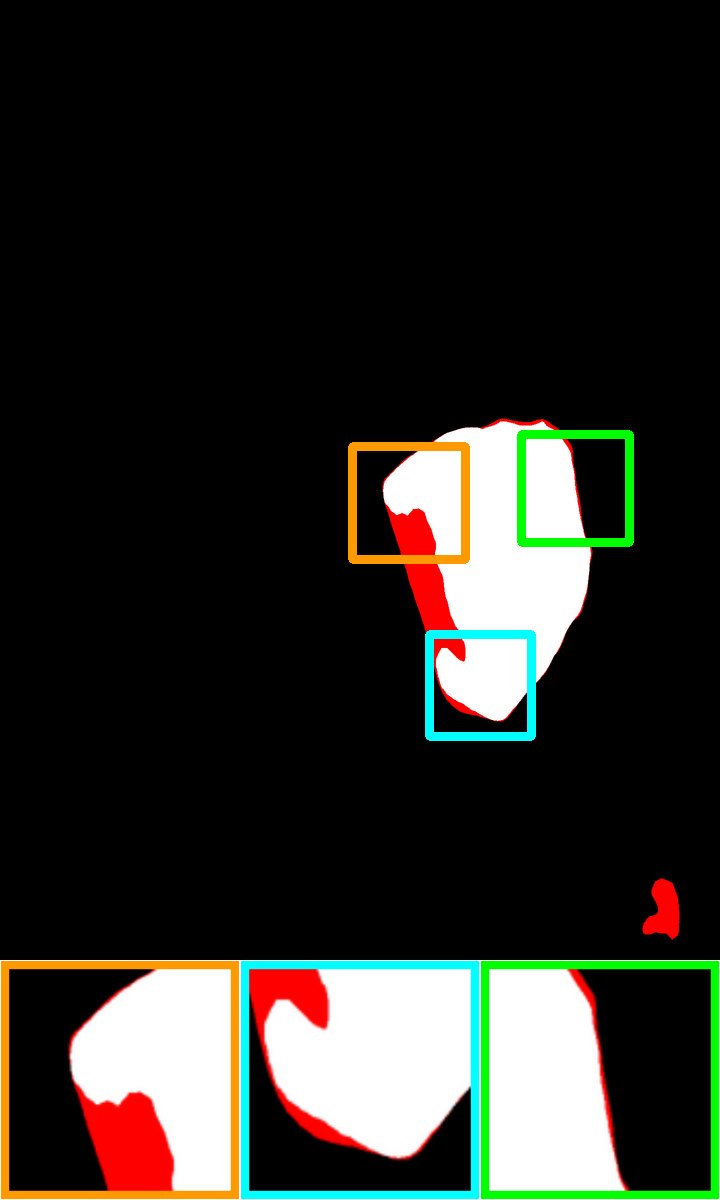} &
\includegraphics[width=0.12\textwidth]{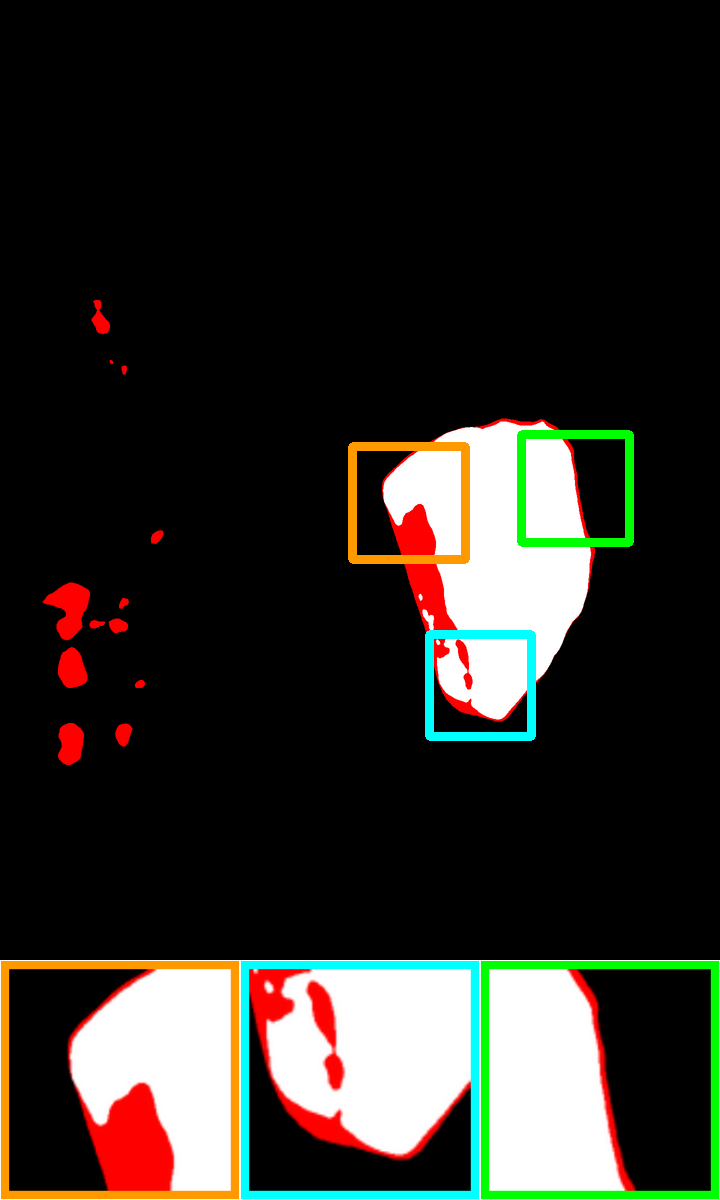} &
\includegraphics[width=0.12\textwidth]{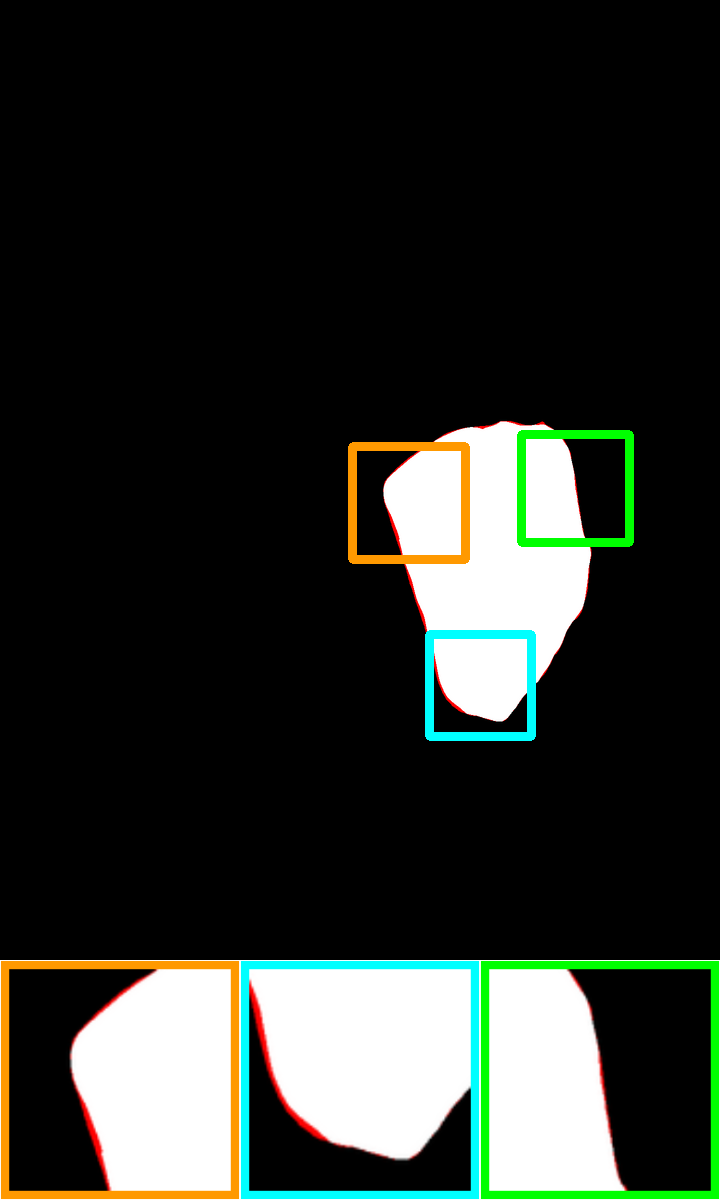} &
\includegraphics[width=0.12\textwidth]{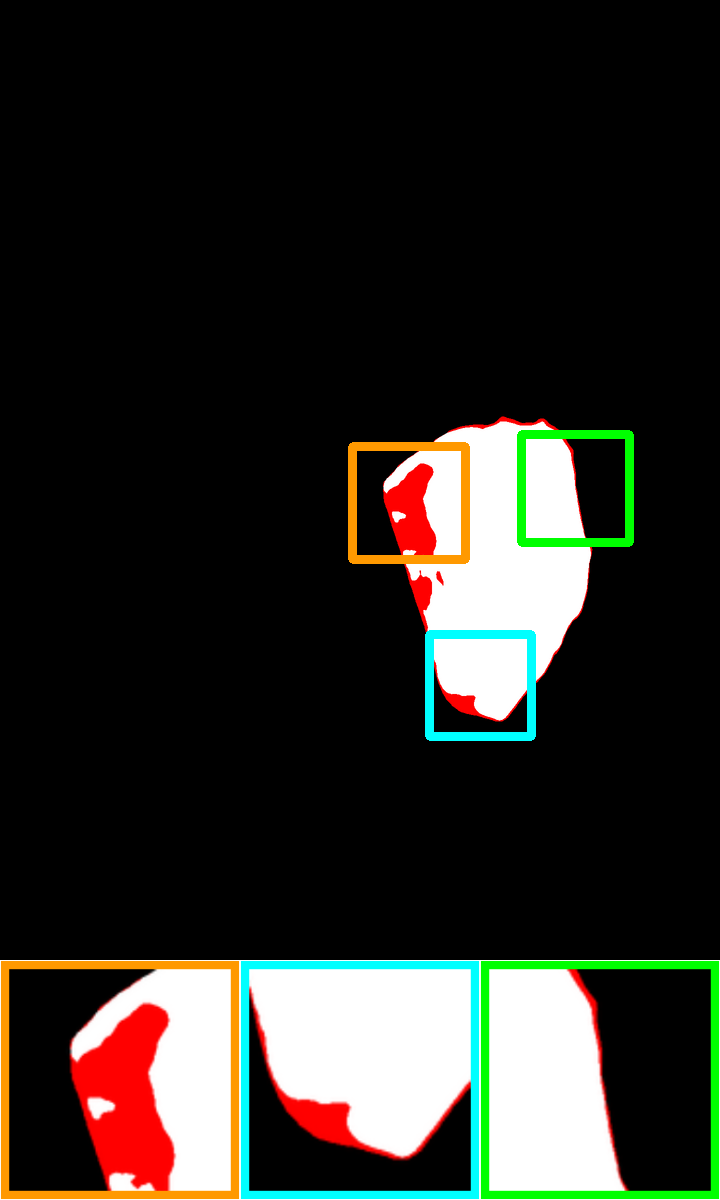} &
\includegraphics[width=0.12\textwidth]{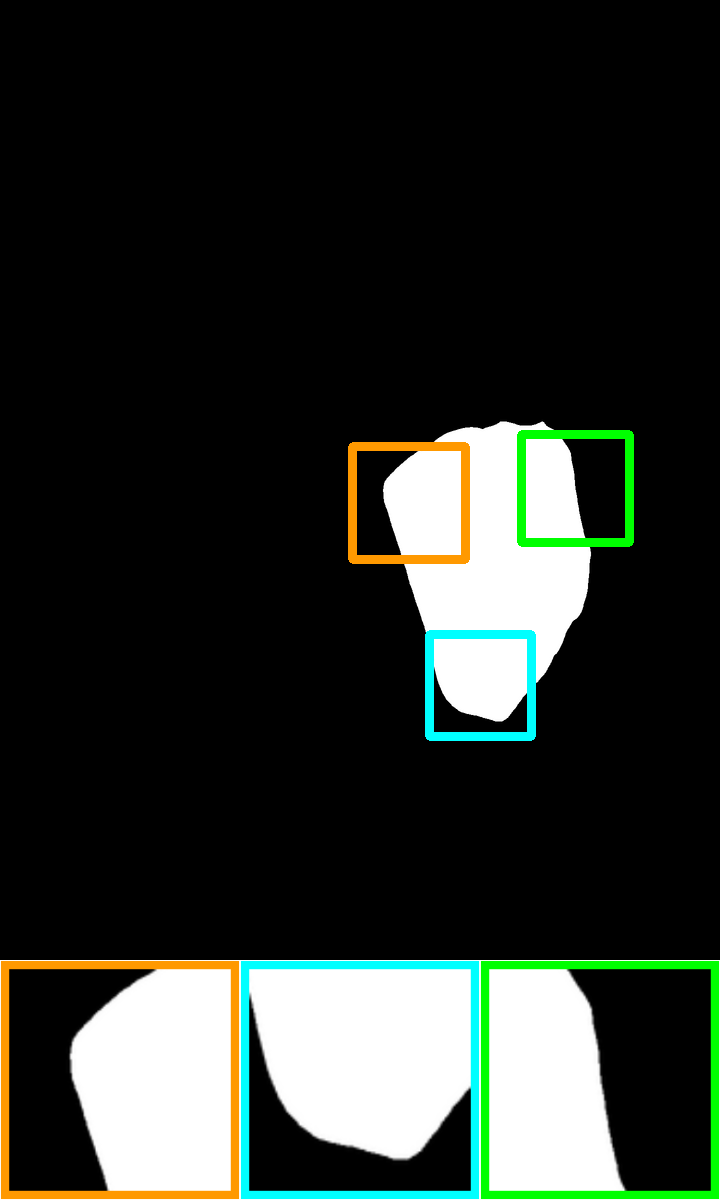} &
\includegraphics[width=0.12\textwidth]{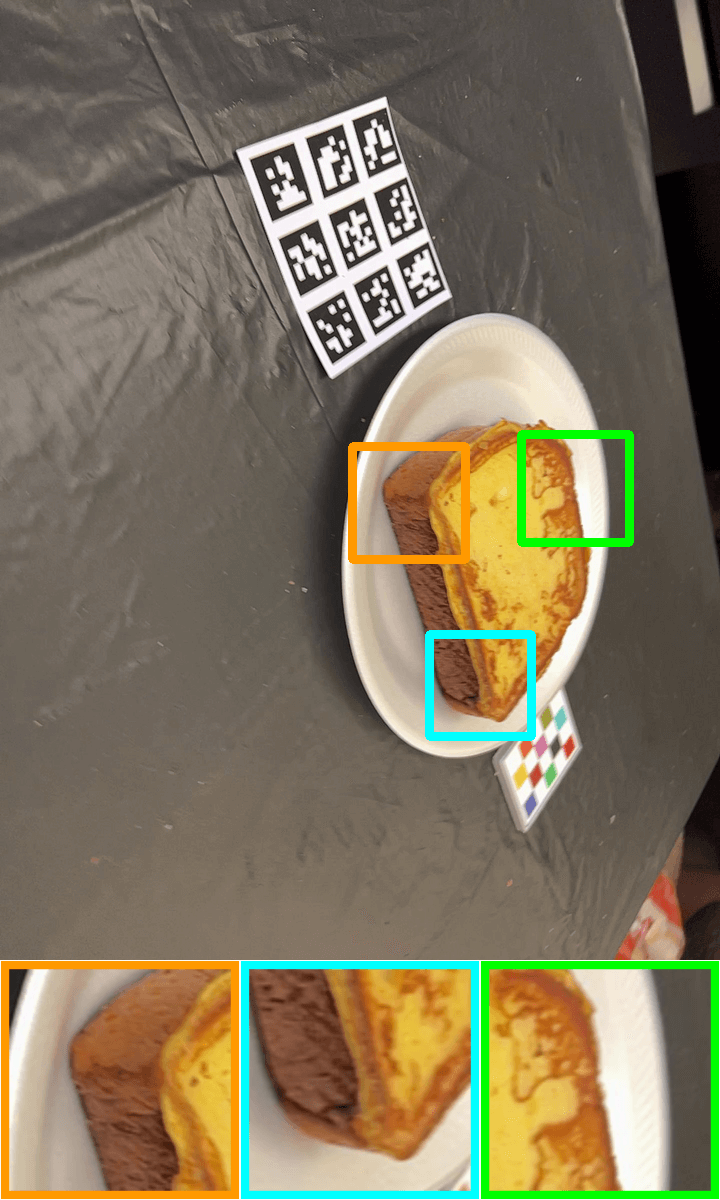} \\

\hline

\raisebox{+1.5cm}{\centering{7}} & 

\includegraphics[width=0.12\textwidth]{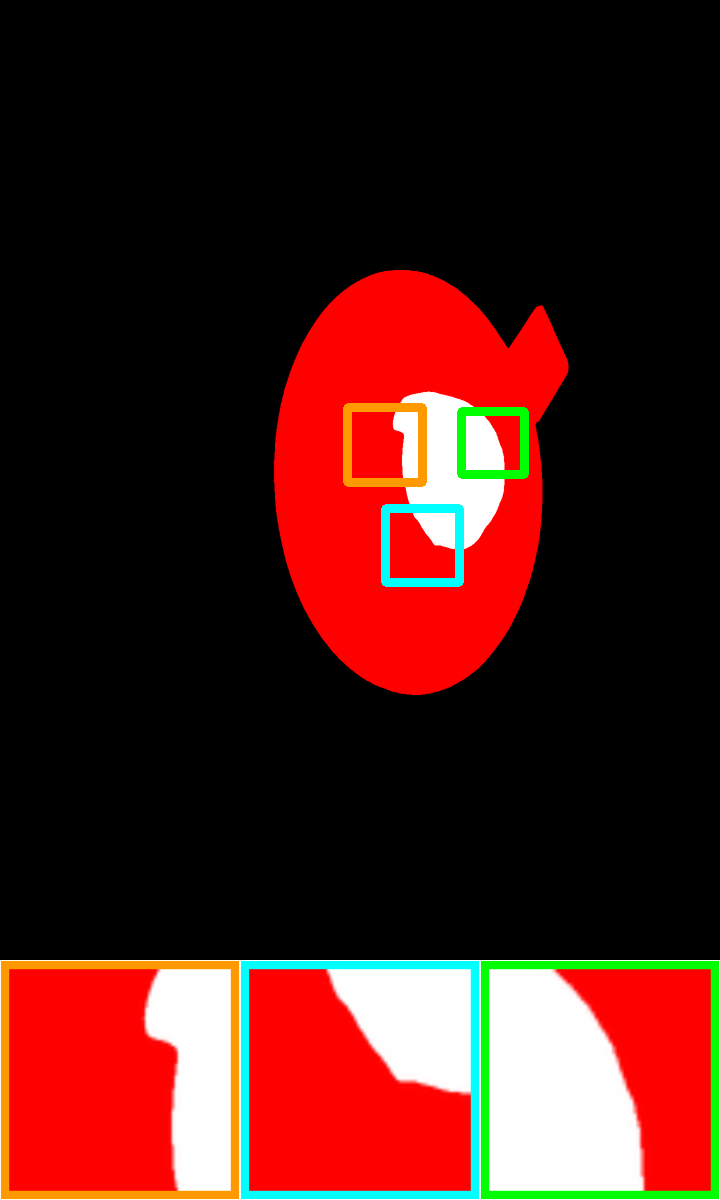} &
\includegraphics[width=0.12\textwidth]{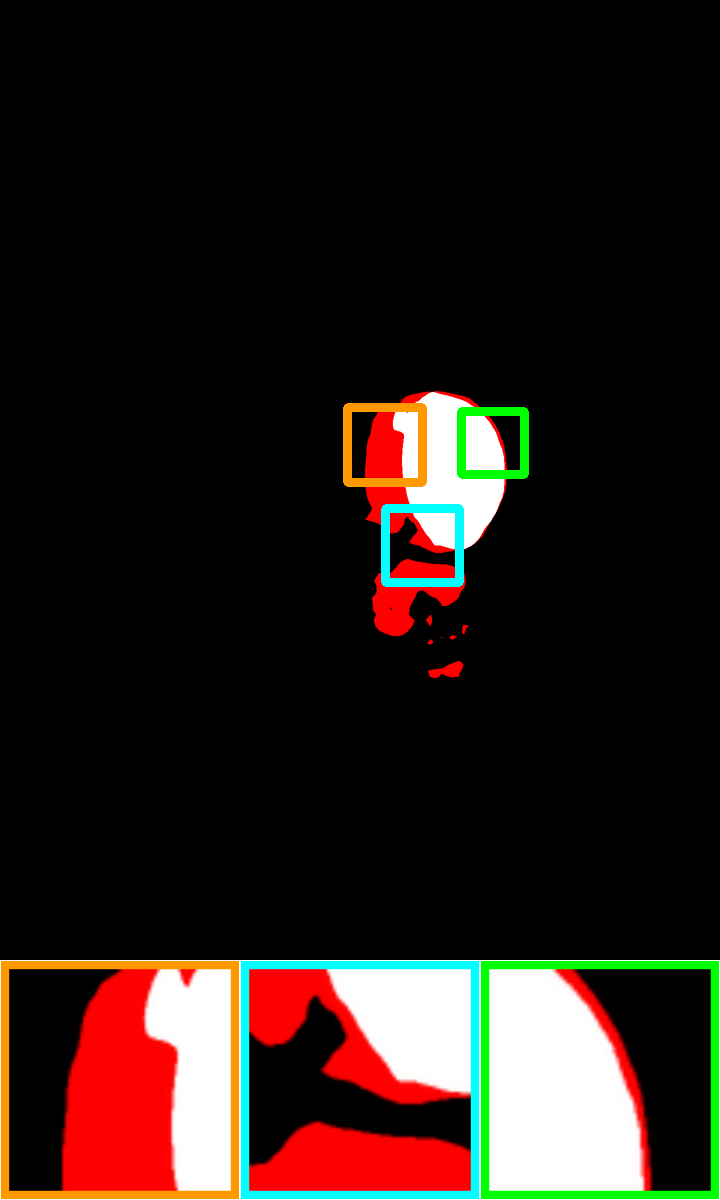} &
\includegraphics[width=0.12\textwidth]{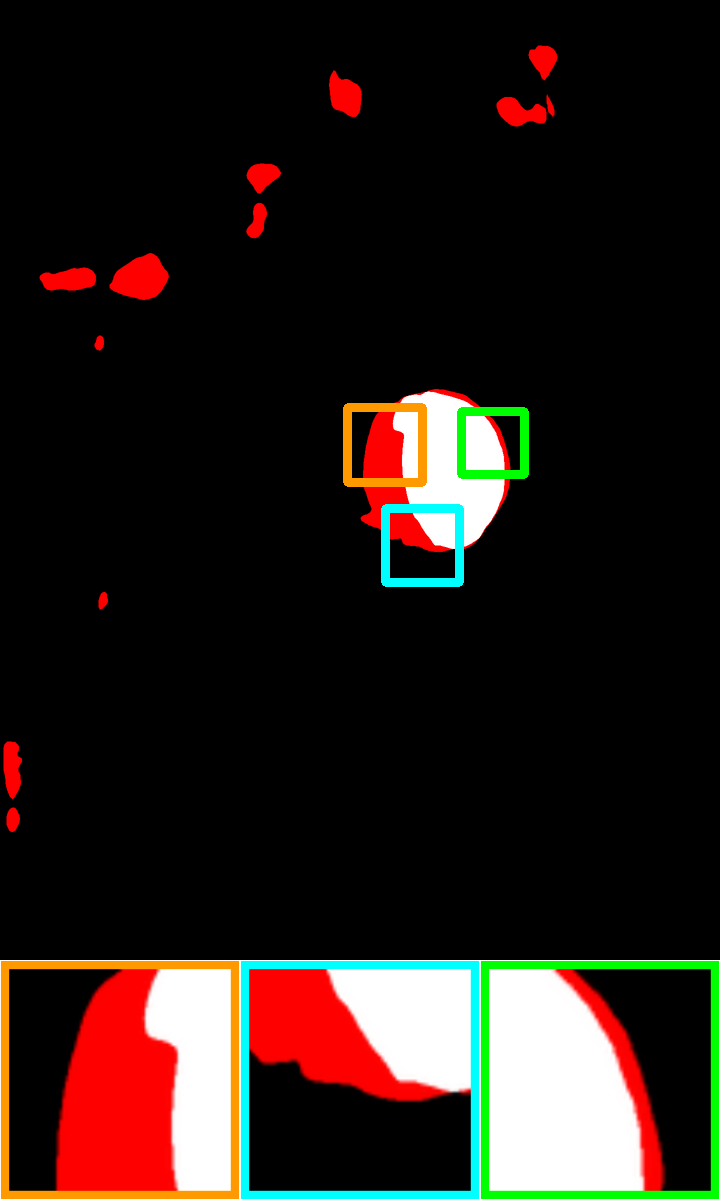} &
\includegraphics[width=0.12\textwidth]{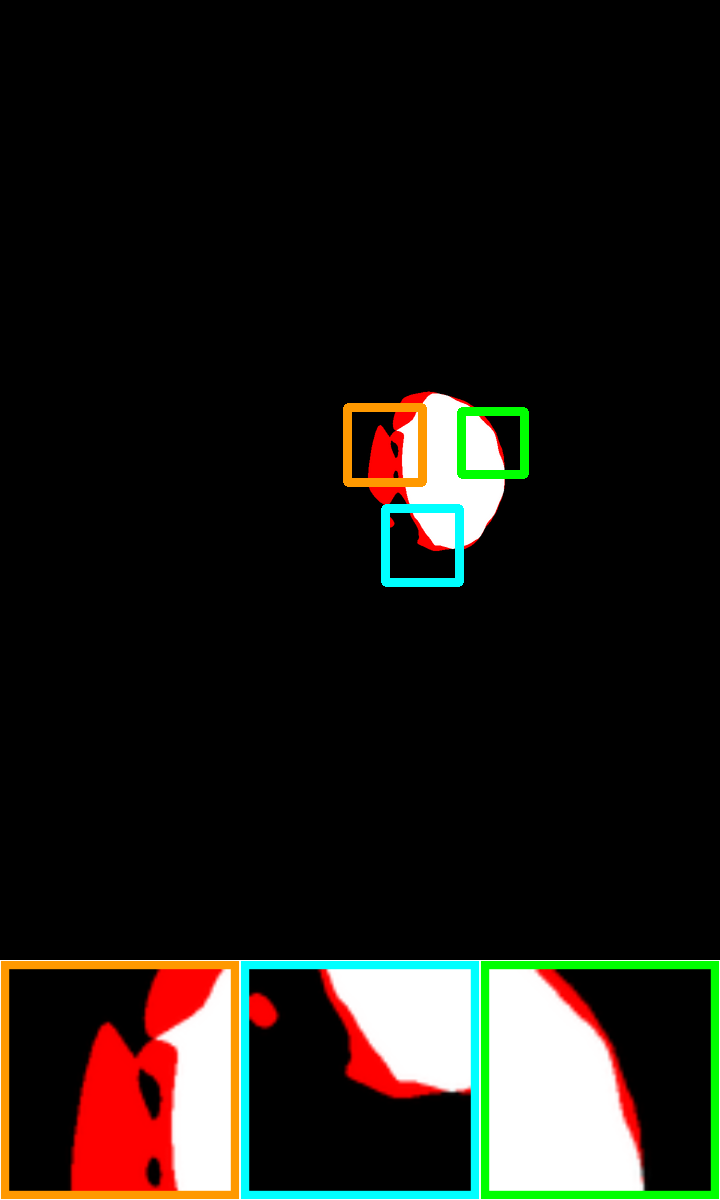} &
\includegraphics[width=0.12\textwidth]{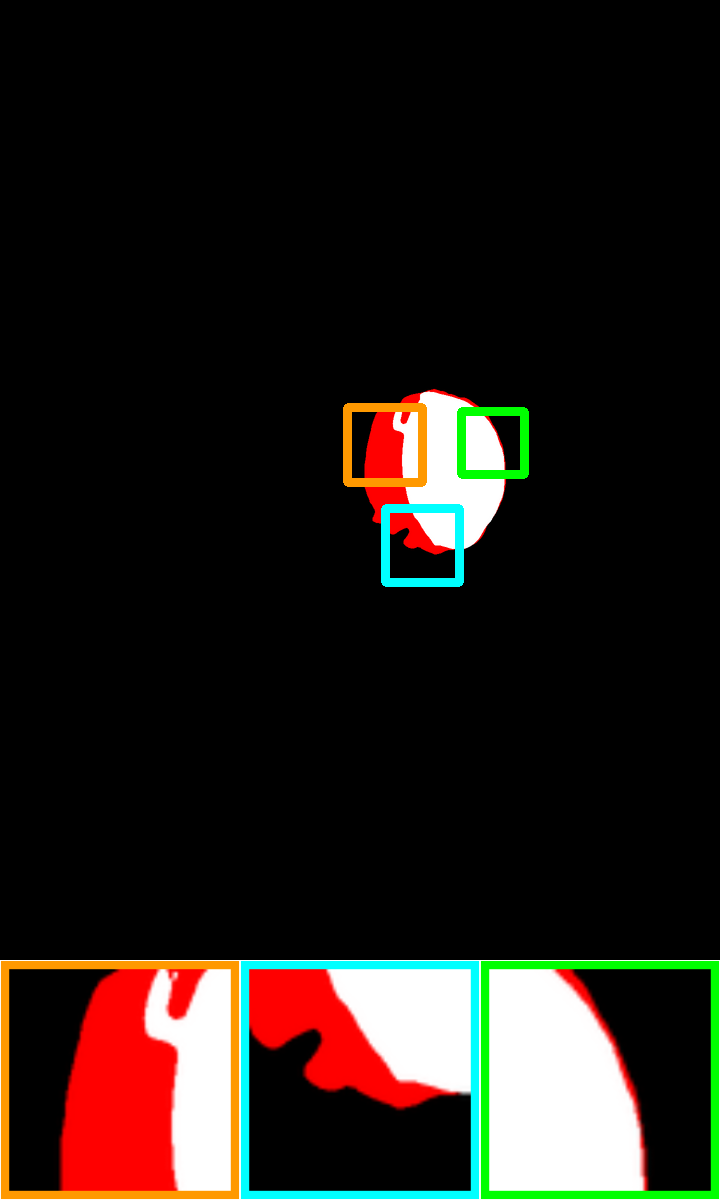} &
\includegraphics[width=0.12\textwidth]{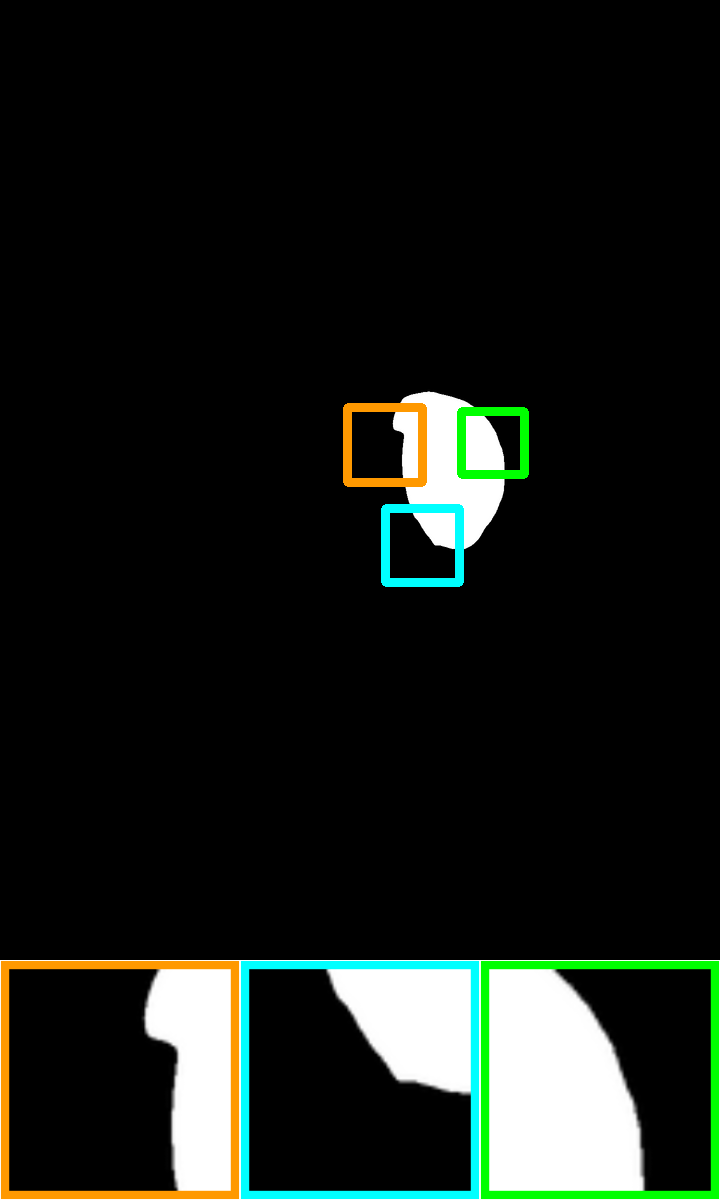} &
\includegraphics[width=0.12\textwidth]{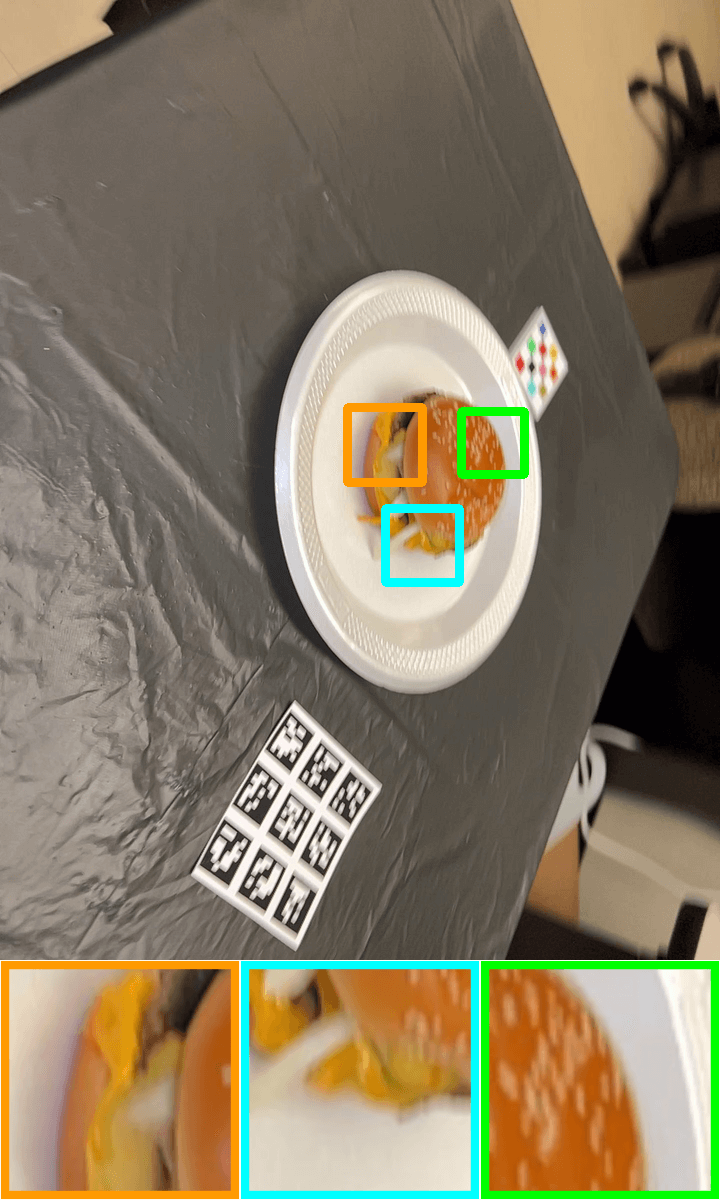} \\

\hline

\raisebox{+1.5cm}{\centering{8}} & 

\includegraphics[width=0.12\textwidth]{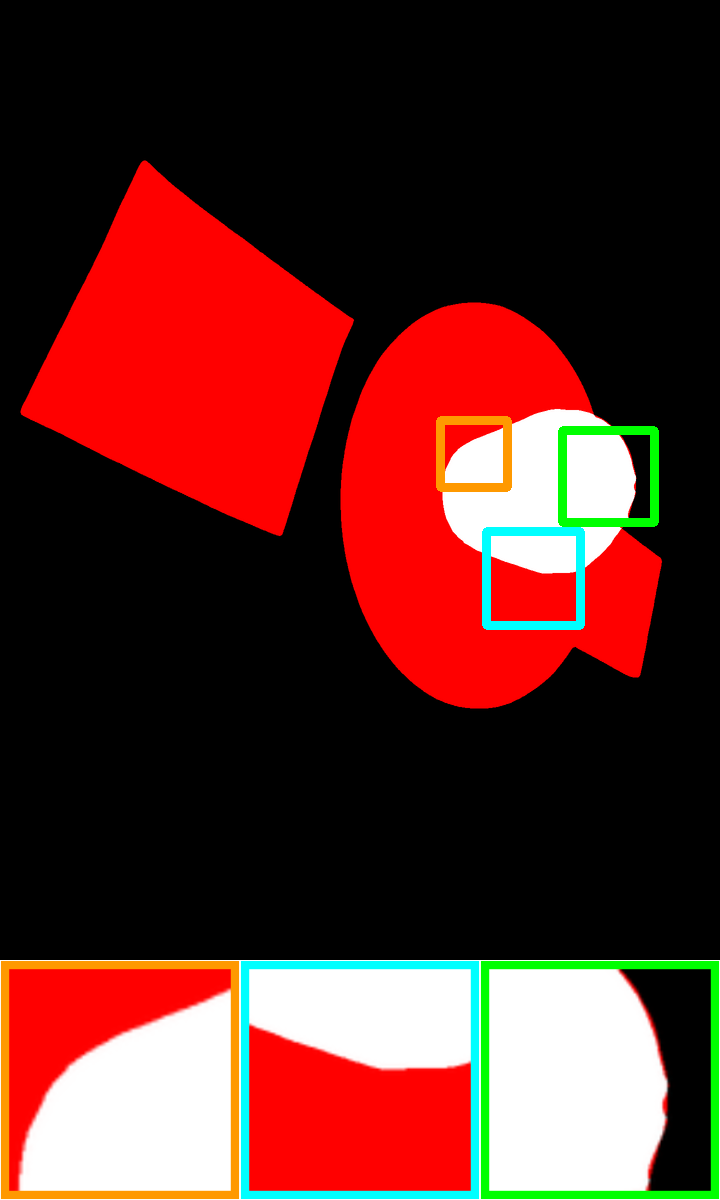} &
\includegraphics[width=0.12\textwidth]{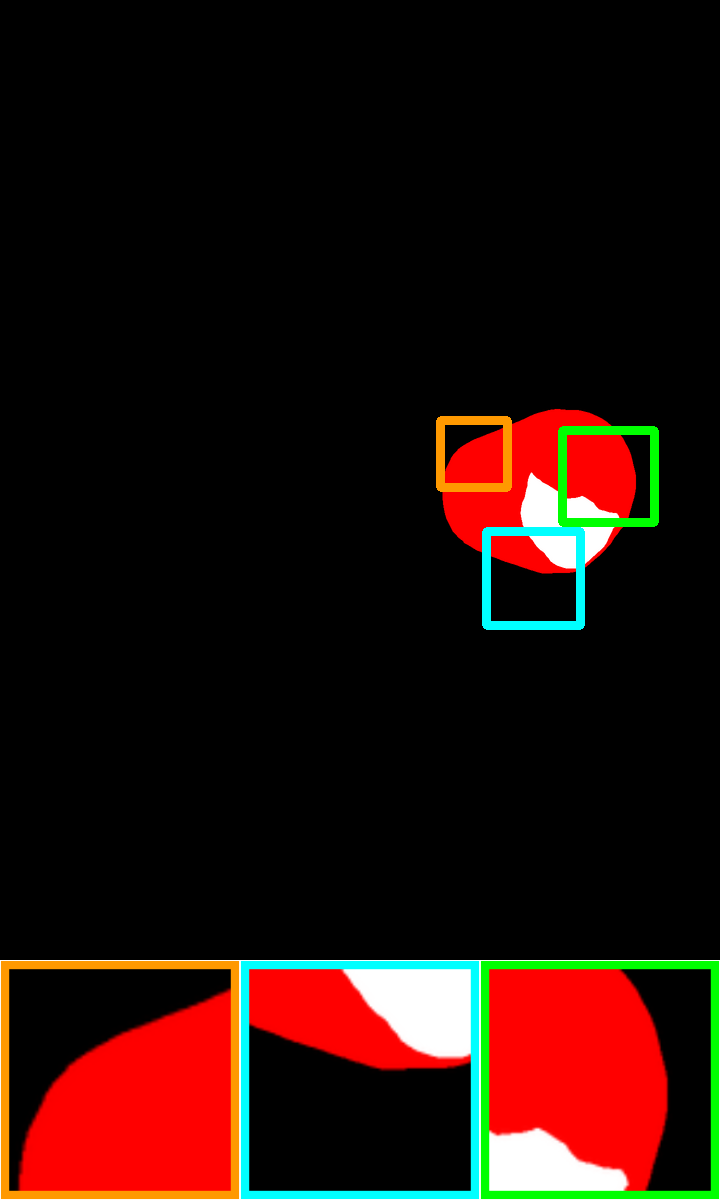} &
\includegraphics[width=0.12\textwidth]{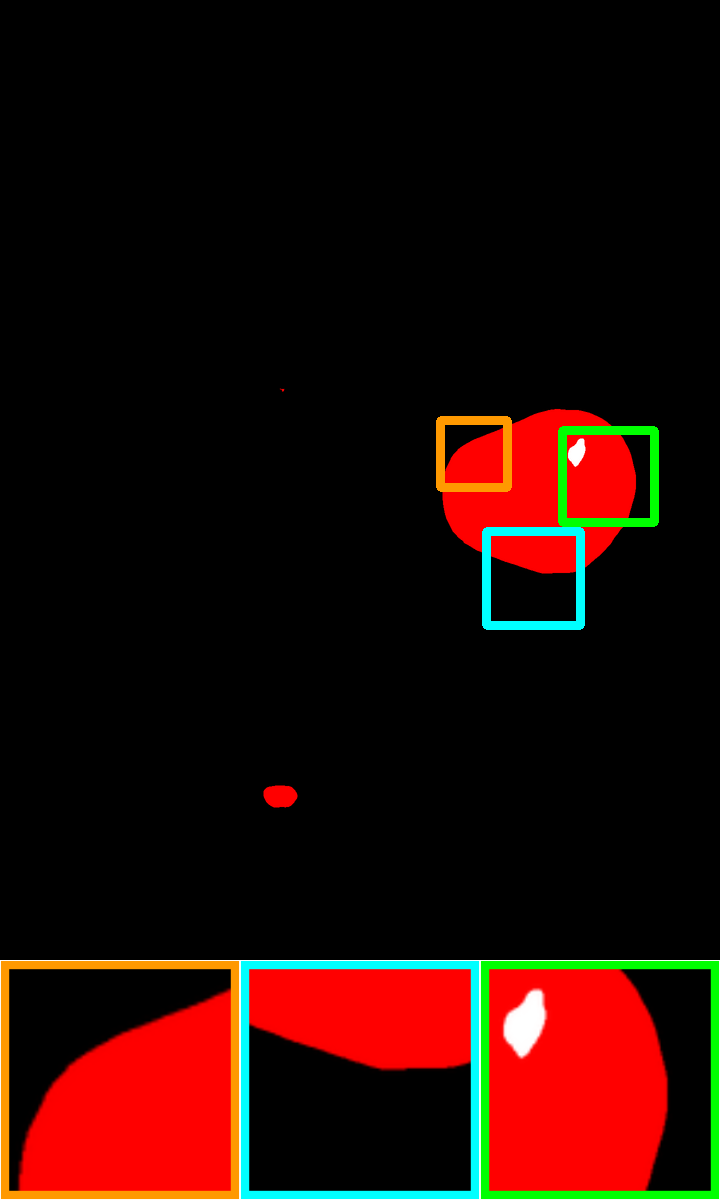} &
\includegraphics[width=0.12\textwidth]{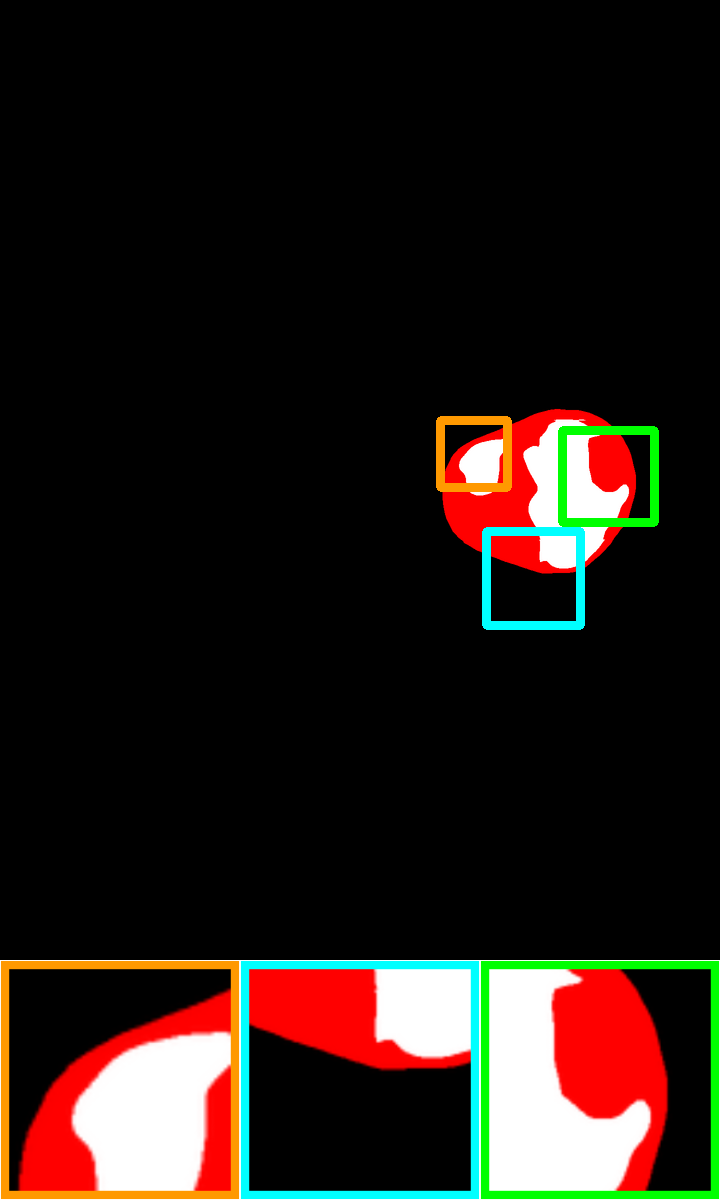} &
\includegraphics[width=0.12\textwidth]{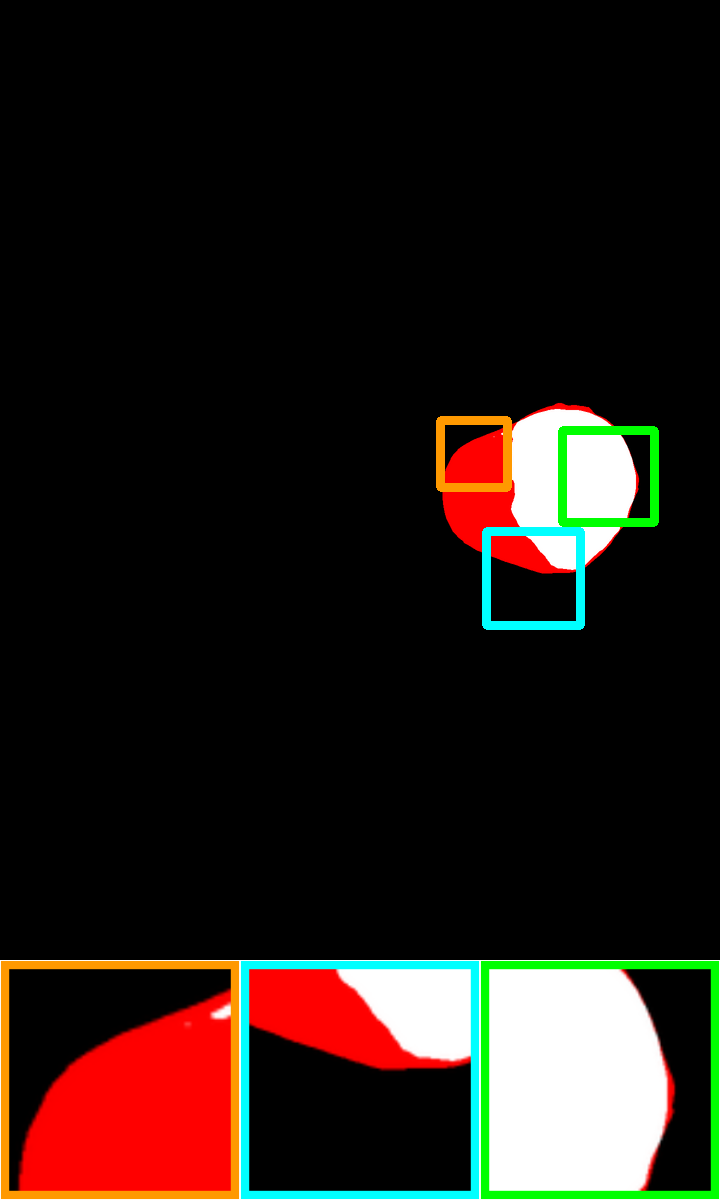} &
\includegraphics[width=0.12\textwidth]{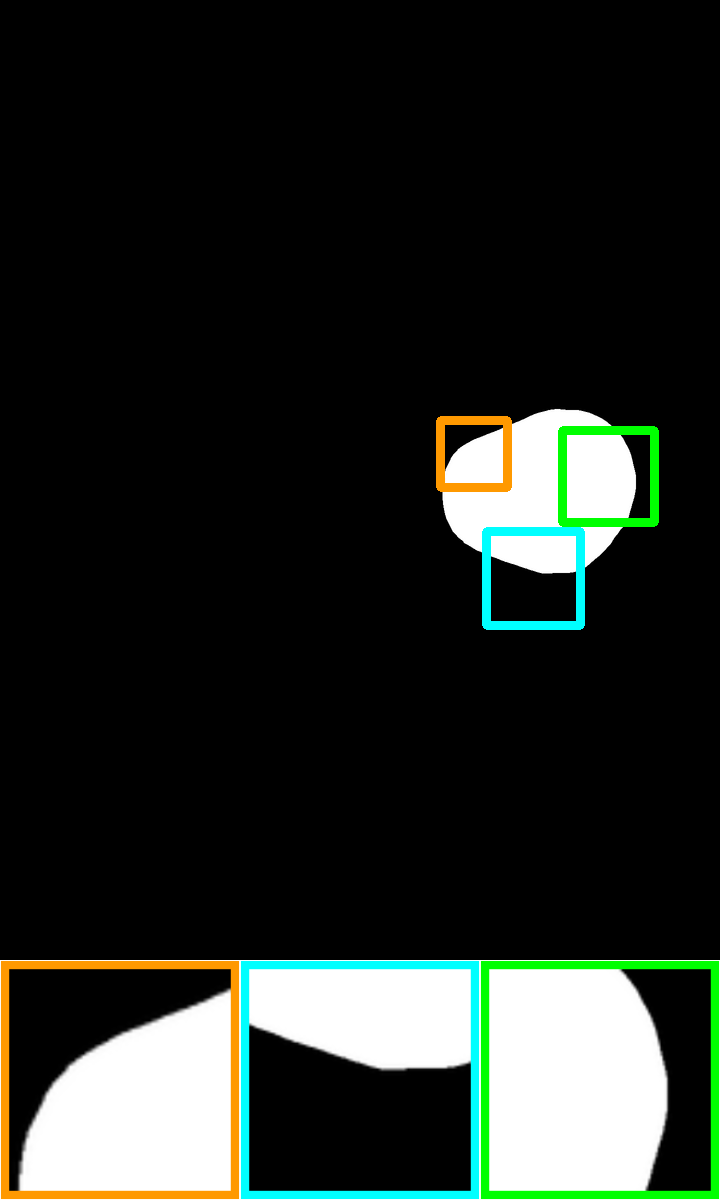} &
\includegraphics[width=0.12\textwidth]{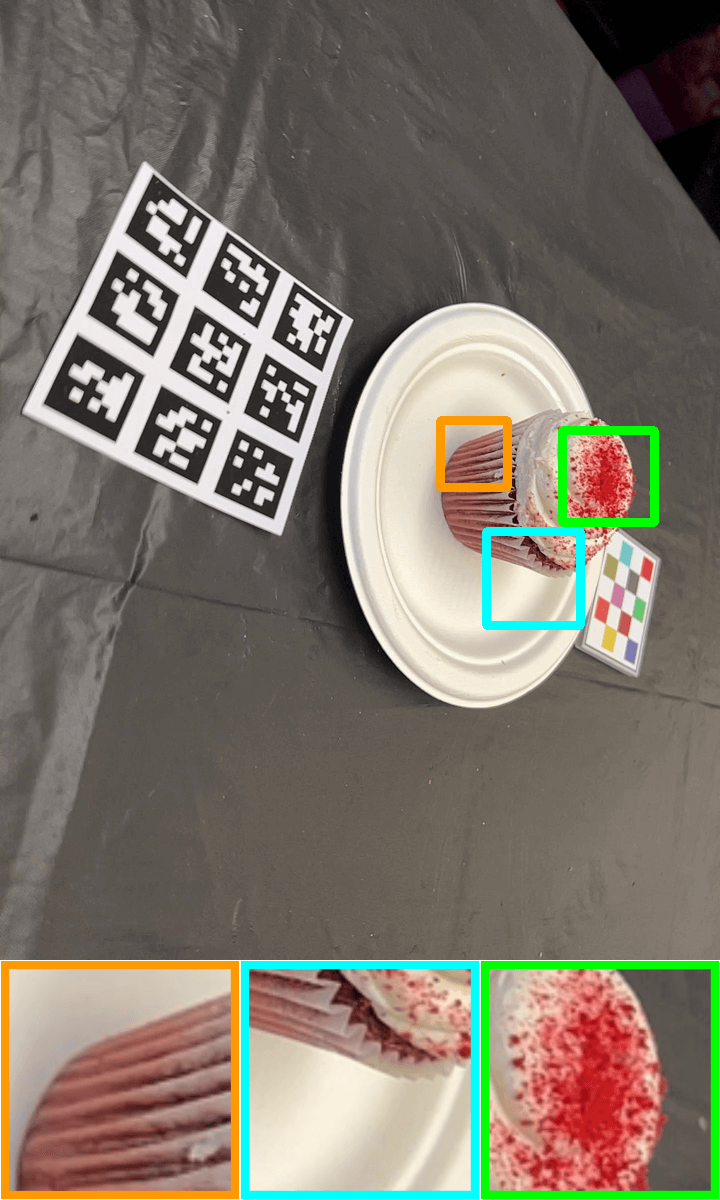} \\

\hline

\raisebox{+1.5cm}{\centering{11}} &

\includegraphics[width=0.12\textwidth]{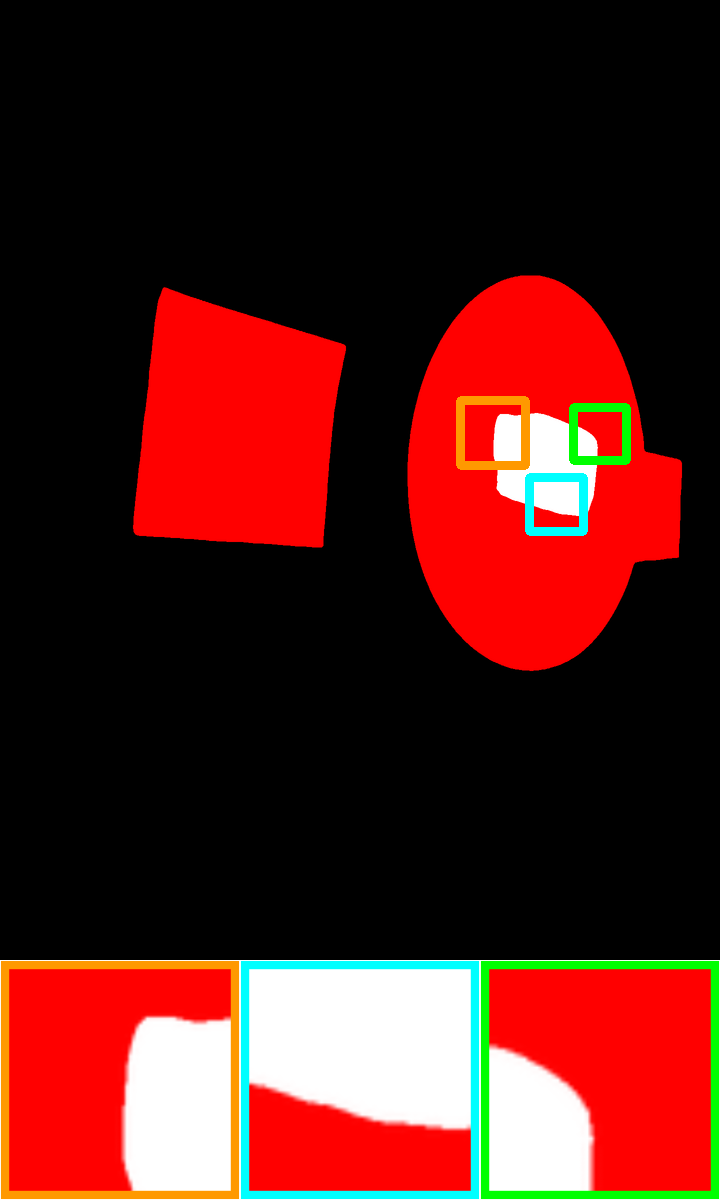} &
\includegraphics[width=0.12\textwidth]{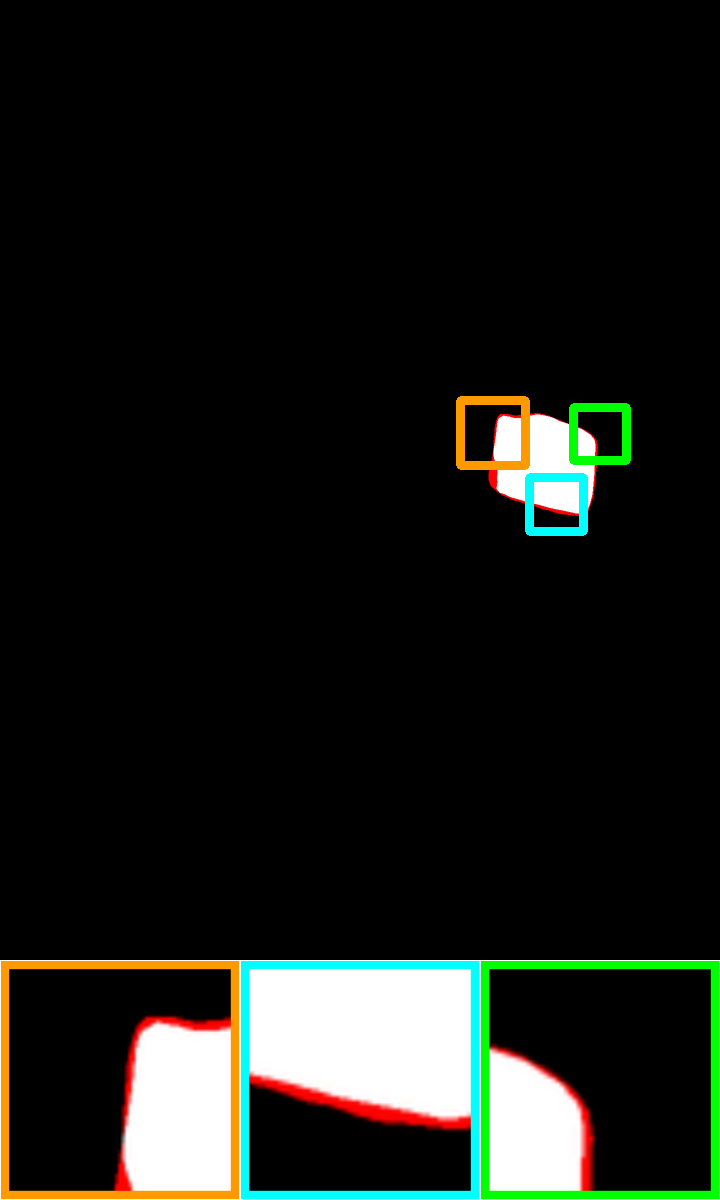} &
\includegraphics[width=0.12\textwidth]{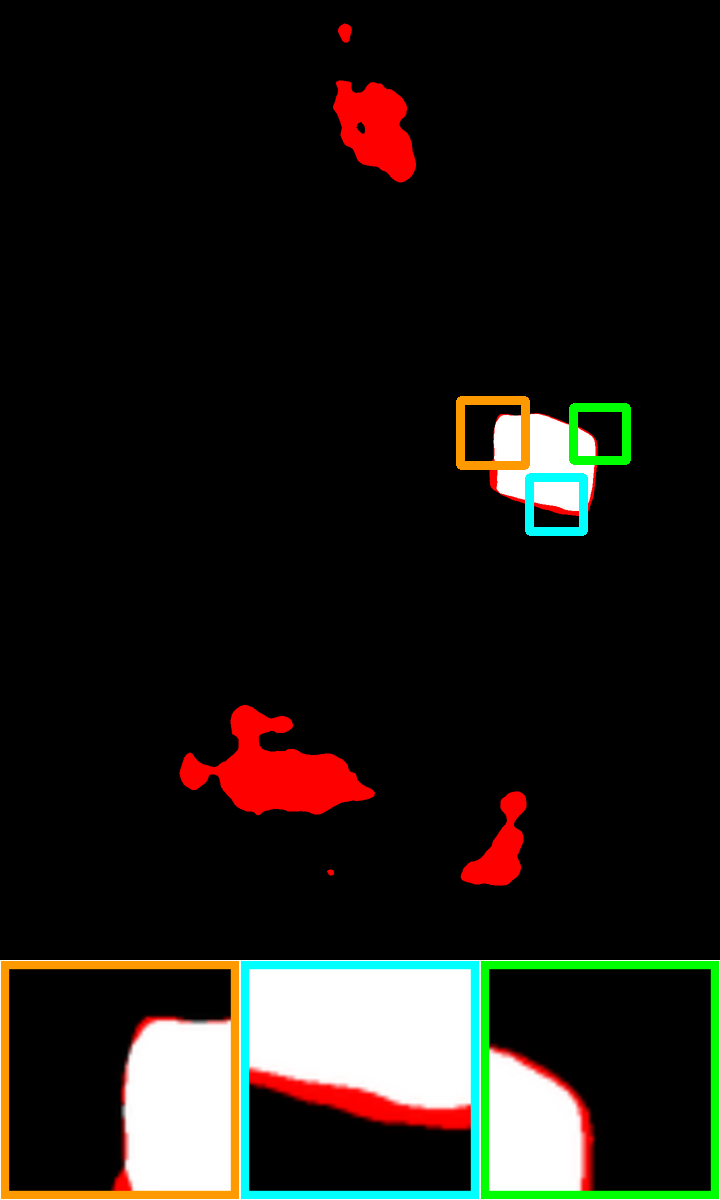} &
\includegraphics[width=0.12\textwidth]{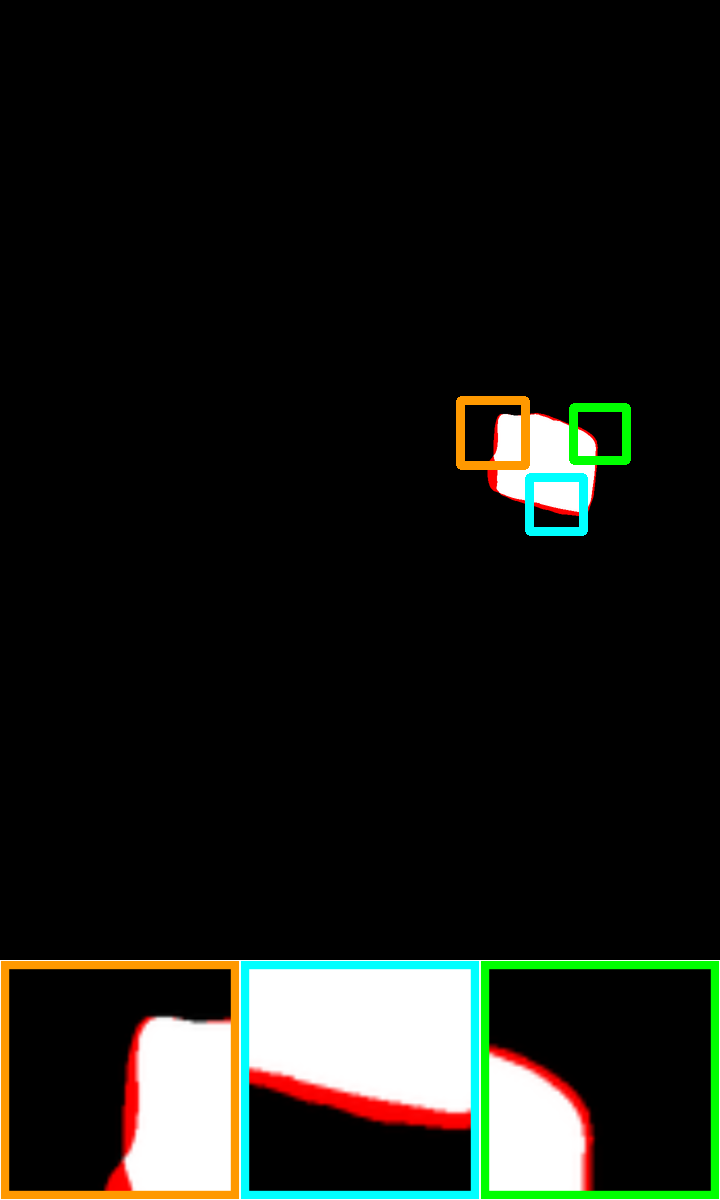} &
\includegraphics[width=0.12\textwidth]{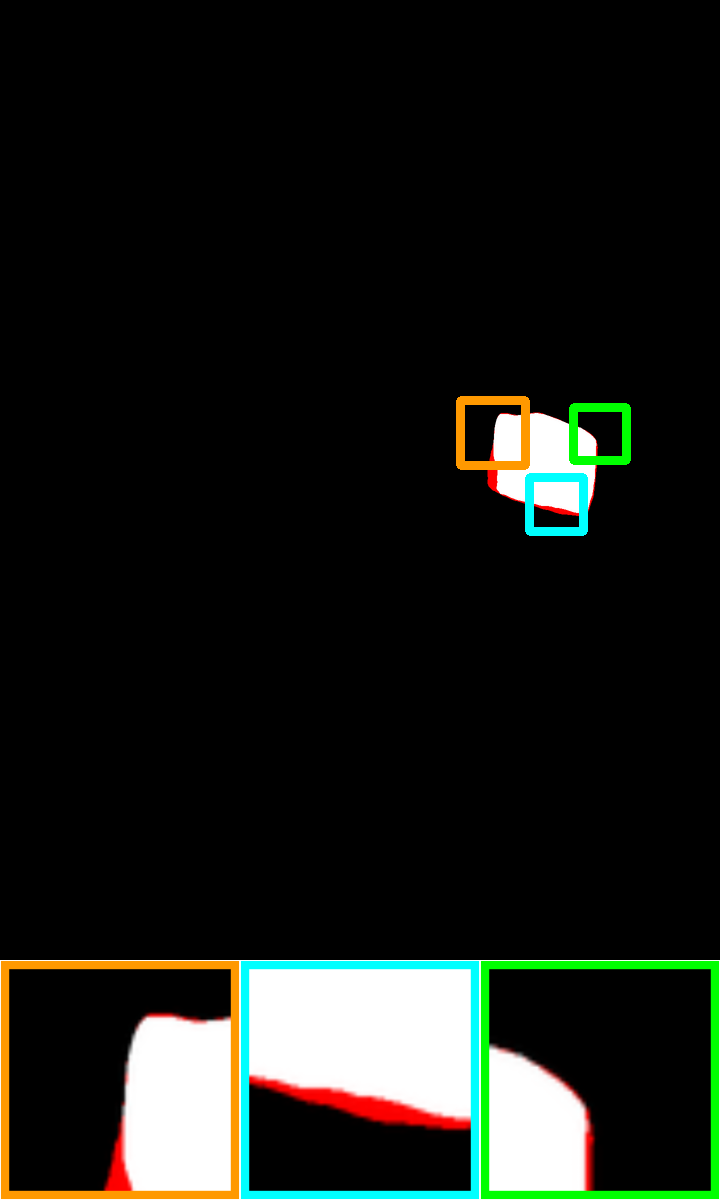} &
\includegraphics[width=0.12\textwidth]{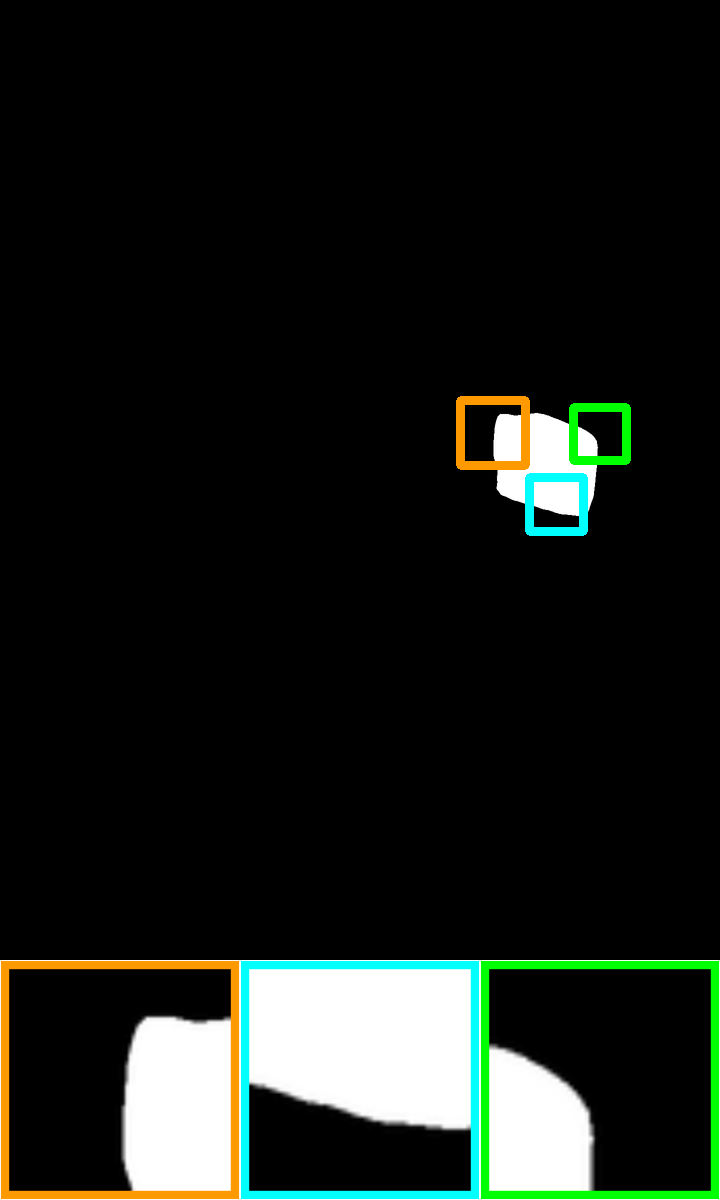} &
\includegraphics[width=0.12\textwidth]{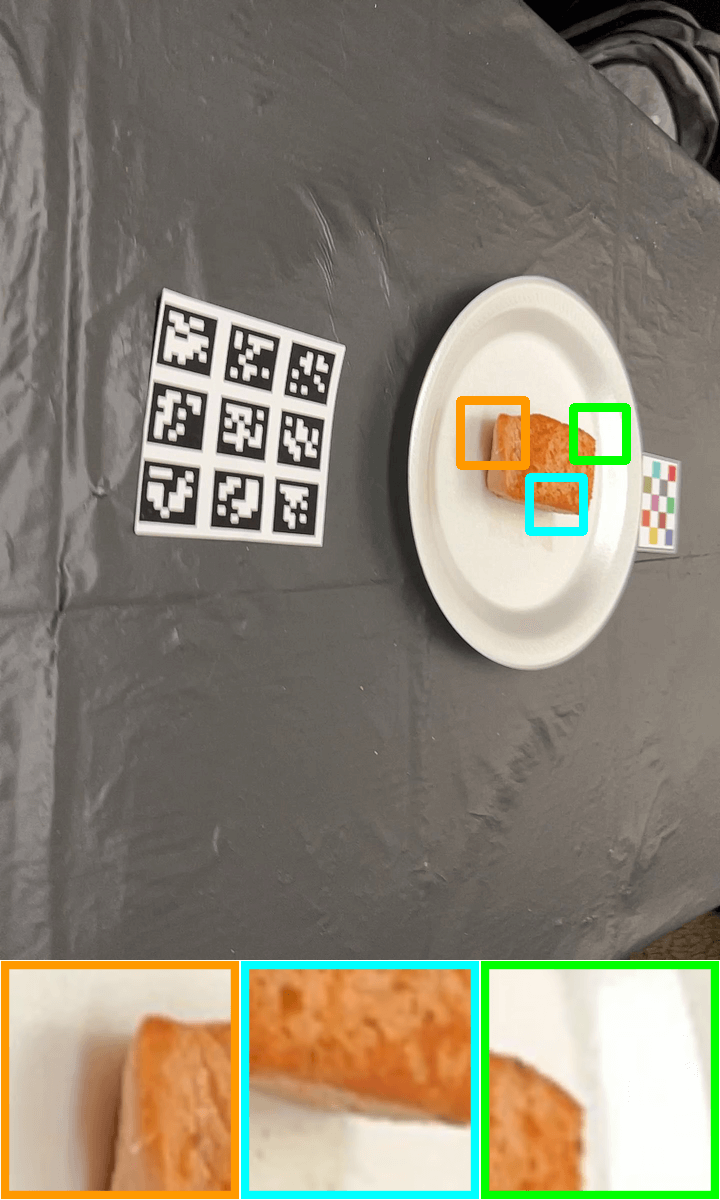} \\

\hline

\raisebox{+1.5cm}{\centering{14}} &

\includegraphics[width=0.12\textwidth]{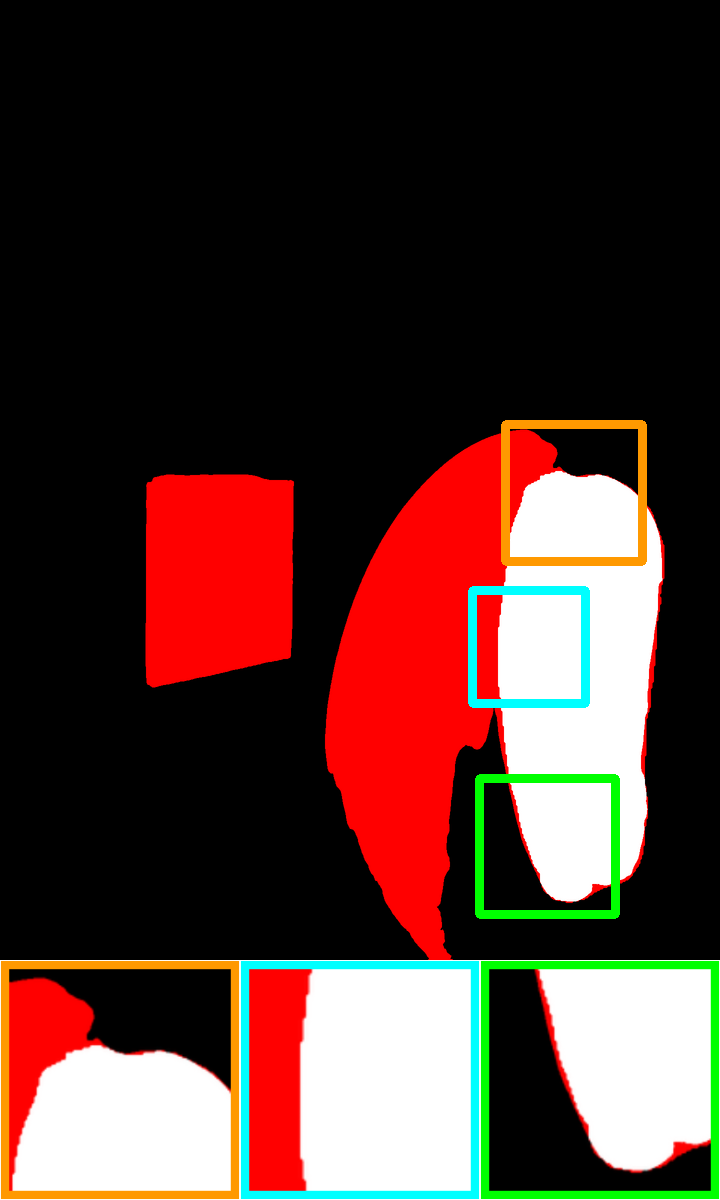} &
\includegraphics[width=0.12\textwidth]{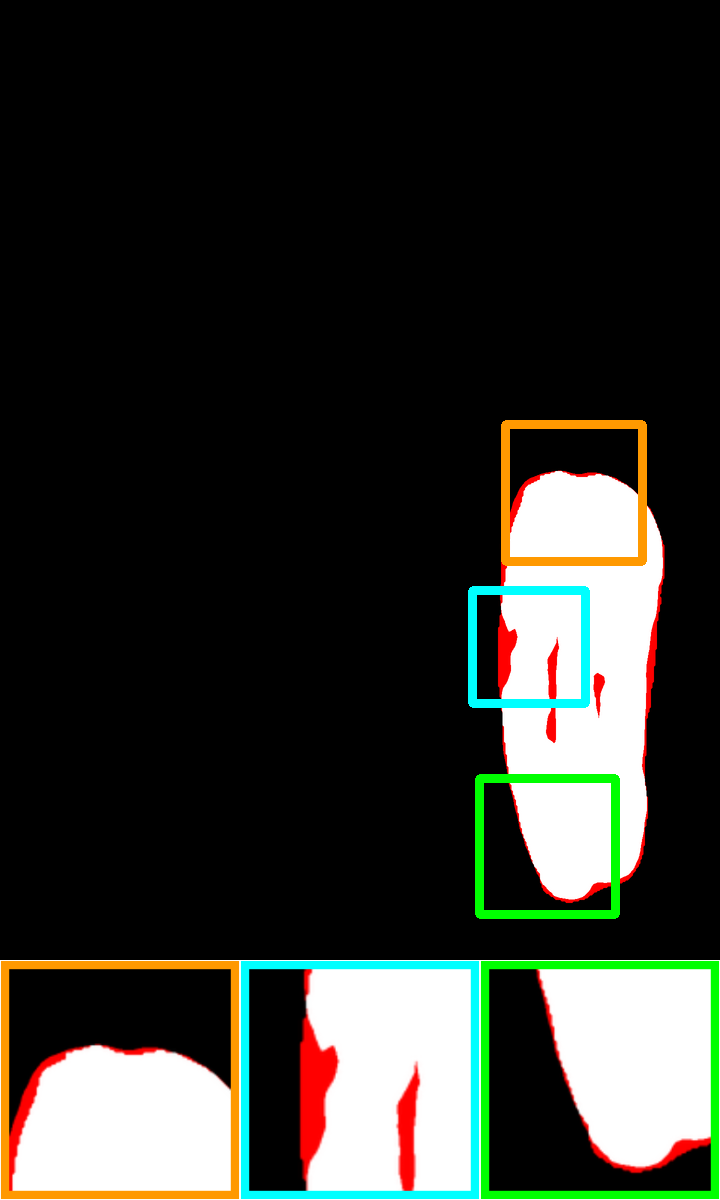} &
\includegraphics[width=0.12\textwidth]{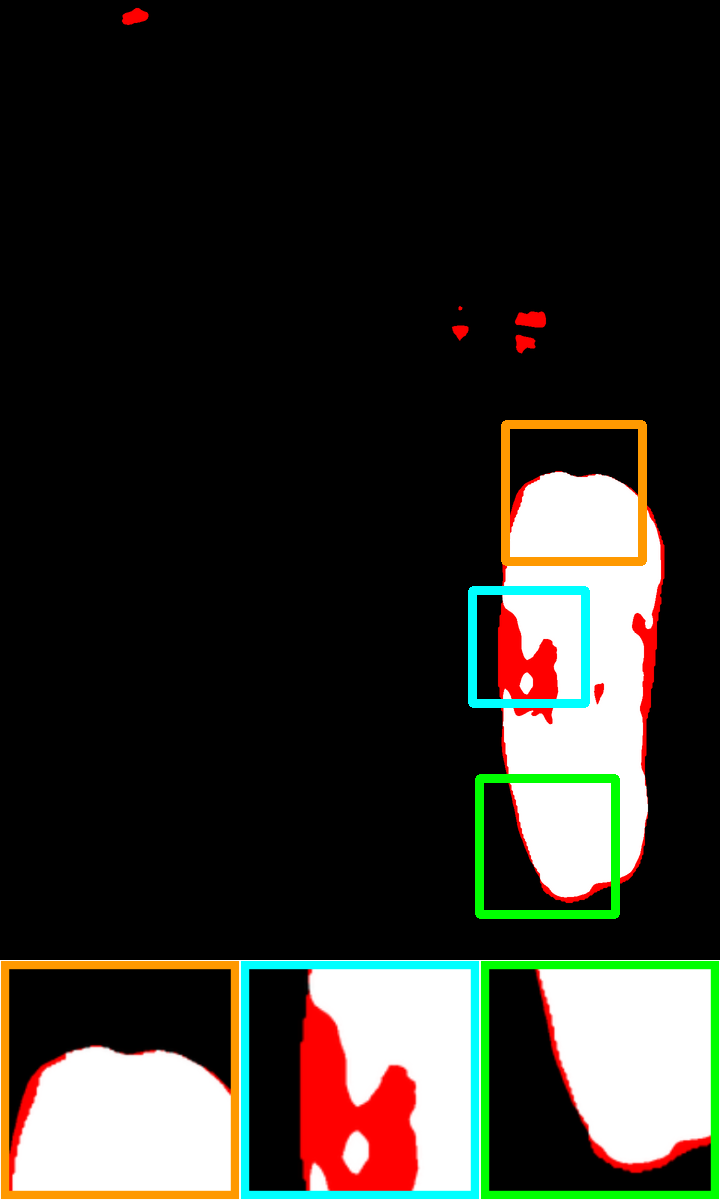} &
\includegraphics[width=0.12\textwidth]{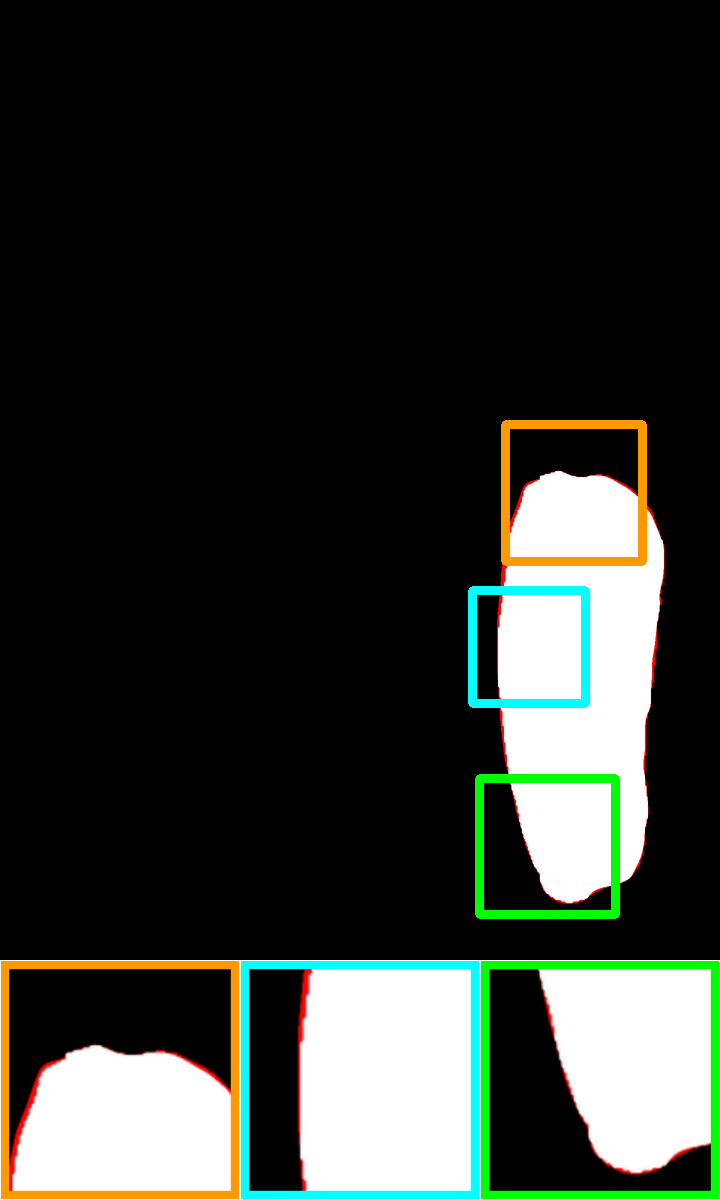} &
\includegraphics[width=0.12\textwidth]{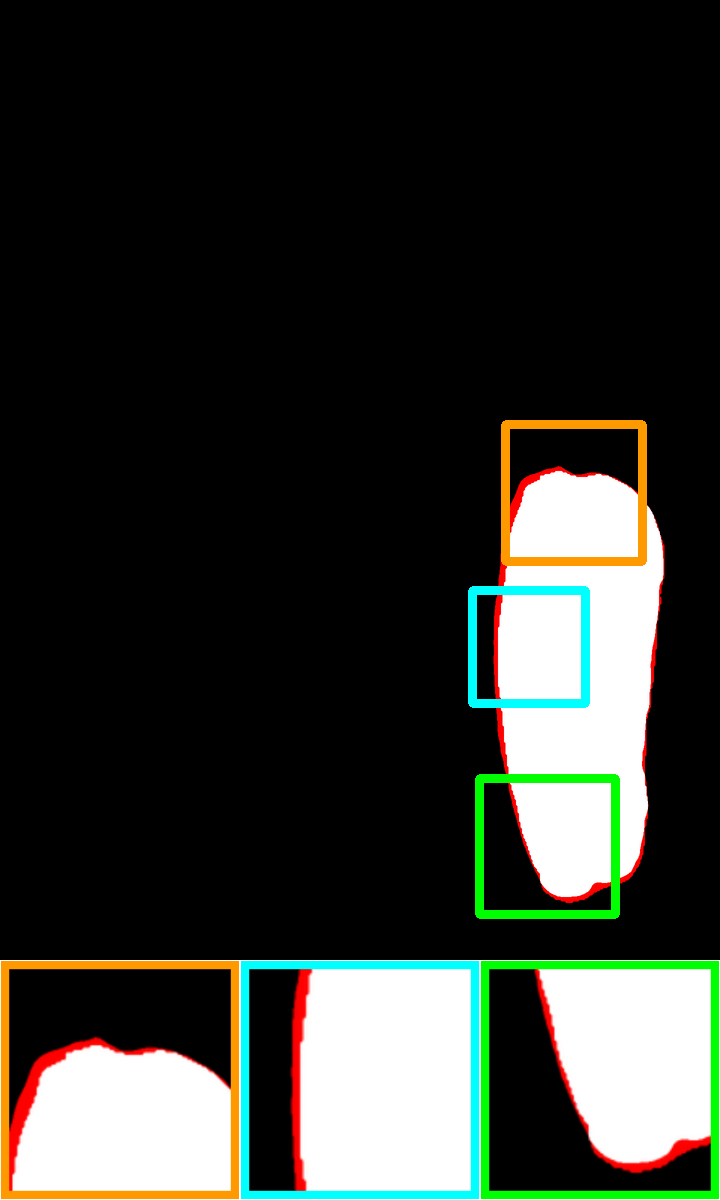} &
\includegraphics[width=0.12\textwidth]{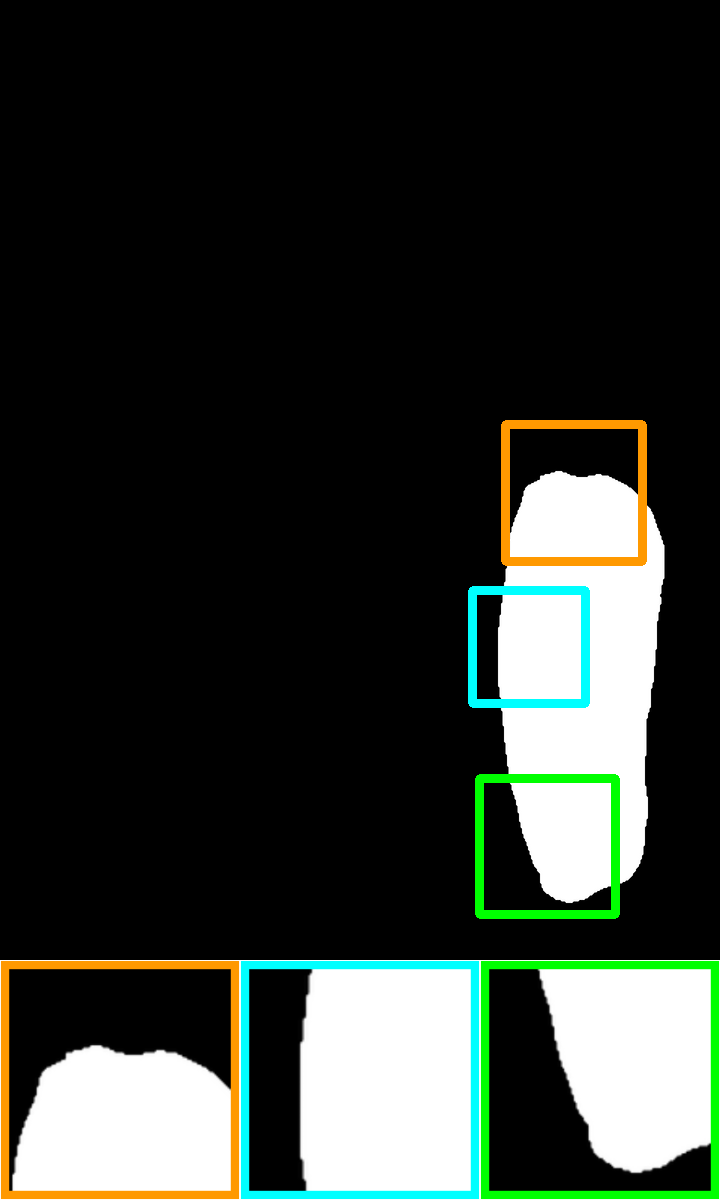} &
\includegraphics[width=0.12\textwidth]{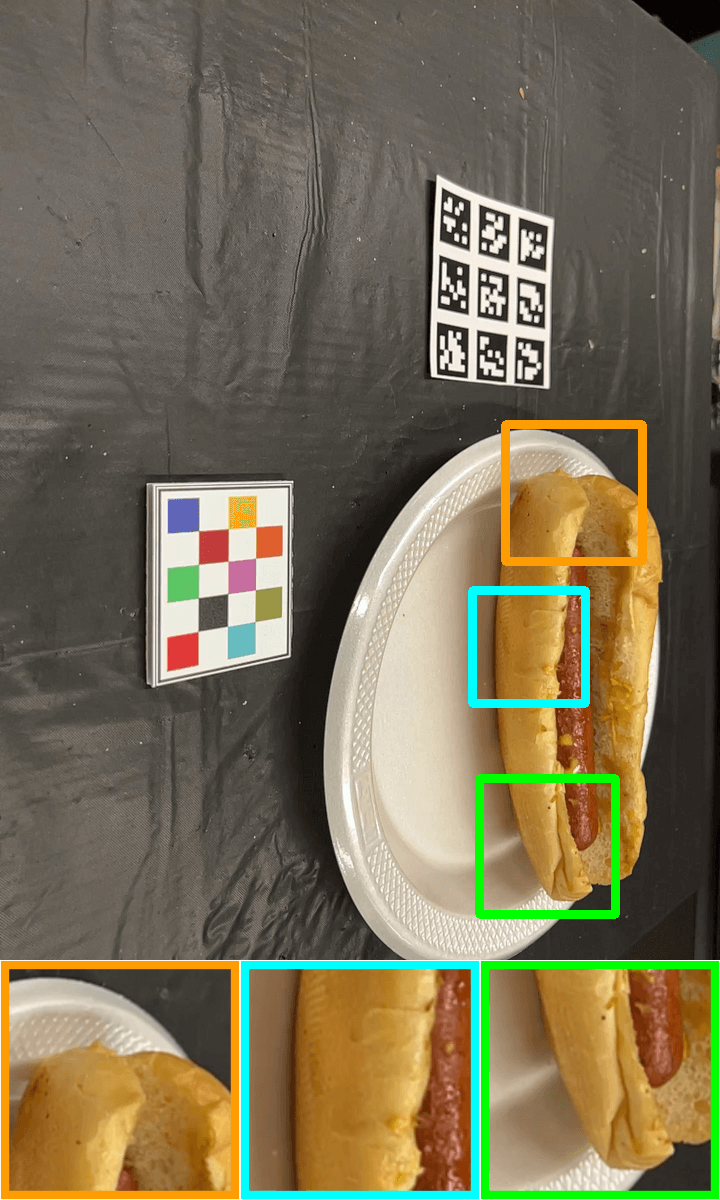} \\

\hline

    \end{tabular}
    \label{tab:MTF_2d_comparisons}
\end{table}

%%%%%%%%%%%%%%%%%%%%%%%%%%%%%%%%%%%%%%%%%%%%%%%%%%%%%

\begin{table}[!htbp]
    \centering
    \small
    \caption{Qualitative comparison of 2D Methods on N5K Dataset}
    \setlength{\tabcolsep}{1pt}
    \begin{tabular}{c|ccccccc}
    \hline
    %Datasets & BiRefNet & CCNET RELEM & FPN RELEM & SETR MLA & SWIN BASE & GT \\
    N5K & BiRefNet & CCNET & FPN & SeTR & SWIN & GT & RGB\\
    \hline
    \hline
    
\raisebox{+0.7cm}{\centering{1}} &

\includegraphics[width=0.12\textwidth]{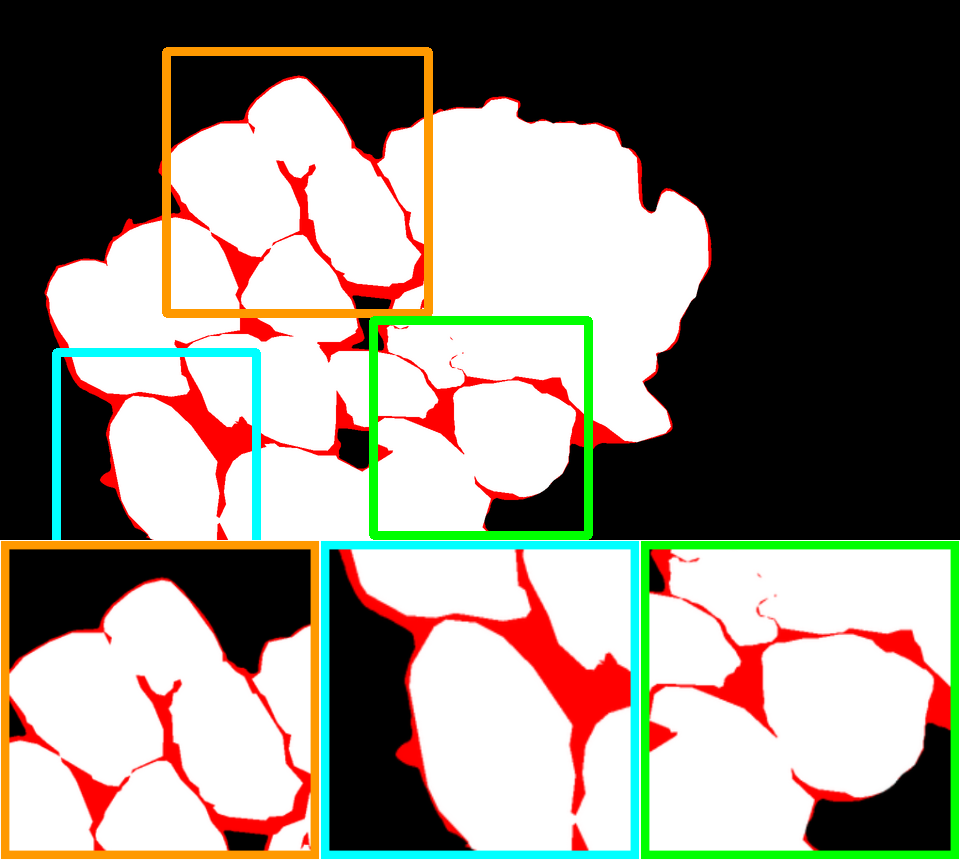} &
\includegraphics[width=0.12\textwidth]{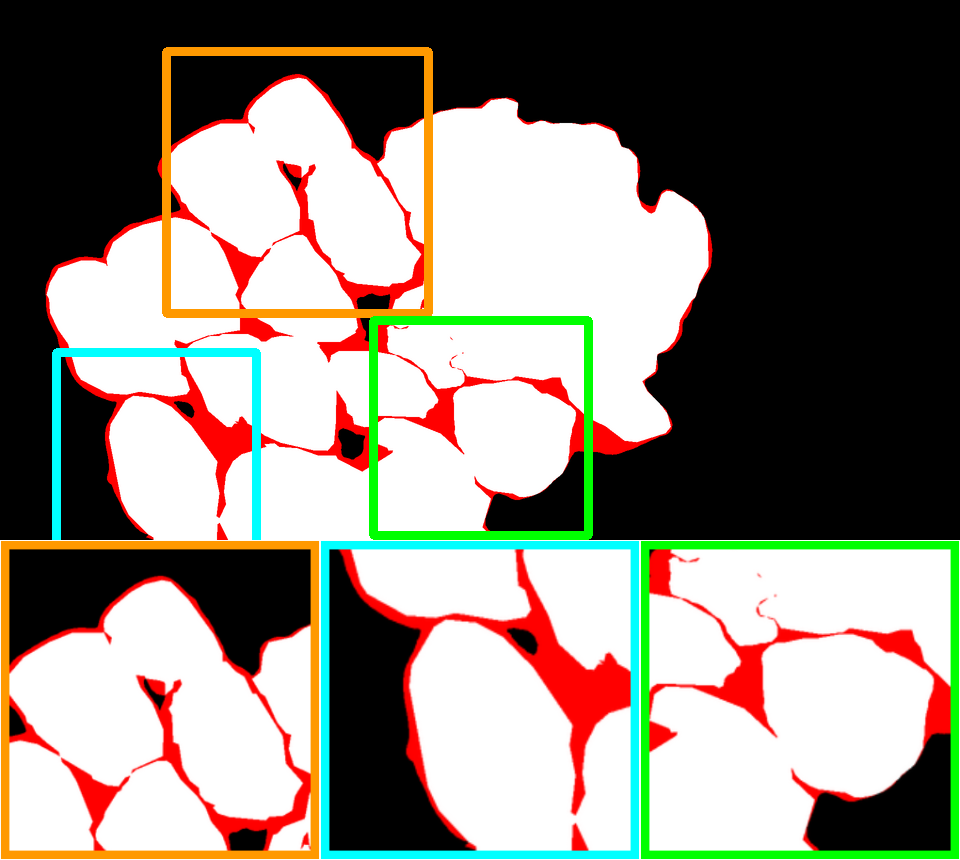} &
\includegraphics[width=0.12\textwidth]{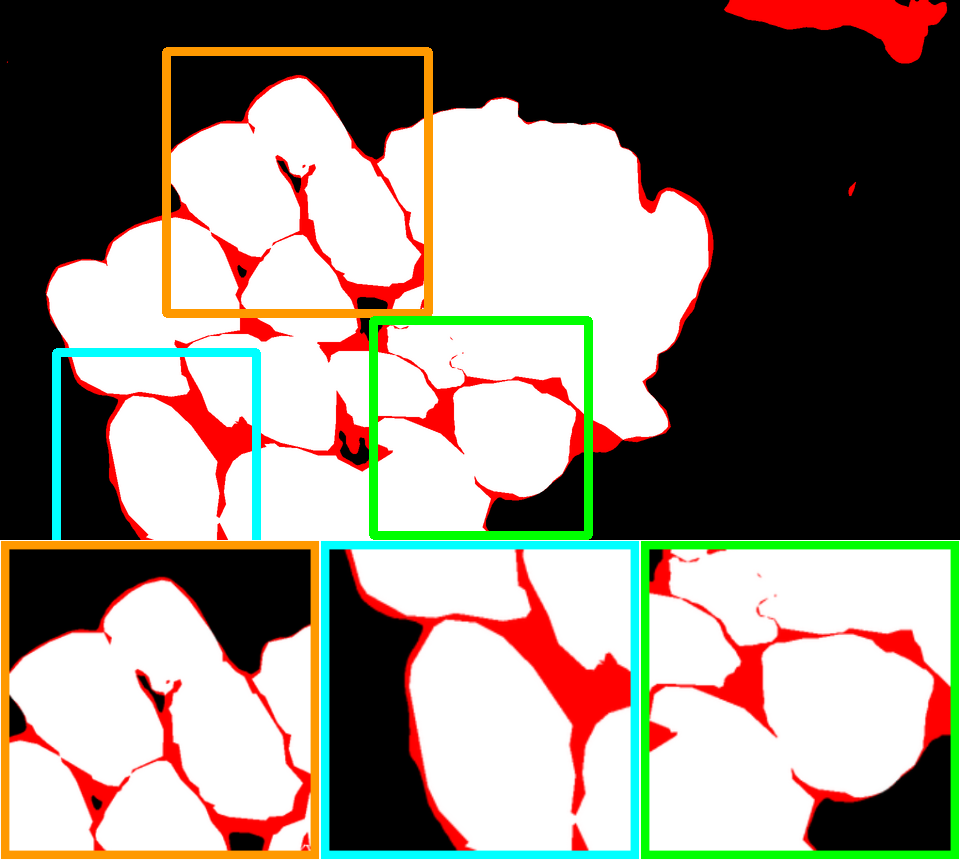} &
\includegraphics[width=0.12\textwidth]{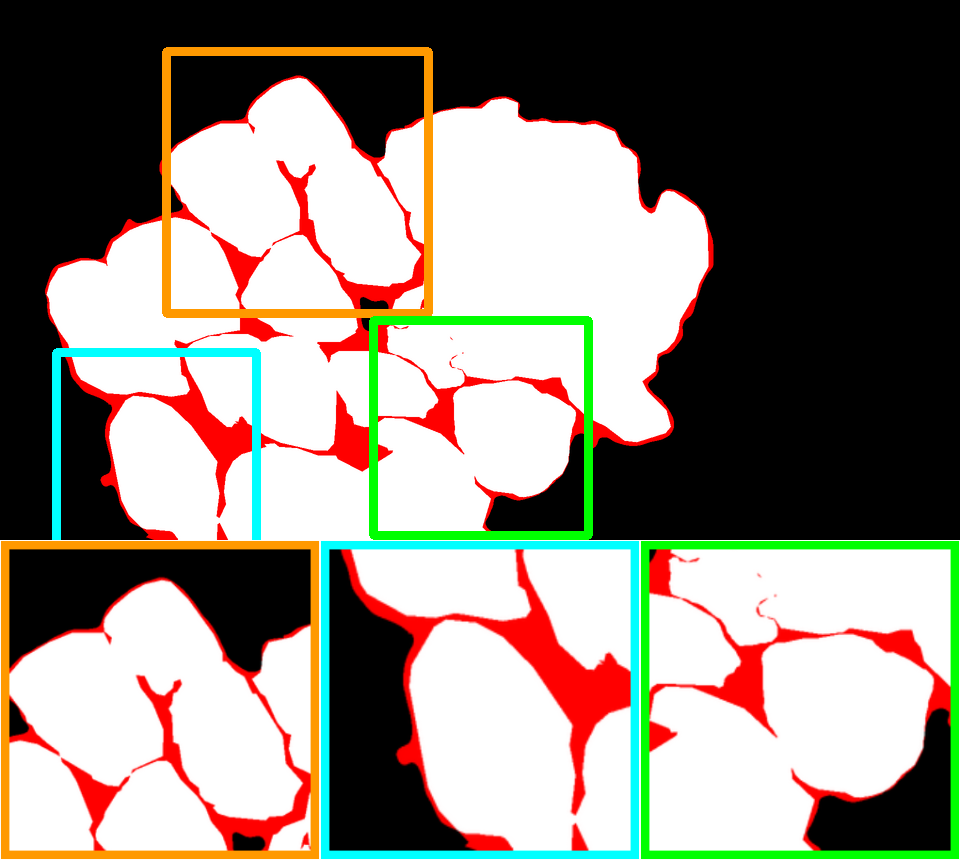} &
\includegraphics[width=0.12\textwidth]{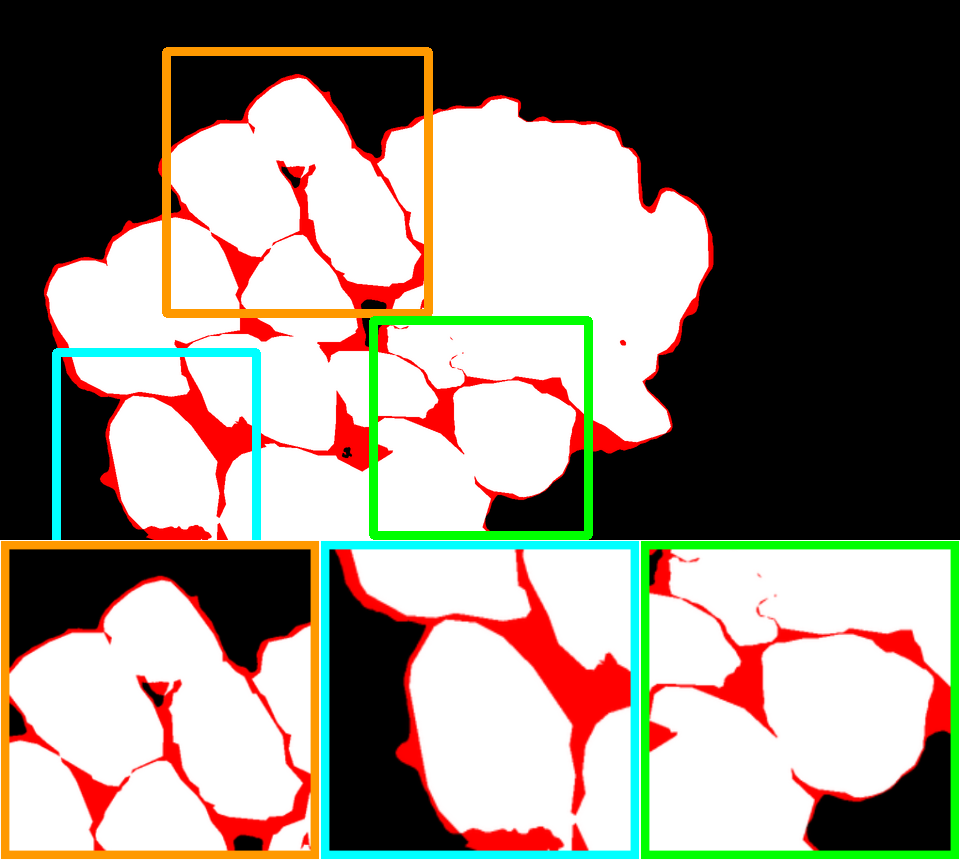} &
\includegraphics[width=0.12\textwidth]{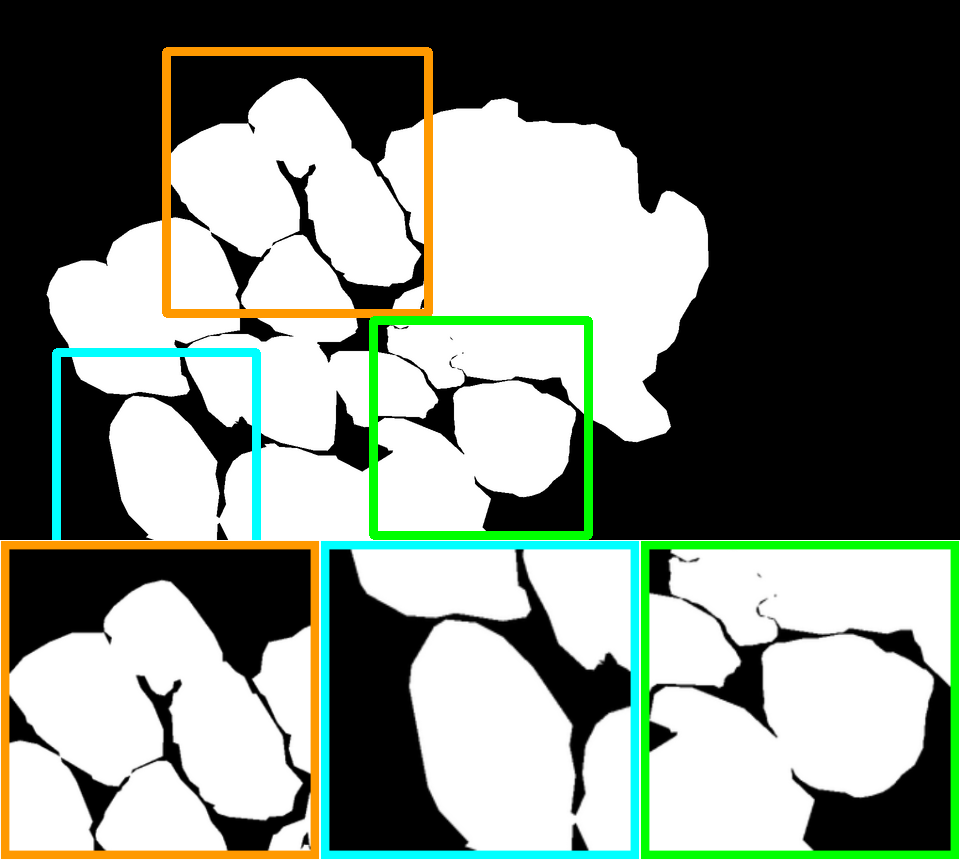} &
\includegraphics[width=0.12\textwidth]{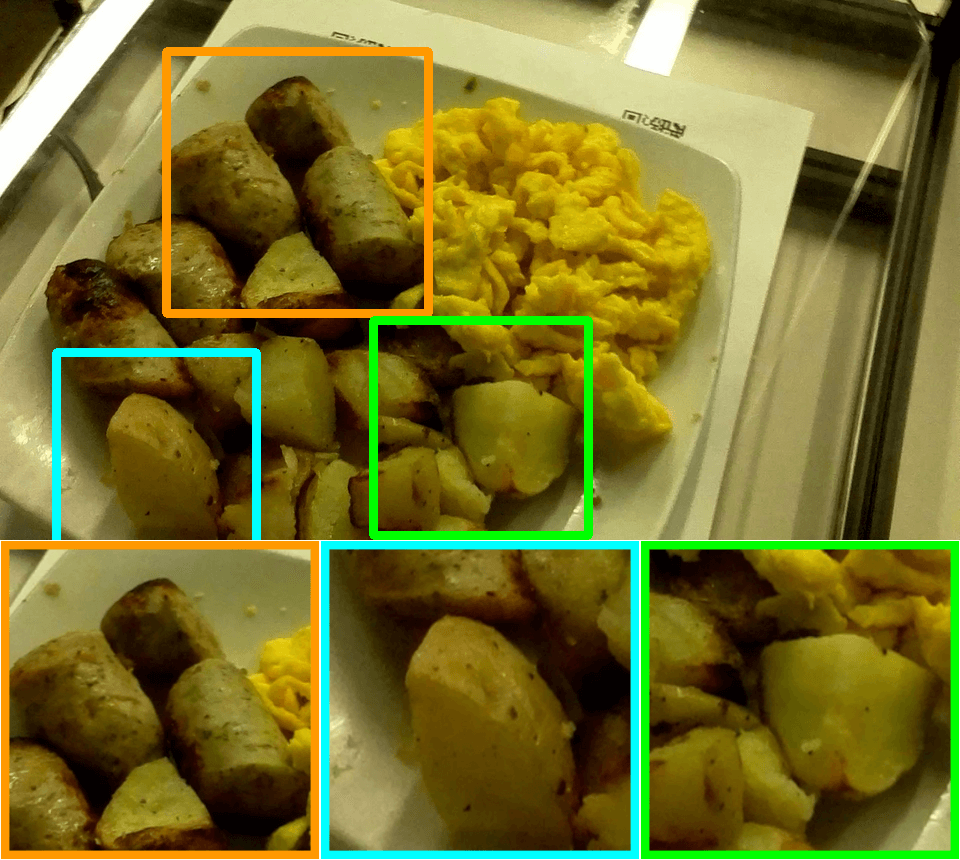} \\

\hline

\raisebox{+0.7cm}{\centering{2}} & 

\includegraphics[width=0.12\textwidth]{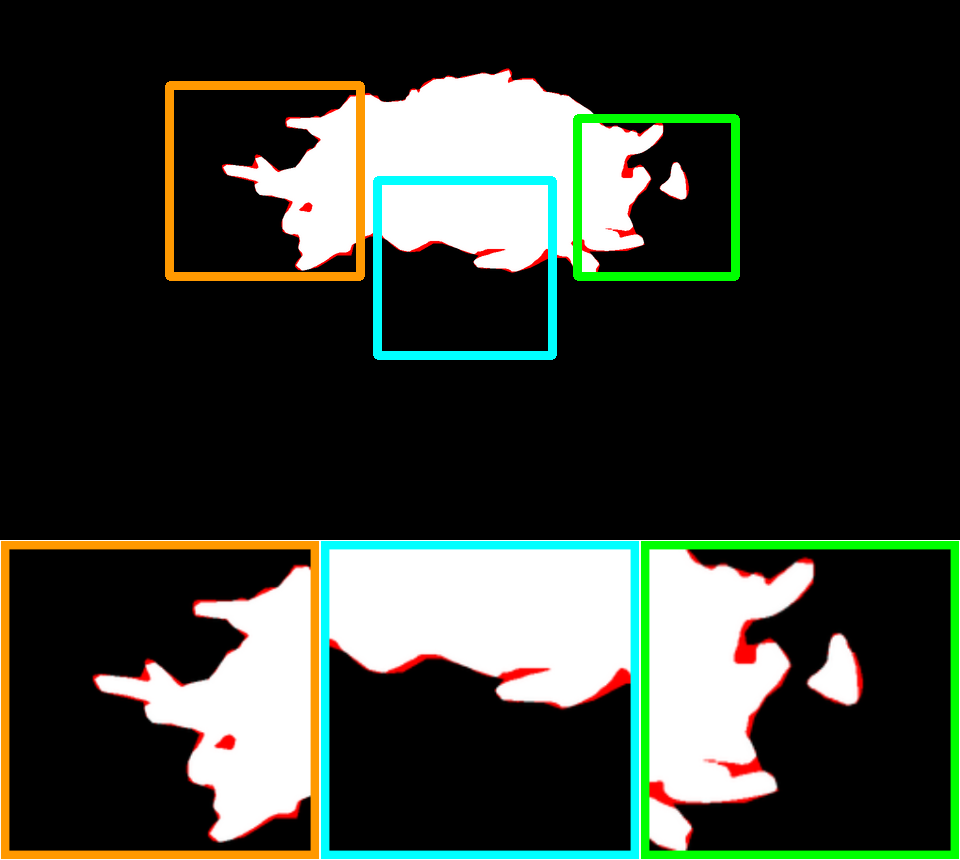} &
\includegraphics[width=0.12\textwidth]{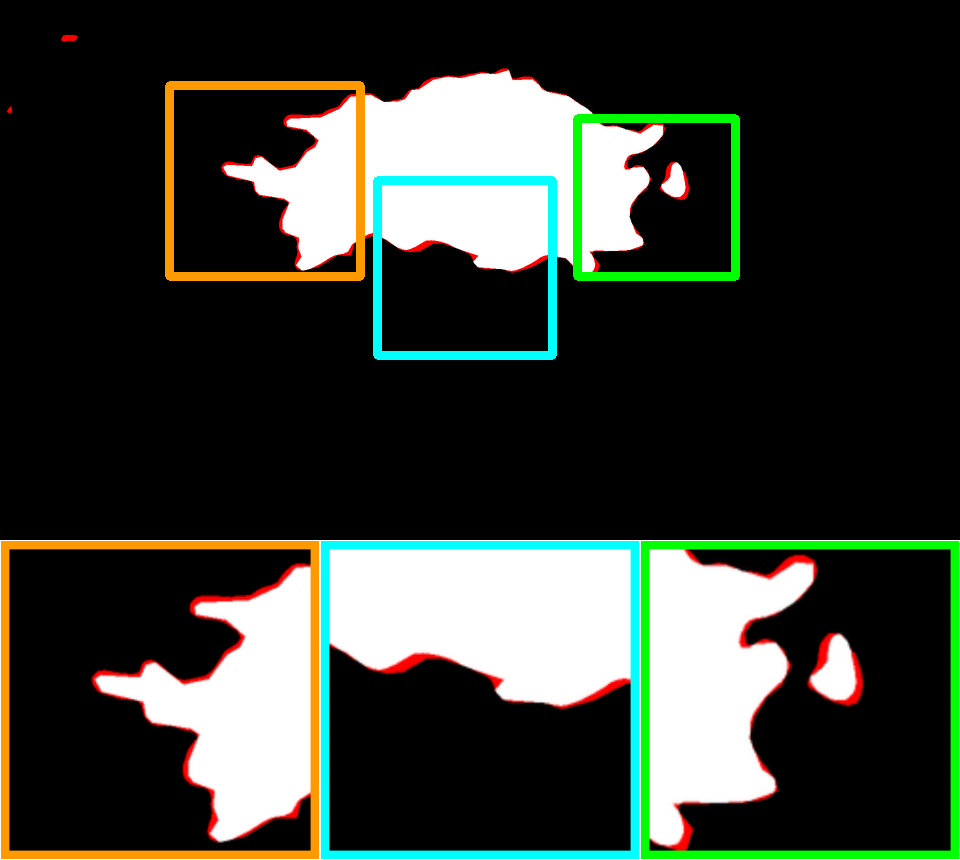} &
\includegraphics[width=0.12\textwidth]{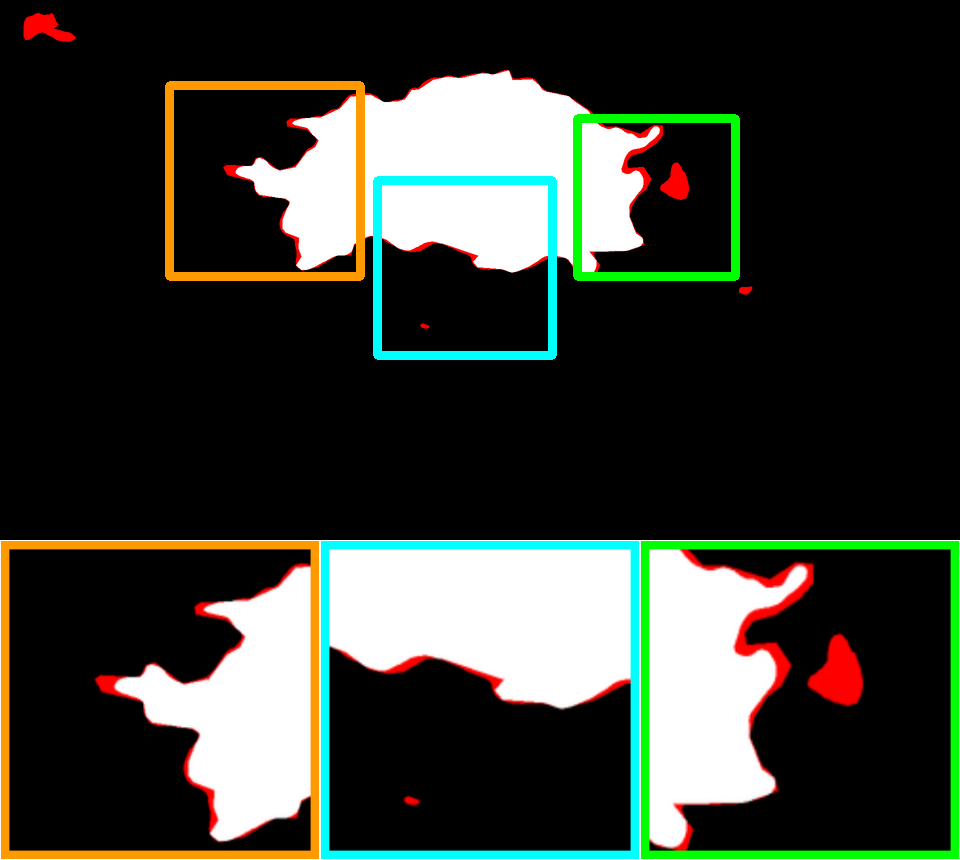} &
\includegraphics[width=0.12\textwidth]{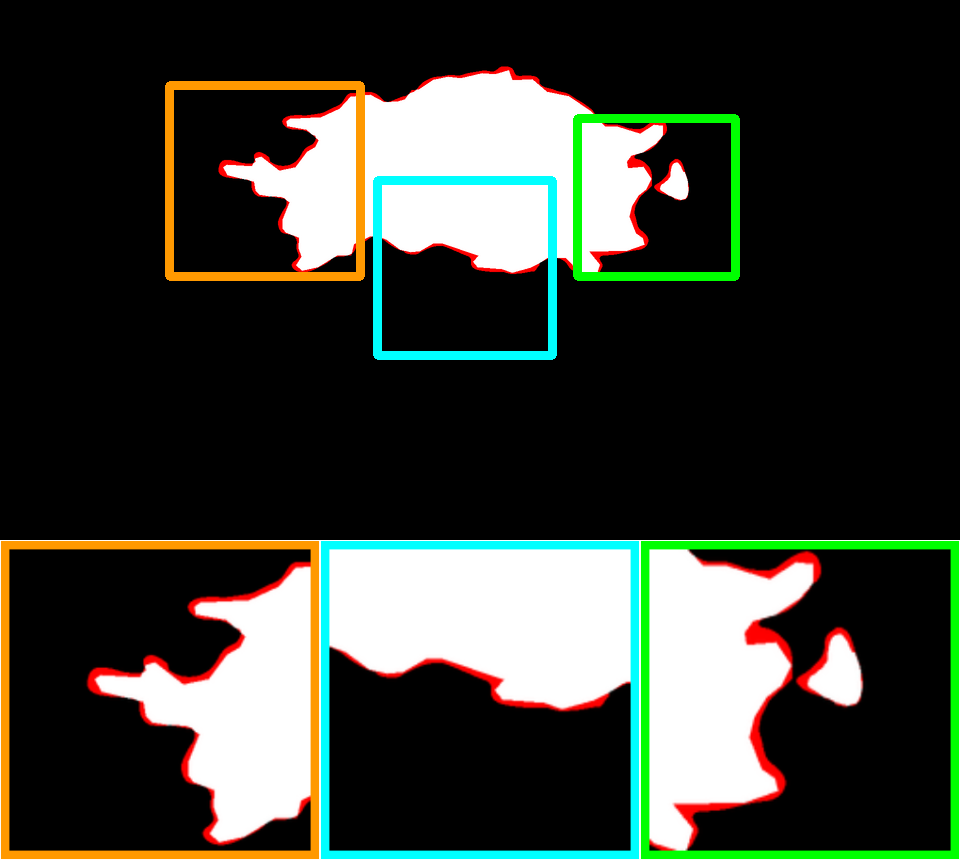} &
\includegraphics[width=0.12\textwidth]{2_056_SE.png} &
\includegraphics[width=0.12\textwidth]{2_056_GT.png} &
\includegraphics[width=0.12\textwidth]{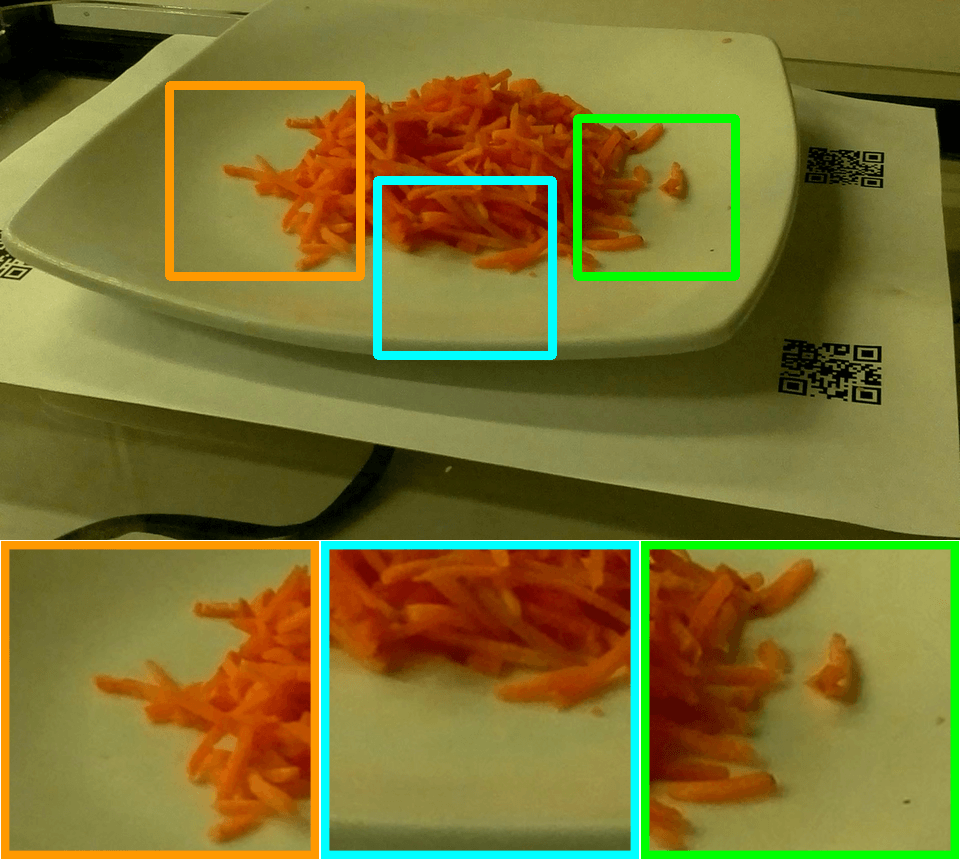} \\

\hline

\raisebox{+0.7cm}{\centering{3}} & 

\includegraphics[width=0.12\textwidth]{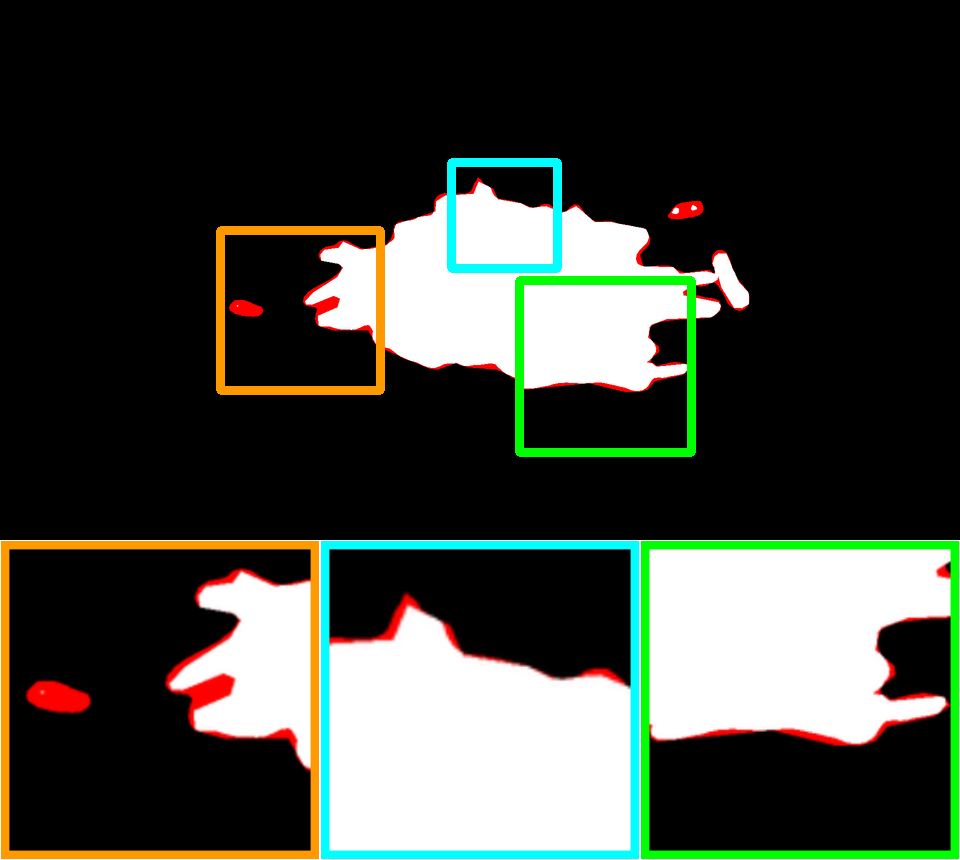} &
\includegraphics[width=0.12\textwidth]{3_038_CCNET_RELEM_binary.png} &
\includegraphics[width=0.12\textwidth]{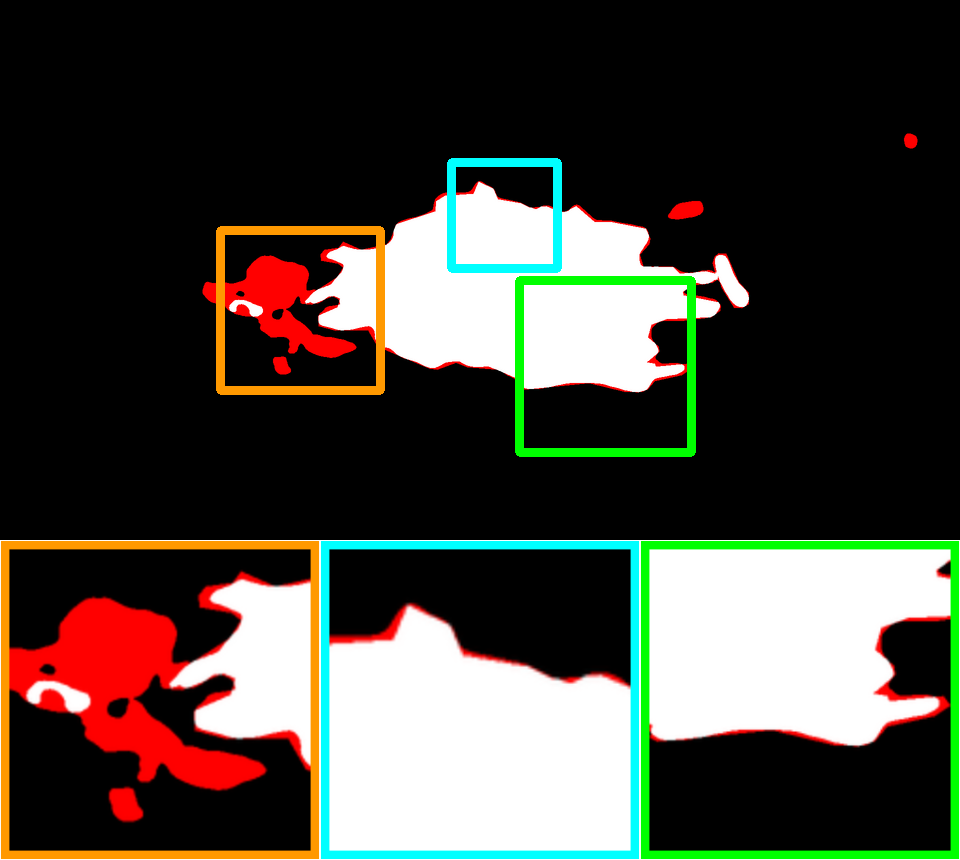} &
\includegraphics[width=0.12\textwidth]{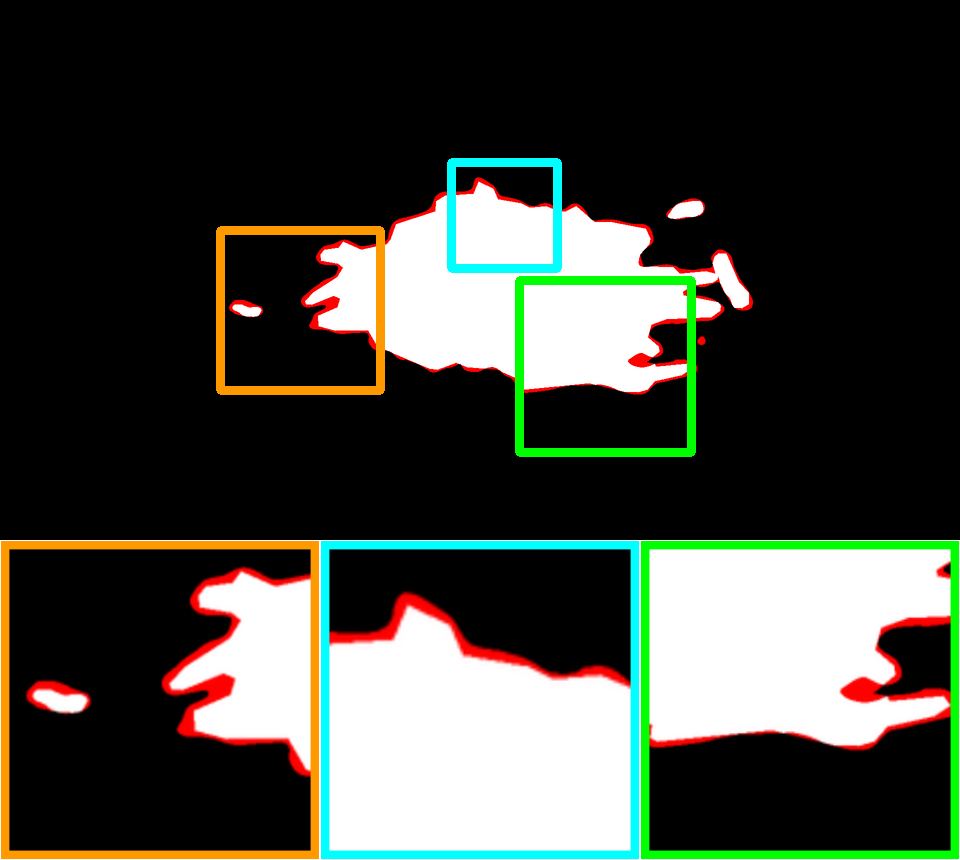} &
\includegraphics[width=0.12\textwidth]{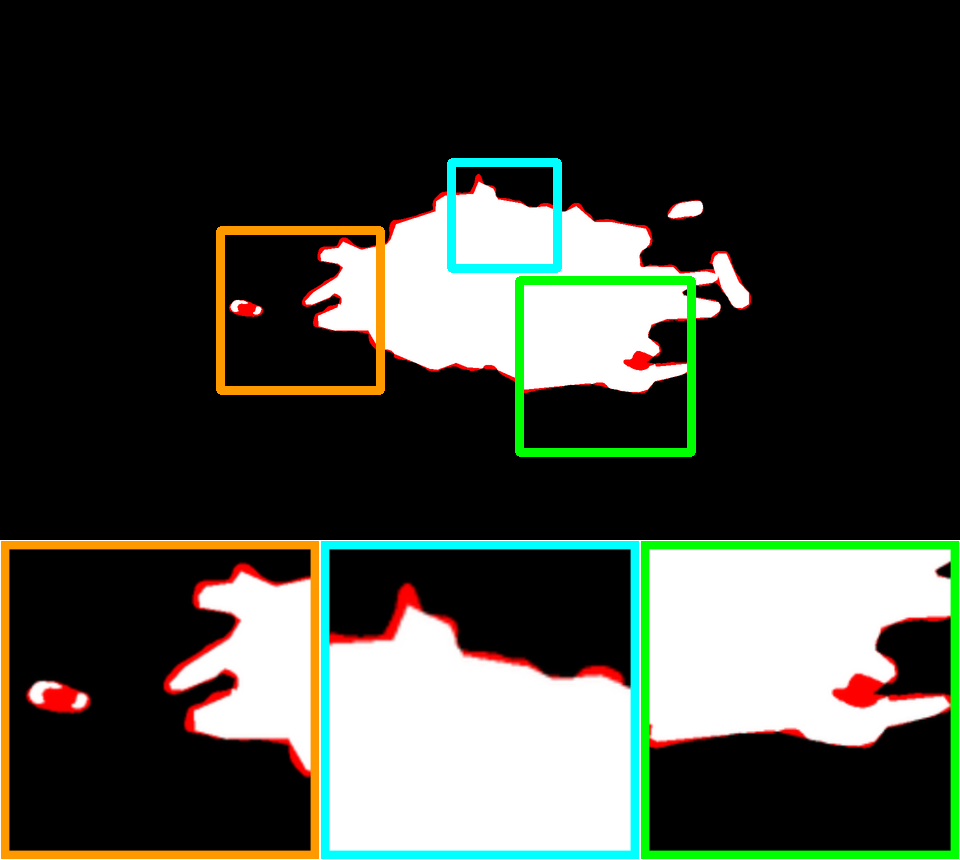} &
\includegraphics[width=0.12\textwidth]{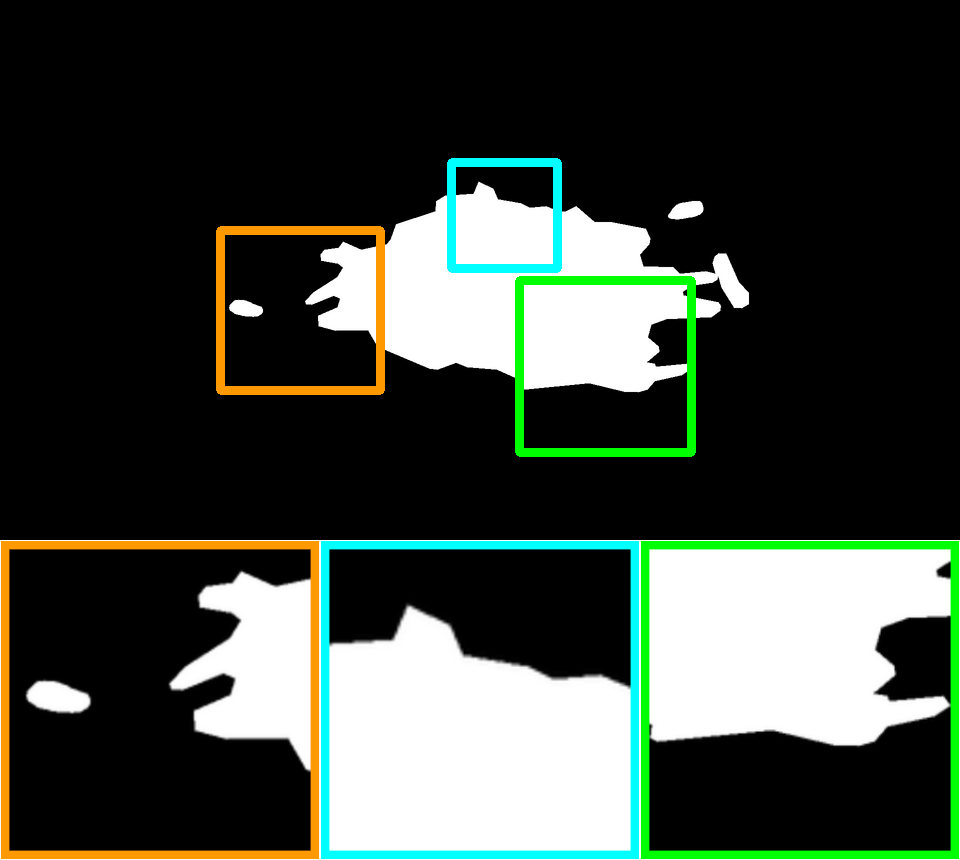} &
\includegraphics[width=0.12\textwidth]{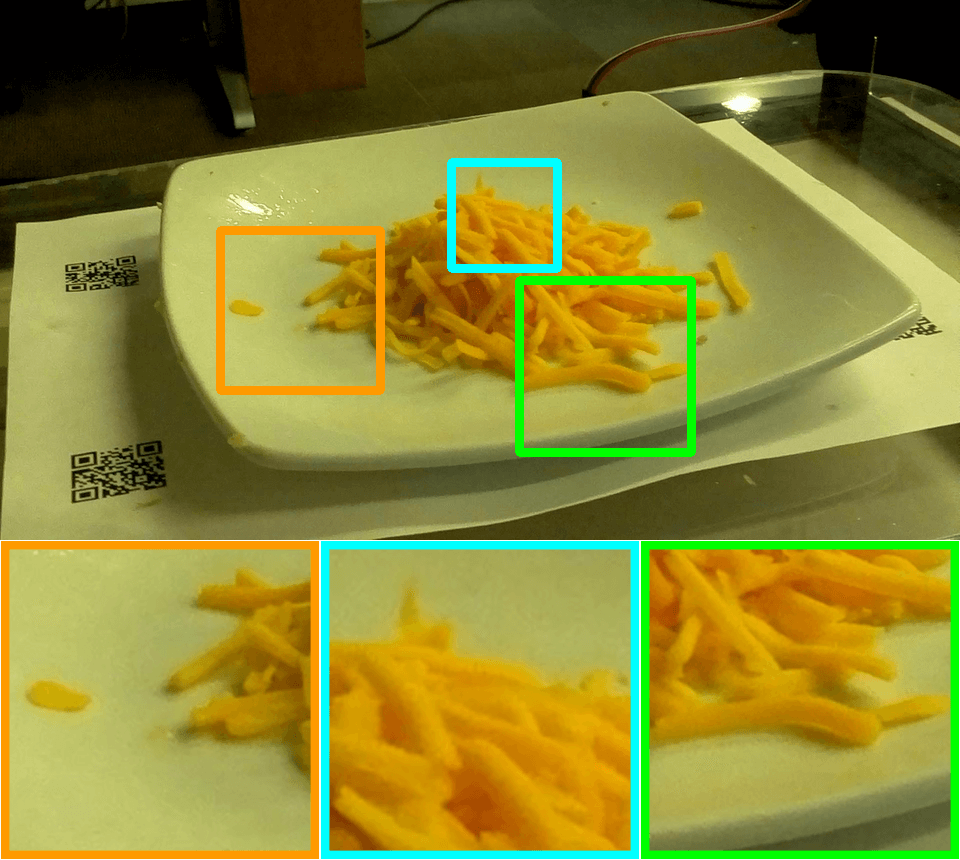} \\

\hline

\raisebox{+0.7cm}{\centering{5}} &

\includegraphics[width=0.12\textwidth]{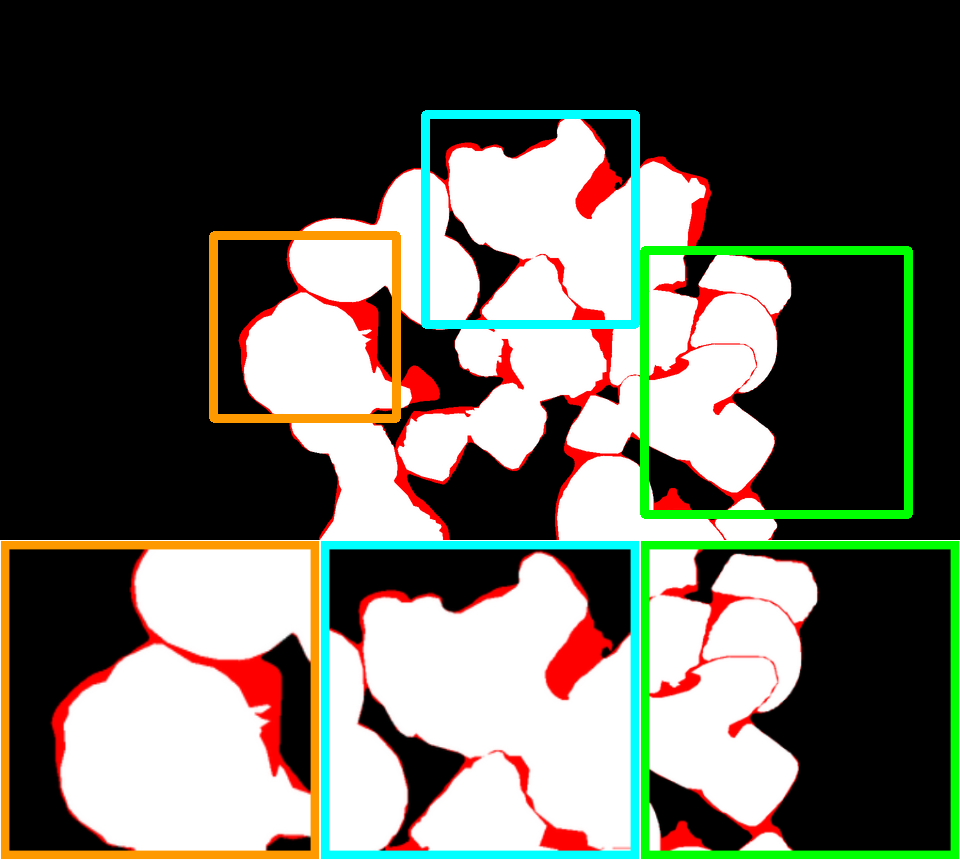} &
\includegraphics[width=0.12\textwidth]{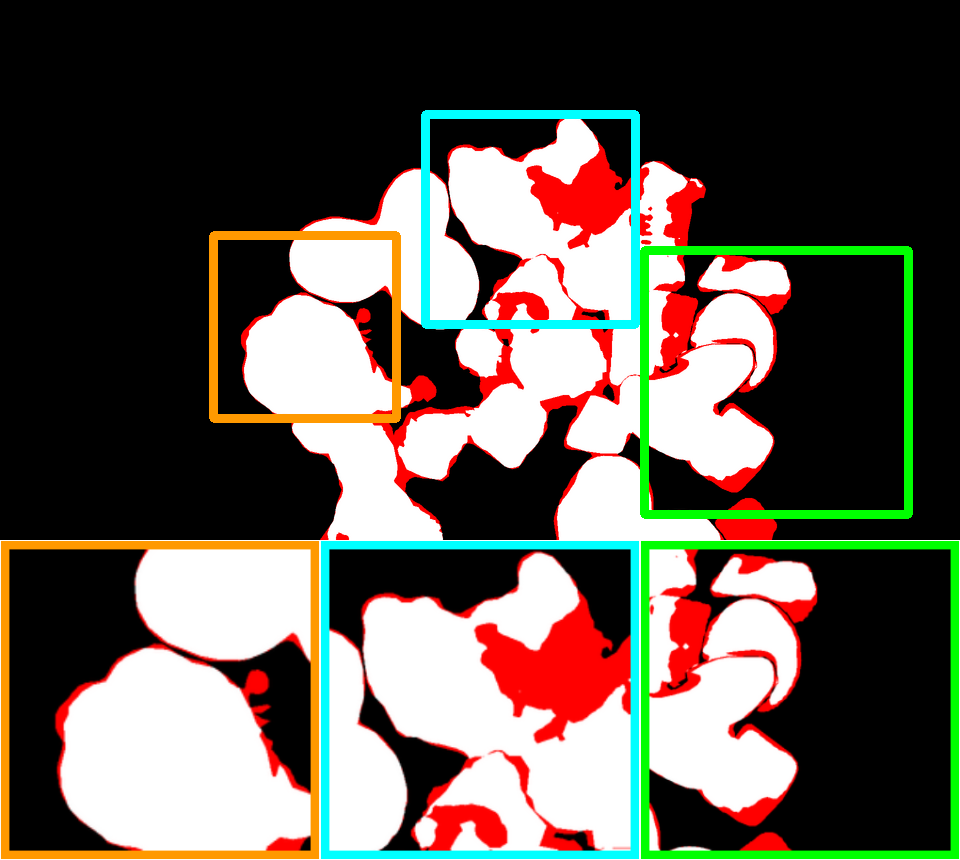} &
\includegraphics[width=0.12\textwidth]{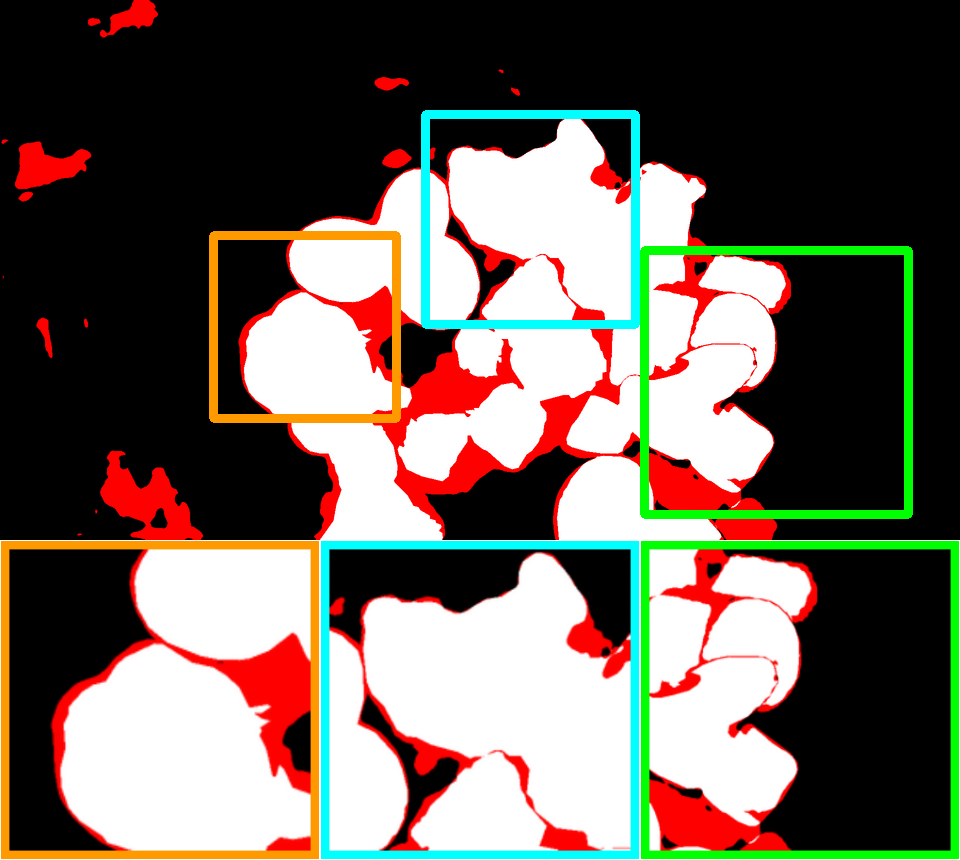} &
\includegraphics[width=0.12\textwidth]{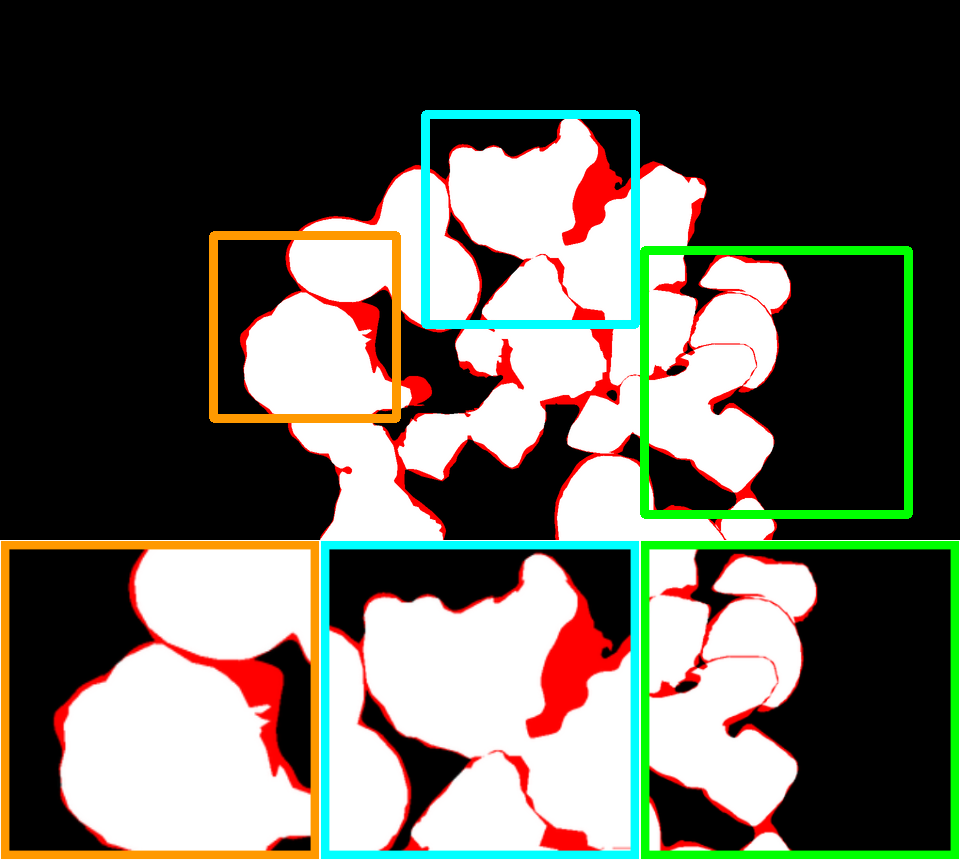} &
\includegraphics[width=0.12\textwidth]{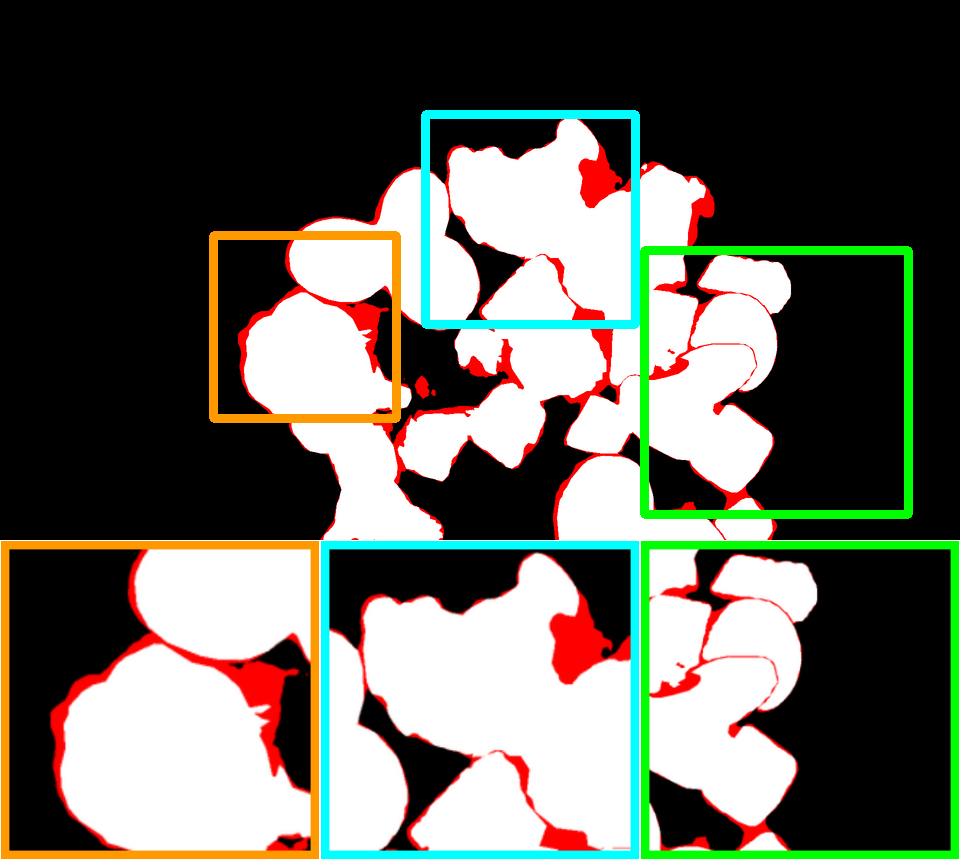} &
\includegraphics[width=0.12\textwidth]{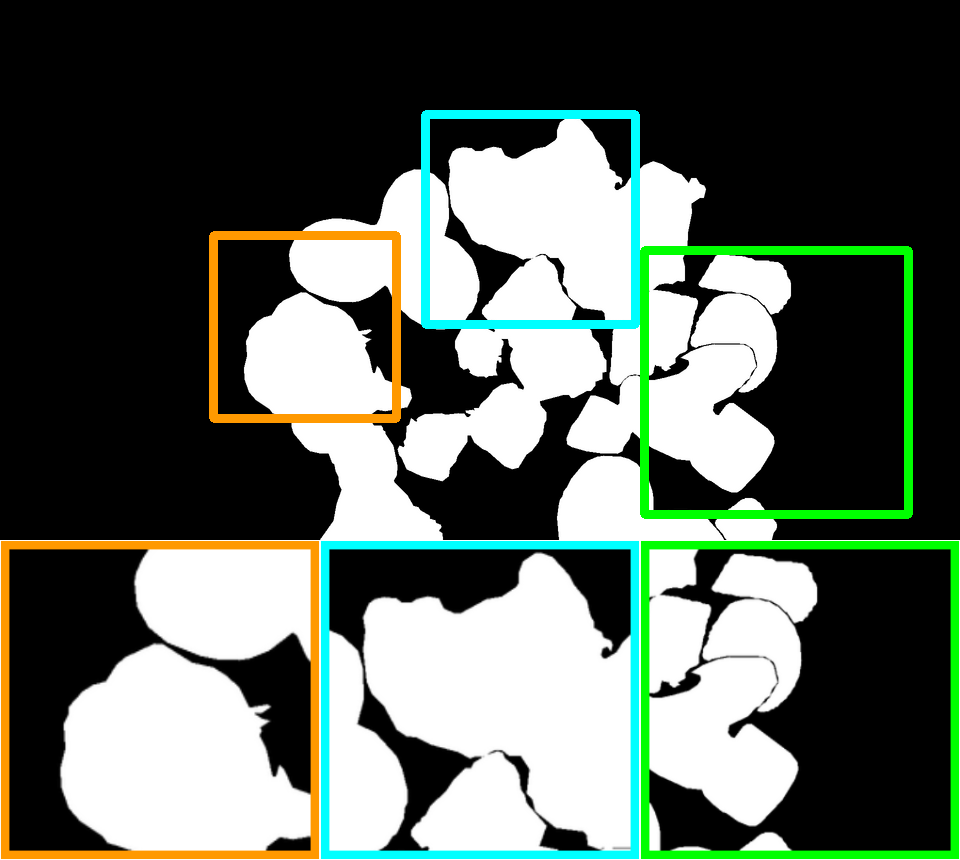} &
\includegraphics[width=0.12\textwidth]{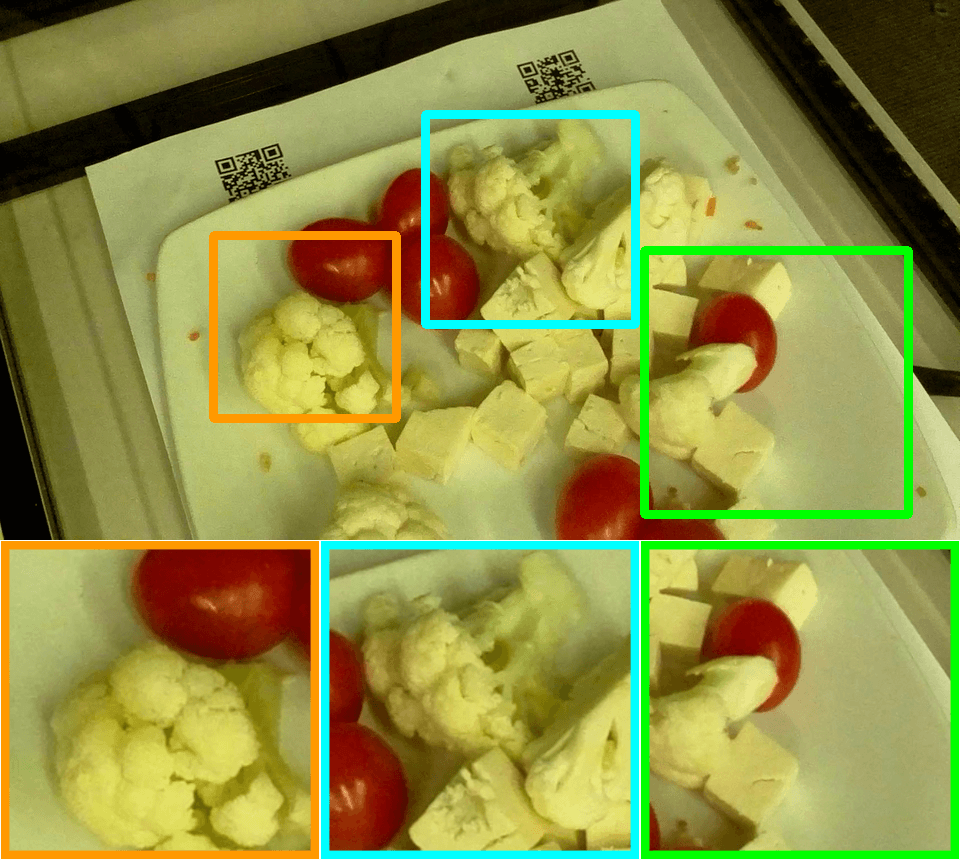} \\

\hline

\raisebox{+0.7cm}{\centering{9}} &

\includegraphics[width=0.12\textwidth]{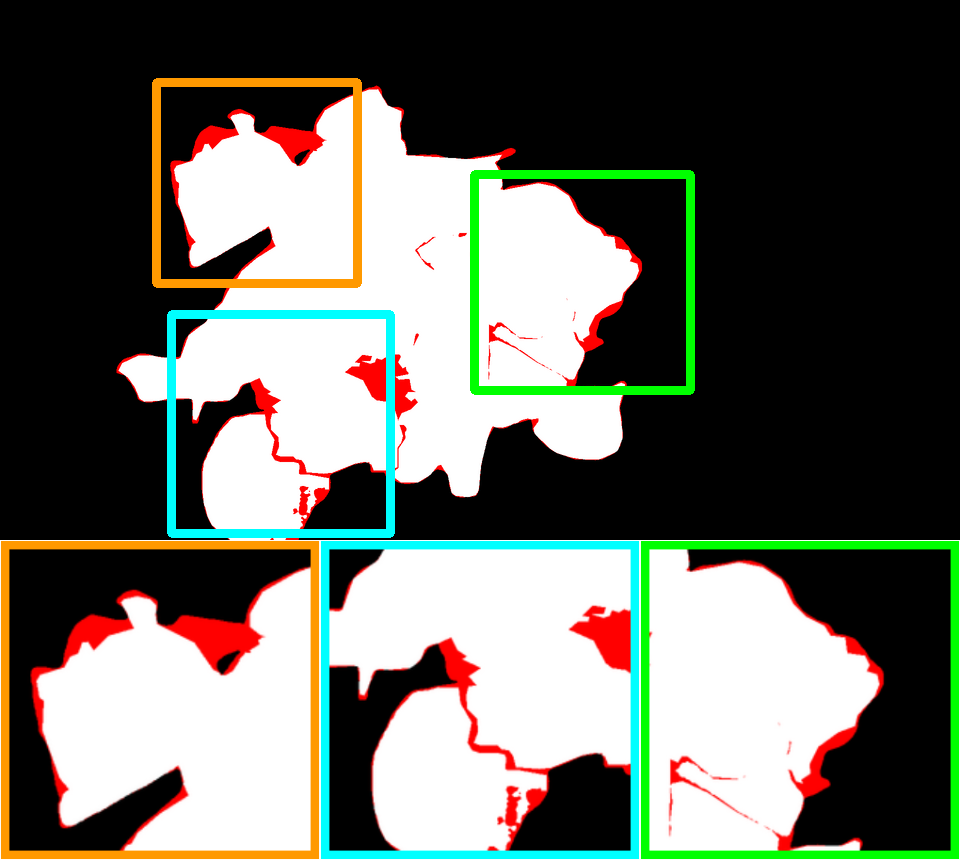} &
\includegraphics[width=0.12\textwidth]{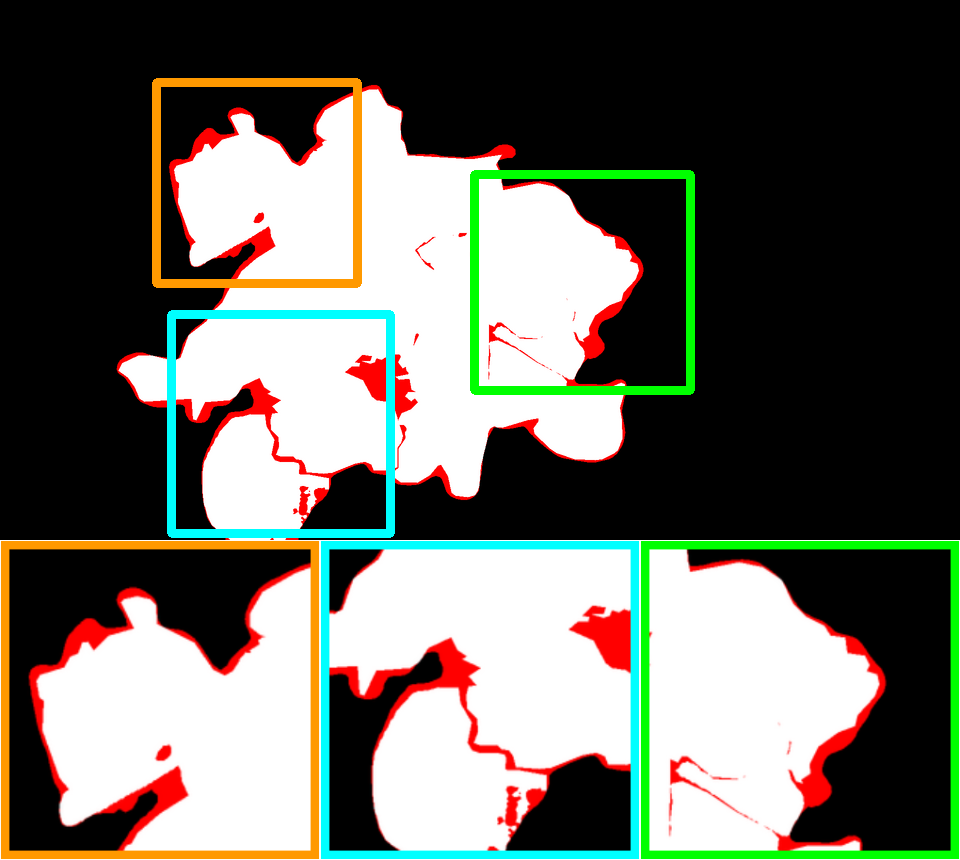} &
\includegraphics[width=0.12\textwidth]{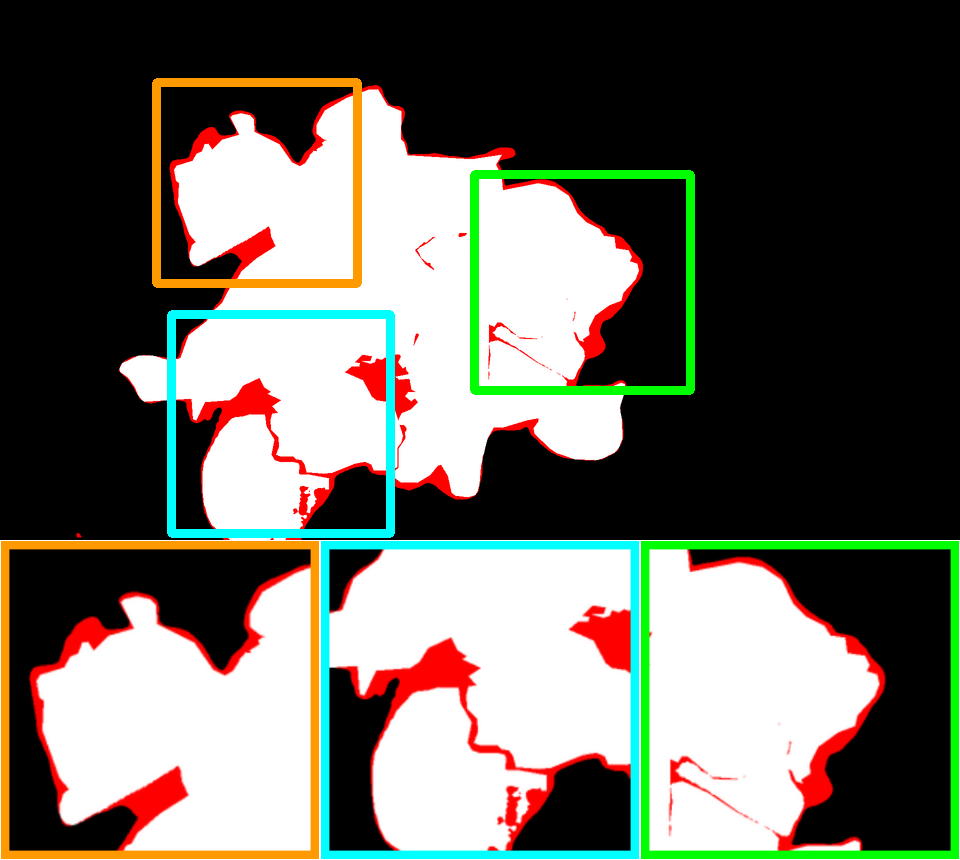} &
\includegraphics[width=0.12\textwidth]{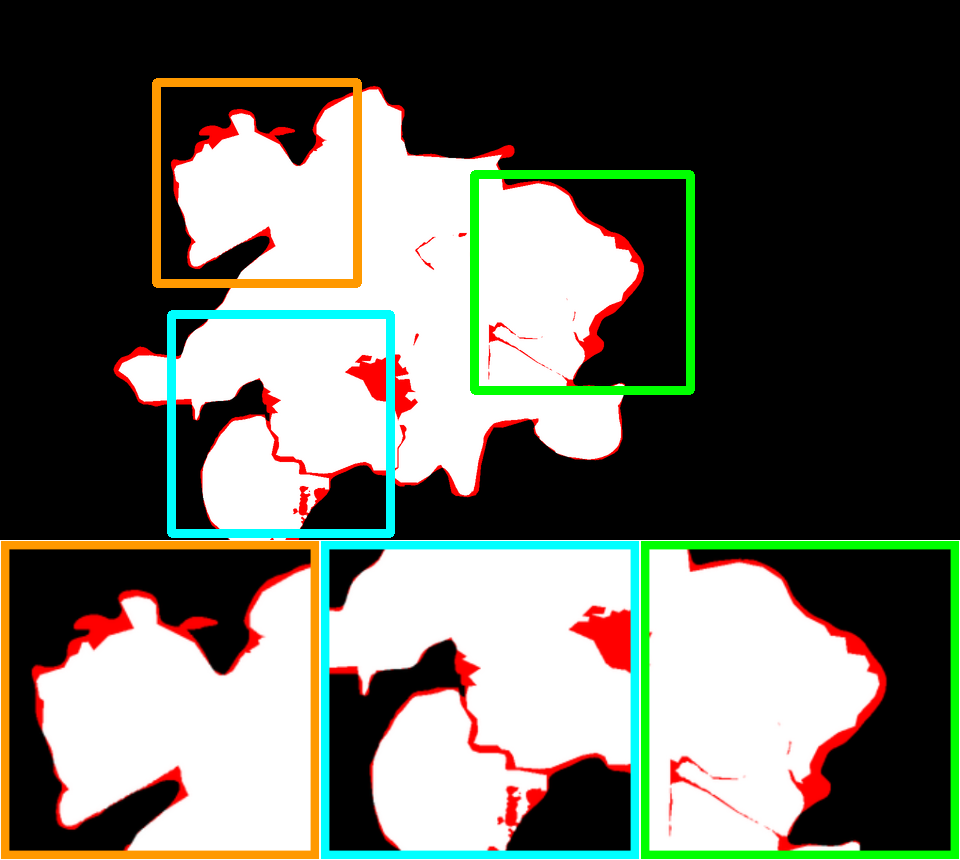} &
\includegraphics[width=0.12\textwidth]{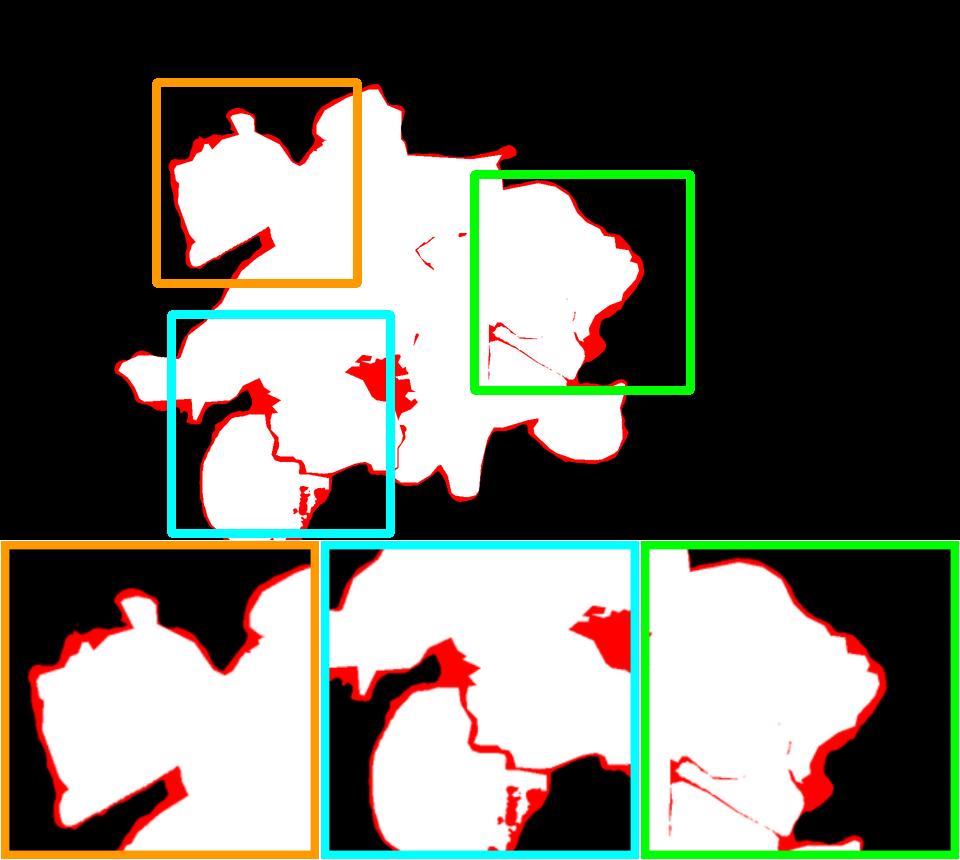} &
\includegraphics[width=0.12\textwidth]{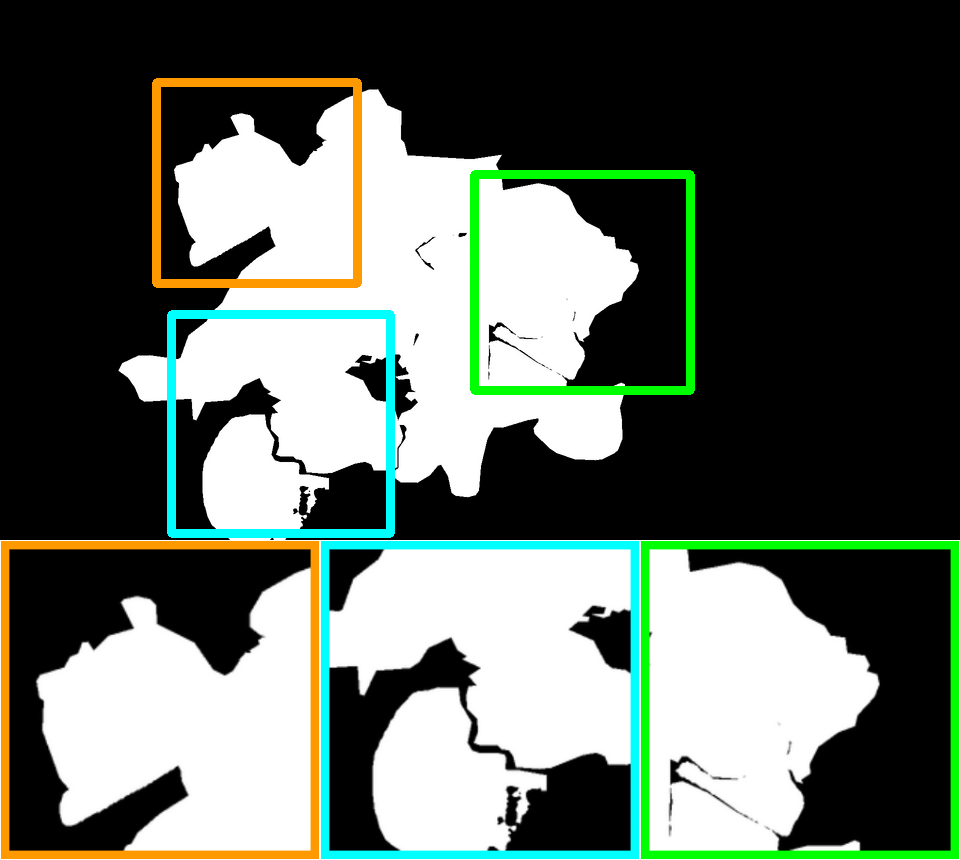} &
\includegraphics[width=0.12\textwidth]{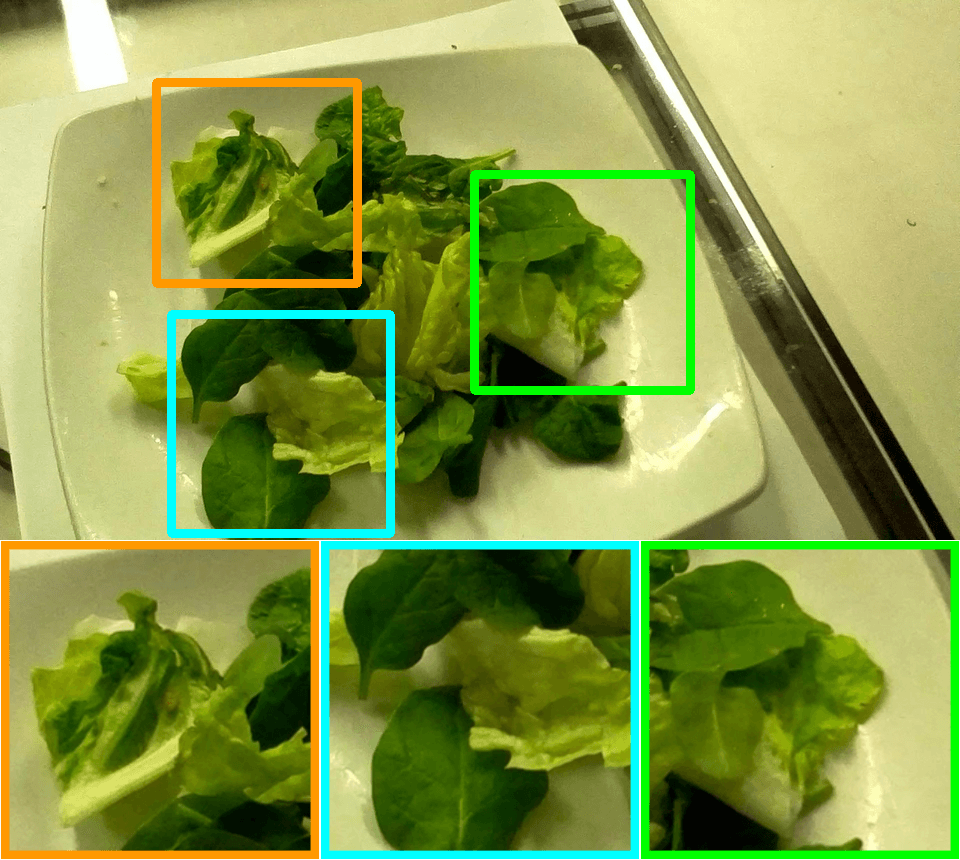} \\

\hline

    \end{tabular}
    \label{tab:N5K_2d_comparisons}
\end{table}

\end{document}